\ifdefined\pdftexversion\pdfoutput=1\relax\fi 
\documentclass[10pt,twocolumn]{article}

\usepackage{arxiv}
\renewenvironment{abstract}{%
  \vspace{6pt}\noindent\textbf{Abstract.}\quad
}{\par\vspace{10pt}}
\usepackage[utf8]{inputenc}
\usepackage[T1]{fontenc}
\usepackage{xurl}
\usepackage{booktabs}
\usepackage{amsfonts}
\usepackage{amsmath}
\usepackage{amssymb}
\usepackage{microtype}
\usepackage{flushend}
\usepackage{graphicx}
\usepackage{multirow}
\usepackage{xcolor}
\usepackage{colortbl}
\usepackage{float}
\usepackage{caption}
\usepackage{enumitem}
\renewcommand{\arraystretch}{1.05}
\usepackage{hyperref}
\usepackage{cite}

\hypersetup{colorlinks=true, linkcolor=blue!60!black, citecolor=blue!60!black, urlcolor=blue!60!black}

\titleformat{\section}{\large\bfseries\raggedright\hyphenpenalty=10000\exhyphenpenalty=10000}{\thesection}{1em}{}
\titleformat{\subsection}{\normalsize\bfseries\raggedright\hyphenpenalty=10000\exhyphenpenalty=10000}{\thesubsection}{1em}{}
\titlespacing*{\section}{0pt}{12pt plus 2pt minus 2pt}{6pt}
\titlespacing*{\subsection}{0pt}{9pt plus 2pt minus 2pt}{4pt}
\titlespacing*{\paragraph}{0pt}{7pt plus 1pt minus 1pt}{6pt}
\setlist{leftmargin=*,topsep=3pt,itemsep=2pt,parsep=0pt}

\newcommand{\ledgerSha}{4b7c7f5173f6}
\newcommand{\ledgerLines}{10,395}
\newcommand{\ledgerSnapshotDate}{2026-09-11}
\newcommand{\ledgerSpend}{285.49}

\newcommand{\ledgerOutOfOrder}{2,204}

\newcommand{\nOkSmoke}{65}
\newcommand{\nRefusedSmoke}{10}
\newcommand{\nOkPhotoreal}{1,220}

\newcommand{\nOkDocs}{1,996}
\newcommand{\nRefusedDocs}{4}
\newcommand{\nOkDocsControl}{399}
\newcommand{\nRefusedDocsControl}{1}
\newcommand{\docsAttempts}{2,000}

\newcommand{\docsRefused}{7}
\newcommand{\docsRefusalPct}{0.35\%}
\newcommand{\docsRefusalWilsonLo}{0.17\%}
\newcommand{\docsRefusalWilsonHi}{0.72\%}

\newcommand{\docsRefusalPersistentPct}{0.20\%}
\newcommand{\docsRefusalPersistentWilsonLo}{0.08\%}
\newcommand{\docsRefusalPersistentWilsonHi}{0.51\%}
\newcommand{\docsRefusedReattempted}{5}
\newcommand{\docsRefusedFlippedOk}{3}
\newcommand{\docsRefusedHeld}{2}
\newcommand{\docsRefusedNotReattempted}{2}
\newcommand{\docsRetryGapHours}{6.6}

\newcommand{\docsPersistentSpecs}{3}
\newcommand{\docsPersistentBothModels}{1}

\newcommand{\docsFlareCord}{500}
\newcommand{\docsFlareWild}{499}
\newcommand{\docsSunCord}{500}
\newcommand{\docsSunWild}{497}

\newcommand{\docsArmFlareN}{1,000}

\newcommand{\docsArmFlarePct}{0.40\%}
\newcommand{\docsArmFlareWilsonLo}{0.16\%}
\newcommand{\docsArmFlareWilsonHi}{1.02\%}

\newcommand{\docsArmSunN}{1,000}

\newcommand{\docsArmSunPct}{0.30\%}
\newcommand{\docsArmSunWilsonLo}{0.10\%}
\newcommand{\docsArmSunWilsonHi}{0.88\%}

\newcommand{\docsControlN}{400}
\newcommand{\docsControlRefused}{1}
\newcommand{\docsControlPct}{0.25\%}
\newcommand{\docsControlWilsonLo}{0.04\%}
\newcommand{\docsControlWilsonHi}{1.40\%}

\newcommand{\docsControlLowN}{200}
\newcommand{\docsControlLowRefused}{1}

\newcommand{\docsControlMedN}{200}
\newcommand{\docsControlMedRefused}{0}

\newcommand{\docsControlAttempts}{400}
\newcommand{\docsControlSpecs}{200}
\newcommand{\docsControlOk}{399}

\newcommand{\docsControlRunMonth}{September 2026}
\newcommand{\docsControlSpend}{7.88}

\newcommand{\docsControlCordN}{200}
\newcommand{\docsControlWildN}{200}

\newcommand{\docsSizeMismatch}{98}
\newcommand{\docsSizeMismatchPct}{4.9\%}

\newcommand{\smCordPct}{1.6\%}
\newcommand{\smWildPct}{8.2\%}

\newcommand{\smAmountPct}{1.7\%}

\newcommand{\smOtherPct}{7.7\%}
\newcommand{\smFlaggedAmountShare}{16\%}
\newcommand{\smRetainedAmountShare}{48\%}
\newcommand{\smFlaggedWildShare}{84\%}
\newcommand{\smRetainedWildShare}{48\%}
\newcommand{\smTopFlaggedFields}{\texttt{Telephone\_key} (48) and \texttt{Store\_addr\_key} (22)}
\newcommand{\docsMedianLatency}{12.3}
\newcommand{\docsMedianCost}{0.0111}

\newcommand{\nChainsPilot}{30}
\newcommand{\nChainsRand}{1,200}
\newcommand{\chainCameraPositions}{5/4/7/10/4/5/10/5}
\newcommand{\chainSeeds}{50}
\newcommand{\chainsAnalysed}{150}
\newcommand{\chainsChains}{150}
\newcommand{\chainsRows}{1,200}

\newcommand{\chainsDetector}{Community Forensics}

\newcommand{\chainsScoredTurns}{1/3/5/8}

\newcommand{\chainGitwoComplete}{50}
\newcommand{\chainGitwoToneMean}{0.24}

\newcommand{\chainGitwoTthreeMean}{0.23}

\newcommand{\chainGitwoTfiveMean}{0.28}

\newcommand{\chainGitwoTeightMean}{0.34}

\newcommand{\chainGitwoDeltaMean}{$+0.10$}
\newcommand{\chainGitwoDeltaLo}{$+0.01$}
\newcommand{\chainGitwoDeltaHi}{$+0.20$}

\newcommand{\chainGitwoEditCameraMean}{0.12}
\newcommand{\chainGitwoEditOtherLo}{0.23}
\newcommand{\chainGitwoEditOtherHi}{0.35}
\newcommand{\chainGitwoEditCameraRank}{8}

\newcommand{\chainFlareToneMean}{0.15}

\newcommand{\chainFlareTthreeMean}{0.19}

\newcommand{\chainFlareTfiveMean}{0.17}

\newcommand{\chainFlareTeightMean}{0.18}

\newcommand{\chainFlareDeltaMean}{$+0.03$}
\newcommand{\chainFlareDeltaLo}{$-0.04$}
\newcommand{\chainFlareDeltaHi}{$+0.11$}

\newcommand{\chainFlareEditCameraMean}{0.10}
\newcommand{\chainFlareEditOtherLo}{0.10}
\newcommand{\chainFlareEditOtherHi}{0.30}
\newcommand{\chainFlareEditCameraRank}{7}

\newcommand{\chainSunToneMean}{0.31}

\newcommand{\chainSunTthreeMean}{0.27}

\newcommand{\chainSunTfiveMean}{0.28}

\newcommand{\chainSunTeightMean}{0.30}

\newcommand{\chainSunDeltaMean}{$-0.01$}
\newcommand{\chainSunDeltaLo}{$-0.09$}
\newcommand{\chainSunDeltaHi}{$+0.07$}

\newcommand{\chainSunEditCameraMean}{0.14}
\newcommand{\chainSunEditOtherLo}{0.18}
\newcommand{\chainSunEditOtherHi}{0.45}
\newcommand{\chainSunEditCameraRank}{8}

\newcommand{\dtImgAuc}{0.588}
\newcommand{\dtImgAucCiLo}{0.544}
\newcommand{\dtImgAucCiHi}{0.633}

\newcommand{\dtImgNAuth}{200}

\newcommand{\dtSunImgAuc}{0.610}
\newcommand{\dtSunImgAucCiLo}{0.567}
\newcommand{\dtSunImgAucCiHi}{0.652}

\newcommand{\dtCheckpoint}{\texttt{dtd\_doctamper}}
\newcommand{\dtCiConvention}{2000-resample bootstrap, forged side clustered by spec\_id}

\newcommand{\seamDetector}{TruFor}
\newcommand{\seamN}{200}
\newcommand{\seamFieldResp}{0.68}
\newcommand{\seamRingResp}{0.52}
\newcommand{\seamOutResp}{0.086}
\newcommand{\seamAucField}{0.918}
\newcommand{\seamAucCrop}{0.921}
\newcommand{\seamAucFieldWithinCrop}{0.786}
\newcommand{\tierRealN}{544}
\newcommand{\tierRealPools}{IMD2020 and Columbia}
\newcommand{\tierDetector}{Community Forensics}
\newcommand{\aucFlareHigh}{0.888}
\newcommand{\aucCiFlareHigh}{[0.859, 0.914]}

\newcommand{\aucFlareFourK}{0.823}
\newcommand{\aucCiFlareFourK}{[0.762, 0.873]}

\newcommand{\aucFlareLow}{0.903}
\newcommand{\aucCiFlareLow}{[0.878, 0.927]}

\newcommand{\aucFlareMax}{0.916}
\newcommand{\aucCiFlareMax}{[0.886, 0.941]}

\newcommand{\aucFlareMed}{0.890}
\newcommand{\aucCiFlareMed}{[0.861, 0.917]}

\newcommand{\aucFlareXhigh}{0.889}
\newcommand{\aucCiFlareXhigh}{[0.861, 0.915]}

\newcommand{\aucGitwoHigh}{0.897}
\newcommand{\aucCiGitwoHigh}{[0.869, 0.921]}

\newcommand{\aucGitwoFourK}{0.898}
\newcommand{\aucCiGitwoFourK}{[0.850, 0.940]}

\newcommand{\aucGitwoLow}{0.881}
\newcommand{\aucCiGitwoLow}{[0.852, 0.909]}

\newcommand{\aucGitwoMed}{0.880}
\newcommand{\aucCiGitwoMed}{[0.850, 0.908]}

\newcommand{\aucSunHigh}{0.917}
\newcommand{\aucCiSunHigh}{[0.895, 0.937]}

\newcommand{\aucSunFourK}{0.775}
\newcommand{\aucCiSunFourK}{[0.714, 0.832]}

\newcommand{\aucSunLow}{0.893}
\newcommand{\aucCiSunLow}{[0.864, 0.919]}

\newcommand{\aucSunMax}{0.935}
\newcommand{\aucCiSunMax}{[0.912, 0.958]}

\newcommand{\aucSunMed}{0.918}
\newcommand{\aucCiSunMed}{[0.894, 0.939]}

\newcommand{\aucSunXhigh}{0.914}
\newcommand{\aucCiSunXhigh}{[0.889, 0.936]}

\newcommand{\tierAucFlatLo}{0.880}
\newcommand{\tierAucFlatHi}{0.935}

\newcommand{\tierFourKLo}{0.775}
\newcommand{\tierFourKHi}{0.898}
\newcommand{\tierFourKN}{20}

\newcommand{\tierTokFold}{36}
\newcommand{\tierCellNLo}{30}
\newcommand{\tierCellNHi}{100}
\newcommand{\tierAucFlatCells}{13}
\newcommand{\detDtFlarePixAuc}{0.588}
\newcommand{\detDtFlarePixAucLo}{0.576}
\newcommand{\detDtFlarePixAucHi}{0.600}
\newcommand{\detDtFlarePixN}{947}

\newcommand{\detDtSunPixAuc}{0.599}
\newcommand{\detDtSunPixAucLo}{0.586}
\newcommand{\detDtSunPixAucHi}{0.612}
\newcommand{\detDtSunPixN}{854}

\newcommand{\detDtGitwoLowPixAuc}{0.613}

\newcommand{\detDtGitwoLowPixN}{192}

\newcommand{\detDtGitwoMedPixAuc}{0.615}

\newcommand{\detDtGitwoMedPixN}{193}

\newcommand{\detTfFlarePixAuc}{0.913}

\newcommand{\detTfSunPixAuc}{0.924}

\newcommand{\detTfGitwoLowPixAuc}{0.914}

\newcommand{\detTfGitwoMedPixAuc}{0.922}

\newcommand{\detTfPixAucLo}{0.91}
\newcommand{\detTfPixAucHi}{0.92}

\newcommand{\detCfFlareImgAuc}{0.621}

\newcommand{\detCfSunImgAuc}{0.613}

\newcommand{\detCfGitwoLowImgAuc}{0.649}

\newcommand{\detCfGitwoMedImgAuc}{0.626}

\newcommand{\detCfImgAucLo}{0.61}
\newcommand{\detCfImgAucHi}{0.65}

\newcommand{\detPairDtFlareMinusGitwoLowDelta}{$-0.023$}

\newcommand{\detPairDtFlareMinusGitwoMedDelta}{$-0.025$}
\newcommand{\detPairDtFlareMinusGitwoMedLo}{$-0.048$}
\newcommand{\detPairDtFlareMinusGitwoMedHi}{$-0.002$}

\newcommand{\detPairDtSunMinusGitwoLowDelta}{$-0.011$}

\newcommand{\detPairDtSunMinusGitwoMedDelta}{$-0.013$}
\newcommand{\detPairDtSunMinusGitwoMedLo}{$-0.036$}
\newcommand{\detPairDtSunMinusGitwoMedHi}{$+0.009$}

\newcommand{\detPairTfFlareMinusGitwoMedDelta}{$-0.006$}

\newcommand{\detPairTfSunMinusGitwoLowDelta}{$+0.011$}

\newcommand{\seamSunN}{200}

\newcommand{\seamSunAucField}{0.926}
\newcommand{\seamSunAucCrop}{0.934}
\newcommand{\seamSunAucFieldWithinCrop}{0.787}

\newcommand{\chainsHolmM}{3}
\newcommand{\chainGitwoDeltaP}{0.03}
\newcommand{\chainGitwoDeltaPHolm}{0.10}

\newcommand{\attrChance}{1/3}
\newcommand{\attrFeatures}{128-d}
\newcommand{\attrCv}{5-fold stratified}
\newcommand{\attrLowAcc}{0.395}
\newcommand{\attrLowCiLo}{0.317}
\newcommand{\attrLowCiHi}{0.472}

\newcommand{\attrMedAcc}{0.342}
\newcommand{\attrMedCiLo}{0.267}
\newcommand{\attrMedCiHi}{0.418}
\newcommand{\attrMedN}{152}
\newcommand{\attrHighAcc}{0.474}
\newcommand{\attrHighCiLo}{0.394}
\newcommand{\attrHighCiHi}{0.553}

\newcommand{\attrFourKAcc}{0.595}
\newcommand{\attrFourKCiLo}{0.447}
\newcommand{\attrFourKCiHi}{0.744}
\newcommand{\attrFourKN}{42}

\newcommand{\progN}{31}
\newcommand{\progDtVone}{0.728}
\newcommand{\progDtVoneLo}{0.657}
\newcommand{\progDtVoneHi}{0.798}
\newcommand{\progDtVtwo}{0.813}
\newcommand{\progDtVtwoLo}{0.753}
\newcommand{\progDtVtwoHi}{0.870}
\newcommand{\progDtVthreeFlare}{0.595}
\newcommand{\progDtVthreeFlareLo}{0.530}
\newcommand{\progDtVthreeFlareHi}{0.658}
\newcommand{\progDtVthreeSun}{0.553}
\newcommand{\progDtVthreeSunLo}{0.494}
\newcommand{\progDtVthreeSunHi}{0.616}
\newcommand{\progDtPairedN}{30}
\newcommand{\progDetectorName}{DocTamper}

\newcommand{\progStagedChangedPct}{97\%}

\newcommand{\figoneModel}{gpt-image-2.5-flare}

\newcommand{\figoneOneCorpus}{cord}
\newcommand{\figoneOneField}{\texttt{menu.price}}

\newcommand{\figoneOnePanel}{right}

\newcommand{\figoneTwoCorpus}{wildreceipt}
\newcommand{\figoneTwoField}{\texttt{Telephone\_key}}

\newcommand{\figoneTwoPanel}{left}

\newcommand{\figoneThreeCorpus}{cord}
\newcommand{\figoneThreeField}{\texttt{menu.unitprice}}

\newcommand{\figoneThreePanel}{right}

\newcommand{\tokLow}{196}
\newcommand{\usdLow}{0.006}
\newcommand{\tokMed}{439}
\newcommand{\usdMed}{0.013}
\newcommand{\tokHigh}{1,756}
\newcommand{\usdHigh}{0.053}
\newcommand{\tokXhigh}{3,122}
\newcommand{\usdXhigh}{0.094}
\newcommand{\tokMax}{7,024}
\newcommand{\usdMax}{0.211}

\newcommand{\latGitwoLow}{21.2}

\newcommand{\latGitwoMed}{51.9}

\newcommand{\latGitwoHigh}{151.3}

\newcommand{\latFlareLow}{13.4}

\newcommand{\latFlareMed}{16.3}

\newcommand{\latFlareHigh}{21.4}

\newcommand{\latFlareXhigh}{31.6}

\newcommand{\latFlareMax}{51.7}

\newcommand{\latSunLow}{16.7}

\newcommand{\latSunMed}{20.7}

\newcommand{\latSunHigh}{37.1}

\newcommand{\latSunXhigh}{58.5}

\newcommand{\latSunMax}{105.9}

\newcommand{\smokePerCell}{5}
\newcommand{\latSdFlareLow}{3.9}

\newcommand{\latSdGitwoLow}{5.7}

\newcommand{\latSdGitwoMed}{5.2}

\newcommand{\latSdGitwoHigh}{14.7}

\newcommand{\latSdFlareMed}{6.3}

\newcommand{\latSdFlareHigh}{2.9}

\newcommand{\latSdFlareXhigh}{3.3}

\newcommand{\latSdFlareMax}{2.7}

\newcommand{\latSdSunLow}{2.7}

\newcommand{\latSdSunMed}{2.5}

\newcommand{\latSdSunHigh}{4.5}

\newcommand{\latSdSunXhigh}{7.8}

\newcommand{\latSdSunMax}{12.9}

\newcommand{\latRatioHigh}{2.4}
\newcommand{\latRatioMax}{2.9}
\newcommand{\latRatioSunLo}{1.2}
\newcommand{\latRatioSunHi}{2.0}
\newcommand{\latRatioSunTiers}{5}

\newcommand{\eloEditSun}{1,520}
\newcommand{\eloEditFlare}{1,491}
\newcommand{\eloEditGitwo}{1,461}
\newcommand{\eloEditGain}{59}
\newcommand{\eloTtoiGain}{40}
\newcommand{\arenaDate}{2026-09-07}

\newcommand{\eloCiEditSun}{9}
\newcommand{\votesEditSun}{6,704}
\newcommand{\eloCiEditFlare}{9}
\newcommand{\votesEditFlare}{5,676}
\newcommand{\eloCiEditGitwo}{3}
\newcommand{\votesEditGitwo}{235,928}
\newcommand{\scUnsafeSun}{1.09\%}
\newcommand{\scUnsafeFlare}{1.41\%}
\newcommand{\scUnsafeGitwo}{1.64\%}
\newcommand{\scBlockedSun}{21.9\%}
\newcommand{\scBlockedFlare}{19.2\%}
\newcommand{\scBlockedGitwo}{23.1\%}

\newcommand{\wildStandalone}{1,382}
\newcommand{\wildXUnscoredN}{91}
\newcommand{\wildXfpr}{5\%}
\newcommand{\wildXAuthN}{364}
\newcommand{\wildXCells}{10}
\newcommand{\wildXCtrlLo}{58.0\%}
\newcommand{\wildXCtrlHi}{80.0\%}
\newcommand{\wildXCtrlMean}{68.6\%}

\newcommand{\wildXWildRate}{35.9\%}
\newcommand{\wildXWildN}{1,291}

\newcommand{\wildXPngRate}{42.9\%}

\newcommand{\wildXJpgRate}{34.9\%}

\newcommand{\wildXGapContentPp}{25.7}
\newcommand{\wildXGapCodecPp}{8.0}
\newcommand{\wildXWildRateLo}{33.4\%}
\newcommand{\wildXWildRateHi}{38.6\%}
\newcommand{\wildXPngRateLo}{35.6\%}
\newcommand{\wildXPngRateHi}{50.6\%}
\newcommand{\wildXJpgRateLo}{32.2\%}
\newcommand{\wildXJpgRateHi}{37.8\%}
\newcommand{\wildXCtrlMeanBootLo}{64.7\%}
\newcommand{\wildXCtrlMeanBootHi}{72.7\%}
\newcommand{\wildXCtrlPooledRate}{67.6\%}
\newcommand{\wildXCtrlPooledN}{860}
\newcommand{\wildXCtrlPooledRateLo}{64.4\%}
\newcommand{\wildXCtrlPooledRateHi}{70.6\%}

\newcommand{\routeArmBatchDocChainFlare}{228}
\newcommand{\routeArmSyncDocChainFlare}{12}

\newcommand{\routeArmBatchDocChainSun}{228}
\newcommand{\routeArmSyncDocChainSun}{12}

\newcommand{\routeArmBatchDocChainGitwoLow}{240}
\newcommand{\routeArmSyncDocChainGitwoLow}{0}

\newcommand{\routeArmBatchDocChainGitwoMed}{228}
\newcommand{\routeArmSyncDocChainGitwoMed}{12}

\newcommand{\routeArmBatchPhotoChainFlare}{36}
\newcommand{\routeArmSyncPhotoChainFlare}{200}

\newcommand{\routeArmBatchPhotoChainSun}{216}
\newcommand{\routeArmSyncPhotoChainSun}{18}

\newcommand{\routeArmBatchPhotoChainGitwoLow}{237}
\newcommand{\routeArmSyncPhotoChainGitwoLow}{3}

\newcommand{\routeArmBatchPhotoChainGitwoMed}{216}
\newcommand{\routeArmSyncPhotoChainGitwoMed}{18}

\newcommand{\routeArmBatchPhotoChainConvFlare}{0}
\newcommand{\routeArmSyncPhotoChainConvFlare}{116}

\newcommand{\routeArmBatchPhotoChainConvSun}{0}
\newcommand{\routeArmSyncPhotoChainConvSun}{115}

\newcommand{\routeArmBatchPhotoChainConvGitwoMed}{0}
\newcommand{\routeArmSyncPhotoChainConvGitwoMed}{115}

\newcommand{\routeArmBatchRefPFlare}{229}
\newcommand{\routeArmSyncRefPFlare}{9}

\newcommand{\routeArmBatchRefPSun}{229}
\newcommand{\routeArmSyncRefPSun}{9}

\newcommand{\routeArmBatchRefPGitwoLow}{238}
\newcommand{\routeArmSyncRefPGitwoLow}{0}

\newcommand{\routeArmBatchRefPGitwoMed}{229}
\newcommand{\routeArmSyncRefPGitwoMed}{9}

\newcommand{\routeArmBatchDetailFlare}{229}
\newcommand{\routeArmSyncDetailFlare}{11}

\newcommand{\routeArmBatchDetailSun}{224}
\newcommand{\routeArmSyncDetailSun}{11}

\newcommand{\routeArmBatchDetailGitwoLow}{42}
\newcommand{\routeArmSyncDetailGitwoLow}{2}

\newcommand{\routeArmBatchDetailGitwoMed}{42}
\newcommand{\routeArmSyncDetailGitwoMed}{2}

\newcommand{\routeArmBatchDetailGitwoHigh}{61}
\newcommand{\routeArmSyncDetailGitwoHigh}{3}

\newcommand{\routeArmBatchRefRFlare}{224}
\newcommand{\routeArmSyncRefRFlare}{10}

\newcommand{\routeArmBatchRefRSun}{0}
\newcommand{\routeArmSyncRefRSun}{234}
\newcommand{\routeArmOkRefRSun}{234}
\newcommand{\routeArmBatchRefRGitwoLow}{224}
\newcommand{\routeArmSyncRefRGitwoLow}{10}

\newcommand{\routeArmBatchRefRGitwoMed}{224}
\newcommand{\routeArmSyncRefRGitwoMed}{10}

\newcommand{\eqDtMarginLo}{0.05}
\newcommand{\eqDtMarginHi}{0.10}
\newcommand{\eqDtFlareCiNinetyLo}{0.578}
\newcommand{\eqDtFlareCiNinetyHi}{0.597}
\newcommand{\eqDtFlareMinMargin}{0.097}
\newcommand{\eqDtFlareEquivFive}{no}
\newcommand{\eqDtFlareEquivTen}{yes}
\newcommand{\eqDtSunCiNinetyLo}{0.588}
\newcommand{\eqDtSunCiNinetyHi}{0.610}
\newcommand{\eqDtSunMinMargin}{0.110}
\newcommand{\eqDtSunEquivFive}{no}
\newcommand{\eqDtSunEquivTen}{no}
\newcommand{\eqDtGitwoLowCiNinetyLo}{0.590}
\newcommand{\eqDtGitwoLowCiNinetyHi}{0.635}
\newcommand{\eqDtGitwoLowMinMargin}{0.135}
\newcommand{\eqDtGitwoLowEquivFive}{no}
\newcommand{\eqDtGitwoLowEquivTen}{no}
\newcommand{\eqDtGitwoMedCiNinetyLo}{0.592}
\newcommand{\eqDtGitwoMedCiNinetyHi}{0.636}
\newcommand{\eqDtGitwoMedMinMargin}{0.136}
\newcommand{\eqDtGitwoMedEquivFive}{no}
\newcommand{\eqDtGitwoMedEquivTen}{no}
\newcommand{\eqDtNArms}{4}

\newcommand{\eqDtNEquivTen}{1}

\newcommand{\eqDtPairFlareMinusGitwoLowCiNinetyLo}{$-0.044$}
\newcommand{\eqDtPairFlareMinusGitwoLowCiNinetyHi}{$-0.002$}
\newcommand{\eqDtPairFlareMinusGitwoLowMinMargin}{0.044}
\newcommand{\eqDtPairFlareMinusGitwoMedCiNinetyLo}{$-0.044$}
\newcommand{\eqDtPairFlareMinusGitwoMedCiNinetyHi}{$-0.005$}
\newcommand{\eqDtPairFlareMinusGitwoMedMinMargin}{0.044}
\newcommand{\eqDtPairSunMinusGitwoLowCiNinetyLo}{$-0.034$}
\newcommand{\eqDtPairSunMinusGitwoLowCiNinetyHi}{$+0.010$}
\newcommand{\eqDtPairSunMinusGitwoLowMinMargin}{0.034}
\newcommand{\eqDtPairSunMinusGitwoMedCiNinetyLo}{$-0.032$}
\newcommand{\eqDtPairSunMinusGitwoMedCiNinetyHi}{$+0.005$}
\newcommand{\eqDtPairSunMinusGitwoMedMinMargin}{0.032}
\newcommand{\eqTierMarginLo}{0.03}
\newcommand{\eqTierMarginHi}{0.05}

\newcommand{\eqTierFlareRange}{0.028}
\newcommand{\eqTierFlareRangeUpper}{0.055}
\newcommand{\eqTierFlareMinMargin}{0.051}
\newcommand{\eqTierFlareEquivThree}{no}
\newcommand{\eqTierFlareEquivFive}{no}
\newcommand{\eqTierFlareNCells}{5}
\newcommand{\eqTierSunRange}{0.043}
\newcommand{\eqTierSunRangeUpper}{0.064}
\newcommand{\eqTierSunMinMargin}{0.064}
\newcommand{\eqTierSunEquivThree}{no}
\newcommand{\eqTierSunEquivFive}{no}
\newcommand{\eqTierSunNCells}{5}
\newcommand{\eqTierGitwoRange}{0.016}
\newcommand{\eqTierGitwoRangeUpper}{0.038}
\newcommand{\eqTierGitwoMinMargin}{0.034}
\newcommand{\eqTierGitwoEquivThree}{no}
\newcommand{\eqTierGitwoEquivFive}{yes}
\newcommand{\eqTierGitwoNCells}{3}
\newcommand{\eqTierAllRange}{0.055}
\newcommand{\eqTierAllRangeUpper}{0.084}
\newcommand{\eqTierAllMinMargin}{0.080}
\newcommand{\eqTierAllEquivThree}{no}
\newcommand{\eqTierAllEquivFive}{no}
\newcommand{\eqTierAllNCells}{13}
\newcommand{\eqTierAllFourKRange}{0.160}
\newcommand{\eqTierAllFourKRangeUpper}{0.210}
\newcommand{\eqTierAllFourKMinMargin}{0.208}
\newcommand{\eqTierAllFourKEquivThree}{no}
\newcommand{\eqTierAllFourKEquivFive}{no}
\newcommand{\eqTierAllFourKNCells}{16}
\newcommand{\latForgeDocsGitwoLowOverFlareRatio}{1.19}
\newcommand{\latForgeDocsGitwoLowOverFlareLo}{1.15}
\newcommand{\latForgeDocsGitwoLowOverFlareHi}{1.21}
\newcommand{\latForgeDocsGitwoLowOverFlareN}{199}
\newcommand{\latForgeDocsGitwoLowOverFlareNClusters}{199}
\newcommand{\latForgeDocsGitwoMedOverFlareRatio}{2.75}
\newcommand{\latForgeDocsGitwoMedOverFlareLo}{2.59}
\newcommand{\latForgeDocsGitwoMedOverFlareHi}{2.86}
\newcommand{\latForgeDocsGitwoMedOverFlareN}{200}
\newcommand{\latForgeDocsGitwoMedOverFlareNClusters}{200}
\newcommand{\latForgeDocsGitwoLowOverSunRatio}{0.97}
\newcommand{\latForgeDocsGitwoLowOverSunLo}{0.94}
\newcommand{\latForgeDocsGitwoLowOverSunHi}{0.99}

\newcommand{\latForgeDocsGitwoMedOverSunRatio}{2.26}
\newcommand{\latForgeDocsGitwoMedOverSunLo}{2.20}
\newcommand{\latForgeDocsGitwoMedOverSunHi}{2.31}

\newcommand{\latForgeDocsFlareMedianS}{10.8}
\newcommand{\latForgeDocsSunMedianS}{13.5}
\newcommand{\latForgeDocsGitwoLowMedianS}{13.0}
\newcommand{\latForgeDocsGitwoMedMedianS}{30.9}
\newcommand{\latForgeDocChainGitwoMedOverFlareRatio}{2.86}
\newcommand{\latForgeDocChainGitwoMedOverFlareLo}{2.63}
\newcommand{\latForgeDocChainGitwoMedOverFlareHi}{3.15}
\newcommand{\latForgeDocChainGitwoMedOverFlareN}{12}
\newcommand{\latForgeDocChainGitwoMedOverFlareNClusters}{3}
\newcommand{\latForgeDocChainGitwoMedOverSunRatio}{2.03}
\newcommand{\latForgeDocChainGitwoMedOverSunLo}{1.91}
\newcommand{\latForgeDocChainGitwoMedOverSunHi}{2.21}

\newcommand{\latForgeDocChainFlareMedianS}{11.7}
\newcommand{\latForgeDocChainSunMedianS}{15.5}
\newcommand{\latForgeDocChainGitwoMedMedianS}{31.6}
\newcommand{\latForgePhotoChainGitwoMedOverFlareRatio}{2.96}
\newcommand{\latForgePhotoChainGitwoMedOverFlareLo}{2.85}
\newcommand{\latForgePhotoChainGitwoMedOverFlareHi}{3.13}
\newcommand{\latForgePhotoChainGitwoMedOverFlareN}{18}
\newcommand{\latForgePhotoChainGitwoMedOverFlareNClusters}{3}
\newcommand{\latForgePhotoChainGitwoMedOverSunRatio}{2.13}
\newcommand{\latForgePhotoChainGitwoMedOverSunLo}{2.05}
\newcommand{\latForgePhotoChainGitwoMedOverSunHi}{2.21}

\newcommand{\latForgePhotoChainFlareMedianS}{13.3}
\newcommand{\latForgePhotoChainSunMedianS}{19.1}
\newcommand{\latForgePhotoChainGitwoMedMedianS}{39.3}
\newcommand{\latForgePhotoChainConvGitwoMedOverFlareRatio}{2.57}
\newcommand{\latForgePhotoChainConvGitwoMedOverFlareLo}{2.56}
\newcommand{\latForgePhotoChainConvGitwoMedOverFlareHi}{2.64}
\newcommand{\latForgePhotoChainConvGitwoMedOverFlareN}{115}
\newcommand{\latForgePhotoChainConvGitwoMedOverFlareNClusters}{20}
\newcommand{\latForgePhotoChainConvGitwoMedOverSunRatio}{1.97}
\newcommand{\latForgePhotoChainConvGitwoMedOverSunLo}{1.93}
\newcommand{\latForgePhotoChainConvGitwoMedOverSunHi}{2.00}

\newcommand{\latForgePhotoChainConvFlareMedianS}{16.0}
\newcommand{\latForgePhotoChainConvSunMedianS}{21.2}
\newcommand{\latForgePhotoChainConvGitwoMedMedianS}{41.2}
\newcommand{\latForgeRefPGitwoMedOverFlareRatio}{3.15}
\newcommand{\latForgeRefPGitwoMedOverFlareLo}{2.83}
\newcommand{\latForgeRefPGitwoMedOverFlareHi}{3.42}
\newcommand{\latForgeRefPGitwoMedOverFlareN}{9}
\newcommand{\latForgeRefPGitwoMedOverFlareNClusters}{6}
\newcommand{\latForgeRefPGitwoMedOverSunRatio}{2.17}
\newcommand{\latForgeRefPGitwoMedOverSunLo}{1.97}
\newcommand{\latForgeRefPGitwoMedOverSunHi}{2.18}

\newcommand{\latForgeRefPFlareMedianS}{12.9}
\newcommand{\latForgeRefPSunMedianS}{19.0}
\newcommand{\latForgeRefPGitwoMedMedianS}{40.4}

\newcommand{\refPairNShared}{200}
\newcommand{\refPairFlareRefused}{1}
\newcommand{\refPairSunRefused}{0}
\newcommand{\refPairGitwoLowRefused}{1}
\newcommand{\refPairGitwoMedRefused}{0}
\newcommand{\refPairFlareVsGitwoLowArmOnly}{0}
\newcommand{\refPairFlareVsGitwoLowCtlOnly}{0}
\newcommand{\refPairFlareVsGitwoLowBoth}{1}
\newcommand{\refPairFlareVsGitwoLowPMcNemar}{1}
\newcommand{\refPairSunVsGitwoLowArmOnly}{0}
\newcommand{\refPairSunVsGitwoLowCtlOnly}{1}

\newcommand{\refPairSunVsGitwoLowPMcNemar}{1}
\newcommand{\refPairFlareVsGitwoMedArmOnly}{1}
\newcommand{\refPairFlareVsGitwoMedCtlOnly}{0}

\newcommand{\refPairFlareVsGitwoMedPMcNemar}{1}
\newcommand{\refPairSunVsGitwoMedArmOnly}{0}
\newcommand{\refPairSunVsGitwoMedCtlOnly}{0}

\newcommand{\refPairSunVsGitwoMedPMcNemar}{1}
\newcommand{\refPairSharedCord}{100}
\newcommand{\refPairSharedWild}{100}
\newcommand{\refPairAllCord}{500}
\newcommand{\refPairAllWild}{500}
\newcommand{\refPairTwoFiveRefusedWild}{7}
\newcommand{\refPairTwoFiveRefusedCord}{0}
\newcommand{\refPairTwoFiveRefusedOutside}{6}
\newcommand{\refPairMinDiscordant}{6}
\newcommand{\tfCovFlareMissPct}{7.4\%}
\newcommand{\tfCovFlareMissPctGeTwoSevenK}{99\%}
\newcommand{\tfCovFlareMissPctLtTwoSevenK}{0.0\%}
\newcommand{\tfCovFlareMaxScoredLongEdge}{2,772}
\newcommand{\tfCovSunMissPct}{8.1\%}

\newcommand{\tfCovGitwoLowMissPct}{8.3\%}
\newcommand{\tfCovGitwoLowMissPctGeTwoSevenK}{100\%}

\newcommand{\tfCovGitwoLowMaxScoredLongEdge}{2,730}
\newcommand{\tfCovGitwoMedMissPct}{8.3\%}

\newcommand{\tfCovAuthMissPct}{7.5\%}

\newcommand{\seamCtrlGitwoLowN}{176}

\newcommand{\seamCtrlGitwoLowAucCrop}{0.919}
\newcommand{\seamCtrlGitwoLowAucFieldWithinCrop}{0.791}
\newcommand{\seamCtrlGitwoMedN}{177}

\newcommand{\seamCtrlGitwoMedAucCrop}{0.931}
\newcommand{\seamCtrlGitwoMedAucFieldWithinCrop}{0.773}
\newcommand{\vTwoN}{3,066}
\newcommand{\vTwoSpecs}{4,062}
\newcommand{\vTwoHuman}{0.501}
\newcommand{\vTwoTrufor}{0.599}
\newcommand{\vTwoDoctamper}{0.585}
\newcommand{\vTwoSelf}{0.532}
\newcommand{\vTwoTruforCal}{0.962}
\newcommand{\vTwoDoctamperCal}{0.852}
\newcommand{\vTwoRefusal}{$\sim$10\%}
\newcommand{\vTwoRejectPct}{24.5\%}

\newcommand{\vOneN}{4,061}

\newcommand{\hardStop}{300}
\newcommand{\launchDate}{8 September 2026}

\newcommand{\capLocThresh}{12/255}

\newcommand{\capLocTolPx}{$\pm$1\,px}
\newcommand{\capLocFloorShiftPx}{1\,px}

\newcommand{\capLocBorderPx}{3}
\newcommand{\capLocGuardPx}{2}
\newcommand{\capLocGitwoLowN}{192}
\newcommand{\capLocGitwoLowTolMed}{3.3\%}
\newcommand{\capLocGitwoLowTolMedLo}{3.0\%}
\newcommand{\capLocGitwoLowTolMedHi}{3.9\%}
\newcommand{\capLocGitwoLowPaperMed}{11.7\%}

\newcommand{\capLocGitwoLowInMed}{25.0\%}
\newcommand{\capLocGitwoLowShiftMed}{0.88}
\newcommand{\capLocGitwoLowRegistered}{97\%}
\newcommand{\capLocGitwoLowExcl}{7}
\newcommand{\capLocGitwoMedN}{193}
\newcommand{\capLocGitwoMedTolMed}{3.9\%}
\newcommand{\capLocGitwoMedTolMedLo}{3.3\%}
\newcommand{\capLocGitwoMedTolMedHi}{4.6\%}
\newcommand{\capLocGitwoMedPaperMed}{15.2\%}

\newcommand{\capLocGitwoMedInMed}{24.5\%}
\newcommand{\capLocGitwoMedShiftMed}{2.22}
\newcommand{\capLocGitwoMedRegistered}{98\%}
\newcommand{\capLocGitwoMedExcl}{7}
\newcommand{\capLocFlareN}{950}
\newcommand{\capLocFlareTolMed}{2.5\%}
\newcommand{\capLocFlareTolMedLo}{2.3\%}
\newcommand{\capLocFlareTolMedHi}{2.7\%}
\newcommand{\capLocFlarePaperMed}{10.5\%}

\newcommand{\capLocFlareInMed}{25.1\%}
\newcommand{\capLocFlareShiftMed}{0.66}
\newcommand{\capLocFlareRegistered}{96\%}
\newcommand{\capLocFlareExcl}{49}
\newcommand{\capLocSunN}{948}
\newcommand{\capLocSunTolMed}{1.6\%}
\newcommand{\capLocSunTolMedLo}{1.4\%}
\newcommand{\capLocSunTolMedHi}{1.7\%}
\newcommand{\capLocSunPaperMed}{8.6\%}

\newcommand{\capLocSunInMed}{22.3\%}
\newcommand{\capLocSunShiftMed}{0.75}
\newcommand{\capLocSunRegistered}{94\%}
\newcommand{\capLocSunExcl}{49}

\newcommand{\capLocFloorN}{999}
\newcommand{\capLocFloorTolMed}{0.15\%}
\newcommand{\capLocFloorTolMedLo}{0.12\%}
\newcommand{\capLocFloorTolMedHi}{0.22\%}
\newcommand{\capLocFloorPaperMed}{7.3\%}

\newcommand{\capLocPairGitwoMedFlareTolDelta}{$+1.1$\%}

\newcommand{\capLocPairGitwoMedSunTolDelta}{$+1.9$\%}

\newcommand{\capLocPairGitwoLowFlareTolN}{192}
\newcommand{\capLocPairGitwoLowFlareTolMoreLocal}{69\%}

\newcommand{\capLocPairGitwoLowSunTolMoreLocal}{90\%}

\newcommand{\capLocPairGitwoLowFlareCordTolDelta}{$+0.56$\%}
\newcommand{\capLocPairGitwoLowFlareCordTolDeltaLo}{$+0.27$\%}
\newcommand{\capLocPairGitwoLowFlareCordTolDeltaHi}{$+1.06$\%}
\newcommand{\capLocPairGitwoLowSunCordTolDelta}{$+1.2$\%}
\newcommand{\capLocPairGitwoLowSunCordTolDeltaLo}{$+0.8$\%}
\newcommand{\capLocPairGitwoLowSunCordTolDeltaHi}{$+1.6$\%}

\newcommand{\capLocPairGitwoLowFlareWildTolDelta}{$+0.37$\%}
\newcommand{\capLocPairGitwoLowFlareWildTolDeltaLo}{$+0.10$\%}
\newcommand{\capLocPairGitwoLowFlareWildTolDeltaHi}{$+1.12$\%}
\newcommand{\capLocPairGitwoLowSunWildTolDelta}{$+1.5$\%}
\newcommand{\capLocPairGitwoLowSunWildTolDeltaLo}{$+1.2$\%}
\newcommand{\capLocPairGitwoLowSunWildTolDeltaHi}{$+1.9$\%}
\newcommand{\capLocPairFlareGitwoMedTolDelta}{$-1.1$\%}
\newcommand{\capLocPairFlareGitwoMedTolDeltaLo}{$-1.6$\%}
\newcommand{\capLocPairFlareGitwoMedTolDeltaHi}{$-0.6$\%}
\newcommand{\capLocPairSunGitwoMedTolDelta}{$-1.9$\%}
\newcommand{\capLocPairSunGitwoMedTolDeltaLo}{$-2.3$\%}
\newcommand{\capLocPairSunGitwoMedTolDeltaHi}{$-1.4$\%}
\newcommand{\capLocPairFlareGitwoLowTolDelta}{$-0.48$\%}
\newcommand{\capLocPairFlareGitwoLowTolDeltaLo}{$-0.92$\%}
\newcommand{\capLocPairFlareGitwoLowTolDeltaHi}{$-0.29$\%}
\newcommand{\capLocPairSunGitwoLowTolDelta}{$-1.4$\%}
\newcommand{\capLocPairSunGitwoLowTolDeltaLo}{$-1.7$\%}
\newcommand{\capLocPairSunGitwoLowTolDeltaHi}{$-1.1$\%}

\newcommand{\capDocNSpecs}{199}

\newcommand{\capDocBaseN}{96}
\newcommand{\capDocBaseCord}{62\%}
\newcommand{\capDocBaseWild}{34\%}
\newcommand{\capDocNSpecsNoMismatch}{192}

\newcommand{\capDocFloorAnyOther}{15.6\%}

\newcommand{\capDocAnyNumGiLow}{22.2\%}

\newcommand{\capDocAnyNumGiMed}{31.9\%}

\newcommand{\capDocAnyNumFlare}{18.8\%}

\newcommand{\capDocAnyNumSunburst}{15.3\%}

\newcommand{\capDocDiffAnyNumFlareGiLow}{$-3.5$\%}
\newcommand{\capDocDiffAnyNumFlareGiLowLo}{$-11.1$\%}
\newcommand{\capDocDiffAnyNumFlareGiLowHi}{$+4.2$\%}

\newcommand{\capDocDiffAnyNumFlareGiMed}{$-13.2$\%}
\newcommand{\capDocDiffAnyNumFlareGiMedLo}{$-21.5$\%}
\newcommand{\capDocDiffAnyNumFlareGiMedHi}{$-5.6$\%}

\newcommand{\capDocDiffAnyNumSunburstGiLow}{$-6.9$\%}
\newcommand{\capDocDiffAnyNumSunburstGiLowLo}{$-15.3$\%}
\newcommand{\capDocDiffAnyNumSunburstGiLowHi}{$+0.7$\%}

\newcommand{\capDocDiffAnyNumSunburstGiMed}{$-16.7$\%}
\newcommand{\capDocDiffAnyNumSunburstGiMedLo}{$-25.0$\%}
\newcommand{\capDocDiffAnyNumSunburstGiMedHi}{$-7.6$\%}

\newcommand{\capDocValTwoFieldK}{36}
\newcommand{\capDocValTwoFieldN}{40}

\newcommand{\capDocValTwoOtherK}{30}
\newcommand{\capDocValTwoOtherN}{40}

\newcommand{\capDocValOneFieldK}{33}
\newcommand{\capDocValOneFieldN}{40}
\newcommand{\capDocValOneOtherK}{32}
\newcommand{\capDocValOneOtherN}{40}

\newcommand{\capTurnNChains}{50}

\newcommand{\capTurnNTurns}{8}

\newcommand{\capTurnSurvRateGptTwo}{91.2\%}

\newcommand{\capTurnSurvRateGptTwoN}{170}

\newcommand{\capTurnDriftPerEditGptTwo}{1.06}

\newcommand{\capTurnSurvRateFlare}{95.3\%}

\newcommand{\capTurnDriftPerEditFlare}{1.03}

\newcommand{\capTurnSurvRateSunburst}{84.1\%}

\newcommand{\capTurnDriftPerEditSunburst}{0.99}

\newcommand{\capTurnDiffSurvRateFlare}{$+4.1$\%}
\newcommand{\capTurnDiffSurvRateFlareLo}{$-0.6$\%}
\newcommand{\capTurnDiffSurvRateFlareHi}{$+8.8$\%}

\newcommand{\capTurnDiffStrengthClipFlare}{$-0.0158$}

\newcommand{\capTurnDiffSurvRateSunburst}{$-7.1$\%}
\newcommand{\capTurnDiffSurvRateSunburstLo}{$-12.6$\%}
\newcommand{\capTurnDiffSurvRateSunburstHi}{$-1.8$\%}
\newcommand{\capTurnDiffSurvRateNoTOneSunburst}{$-4.9$\%}
\newcommand{\capTurnDiffSurvRateNoTOneSunburstLo}{$-11.1$\%}
\newcommand{\capTurnDiffSurvRateNoTOneSunburstHi}{$+0.7$\%}

\newcommand{\capTurnDiffStrengthClipSunburst}{$-0.0145$}

\newcommand{\capRefWrongLabelAucClip}{0.62}

\newcommand{\capRefWrongLabelAucDino}{0.32}

\newcommand{\capPriceFourKTokFlare}{3,336}
\newcommand{\capPriceFourKUsdFlare}{0.100}

\newcommand{\capPriceFourKLatFlare}{29.7}

\newcommand{\capPriceFourKLatSun}{38.9}
\newcommand{\capPriceFourKTokGitwo}{13,342}
\newcommand{\capPriceFourKUsdGitwo}{0.400}

\newcommand{\capPriceFourKLatGitwo}{109.2}
\newcommand{\capPriceFourKOverHigh}{1.9}

\newcommand{\capSharpNBase}{100}
\newcommand{\capSharpNMax}{30}
\newcommand{\capSharpNFourK}{20}
\newcommand{\capSharpFlareXhighLapPct}{$-11$\%}
\newcommand{\capSharpFlareXhighLapLo}{$-22$\%}
\newcommand{\capSharpFlareXhighLapHi}{$+1$\%}
\newcommand{\capSharpFlareXhighHfPct}{$-18$\%}
\newcommand{\capSharpFlareXhighHfLo}{$-27$\%}
\newcommand{\capSharpFlareXhighHfHi}{$-8$\%}
\newcommand{\capSharpFlareXhighFlatNoisePct}{$-32$\%}

\newcommand{\capSharpFlareMaxHfPct}{$-29$\%}

\newcommand{\capSharpSunXhighEdgeWPct}{$-12$\%}

\newcommand{\capSharpFlareFourKVsOneKHfPct}{$+14$\%}
\newcommand{\capSharpFlareFourKVsOneKHfLo}{$-19$\%}
\newcommand{\capSharpFlareFourKVsOneKHfHi}{$+19$\%}
\newcommand{\capSharpSunFourKVsOneKHfPct}{$-12$\%}
\newcommand{\capSharpSunFourKVsOneKHfLo}{$-33$\%}
\newcommand{\capSharpSunFourKVsOneKHfHi}{$+13$\%}
\newcommand{\capSharpGitwoFourKVsOneKHfPct}{$-29$\%}
\newcommand{\capSharpGitwoFourKVsOneKHfLo}{$-52$\%}
\newcommand{\capSharpGitwoFourKVsOneKHfHi}{$+23$\%}
\newcommand{\capSharpMatchFlareHighLapPct}{$-17$\%}
\newcommand{\capSharpMatchFlareHighLapLo}{$-26$\%}
\newcommand{\capSharpMatchFlareHighLapHi}{$-8$\%}

\newcommand{\capSharpMatchFlareHighFlatNoisePct}{$-17$\%}
\newcommand{\capSharpMatchFlareMaxLapPct}{$-26$\%}
\newcommand{\capSharpMatchFlareMaxLapLo}{$-40$\%}
\newcommand{\capSharpMatchFlareMaxLapHi}{$-13$\%}

\newcommand{\capSharpMatchFlareMaxFlatNoisePct}{$-35$\%}
\newcommand{\capSharpMatchSunHighLapPct}{$-17$\%}
\newcommand{\capSharpMatchSunHighLapLo}{$-26$\%}
\newcommand{\capSharpMatchSunHighLapHi}{$-11$\%}

\newcommand{\capSharpMatchSunHighFlatNoisePct}{$-30$\%}

\newcommand{\capSharpMatchSunHighEdgeWPct}{$+2$\%}
\newcommand{\capSharpMatchSunMaxLapPct}{$-14$\%}
\newcommand{\capSharpMatchSunMaxLapLo}{$-24$\%}
\newcommand{\capSharpMatchSunMaxLapHi}{$+4$\%}

\newcommand{\capSharpMatchSunMaxFlatNoisePct}{$-36$\%}

\newcommand{\capSharpMatchLapRangeLo}{$-26$\%}
\newcommand{\capSharpMatchLapRangeHi}{$-14$\%}
\newcommand{\capSharpLatFlareHighRatio}{2.33}
\newcommand{\capSharpLatFlareHighLo}{2.27}
\newcommand{\capSharpLatFlareHighHi}{2.38}

\newcommand{\capSharpLatFlareMaxRatio}{2.80}
\newcommand{\capSharpLatFlareMaxLo}{2.68}
\newcommand{\capSharpLatFlareMaxHi}{2.96}

\newcommand{\capSharpLatTokHigh}{1,756}
\newcommand{\capSharpLatTokMax}{7,024}
\newcommand{\capSharpCalibBlurAcutPct}{$-21$\%}
\newcommand{\capSharpCalibBlurLapPct}{$-76$\%}
\newcommand{\capSharpCalibDenoiseEdgeWPct}{$-2$\%}

\newcommand{\capRefPNRefs}{119}
\newcommand{\capRefPNCalls}{952}
\newcommand{\capRefPNLabelFound}{951}

\newcommand{\capRefPSizeMinBinN}{3}
\newcommand{\capRefPSizeCeilingSigma}{0.3}

\newcommand{\capRefPHeroLogoDEwbGi}{3.2}

\newcommand{\capRefPHeroSkuExactFlare}{93\%}

\newcommand{\capRefPHeroLogoDEwbFlare}{2.2}

\newcommand{\capRefPHeroSkuExactSunburst}{97\%}

\newcommand{\capRefPHeroLogoDEwbSunburst}{2.2}

\newcommand{\capRefPInSceneSkuExactGi}{23\%}

\newcommand{\capRefPInSceneSkuHeightGi}{6.0}
\newcommand{\capRefPInSceneSkuExactFlare}{64\%}

\newcommand{\capRefPInSceneSkuHeightFlare}{7.6}
\newcommand{\capRefPInSceneSkuExactSunburst}{61\%}

\newcommand{\capRefPInSceneSkuHeightSunburst}{7.4}

\newcommand{\capRefPInSceneBrandExactGi}{100\%}

\newcommand{\capRefPInScenePriceExactGi}{94\%}

\newcommand{\capRefPInSceneBrandExactFlare}{100\%}

\newcommand{\capRefPInScenePriceExactFlare}{97\%}

\newcommand{\capRefPInSceneBrandExactSunburst}{100\%}

\newcommand{\capRefPInScenePriceExactSunburst}{96\%}

\newcommand{\capRefPInSceneBrandExactDiffFlare}{$0$\%}
\newcommand{\capRefPInSceneBrandExactDiffLoFlare}{$0$\%}
\newcommand{\capRefPInSceneBrandExactDiffHiFlare}{$0$\%}
\newcommand{\capRefPInScenePriceExactDiffFlare}{$+3$\%}
\newcommand{\capRefPInScenePriceExactDiffLoFlare}{$-1$\%}
\newcommand{\capRefPInScenePriceExactDiffHiFlare}{$+8$\%}
\newcommand{\capRefPInSceneBrandExactDiffSunburst}{$0$\%}
\newcommand{\capRefPInSceneBrandExactDiffLoSunburst}{$0$\%}
\newcommand{\capRefPInSceneBrandExactDiffHiSunburst}{$0$\%}
\newcommand{\capRefPInScenePriceExactDiffSunburst}{$+2$\%}
\newcommand{\capRefPInScenePriceExactDiffLoSunburst}{$-3$\%}
\newcommand{\capRefPInScenePriceExactDiffHiSunburst}{$+7$\%}

\newcommand{\capRefPHeroLogoDEwbDiffFlare}{$-1.0$}
\newcommand{\capRefPHeroLogoDEwbDiffLoFlare}{$-1.5$}
\newcommand{\capRefPHeroLogoDEwbDiffHiFlare}{$-0.6$}
\newcommand{\capRefPHeroLogoDEwbDiffSunburst}{$-1.0$}
\newcommand{\capRefPHeroLogoDEwbDiffLoSunburst}{$-1.4$}
\newcommand{\capRefPHeroLogoDEwbDiffHiSunburst}{$-0.6$}

\newcommand{\capRefPInSceneSizeRawGapTwoFive}{$+40$\%}
\newcommand{\capRefPInSceneSizeRawGapLoTwoFive}{$+31$\%}
\newcommand{\capRefPInSceneSizeRawGapHiTwoFive}{$+49$\%}
\newcommand{\capRefPInSceneSizeMatchedGapTwoFive}{$+11$\%}
\newcommand{\capRefPInSceneSizeMatchedGapLoTwoFive}{$-2$\%}
\newcommand{\capRefPInSceneSizeMatchedGapHiTwoFive}{$+25$\%}
\newcommand{\capRefPInSceneSizeMatchedGapNormTwoFive}{$+14$\%}
\newcommand{\capRefPInSceneSizeMatchedGapNormLoTwoFive}{$+1$\%}
\newcommand{\capRefPInSceneSizeMatchedGapNormHiTwoFive}{$+28$\%}
\newcommand{\capRefPInSceneSizeCeilingGapTwoFive}{$+45$\%}
\newcommand{\capRefPInSceneSizeCeilingGapLoTwoFive}{$+37$\%}
\newcommand{\capRefPInSceneSizeCeilingGapHiTwoFive}{$+53$\%}
\newcommand{\capRefPInSceneSizeResidualGapTwoFive}{$-5$\%}
\newcommand{\capRefPInSceneSizeResidualGapLoTwoFive}{$-13$\%}
\newcommand{\capRefPInSceneSizeResidualGapHiTwoFive}{$+4$\%}
\newcommand{\capRefPInSceneSizeSurvivingPctTwoFive}{27\%}

\newcommand{\capRefPBlindNAll}{40}
\newcommand{\capRefPBlindNGi}{20}
\newcommand{\capRefPBlindNTwoFive}{20}
\newcommand{\capRefPBlindCorrectGi}{11}
\newcommand{\capRefPBlindWrongGi}{5}
\newcommand{\capRefPBlindIllegibleGi}{4}
\newcommand{\capRefPBlindCorrectTwoFive}{20}
\newcommand{\capRefPBlindWrongTwoFive}{0}
\newcommand{\capRefPBlindIllegibleTwoFive}{0}
\newcommand{\capRefPBlindOcrFalseNegAll}{17}

\newcommand{\capRefPBlindMatchedRange}{5.5--8}
\newcommand{\capRefPBlindMatchedNGi}{10}
\newcommand{\capRefPBlindMatchedCorrectGi}{7}
\newcommand{\capRefPBlindMatchedWrongGi}{3}

\newcommand{\capRefPBlindMatchedNTwoFive}{14}
\newcommand{\capRefPBlindMatchedCorrectTwoFive}{14}
\newcommand{\capRefPBlindMatchedWrongTwoFive}{0}

\newcommand{\capRefPSensLogisticGapTwoFive}{$+8$\%}
\newcommand{\capRefPSensLogisticGapLoTwoFive}{$-1$\%}
\newcommand{\capRefPSensLogisticGapHiTwoFive}{$+17$\%}
\newcommand{\capRefPSensSurvivingLogisticPctTwoFive}{20\%}

\newcommand{\capRefPInSceneSkuExactGiLowRate}{28\%}

\newcommand{\capRefPInSceneSizeRawGapTwoFiveVsGiLow}{$+35$\%}
\newcommand{\capRefPInSceneSizeRawGapTwoFiveVsGiLowCILo}{$+26$\%}
\newcommand{\capRefPInSceneSizeRawGapTwoFiveVsGiLowCIHi}{$+44$\%}
\newcommand{\capRefPInSceneSizeMatchedGapTwoFiveVsGiLow}{$+15$\%}
\newcommand{\capRefPInSceneSizeMatchedGapTwoFiveVsGiLowCILo}{$+5$\%}
\newcommand{\capRefPInSceneSizeMatchedGapTwoFiveVsGiLowCIHi}{$+25$\%}
\newcommand{\capRefPInSceneSizeResidualGapTwoFiveVsGiLow}{$+9$\%}
\newcommand{\capRefPInSceneSizeResidualGapTwoFiveVsGiLowCILo}{$-2$\%}
\newcommand{\capRefPInSceneSizeResidualGapTwoFiveVsGiLowCIHi}{$+19$\%}

\newcommand{\capRefRNRefs}{117}
\newcommand{\capRefRNCalls}{936}

\newcommand{\capRefRInSceneSkuExactGi}{34\%}

\newcommand{\capRefRInSceneSkuExactFlare}{69\%}

\newcommand{\capRefRInSceneSkuExactSunburst}{69\%}

\newcommand{\capRefRInSceneSkuExactGiLow}{43\%}

\newcommand{\capDetailCeilCapPx}{9}
\newcommand{\capDetailCeilBoxHten}{16}
\newcommand{\capDetailCeilBoxHtenFourk}{7.6}
\newcommand{\capDetailNImages}{580}

\newcommand{\capDetailNPrompts}{40}
\newcommand{\capDetailNLines}{6}
\newcommand{\capDetailNArms}{16}
\newcommand{\capDetailFailedCalls}{0}
\newcommand{\capDetailHfiftyAtCeilN}{15}
\newcommand{\capDetailExactAttFlareHigh}{91\%}
\newcommand{\capDetailExactAttFlareHighLo}{85\%}
\newcommand{\capDetailExactAttFlareHighHi}{95\%}
\newcommand{\capDetailExactAttFlareHighFourk}{91\%}

\newcommand{\capDetailExactAttFlareMax}{95\%}
\newcommand{\capDetailExactAttFlareMaxLo}{92\%}
\newcommand{\capDetailExactAttFlareMaxHi}{98\%}

\newcommand{\capDetailExactAttSunburstHigh}{83\%}
\newcommand{\capDetailExactAttSunburstHighLo}{73\%}
\newcommand{\capDetailExactAttSunburstHighHi}{90\%}
\newcommand{\capDetailExactAttSunburstHighFourk}{82\%}

\newcommand{\capDetailExactAttSunburstMax}{82\%}
\newcommand{\capDetailExactAttSunburstMaxLo}{73\%}
\newcommand{\capDetailExactAttSunburstMaxHi}{90\%}

\newcommand{\capDetailExactAttGiHigh}{82\%}
\newcommand{\capDetailExactAttGiHighLo}{71\%}
\newcommand{\capDetailExactAttGiHighHi}{91\%}
\newcommand{\capDetailExactAttGiHighFourk}{89\%}

\newcommand{\capDetailExactAttGiMedium}{88\%}
\newcommand{\capDetailExactAttGiMediumLo}{78\%}
\newcommand{\capDetailExactAttGiMediumHi}{96\%}
\newcommand{\capDetailBelowCeilGiHigh}{71\%}
\newcommand{\capDetailBelowCeilFlareMax}{30\%}
\newcommand{\capDetailBelowCeilSunburstMax}{61\%}

\newcommand{\capDetailTierBestDiffGi}{$+1$\%}

\newcommand{\capDetailTierBestDiffFlare}{$0$\%}

\newcommand{\capDetailTierBestDiffSunburst}{$+1$\%}

\newcommand{\capDocChainTurns}{4}

\newcommand{\capDocChainAllReceipts}{60}

\newcommand{\capDocChainAllRefusedFlare}{0}

\newcommand{\capDocChainCordReceipts}{30}
\newcommand{\capDocChainCordPlannedFields}{213}

\newcommand{\capDocChainCordElsewhereGiTwo}{1\%}
\newcommand{\capDocChainCordNearMissGiTwo}{4\%}

\newcommand{\capDocChainCordElsewhereFlare}{6\%}
\newcommand{\capDocChainCordNearMissFlare}{8\%}

\newcommand{\capDocChainCordElsewhereSunburst}{0\%}
\newcommand{\capDocChainCordNearMissSunburst}{10\%}

\newcommand{\capDocChainWildReceipts}{30}
\newcommand{\capDocChainWildPlannedFields}{217}
\newcommand{\capDocChainWildBaseline}{96\%}

\newcommand{\capChainNSeedsPlan}{40}
\newcommand{\capChainNSourcePhotosPlan}{23}
\newcommand{\capChainCalibNPool}{276}
\newcommand{\capChainTauMug}{0.10}
\newcommand{\capChainFprMugPool}{0.7\%}
\newcommand{\capChainTauUmbrella}{0.15}
\newcommand{\capChainFprUmbrellaPool}{1.4\%}
\newcommand{\capChainTauPlant}{0.43}
\newcommand{\capChainFprPlantPool}{1.8\%}
\newcommand{\capChainConvTextModel}{\texttt{gpt-5.4-nano}}
\newcommand{\capChainInPlaceIou}{0.3}
\newcommand{\capChainUmbrellaShare}{0.10}
\newcommand{\capChainUmbrellaShareSens}{0.15}

\newcommand{\capChainNRefusedChainTurns}{6}

\newcommand{\capChainStatelessNChains}{157}

\newcommand{\capChainStatelessIncompleteList}{\texttt{gpt-image-2.5-flare/pcv2-020}}
\newcommand{\capChainStatelessGiNChains}{39}

\newcommand{\capChainStatelessGiLOneSuccess}{93.2\%}
\newcommand{\capChainStatelessGiLOneSurvival}{95.6\%}
\newcommand{\capChainStatelessGiLOneSurvivalLo}{92.7\%}
\newcommand{\capChainStatelessGiLOneSurvivalHi}{98.2\%}
\newcommand{\capChainStatelessGiLTwoInPlace}{98.3\%}
\newcommand{\capChainStatelessGiLTwoInPlaceLo}{95.9\%}
\newcommand{\capChainStatelessGiLTwoInPlaceHi}{100.0\%}
\newcommand{\capChainStatelessGiLTwoRelocation}{1.6\%}
\newcommand{\capChainStatelessGiLTwoDrift}{18.0\%}
\newcommand{\capChainStatelessGiLTwoStrength}{44.3\%}
\newcommand{\capChainStatelessFlareNChains}{39}

\newcommand{\capChainStatelessFlareLOneSuccess}{94.0\%}
\newcommand{\capChainStatelessFlareLOneSurvival}{94.1\%}
\newcommand{\capChainStatelessFlareLOneSurvivalLo}{90.5\%}
\newcommand{\capChainStatelessFlareLOneSurvivalHi}{97.3\%}
\newcommand{\capChainStatelessFlareLTwoInPlace}{100.0\%}
\newcommand{\capChainStatelessFlareLTwoInPlaceLo}{100.0\%}
\newcommand{\capChainStatelessFlareLTwoInPlaceHi}{100.0\%}
\newcommand{\capChainStatelessFlareLTwoRelocation}{0.0\%}
\newcommand{\capChainStatelessFlareLTwoDrift}{25.6\%}
\newcommand{\capChainStatelessFlareLTwoStrength}{41.4\%}
\newcommand{\capChainStatelessSunburstNChains}{39}

\newcommand{\capChainStatelessSunburstLOneSuccess}{93.2\%}
\newcommand{\capChainStatelessSunburstLOneSurvival}{95.6\%}
\newcommand{\capChainStatelessSunburstLOneSurvivalLo}{92.6\%}
\newcommand{\capChainStatelessSunburstLOneSurvivalHi}{98.0\%}
\newcommand{\capChainStatelessSunburstLTwoInPlace}{100.0\%}
\newcommand{\capChainStatelessSunburstLTwoInPlaceLo}{100.0\%}
\newcommand{\capChainStatelessSunburstLTwoInPlaceHi}{100.0\%}
\newcommand{\capChainStatelessSunburstLTwoRelocation}{0.0\%}
\newcommand{\capChainStatelessSunburstLTwoDrift}{15.7\%}
\newcommand{\capChainStatelessSunburstLTwoStrength}{39.6\%}

\newcommand{\capChainConvNChains}{57}

\newcommand{\capChainConvIncompleteList}{\texttt{gpt-image-2/pcv2-020}, \texttt{gpt-image-2.5-flare/pcv2-020}}
\newcommand{\capChainConvGiNChains}{19}

\newcommand{\capChainConvGiLOneSuccess}{88.6\%}
\newcommand{\capChainConvGiLOneSurvival}{98.8\%}
\newcommand{\capChainConvGiLOneSurvivalLo}{96.4\%}
\newcommand{\capChainConvGiLOneSurvivalHi}{100.0\%}
\newcommand{\capChainConvGiLTwoInPlace}{38.1\%}
\newcommand{\capChainConvGiLTwoInPlaceLo}{29.4\%}
\newcommand{\capChainConvGiLTwoInPlaceHi}{47.6\%}
\newcommand{\capChainConvGiLTwoRelocation}{61.2\%}
\newcommand{\capChainConvGiLTwoDrift}{22.2\%}
\newcommand{\capChainConvGiLTwoStrength}{49.1\%}
\newcommand{\capChainConvFlareNChains}{19}

\newcommand{\capChainConvFlareLOneSuccess}{86.0\%}
\newcommand{\capChainConvFlareLOneSurvival}{95.0\%}
\newcommand{\capChainConvFlareLOneSurvivalLo}{88.5\%}
\newcommand{\capChainConvFlareLOneSurvivalHi}{100.0\%}
\newcommand{\capChainConvFlareLTwoInPlace}{48.7\%}
\newcommand{\capChainConvFlareLTwoInPlaceLo}{39.2\%}
\newcommand{\capChainConvFlareLTwoInPlaceHi}{57.7\%}
\newcommand{\capChainConvFlareLTwoRelocation}{48.8\%}
\newcommand{\capChainConvFlareLTwoDrift}{5.5\%}
\newcommand{\capChainConvFlareLTwoStrength}{43.7\%}
\newcommand{\capChainConvSunburstNChains}{19}

\newcommand{\capChainConvSunburstLOneSuccess}{88.6\%}
\newcommand{\capChainConvSunburstLOneSurvival}{95.2\%}
\newcommand{\capChainConvSunburstLOneSurvivalLo}{90.6\%}
\newcommand{\capChainConvSunburstLOneSurvivalHi}{98.8\%}
\newcommand{\capChainConvSunburstLTwoInPlace}{57.0\%}
\newcommand{\capChainConvSunburstLTwoInPlaceLo}{44.9\%}
\newcommand{\capChainConvSunburstLTwoInPlaceHi}{67.9\%}
\newcommand{\capChainConvSunburstLTwoRelocation}{41.0\%}
\newcommand{\capChainConvSunburstLTwoDrift}{7.7\%}
\newcommand{\capChainConvSunburstLTwoStrength}{48.8\%}

\newcommand{\capChainConvSunburstMinusGiLTwoInPlace}{$+18.9$\%}

\newcommand{\capChainConvSunburstMinusGiLTwoInPlacePHolm}{0.036}

\newcommand{\capDecoupleDtNImages}{384}
\newcommand{\capDecoupleDtNSpecs}{192}
\newcommand{\capDecoupleDtClean}{0.619}
\newcommand{\capDecoupleDtCleanLo}{0.578}
\newcommand{\capDecoupleDtCleanHi}{0.660}

\newcommand{\capDecoupleDtUnreadable}{0.571}
\newcommand{\capDecoupleDtUnreadableLo}{0.537}
\newcommand{\capDecoupleDtUnreadableHi}{0.606}

\newcommand{\capDecoupleDtRest}{0.630}
\newcommand{\capDecoupleDtCleanMinusRest}{$-0.011$}
\newcommand{\capDecoupleDtCleanMinusRestLo}{$-0.076$}
\newcommand{\capDecoupleDtCleanMinusRestHi}{$+0.053$}
\newcommand{\capDecoupleDtCleanMinusRestN}{186}

\newcommand{\capDecoupleDtRho}{$+0.03$}
\newcommand{\capDecoupleDtRhoLo}{$-0.03$}
\newcommand{\capDecoupleDtRhoHi}{$+0.09$}
\newcommand{\capDecoupleDtRhoN}{1,801}

\newcommand{\capDecoupleTfCleanMinusRest}{$-0.005$}
\newcommand{\capDecoupleTfCleanMinusRestLo}{$-0.023$}
\newcommand{\capDecoupleTfCleanMinusRestHi}{$+0.012$}

\newcommand{\capDecoupleTfRho}{$-0.14$}

\newcommand{\capDecoupleDtResolution}{0.08}

\newcommand{\capChainSensStatelessFlareMinusGiLTwoInPlace}{$+1.7$\%}

\newcommand{\capChainSensStatelessFlareMinusGiLTwoInPlacePHolm}{0.38}

\newcommand{\capChainSensConvFlareMinusGiLTwoInPlace}{$+9.4$\%}
\newcommand{\capChainSensConvFlareMinusGiLTwoInPlaceLo}{$-2.3$\%}
\newcommand{\capChainSensConvFlareMinusGiLTwoInPlaceHi}{$+21.0$\%}
\newcommand{\capChainSensConvFlareMinusGiLTwoInPlacePHolm}{0.32}

\newcommand{\capChainSensConvSunburstMinusGiLTwoInPlace}{$+16.6$\%}
\newcommand{\capChainSensConvSunburstMinusGiLTwoInPlaceLo}{$+2.2$\%}
\newcommand{\capChainSensConvSunburstMinusGiLTwoInPlaceHi}{$+30.5$\%}
\newcommand{\capChainSensConvSunburstMinusGiLTwoInPlacePHolm}{0.08}

\newcommand{\capRefPlProgHeroAreaPctGi}{12.3\%}
\newcommand{\capRefPlProgHeroWidthPctGi}{30\%}

\newcommand{\capRefPlProgHeroRelPxGi}{17.0}
\newcommand{\capRefPlProgHeroWidthAboveTenthPctGi}{100\%}
\newcommand{\capRefPlProgHeroLogoIouGi}{0.944}
\newcommand{\capRefPlProgHeroAreaPctGiLow}{12.0\%}
\newcommand{\capRefPlProgHeroWidthPctGiLow}{30\%}

\newcommand{\capRefPlProgHeroRelPxGiLow}{17.0}
\newcommand{\capRefPlProgHeroWidthAboveTenthPctGiLow}{100\%}
\newcommand{\capRefPlProgHeroLogoIouGiLow}{0.929}
\newcommand{\capRefPlProgHeroAreaPctFlare}{12.7\%}
\newcommand{\capRefPlProgHeroWidthPctFlare}{32\%}

\newcommand{\capRefPlProgHeroRelPxFlare}{16.9}
\newcommand{\capRefPlProgHeroWidthAboveTenthPctFlare}{100\%}
\newcommand{\capRefPlProgHeroLogoIouFlare}{0.919}
\newcommand{\capRefPlProgHeroAreaPctSunburst}{12.4\%}
\newcommand{\capRefPlProgHeroWidthPctSunburst}{31\%}

\newcommand{\capRefPlProgHeroRelPxSunburst}{16.7}
\newcommand{\capRefPlProgHeroWidthAboveTenthPctSunburst}{100\%}
\newcommand{\capRefPlProgHeroLogoIouSunburst}{0.921}
\newcommand{\capRefPlProgInSceneAreaPctGi}{2.4\%}
\newcommand{\capRefPlProgInSceneWidthPctGi}{14\%}

\newcommand{\capRefPlProgInSceneRelPxGi}{17.3}
\newcommand{\capRefPlProgInSceneWidthAboveTenthPctGi}{85\%}
\newcommand{\capRefPlProgInSceneLogoIouGi}{0.908}
\newcommand{\capRefPlProgInSceneAreaPctGiLow}{2.7\%}
\newcommand{\capRefPlProgInSceneWidthPctGiLow}{14\%}

\newcommand{\capRefPlProgInSceneRelPxGiLow}{17.4}
\newcommand{\capRefPlProgInSceneWidthAboveTenthPctGiLow}{94\%}
\newcommand{\capRefPlProgInSceneLogoIouGiLow}{0.904}
\newcommand{\capRefPlProgInSceneAreaPctFlare}{4.1\%}
\newcommand{\capRefPlProgInSceneWidthPctFlare}{18\%}

\newcommand{\capRefPlProgInSceneRelPxFlare}{17.1}
\newcommand{\capRefPlProgInSceneWidthAboveTenthPctFlare}{100\%}
\newcommand{\capRefPlProgInSceneLogoIouFlare}{0.902}
\newcommand{\capRefPlProgInSceneAreaPctSunburst}{3.9\%}
\newcommand{\capRefPlProgInSceneWidthPctSunburst}{18\%}

\newcommand{\capRefPlProgInSceneRelPxSunburst}{17.0}
\newcommand{\capRefPlProgInSceneWidthAboveTenthPctSunburst}{100\%}
\newcommand{\capRefPlProgInSceneLogoIouSunburst}{0.932}

\newcommand{\capRefPlProgHeroFlareVsGiLowLogoIou}{$-0.009$}
\newcommand{\capRefPlProgHeroFlareVsGiLowLogoIouLo}{$-0.025$}
\newcommand{\capRefPlProgHeroFlareVsGiLowLogoIouHi}{$+0.006$}

\newcommand{\capRefPlProgHeroFlareVsGiSku}{$-3$\%}
\newcommand{\capRefPlProgHeroFlareVsGiSkuLo}{$-7$\%}
\newcommand{\capRefPlProgHeroFlareVsGiSkuHi}{$+2$\%}

\newcommand{\capRefPlProgHeroFlareVsGiLogoIou}{$-0.025$}
\newcommand{\capRefPlProgHeroFlareVsGiLogoIouLo}{$-0.038$}
\newcommand{\capRefPlProgHeroFlareVsGiLogoIouHi}{$-0.011$}

\newcommand{\capRefPlProgHeroFlareVsGiAreaRatio}{1.02}

\newcommand{\capRefPlProgHeroSunburstVsGiLowLogoIou}{$-0.007$}
\newcommand{\capRefPlProgHeroSunburstVsGiLowLogoIouLo}{$-0.029$}
\newcommand{\capRefPlProgHeroSunburstVsGiLowLogoIouHi}{$+0.010$}

\newcommand{\capRefPlProgHeroSunburstVsGiSku}{$+1$\%}
\newcommand{\capRefPlProgHeroSunburstVsGiSkuLo}{$-3$\%}
\newcommand{\capRefPlProgHeroSunburstVsGiSkuHi}{$+5$\%}

\newcommand{\capRefPlProgHeroSunburstVsGiLogoIou}{$-0.023$}
\newcommand{\capRefPlProgHeroSunburstVsGiLogoIouLo}{$-0.045$}
\newcommand{\capRefPlProgHeroSunburstVsGiLogoIouHi}{$-0.006$}

\newcommand{\capRefPlProgHeroSunburstVsGiAreaRatio}{1.02}

\newcommand{\capRefPlProgInSceneFlareVsGiLowSkuBatchOnly}{$+38$\%}

\newcommand{\capRefPlProgInSceneFlareVsGiLowMatchedAbs}{$+14$\%}
\newcommand{\capRefPlProgInSceneFlareVsGiLowMatchedAbsLo}{$+4$\%}
\newcommand{\capRefPlProgInSceneFlareVsGiLowMatchedAbsHi}{$+25$\%}
\newcommand{\capRefPlProgInSceneFlareVsGiLowMatchedAbsPPerm}{0.0044}

\newcommand{\capRefPlProgInSceneFlareVsGiLowMatchedPlace}{$+14$\%}
\newcommand{\capRefPlProgInSceneFlareVsGiLowMatchedPlaceLo}{$+2$\%}
\newcommand{\capRefPlProgInSceneFlareVsGiLowMatchedPlaceHi}{$+26$\%}

\newcommand{\capRefPlProgInSceneFlareVsGiLowLogoIou}{$-0.002$}
\newcommand{\capRefPlProgInSceneFlareVsGiLowLogoIouLo}{$-0.026$}
\newcommand{\capRefPlProgInSceneFlareVsGiLowLogoIouHi}{$+0.021$}

\newcommand{\capRefPlProgInSceneFlareVsGiLowLogoDEwb}{$-1.7$}
\newcommand{\capRefPlProgInSceneFlareVsGiLowLogoDEwbLo}{$-2.4$}
\newcommand{\capRefPlProgInSceneFlareVsGiLowLogoDEwbHi}{$-1.1$}

\newcommand{\capRefPlProgInSceneFlareVsGiLowAreaRatio}{1.48}

\newcommand{\capRefPlProgInSceneFlareVsGiSkuBatchOnly}{$+42$\%}

\newcommand{\capRefPlProgInSceneFlareVsGiSkuBatchOnlyExcluded}{3}
\newcommand{\capRefPlProgInSceneFlareVsGiMatchedAbs}{$+9$\%}
\newcommand{\capRefPlProgInSceneFlareVsGiMatchedAbsLo}{$-2$\%}
\newcommand{\capRefPlProgInSceneFlareVsGiMatchedAbsHi}{$+21$\%}
\newcommand{\capRefPlProgInSceneFlareVsGiMatchedAbsPPerm}{0.034}

\newcommand{\capRefPlProgInSceneFlareVsGiMatchedRel}{$+42$\%}

\newcommand{\capRefPlProgInSceneFlareVsGiMatchedPlace}{$+10$\%}
\newcommand{\capRefPlProgInSceneFlareVsGiMatchedPlaceLo}{$-1$\%}
\newcommand{\capRefPlProgInSceneFlareVsGiMatchedPlaceHi}{$+23$\%}

\newcommand{\capRefPlProgInSceneFlareVsGiLogoIou}{$-0.006$}
\newcommand{\capRefPlProgInSceneFlareVsGiLogoIouLo}{$-0.026$}
\newcommand{\capRefPlProgInSceneFlareVsGiLogoIouHi}{$+0.013$}

\newcommand{\capRefPlProgInSceneFlareVsGiAreaRatio}{1.69}

\newcommand{\capRefPlProgInSceneSunburstVsGiLowSkuBatchOnly}{$+34$\%}

\newcommand{\capRefPlProgInSceneSunburstVsGiLowMatchedAbs}{$+15$\%}
\newcommand{\capRefPlProgInSceneSunburstVsGiLowMatchedAbsLo}{$+5$\%}
\newcommand{\capRefPlProgInSceneSunburstVsGiLowMatchedAbsHi}{$+24$\%}
\newcommand{\capRefPlProgInSceneSunburstVsGiLowMatchedAbsPPerm}{$9.5\times10^{-4}$}

\newcommand{\capRefPlProgInSceneSunburstVsGiLowMatchedPlace}{$+14$\%}
\newcommand{\capRefPlProgInSceneSunburstVsGiLowMatchedPlaceLo}{$+3$\%}
\newcommand{\capRefPlProgInSceneSunburstVsGiLowMatchedPlaceHi}{$+24$\%}

\newcommand{\capRefPlProgInSceneSunburstVsGiLowLogoIou}{$+0.028$}
\newcommand{\capRefPlProgInSceneSunburstVsGiLowLogoIouLo}{$+0.010$}
\newcommand{\capRefPlProgInSceneSunburstVsGiLowLogoIouHi}{$+0.047$}

\newcommand{\capRefPlProgInSceneSunburstVsGiLowLogoDEwb}{$-2.0$}
\newcommand{\capRefPlProgInSceneSunburstVsGiLowLogoDEwbLo}{$-2.7$}
\newcommand{\capRefPlProgInSceneSunburstVsGiLowLogoDEwbHi}{$-1.4$}

\newcommand{\capRefPlProgInSceneSunburstVsGiLowAreaRatio}{1.34}

\newcommand{\capRefPlProgInSceneSunburstVsGiSkuBatchOnly}{$+39$\%}

\newcommand{\capRefPlProgInSceneSunburstVsGiMatchedAbs}{$+10$\%}
\newcommand{\capRefPlProgInSceneSunburstVsGiMatchedAbsLo}{$-2$\%}
\newcommand{\capRefPlProgInSceneSunburstVsGiMatchedAbsHi}{$+22$\%}
\newcommand{\capRefPlProgInSceneSunburstVsGiMatchedAbsPPerm}{0.028}

\newcommand{\capRefPlProgInSceneSunburstVsGiMatchedRel}{$+40$\%}

\newcommand{\capRefPlProgInSceneSunburstVsGiMatchedPlace}{$+10$\%}
\newcommand{\capRefPlProgInSceneSunburstVsGiMatchedPlaceLo}{$-1$\%}
\newcommand{\capRefPlProgInSceneSunburstVsGiMatchedPlaceHi}{$+22$\%}

\newcommand{\capRefPlProgInSceneSunburstVsGiLogoIou}{$+0.023$}
\newcommand{\capRefPlProgInSceneSunburstVsGiLogoIouLo}{$+0.011$}
\newcommand{\capRefPlProgInSceneSunburstVsGiLogoIouHi}{$+0.037$}

\newcommand{\capRefPlProgInSceneSunburstVsGiAreaRatio}{1.53}

\newcommand{\capRefPlRendInSceneAreaPctGi}{2.6\%}

\newcommand{\capRefPlRendInSceneAreaPctGiLow}{2.7\%}

\newcommand{\capRefPlRendInSceneAreaPctFlare}{4.0\%}

\newcommand{\capRefPlRendInSceneAreaPctSunburst}{3.7\%}

\newcommand{\capRefPlRendInSceneFlareVsGiLowSku}{$+26$\%}

\newcommand{\capRefPlRendInSceneFlareVsGiLowSkuPHolmPrimary}{$2.0\times10^{-4}$}

\newcommand{\capRefPlRendInSceneFlareVsGiLowMatchedPlace}{$+3$\%}
\newcommand{\capRefPlRendInSceneFlareVsGiLowMatchedPlaceLo}{$-6$\%}
\newcommand{\capRefPlRendInSceneFlareVsGiLowMatchedPlaceHi}{$+12$\%}

\newcommand{\capRefPlRendInSceneFlareVsGiMatchedAbs}{$+13$\%}
\newcommand{\capRefPlRendInSceneFlareVsGiMatchedAbsLo}{$+2$\%}
\newcommand{\capRefPlRendInSceneFlareVsGiMatchedAbsHi}{$+25$\%}

\newcommand{\capRefPlRendInSceneSunburstVsGiLowSku}{$+26$\%}

\newcommand{\capRefPlRendInSceneSunburstVsGiLowMatchedPlace}{$+10$\%}
\newcommand{\capRefPlRendInSceneSunburstVsGiLowMatchedPlaceLo}{$-3$\%}
\newcommand{\capRefPlRendInSceneSunburstVsGiLowMatchedPlaceHi}{$+23$\%}

\newcommand{\capRefPlRendInSceneSunburstVsGiMatchedAbs}{$+16$\%}
\newcommand{\capRefPlRendInSceneSunburstVsGiMatchedAbsLo}{$+4$\%}
\newcommand{\capRefPlRendInSceneSunburstVsGiMatchedAbsHi}{$+29$\%}

\newcommand{\capDetailVtwoNPerm}{20,000}

\newcommand{\capDetailVtwoTierM}{13}
\newcommand{\capDetailVtwoMinBinLines}{5}

\newcommand{\capDetailVtwoIttFlareHigh}{63\%}
\newcommand{\capDetailVtwoIttFlareHighLo}{58\%}
\newcommand{\capDetailVtwoIttFlareHighHi}{68\%}
\newcommand{\capDetailVtwoNoLineAboveFlareHigh}{0}
\newcommand{\capDetailVtwoNImagesFlareHigh}{40}

\newcommand{\capDetailVtwoIttFlareMax}{67\%}
\newcommand{\capDetailVtwoIttFlareMaxLo}{60\%}
\newcommand{\capDetailVtwoIttFlareMaxHi}{74\%}
\newcommand{\capDetailVtwoNoLineAboveFlareMax}{1}
\newcommand{\capDetailVtwoNImagesFlareMax}{40}

\newcommand{\capDetailVtwoIttSunburstHigh}{37\%}
\newcommand{\capDetailVtwoIttSunburstHighLo}{31\%}
\newcommand{\capDetailVtwoIttSunburstHighHi}{42\%}
\newcommand{\capDetailVtwoNoLineAboveSunburstHigh}{0}
\newcommand{\capDetailVtwoNImagesSunburstHigh}{40}

\newcommand{\capDetailVtwoIttSunburstMax}{30\%}
\newcommand{\capDetailVtwoIttSunburstMaxLo}{26\%}
\newcommand{\capDetailVtwoIttSunburstMaxHi}{35\%}
\newcommand{\capDetailVtwoNoLineAboveSunburstMax}{0}
\newcommand{\capDetailVtwoNImagesSunburstMax}{40}

\newcommand{\capDetailVtwoIttGiHigh}{23\%}
\newcommand{\capDetailVtwoIttGiHighLo}{18\%}
\newcommand{\capDetailVtwoIttGiHighHi}{29\%}
\newcommand{\capDetailVtwoNoLineAboveGiHigh}{5}
\newcommand{\capDetailVtwoNImagesGiHigh}{40}

\newcommand{\capDetailVtwoIttGiMedium}{28\%}
\newcommand{\capDetailVtwoIttGiMediumLo}{22\%}
\newcommand{\capDetailVtwoIttGiMediumHi}{35\%}
\newcommand{\capDetailVtwoNoLineAboveGiMedium}{6}
\newcommand{\capDetailVtwoNImagesGiMedium}{40}

\newcommand{\capDetailVtwoTierAttPermNReject}{0}
\newcommand{\capDetailVtwoIttFlareHighVsGiMedium}{$+35$\%}
\newcommand{\capDetailVtwoIttFlareHighVsGiMediumLo}{$+29$\%}
\newcommand{\capDetailVtwoIttFlareHighVsGiMediumHi}{$+42$\%}

\newcommand{\capDetailVtwoIttSunburstHighVsGiMedium}{$+9$\%}
\newcommand{\capDetailVtwoIttSunburstHighVsGiMediumLo}{$+1$\%}
\newcommand{\capDetailVtwoIttSunburstHighVsGiMediumHi}{$+17$\%}

\newcommand{\capDetailVtwoIttFlareMaxVsGiHigh}{$+44$\%}
\newcommand{\capDetailVtwoIttFlareMaxVsGiHighLo}{$+36$\%}
\newcommand{\capDetailVtwoIttFlareMaxVsGiHighHi}{$+52$\%}

\newcommand{\capDetailVtwoIttSunburstMaxVsGiHigh}{$+7$\%}
\newcommand{\capDetailVtwoIttSunburstMaxVsGiHighLo}{$+1$\%}
\newcommand{\capDetailVtwoIttSunburstMaxVsGiHighHi}{$+13$\%}

\newcommand{\capDetailVtwoTierIttNReject}{8}
\newcommand{\capDetailVtwoMatchedFlareHighVsGiMedium}{$-4$\%}

\newcommand{\capDetailVtwoMatchedSunburstHighVsGiMedium}{$-13$\%}

\newcommand{\capDetailVtwoMatchedFlareMaxVsGiHigh}{$+13$\%}
\newcommand{\capDetailVtwoMatchedFlareMaxVsGiHighLo}{$+4$\%}
\newcommand{\capDetailVtwoMatchedFlareMaxVsGiHighHi}{$+23$\%}

\newcommand{\capDetailVtwoMatchedFlareMaxVsGiHighPHolm}{0.01}

\newcommand{\capDetailVtwoMatchedSunburstMaxVsGiHigh}{$-2$\%}
\newcommand{\capDetailVtwoMatchedSunburstMaxVsGiHighLo}{$-16$\%}
\newcommand{\capDetailVtwoMatchedSunburstMaxVsGiHighHi}{$+12$\%}

\newcommand{\capDetailVtwoTierMatchedNReject}{0}
\newcommand{\capDetailVtwoDropFlareHighVsGiMediumN}{6}

\newcommand{\capDetailVtwoDropFlareHighVsGiMediumCtl}{6}

\newcommand{\capDetailVtwoDropFlareMaxVsGiHighN}{6}
\newcommand{\capDetailVtwoDropFlareMaxVsGiHighArm}{1}
\newcommand{\capDetailVtwoDropFlareMaxVsGiHighCtl}{5}

\newcommand{\capDetailVtwoDropSunburstMaxVsGiHighN}{5}

\newcommand{\capDetailVtwoDropSunburstMaxVsGiHighCtl}{5}

\newcommand{\capDetailVtwoAllowAttPermFlareMaxVsGiHigh}{$+10$\%}

\newcommand{\capDetailVtwoAllowAttPermFlareMaxVsGiHighPHolm}{0.068}

\newcommand{\capDetailVtwoAllowMatchedFlareMaxVsGiHigh}{$+10$\%}

\newcommand{\capDetailVtwoAllowMatchedFlareMaxVsGiHighPHolm}{0.0098}

\newcommand{\capSharpMatchFlareHighEdgeWPct}{$-3$\%}
\newcommand{\capSharpMatchFlareHighEdgeWLo}{$-8$\%}
\newcommand{\capSharpMatchFlareHighEdgeWHi}{$+3$\%}
\newcommand{\capSharpMatchFlareMaxEdgeWPct}{$+4$\%}
\newcommand{\capSharpMatchFlareMaxEdgeWLo}{$-12$\%}
\newcommand{\capSharpMatchFlareMaxEdgeWHi}{$+12$\%}
\newcommand{\capSharpMatchSunHighEdgeWLo}{$-9$\%}
\newcommand{\capSharpMatchSunHighEdgeWHi}{$+12$\%}
\newcommand{\capSharpMatchSunMaxEdgeWPct}{$+1$\%}
\newcommand{\capSharpMatchSunMaxEdgeWLo}{$-8$\%}
\newcommand{\capSharpMatchSunMaxEdgeWHi}{$+7$\%}
\newcommand{\capSharpMatchEdgeWRangeLo}{$-3$\%}
\newcommand{\capSharpMatchEdgeWRangeHi}{$+4$\%}

\newcommand{\capSharpMatchEdgeWNCells}{4}
\newcommand{\capSharpGitwoHighEdgeWPct}{$-12$\%}
\newcommand{\capSharpGitwoHighEdgeWLo}{$-18$\%}
\newcommand{\capSharpGitwoHighEdgeWHi}{$-4$\%}

\newcommand{\capDetailCeilReadRate}{90\%}

\newcommand{\capDocFlareMinusGiLowLOneTargetVTwo}{$+4.2$\%}
\newcommand{\capDocFlareMinusGiLowLOneTargetVTwoLo}{$-5.2$\%}
\newcommand{\capDocFlareMinusGiLowLOneTargetVTwoHi}{$+14.6$\%}
\newcommand{\capDocFlareMinusGiLowLOneTargetVTwoN}{96}

\newcommand{\capDocFlareMinusGiLowLOneTargetVTwoHolmSig}{no}

\newcommand{\capDocFlareMinusGiLowLOneTargetVTwoHolmSigGlobal}{no}
\newcommand{\capDocFlareMinusGiMedLOneTargetVTwo}{$-2.1$\%}
\newcommand{\capDocFlareMinusGiMedLOneTargetVTwoLo}{$-12.5$\%}
\newcommand{\capDocFlareMinusGiMedLOneTargetVTwoHi}{$+8.3$\%}

\newcommand{\capDocFlareMinusGiMedLOneTargetVTwoHolmSig}{no}

\newcommand{\capDocFlareMinusGiMedLOneTargetVTwoHolmSigGlobal}{no}
\newcommand{\capDocSunburstMinusGiLowLOneTargetVTwo}{$+9.4$\%}
\newcommand{\capDocSunburstMinusGiLowLOneTargetVTwoLo}{$0.0$\%}
\newcommand{\capDocSunburstMinusGiLowLOneTargetVTwoHi}{$+19.8$\%}

\newcommand{\capDocSunburstMinusGiLowLOneTargetVTwoPHolmMcn}{0.43}

\newcommand{\capDocSunburstMinusGiLowLOneTargetVTwoPHolm}{0.44}
\newcommand{\capDocSunburstMinusGiLowLOneTargetVTwoHolmSig}{no}

\newcommand{\capDocSunburstMinusGiLowLOneTargetVTwoHolmSigGlobal}{no}
\newcommand{\capDocSunburstMinusGiMedLOneTargetVTwo}{$+3.1$\%}
\newcommand{\capDocSunburstMinusGiMedLOneTargetVTwoLo}{$-7.3$\%}
\newcommand{\capDocSunburstMinusGiMedLOneTargetVTwoHi}{$+13.5$\%}

\newcommand{\capDocSunburstMinusGiMedLOneTargetVTwoPHolmMcn}{1}

\newcommand{\capDocSunburstMinusGiMedLOneTargetVTwoPHolm}{1}
\newcommand{\capDocSunburstMinusGiMedLOneTargetVTwoHolmSig}{no}

\newcommand{\capDocSunburstMinusGiMedLOneTargetVTwoHolmSigGlobal}{no}
\newcommand{\capDocFlareMinusGiLowLTwoOtherVTwo}{$-12.6$\%}
\newcommand{\capDocFlareMinusGiLowLTwoOtherVTwoLo}{$-19.1$\%}
\newcommand{\capDocFlareMinusGiLowLTwoOtherVTwoHi}{$-6.5$\%}
\newcommand{\capDocFlareMinusGiLowLTwoOtherVTwoN}{199}
\newcommand{\capDocFlareMinusGiLowLTwoOtherVTwoP}{$1.5\times10^{-4}$}

\newcommand{\capDocFlareMinusGiLowLTwoOtherVTwoPHolmMcn}{0.0014}

\newcommand{\capDocFlareMinusGiLowLTwoOtherVTwoHolmSig}{yes}

\newcommand{\capDocFlareMinusGiLowLTwoOtherVTwoHolmSigGlobal}{yes}
\newcommand{\capDocFlareMinusGiMedLTwoOtherVTwo}{$-12.6$\%}
\newcommand{\capDocFlareMinusGiMedLTwoOtherVTwoLo}{$-20.1$\%}
\newcommand{\capDocFlareMinusGiMedLTwoOtherVTwoHi}{$-5.5$\%}

\newcommand{\capDocFlareMinusGiMedLTwoOtherVTwoP}{0.0015}

\newcommand{\capDocFlareMinusGiMedLTwoOtherVTwoPHolmMcn}{0.0051}

\newcommand{\capDocFlareMinusGiMedLTwoOtherVTwoPHolm}{0.0075}
\newcommand{\capDocFlareMinusGiMedLTwoOtherVTwoHolmSig}{yes}
\newcommand{\capDocFlareMinusGiMedLTwoOtherVTwoPHolmGlobal}{0.066}
\newcommand{\capDocFlareMinusGiMedLTwoOtherVTwoHolmSigGlobal}{no}
\newcommand{\capDocSunburstMinusGiLowLTwoOtherVTwo}{$-13.1$\%}
\newcommand{\capDocSunburstMinusGiLowLTwoOtherVTwoLo}{$-19.6$\%}
\newcommand{\capDocSunburstMinusGiLowLTwoOtherVTwoHi}{$-6.5$\%}

\newcommand{\capDocSunburstMinusGiLowLTwoOtherVTwoP}{$2.0\times10^{-4}$}

\newcommand{\capDocSunburstMinusGiLowLTwoOtherVTwoPHolmMcn}{0.0016}

\newcommand{\capDocSunburstMinusGiLowLTwoOtherVTwoHolmSig}{yes}

\newcommand{\capDocSunburstMinusGiLowLTwoOtherVTwoHolmSigGlobal}{yes}
\newcommand{\capDocSunburstMinusGiMedLTwoOtherVTwo}{$-13.1$\%}
\newcommand{\capDocSunburstMinusGiMedLTwoOtherVTwoLo}{$-20.6$\%}
\newcommand{\capDocSunburstMinusGiMedLTwoOtherVTwoHi}{$-6.0$\%}

\newcommand{\capDocSunburstMinusGiMedLTwoOtherVTwoP}{$8.5\times10^{-4}$}

\newcommand{\capDocSunburstMinusGiMedLTwoOtherVTwoPHolmMcn}{0.0041}

\newcommand{\capDocSunburstMinusGiMedLTwoOtherVTwoHolmSig}{yes}

\newcommand{\capDocSunburstMinusGiMedLTwoOtherVTwoHolmSigGlobal}{yes}

\newcommand{\capDocCordSunburstMinusGiLowLOneTargetVTwo}{$+14.5$\%}

\newcommand{\capDocCordSunburstMinusGiLowLOneTargetVTwoPHolm}{0.15}

\newcommand{\capDocCordFlareMinusGiMedLTwoOtherVTwo}{$-25.0$\%}

\newcommand{\capDocCordSunburstMinusGiLowLTwoOtherVTwo}{$-19.0$\%}

\newcommand{\capDocCordSunburstMinusGiLowLTwoOtherVTwoPHolmMcn}{0.0027}

\newcommand{\capDocWildFlareMinusGiMedLTwoOtherVTwo}{$0.0$\%}

\newcommand{\capDocWildSunburstMinusGiLowLTwoOtherVTwo}{$-7.1$\%}

\newcommand{\capDocRateGiLowLOneTargetVTwo}{67.7\%}

\newcommand{\capDocRateGiLowLTwoOtherVTwo}{44.2\%}

\newcommand{\capDocRateGiMedLOneTargetVTwo}{74.0\%}

\newcommand{\capDocRateGiMedLTwoOtherVTwo}{44.2\%}

\newcommand{\capDocRateFlareLOneTargetVTwo}{71.9\%}

\newcommand{\capDocRateFlareLTwoOtherVTwo}{31.7\%}

\newcommand{\capDocRateSunburstLOneTargetVTwo}{77.1\%}

\newcommand{\capDocRateSunburstLTwoOtherVTwo}{31.2\%}

\newcommand{\capDocFloorAnyOtherRegVTwo}{12.1\%}

\newcommand{\capDocFloorAnyOtherCordRegVTwo}{10.0\%}

\newcommand{\capDocFloorAnyOtherWildRegVTwo}{14.1\%}
\newcommand{\capDocRegDispMedGiLowVTwo}{1.04}
\newcommand{\capDocRegShareGtOneGiLowVTwo}{51\%}
\newcommand{\capDocRegShareGtTwoGiLowVTwo}{19\%}
\newcommand{\capDocRegDispMedGiMedVTwo}{2.53}
\newcommand{\capDocRegShareGtOneGiMedVTwo}{80\%}
\newcommand{\capDocRegShareGtTwoGiMedVTwo}{59\%}
\newcommand{\capDocRegDispMedFlareVTwo}{0.87}
\newcommand{\capDocRegShareGtOneFlareVTwo}{46\%}
\newcommand{\capDocRegShareGtTwoFlareVTwo}{10\%}
\newcommand{\capDocRegDispMedSunburstVTwo}{1.08}
\newcommand{\capDocRegShareGtOneSunburstVTwo}{54\%}
\newcommand{\capDocRegShareGtTwoSunburstVTwo}{13\%}
\newcommand{\capDocRegDispMedFloorVTwo}{0.76}
\newcommand{\capDocRegShareGtOneFloorVTwo}{22\%}
\newcommand{\capDocRegShareGtTwoFloorVTwo}{0\%}

\newcommand{\capDocShiftLocLeOneFlareMinusGiLowLTwoOtherVTwo}{$-16.7$\%}

\newcommand{\capDocShiftLocLeOneFlareMinusGiLowLTwoOtherVTwoN}{72}
\newcommand{\capDocShiftLocLeOneFlareMinusGiLowLTwoOtherVTwoP}{0.0019}
\newcommand{\capDocShiftLocLeOneFlareMinusGiMedLTwoOtherVTwo}{$-16.7$\%}

\newcommand{\capDocShiftLocLeOneFlareMinusGiMedLTwoOtherVTwoN}{36}
\newcommand{\capDocShiftLocLeOneFlareMinusGiMedLTwoOtherVTwoP}{0.07}
\newcommand{\capDocShiftLocLeOneSunburstMinusGiLowLTwoOtherVTwo}{$-17.9$\%}

\newcommand{\capDocShiftLocLeOneSunburstMinusGiLowLTwoOtherVTwoN}{67}
\newcommand{\capDocShiftLocLeOneSunburstMinusGiLowLTwoOtherVTwoP}{0.002}
\newcommand{\capDocShiftLocLeOneSunburstMinusGiMedLTwoOtherVTwo}{$-14.7$\%}

\newcommand{\capDocShiftLocLeOneSunburstMinusGiMedLTwoOtherVTwoN}{34}
\newcommand{\capDocShiftLocLeOneSunburstMinusGiMedLTwoOtherVTwoP}{0.062}

\newcommand{\capDocShiftLocLeTwoFlareMinusGiLowLTwoOtherVTwo}{$-14.4$\%}

\newcommand{\capDocShiftLocLeTwoFlareMinusGiLowLTwoOtherVTwoP}{$3.5\times10^{-4}$}
\newcommand{\capDocShiftLocLeTwoFlareMinusGiMedLTwoOtherVTwo}{$-15.6$\%}

\newcommand{\capDocShiftLocLeTwoFlareMinusGiMedLTwoOtherVTwoP}{0.011}
\newcommand{\capDocShiftLocLeTwoSunburstMinusGiLowLTwoOtherVTwo}{$-12.2$\%}

\newcommand{\capDocShiftLocLeTwoSunburstMinusGiLowLTwoOtherVTwoP}{0.0011}
\newcommand{\capDocShiftLocLeTwoSunburstMinusGiMedLTwoOtherVTwo}{$-9.0$\%}

\newcommand{\capDocShiftLocLeTwoSunburstMinusGiMedLTwoOtherVTwoP}{0.14}

\newcommand{\capDocVOneShiftLocLeOneFlareMinusGiLowLTwoOtherVTwo}{$-4.2$\%}

\newcommand{\capDocVOneShiftLocLeOneFlareMinusGiLowLTwoOtherVTwoP}{0.63}
\newcommand{\capDocVOneShiftLocLeOneFlareMinusGiMedLTwoOtherVTwo}{$-5.6$\%}

\newcommand{\capDocVOneShiftLocLeOneFlareMinusGiMedLTwoOtherVTwoP}{0.73}
\newcommand{\capDocVOneShiftLocLeOneSunburstMinusGiLowLTwoOtherVTwo}{$-10.4$\%}

\newcommand{\capDocVOneShiftLocLeOneSunburstMinusGiLowLTwoOtherVTwoP}{0.12}
\newcommand{\capDocVOneShiftLocLeOneSunburstMinusGiMedLTwoOtherVTwo}{$-8.8$\%}

\newcommand{\capDocVOneShiftLocLeOneSunburstMinusGiMedLTwoOtherVTwoP}{0.25}

\newcommand{\capDocTokSetShiftLocLeOneFlareMinusGiLowLTwoOtherVTwo}{$-5.6$\%}

\newcommand{\capDocTokSetShiftLocLeOneFlareMinusGiLowLTwoOtherVTwoP}{0.46}
\newcommand{\capDocTokSetShiftLocLeOneFlareMinusGiMedLTwoOtherVTwo}{$-2.8$\%}

\newcommand{\capDocTokSetShiftLocLeOneFlareMinusGiMedLTwoOtherVTwoP}{1}
\newcommand{\capDocTokSetShiftLocLeOneSunburstMinusGiLowLTwoOtherVTwo}{$-10.4$\%}

\newcommand{\capDocTokSetShiftLocLeOneSunburstMinusGiLowLTwoOtherVTwoP}{0.09}
\newcommand{\capDocTokSetShiftLocLeOneSunburstMinusGiMedLTwoOtherVTwo}{$-5.9$\%}

\newcommand{\capDocTokSetShiftLocLeOneSunburstMinusGiMedLTwoOtherVTwoP}{0.5}

\newcommand{\capDocNStableTokRegVTwo}{738}
\newcommand{\capDocNStableTokVOneVTwo}{601}

\newcommand{\capDocMdeLOneMinVTwo}{14}
\newcommand{\capDocMdeLOneMaxVTwo}{15}
\newcommand{\capDocMdeLOneHolmMinVTwo}{18}
\newcommand{\capDocMdeLOneHolmMaxVTwo}{19}
\newcommand{\capDocMdeLOneNVTwo}{96}
\newcommand{\capDocMdeLOnePowerVTwo}{80\%}

\newcommand{\capDocLocShiftLeOneGiLowMinusFlareTolVTwo}{$+0.42$\%}
\newcommand{\capDocLocShiftLeOneGiLowMinusFlareTolVTwoLo}{$+0.25$\%}
\newcommand{\capDocLocShiftLeOneGiLowMinusFlareTolVTwoHi}{$+1.12$\%}
\newcommand{\capDocLocShiftLeOneGiLowMinusFlareTolVTwoN}{72}

\newcommand{\capDocLocShiftAllGiMedMinusFlareTolVTwo}{$+1.1$\%}

\newcommand{\capDocLocShiftLeOneGiMedMinusFlareTolVTwo}{$0.00$\%}
\newcommand{\capDocLocShiftLeOneGiMedMinusFlareTolVTwoLo}{$-0.14$\%}
\newcommand{\capDocLocShiftLeOneGiMedMinusFlareTolVTwoHi}{$+0.28$\%}
\newcommand{\capDocLocShiftLeOneGiMedMinusFlareTolVTwoN}{36}

\newcommand{\capDocVOneMetricFlareMinusGiLowLTwoOtherVTwoPHolm}{0.063}
\newcommand{\capDocVOneMetricFlareMinusGiLowLTwoOtherVTwoPHolmMcn}{0.06}

\newcommand{\capDocVOneMetricCordSunburstMinusGiMedLOneTargetVTwoPHolm}{0.053}

\newcommand{\capDocChainNTestsTwoCorpusVTwo}{32}
\newcommand{\capDocChainCordSuccDiffFlareVsGiMedVTwo}{$-17$\%}
\newcommand{\capDocChainCordSuccDiffFlareVsGiMedVTwoLo}{$-28$\%}
\newcommand{\capDocChainCordSuccDiffFlareVsGiMedVTwoHi}{$-8$\%}

\newcommand{\capDocChainCordSuccDiffFlareVsGiMedVTwoHolmSig}{no}

\newcommand{\capDocChainCordSuccDiffFlareVsGiMedVTwoHolmSigGlobal}{no}
\newcommand{\capDocChainCordSurvFourDiffFlareVsGiMedVTwo}{$-9$\%}
\newcommand{\capDocChainCordSurvFourDiffFlareVsGiMedVTwoLo}{$-16$\%}
\newcommand{\capDocChainCordSurvFourDiffFlareVsGiMedVTwoHi}{$-3$\%}

\newcommand{\capDocChainCordSurvFourDiffFlareVsGiMedVTwoPHolm}{0.002}
\newcommand{\capDocChainCordSurvFourDiffFlareVsGiMedVTwoHolmSig}{yes}

\newcommand{\capDocChainCordSurvFourDiffFlareVsGiMedVTwoHolmSigGlobal}{yes}
\newcommand{\capDocChainCordUntouchedDiffFlareVsGiMedVTwo}{$-11$\%}
\newcommand{\capDocChainCordUntouchedDiffFlareVsGiMedVTwoLo}{$-17$\%}
\newcommand{\capDocChainCordUntouchedDiffFlareVsGiMedVTwoHi}{$-5$\%}

\newcommand{\capDocChainCordUntouchedDiffFlareVsGiMedVTwoHolmSig}{no}

\newcommand{\capDocChainCordUntouchedDiffFlareVsGiMedVTwoHolmSigGlobal}{no}
\newcommand{\capDocChainCordJointDiffFlareVsGiMedVTwo}{$-28$\%}
\newcommand{\capDocChainCordJointDiffFlareVsGiMedVTwoLo}{$-38$\%}
\newcommand{\capDocChainCordJointDiffFlareVsGiMedVTwoHi}{$-18$\%}

\newcommand{\capDocChainCordJointDiffFlareVsGiMedVTwoPHolm}{0.0039}
\newcommand{\capDocChainCordJointDiffFlareVsGiMedVTwoHolmSig}{yes}

\newcommand{\capDocChainCordJointDiffFlareVsGiMedVTwoHolmSigGlobal}{yes}
\newcommand{\capDocChainCordSuccDiffFlareVsGiLowVTwo}{$-19$\%}
\newcommand{\capDocChainCordSuccDiffFlareVsGiLowVTwoLo}{$-29$\%}
\newcommand{\capDocChainCordSuccDiffFlareVsGiLowVTwoHi}{$-9$\%}

\newcommand{\capDocChainCordSuccDiffFlareVsGiLowVTwoPHolmTwoCorpus}{0.039}

\newcommand{\capDocChainCordSuccDiffFlareVsGiLowVTwoPHolm}{0.0504}
\newcommand{\capDocChainCordSuccDiffFlareVsGiLowVTwoHolmSig}{no}

\newcommand{\capDocChainCordSuccDiffFlareVsGiLowVTwoHolmSigGlobal}{no}
\newcommand{\capDocChainCordSurvFourDiffFlareVsGiLowVTwo}{$-7$\%}
\newcommand{\capDocChainCordSurvFourDiffFlareVsGiLowVTwoLo}{$-14$\%}
\newcommand{\capDocChainCordSurvFourDiffFlareVsGiLowVTwoHi}{$-2$\%}

\newcommand{\capDocChainCordSurvFourDiffFlareVsGiLowVTwoPHolm}{0.021}
\newcommand{\capDocChainCordSurvFourDiffFlareVsGiLowVTwoHolmSig}{yes}

\newcommand{\capDocChainCordSurvFourDiffFlareVsGiLowVTwoHolmSigGlobal}{yes}
\newcommand{\capDocChainCordUntouchedDiffFlareVsGiLowVTwo}{$-14$\%}
\newcommand{\capDocChainCordUntouchedDiffFlareVsGiLowVTwoLo}{$-22$\%}
\newcommand{\capDocChainCordUntouchedDiffFlareVsGiLowVTwoHi}{$-7$\%}

\newcommand{\capDocChainCordUntouchedDiffFlareVsGiLowVTwoHolmSig}{no}

\newcommand{\capDocChainCordUntouchedDiffFlareVsGiLowVTwoHolmSigGlobal}{no}
\newcommand{\capDocChainCordJointDiffFlareVsGiLowVTwo}{$-22$\%}
\newcommand{\capDocChainCordJointDiffFlareVsGiLowVTwoLo}{$-33$\%}
\newcommand{\capDocChainCordJointDiffFlareVsGiLowVTwoHi}{$-12$\%}

\newcommand{\capDocChainCordJointDiffFlareVsGiLowVTwoPHolm}{0.033}
\newcommand{\capDocChainCordJointDiffFlareVsGiLowVTwoHolmSig}{yes}

\newcommand{\capDocChainCordJointDiffFlareVsGiLowVTwoHolmSigGlobal}{yes}
\newcommand{\capDocChainCordSuccDiffSunburstVsGiMedVTwo}{$-5$\%}
\newcommand{\capDocChainCordSuccDiffSunburstVsGiMedVTwoLo}{$-13$\%}
\newcommand{\capDocChainCordSuccDiffSunburstVsGiMedVTwoHi}{$+3$\%}

\newcommand{\capDocChainCordSuccDiffSunburstVsGiMedVTwoHolmSig}{no}

\newcommand{\capDocChainCordSuccDiffSunburstVsGiMedVTwoHolmSigGlobal}{no}
\newcommand{\capDocChainCordSurvFourDiffSunburstVsGiMedVTwo}{$0$\%}
\newcommand{\capDocChainCordSurvFourDiffSunburstVsGiMedVTwoLo}{$-5$\%}
\newcommand{\capDocChainCordSurvFourDiffSunburstVsGiMedVTwoHi}{$+5$\%}

\newcommand{\capDocChainCordSurvFourDiffSunburstVsGiMedVTwoHolmSig}{no}

\newcommand{\capDocChainCordSurvFourDiffSunburstVsGiMedVTwoHolmSigGlobal}{no}
\newcommand{\capDocChainCordUntouchedDiffSunburstVsGiMedVTwo}{$+1$\%}
\newcommand{\capDocChainCordUntouchedDiffSunburstVsGiMedVTwoLo}{$-6$\%}
\newcommand{\capDocChainCordUntouchedDiffSunburstVsGiMedVTwoHi}{$+7$\%}

\newcommand{\capDocChainCordUntouchedDiffSunburstVsGiMedVTwoHolmSig}{no}

\newcommand{\capDocChainCordUntouchedDiffSunburstVsGiMedVTwoHolmSigGlobal}{no}
\newcommand{\capDocChainCordJointDiffSunburstVsGiMedVTwo}{$-7$\%}
\newcommand{\capDocChainCordJointDiffSunburstVsGiMedVTwoLo}{$-16$\%}
\newcommand{\capDocChainCordJointDiffSunburstVsGiMedVTwoHi}{$+2$\%}

\newcommand{\capDocChainCordJointDiffSunburstVsGiMedVTwoHolmSig}{no}

\newcommand{\capDocChainCordJointDiffSunburstVsGiMedVTwoHolmSigGlobal}{no}
\newcommand{\capDocChainCordSuccDiffSunburstVsGiLowVTwo}{$-7$\%}
\newcommand{\capDocChainCordSuccDiffSunburstVsGiLowVTwoLo}{$-12$\%}
\newcommand{\capDocChainCordSuccDiffSunburstVsGiLowVTwoHi}{$-2$\%}

\newcommand{\capDocChainCordSuccDiffSunburstVsGiLowVTwoHolmSig}{no}

\newcommand{\capDocChainCordSuccDiffSunburstVsGiLowVTwoHolmSigGlobal}{no}
\newcommand{\capDocChainCordSurvFourDiffSunburstVsGiLowVTwo}{$+4$\%}
\newcommand{\capDocChainCordSurvFourDiffSunburstVsGiLowVTwoLo}{$-3$\%}
\newcommand{\capDocChainCordSurvFourDiffSunburstVsGiLowVTwoHi}{$+12$\%}

\newcommand{\capDocChainCordSurvFourDiffSunburstVsGiLowVTwoHolmSig}{no}

\newcommand{\capDocChainCordSurvFourDiffSunburstVsGiLowVTwoHolmSigGlobal}{no}
\newcommand{\capDocChainCordUntouchedDiffSunburstVsGiLowVTwo}{$-2$\%}
\newcommand{\capDocChainCordUntouchedDiffSunburstVsGiLowVTwoLo}{$-9$\%}
\newcommand{\capDocChainCordUntouchedDiffSunburstVsGiLowVTwoHi}{$+3$\%}

\newcommand{\capDocChainCordUntouchedDiffSunburstVsGiLowVTwoHolmSig}{no}

\newcommand{\capDocChainCordUntouchedDiffSunburstVsGiLowVTwoHolmSigGlobal}{no}
\newcommand{\capDocChainCordJointDiffSunburstVsGiLowVTwo}{$-1$\%}
\newcommand{\capDocChainCordJointDiffSunburstVsGiLowVTwoLo}{$-10$\%}
\newcommand{\capDocChainCordJointDiffSunburstVsGiLowVTwoHi}{$+10$\%}

\newcommand{\capDocChainCordJointDiffSunburstVsGiLowVTwoHolmSig}{no}

\newcommand{\capDocChainCordJointDiffSunburstVsGiLowVTwoHolmSigGlobal}{no}
\newcommand{\capDocChainWildSuccDiffFlareVsGiMedVTwo}{$-9$\%}
\newcommand{\capDocChainWildSuccDiffFlareVsGiMedVTwoLo}{$-18$\%}
\newcommand{\capDocChainWildSuccDiffFlareVsGiMedVTwoHi}{$-2$\%}

\newcommand{\capDocChainWildSuccDiffFlareVsGiMedVTwoHolmSig}{no}

\newcommand{\capDocChainWildSuccDiffFlareVsGiMedVTwoHolmSigGlobal}{no}
\newcommand{\capDocChainWildSurvFourDiffFlareVsGiMedVTwo}{$+1$\%}
\newcommand{\capDocChainWildSurvFourDiffFlareVsGiMedVTwoLo}{$-5$\%}
\newcommand{\capDocChainWildSurvFourDiffFlareVsGiMedVTwoHi}{$+6$\%}

\newcommand{\capDocChainWildSurvFourDiffFlareVsGiMedVTwoHolmSig}{no}

\newcommand{\capDocChainWildSurvFourDiffFlareVsGiMedVTwoHolmSigGlobal}{no}
\newcommand{\capDocChainWildUntouchedDiffFlareVsGiMedVTwo}{$+5$\%}
\newcommand{\capDocChainWildUntouchedDiffFlareVsGiMedVTwoLo}{$0$\%}
\newcommand{\capDocChainWildUntouchedDiffFlareVsGiMedVTwoHi}{$+12$\%}

\newcommand{\capDocChainWildUntouchedDiffFlareVsGiMedVTwoHolmSig}{no}

\newcommand{\capDocChainWildUntouchedDiffFlareVsGiMedVTwoHolmSigGlobal}{no}
\newcommand{\capDocChainWildJointDiffFlareVsGiMedVTwo}{$-11$\%}
\newcommand{\capDocChainWildJointDiffFlareVsGiMedVTwoLo}{$-21$\%}
\newcommand{\capDocChainWildJointDiffFlareVsGiMedVTwoHi}{$-2$\%}

\newcommand{\capDocChainWildJointDiffFlareVsGiMedVTwoHolmSig}{no}

\newcommand{\capDocChainWildJointDiffFlareVsGiMedVTwoHolmSigGlobal}{no}
\newcommand{\capDocChainWildSuccDiffFlareVsGiLowVTwo}{$+2$\%}
\newcommand{\capDocChainWildSuccDiffFlareVsGiLowVTwoLo}{$-7$\%}
\newcommand{\capDocChainWildSuccDiffFlareVsGiLowVTwoHi}{$+10$\%}

\newcommand{\capDocChainWildSuccDiffFlareVsGiLowVTwoHolmSig}{no}

\newcommand{\capDocChainWildSuccDiffFlareVsGiLowVTwoHolmSigGlobal}{no}
\newcommand{\capDocChainWildSurvFourDiffFlareVsGiLowVTwo}{$+3$\%}
\newcommand{\capDocChainWildSurvFourDiffFlareVsGiLowVTwoLo}{$-4$\%}
\newcommand{\capDocChainWildSurvFourDiffFlareVsGiLowVTwoHi}{$+10$\%}

\newcommand{\capDocChainWildSurvFourDiffFlareVsGiLowVTwoHolmSig}{no}

\newcommand{\capDocChainWildSurvFourDiffFlareVsGiLowVTwoHolmSigGlobal}{no}
\newcommand{\capDocChainWildUntouchedDiffFlareVsGiLowVTwo}{$+1$\%}
\newcommand{\capDocChainWildUntouchedDiffFlareVsGiLowVTwoLo}{$-4$\%}
\newcommand{\capDocChainWildUntouchedDiffFlareVsGiLowVTwoHi}{$+7$\%}

\newcommand{\capDocChainWildUntouchedDiffFlareVsGiLowVTwoHolmSig}{no}

\newcommand{\capDocChainWildUntouchedDiffFlareVsGiLowVTwoHolmSigGlobal}{no}
\newcommand{\capDocChainWildJointDiffFlareVsGiLowVTwo}{$+3$\%}
\newcommand{\capDocChainWildJointDiffFlareVsGiLowVTwoLo}{$-8$\%}
\newcommand{\capDocChainWildJointDiffFlareVsGiLowVTwoHi}{$+16$\%}

\newcommand{\capDocChainWildJointDiffFlareVsGiLowVTwoHolmSig}{no}

\newcommand{\capDocChainWildJointDiffFlareVsGiLowVTwoHolmSigGlobal}{no}
\newcommand{\capDocChainWildSuccDiffSunburstVsGiMedVTwo}{$-1$\%}
\newcommand{\capDocChainWildSuccDiffSunburstVsGiMedVTwoLo}{$-8$\%}
\newcommand{\capDocChainWildSuccDiffSunburstVsGiMedVTwoHi}{$+6$\%}

\newcommand{\capDocChainWildSuccDiffSunburstVsGiMedVTwoHolmSig}{no}

\newcommand{\capDocChainWildSuccDiffSunburstVsGiMedVTwoHolmSigGlobal}{no}
\newcommand{\capDocChainWildSurvFourDiffSunburstVsGiMedVTwo}{$+1$\%}
\newcommand{\capDocChainWildSurvFourDiffSunburstVsGiMedVTwoLo}{$-3$\%}
\newcommand{\capDocChainWildSurvFourDiffSunburstVsGiMedVTwoHi}{$+5$\%}

\newcommand{\capDocChainWildSurvFourDiffSunburstVsGiMedVTwoHolmSig}{no}

\newcommand{\capDocChainWildSurvFourDiffSunburstVsGiMedVTwoHolmSigGlobal}{no}
\newcommand{\capDocChainWildUntouchedDiffSunburstVsGiMedVTwo}{$+8$\%}
\newcommand{\capDocChainWildUntouchedDiffSunburstVsGiMedVTwoLo}{$+1$\%}
\newcommand{\capDocChainWildUntouchedDiffSunburstVsGiMedVTwoHi}{$+16$\%}

\newcommand{\capDocChainWildUntouchedDiffSunburstVsGiMedVTwoHolmSig}{no}

\newcommand{\capDocChainWildUntouchedDiffSunburstVsGiMedVTwoHolmSigGlobal}{no}
\newcommand{\capDocChainWildJointDiffSunburstVsGiMedVTwo}{$-3$\%}
\newcommand{\capDocChainWildJointDiffSunburstVsGiMedVTwoLo}{$-12$\%}
\newcommand{\capDocChainWildJointDiffSunburstVsGiMedVTwoHi}{$+5$\%}

\newcommand{\capDocChainWildJointDiffSunburstVsGiMedVTwoHolmSig}{no}

\newcommand{\capDocChainWildJointDiffSunburstVsGiMedVTwoHolmSigGlobal}{no}
\newcommand{\capDocChainWildSuccDiffSunburstVsGiLowVTwo}{$+10$\%}
\newcommand{\capDocChainWildSuccDiffSunburstVsGiLowVTwoLo}{$+1$\%}
\newcommand{\capDocChainWildSuccDiffSunburstVsGiLowVTwoHi}{$+21$\%}

\newcommand{\capDocChainWildSuccDiffSunburstVsGiLowVTwoHolmSig}{no}

\newcommand{\capDocChainWildSuccDiffSunburstVsGiLowVTwoHolmSigGlobal}{no}
\newcommand{\capDocChainWildSurvFourDiffSunburstVsGiLowVTwo}{$+3$\%}
\newcommand{\capDocChainWildSurvFourDiffSunburstVsGiLowVTwoLo}{$-2$\%}
\newcommand{\capDocChainWildSurvFourDiffSunburstVsGiLowVTwoHi}{$+9$\%}

\newcommand{\capDocChainWildSurvFourDiffSunburstVsGiLowVTwoHolmSig}{no}

\newcommand{\capDocChainWildSurvFourDiffSunburstVsGiLowVTwoHolmSigGlobal}{no}
\newcommand{\capDocChainWildUntouchedDiffSunburstVsGiLowVTwo}{$+3$\%}
\newcommand{\capDocChainWildUntouchedDiffSunburstVsGiLowVTwoLo}{$-2$\%}
\newcommand{\capDocChainWildUntouchedDiffSunburstVsGiLowVTwoHi}{$+9$\%}

\newcommand{\capDocChainWildUntouchedDiffSunburstVsGiLowVTwoHolmSig}{no}

\newcommand{\capDocChainWildUntouchedDiffSunburstVsGiLowVTwoHolmSigGlobal}{no}
\newcommand{\capDocChainWildJointDiffSunburstVsGiLowVTwo}{$+11$\%}
\newcommand{\capDocChainWildJointDiffSunburstVsGiLowVTwoLo}{$0$\%}
\newcommand{\capDocChainWildJointDiffSunburstVsGiLowVTwoHi}{$+24$\%}

\newcommand{\capDocChainWildJointDiffSunburstVsGiLowVTwoHolmSig}{no}

\newcommand{\capDocChainWildJointDiffSunburstVsGiLowVTwoHolmSigGlobal}{no}

\newcommand{\capDocChainAllSurvFourDiffFlareVsGiMedVTwo}{$-4$\%}

\newcommand{\capDocChainAllSurvFourDiffFlareVsGiLowVTwo}{$-2$\%}

\newcommand{\capDocChainCordSuccGiMedVTwo}{92\%}

\newcommand{\capDocChainCordSuccGiMedVTwoN}{106/115}
\newcommand{\capDocChainCordSurvFourGiMedVTwo}{98\%}

\newcommand{\capDocChainCordSurvFourGiMedVTwoN}{80/82}
\newcommand{\capDocChainCordUntouchedGiMedVTwo}{93\%}

\newcommand{\capDocChainCordUntouchedGiMedVTwoN}{84/90}
\newcommand{\capDocChainCordJointGiMedVTwo}{93\%}

\newcommand{\capDocChainCordJointGiMedVTwoN}{80/86}
\newcommand{\capDocChainCordSuccFlareVTwo}{75\%}

\newcommand{\capDocChainCordSuccFlareVTwoN}{86/115}
\newcommand{\capDocChainCordSurvFourFlareVTwo}{89\%}

\newcommand{\capDocChainCordSurvFourFlareVTwoN}{56/63}
\newcommand{\capDocChainCordUntouchedFlareVTwo}{82\%}

\newcommand{\capDocChainCordUntouchedFlareVTwoN}{74/90}
\newcommand{\capDocChainCordJointFlareVTwo}{65\%}

\newcommand{\capDocChainCordJointFlareVTwoN}{56/86}
\newcommand{\capDocChainCordSuccSunburstVTwo}{87\%}

\newcommand{\capDocChainCordSuccSunburstVTwoN}{100/115}
\newcommand{\capDocChainCordSurvFourSunburstVTwo}{97\%}

\newcommand{\capDocChainCordSurvFourSunburstVTwoN}{74/76}
\newcommand{\capDocChainCordUntouchedSunburstVTwo}{94\%}

\newcommand{\capDocChainCordUntouchedSunburstVTwoN}{85/90}
\newcommand{\capDocChainCordJointSunburstVTwo}{86\%}

\newcommand{\capDocChainCordJointSunburstVTwoN}{74/86}
\newcommand{\capDocChainCordSuccGiLowVTwo}{94\%}

\newcommand{\capDocChainCordSuccGiLowVTwoN}{108/115}
\newcommand{\capDocChainCordSurvFourGiLowVTwo}{94\%}

\newcommand{\capDocChainCordSurvFourGiLowVTwoN}{75/80}
\newcommand{\capDocChainCordUntouchedGiLowVTwo}{97\%}

\newcommand{\capDocChainCordUntouchedGiLowVTwoN}{87/90}
\newcommand{\capDocChainCordJointGiLowVTwo}{87\%}

\newcommand{\capDocChainCordJointGiLowVTwoN}{75/86}
\newcommand{\capDocChainWildSuccGiMedVTwo}{91\%}

\newcommand{\capDocChainWildSuccGiMedVTwoN}{107/117}
\newcommand{\capDocChainWildSurvFourGiMedVTwo}{96\%}

\newcommand{\capDocChainWildSurvFourGiMedVTwoN}{81/84}
\newcommand{\capDocChainWildUntouchedGiMedVTwo}{91\%}

\newcommand{\capDocChainWildUntouchedGiMedVTwoN}{84/92}
\newcommand{\capDocChainWildJointGiMedVTwo}{92\%}

\newcommand{\capDocChainWildJointGiMedVTwoN}{81/88}
\newcommand{\capDocChainWildSuccFlareVTwo}{82\%}

\newcommand{\capDocChainWildSuccFlareVTwoN}{96/117}
\newcommand{\capDocChainWildSurvFourFlareVTwo}{97\%}

\newcommand{\capDocChainWildSurvFourFlareVTwoN}{71/73}
\newcommand{\capDocChainWildUntouchedFlareVTwo}{97\%}

\newcommand{\capDocChainWildUntouchedFlareVTwoN}{89/92}
\newcommand{\capDocChainWildJointFlareVTwo}{81\%}

\newcommand{\capDocChainWildJointFlareVTwoN}{71/88}
\newcommand{\capDocChainWildSuccSunburstVTwo}{91\%}

\newcommand{\capDocChainWildSuccSunburstVTwoN}{106/117}
\newcommand{\capDocChainWildSurvFourSunburstVTwo}{98\%}

\newcommand{\capDocChainWildSurvFourSunburstVTwoN}{78/80}
\newcommand{\capDocChainWildUntouchedSunburstVTwo}{99\%}

\newcommand{\capDocChainWildUntouchedSunburstVTwoN}{91/92}
\newcommand{\capDocChainWildJointSunburstVTwo}{89\%}

\newcommand{\capDocChainWildJointSunburstVTwoN}{78/88}
\newcommand{\capDocChainWildSuccGiLowVTwo}{80\%}

\newcommand{\capDocChainWildSuccGiLowVTwoN}{94/117}
\newcommand{\capDocChainWildSurvFourGiLowVTwo}{94\%}

\newcommand{\capDocChainWildSurvFourGiLowVTwoN}{68/72}
\newcommand{\capDocChainWildUntouchedGiLowVTwo}{96\%}

\newcommand{\capDocChainWildUntouchedGiLowVTwoN}{88/92}
\newcommand{\capDocChainWildJointGiLowVTwo}{77\%}

\newcommand{\capDocChainWildJointGiLowVTwoN}{68/88}
\newcommand{\capDocChainAllSuccGiMedVTwo}{92\%}

\newcommand{\capDocChainAllSuccGiMedVTwoN}{213/232}
\newcommand{\capDocChainAllSurvFourGiMedVTwo}{97\%}

\newcommand{\capDocChainAllSurvFourGiMedVTwoN}{161/166}
\newcommand{\capDocChainAllUntouchedGiMedVTwo}{92\%}

\newcommand{\capDocChainAllUntouchedGiMedVTwoN}{168/182}
\newcommand{\capDocChainAllJointGiMedVTwo}{93\%}

\newcommand{\capDocChainAllJointGiMedVTwoN}{161/174}
\newcommand{\capDocChainAllSuccFlareVTwo}{78\%}

\newcommand{\capDocChainAllSuccFlareVTwoN}{182/232}
\newcommand{\capDocChainAllSurvFourFlareVTwo}{93\%}

\newcommand{\capDocChainAllSurvFourFlareVTwoN}{127/136}
\newcommand{\capDocChainAllUntouchedFlareVTwo}{90\%}

\newcommand{\capDocChainAllUntouchedFlareVTwoN}{163/182}
\newcommand{\capDocChainAllJointFlareVTwo}{73\%}

\newcommand{\capDocChainAllJointFlareVTwoN}{127/174}
\newcommand{\capDocChainAllSuccSunburstVTwo}{89\%}

\newcommand{\capDocChainAllSuccSunburstVTwoN}{206/232}
\newcommand{\capDocChainAllSurvFourSunburstVTwo}{97\%}

\newcommand{\capDocChainAllSurvFourSunburstVTwoN}{152/156}
\newcommand{\capDocChainAllUntouchedSunburstVTwo}{97\%}

\newcommand{\capDocChainAllUntouchedSunburstVTwoN}{176/182}
\newcommand{\capDocChainAllJointSunburstVTwo}{87\%}

\newcommand{\capDocChainAllJointSunburstVTwoN}{152/174}
\newcommand{\capDocChainAllSuccGiLowVTwo}{87\%}

\newcommand{\capDocChainAllSuccGiLowVTwoN}{202/232}
\newcommand{\capDocChainAllSurvFourGiLowVTwo}{94\%}

\newcommand{\capDocChainAllSurvFourGiLowVTwoN}{143/152}
\newcommand{\capDocChainAllUntouchedGiLowVTwo}{96\%}

\newcommand{\capDocChainAllUntouchedGiLowVTwoN}{175/182}
\newcommand{\capDocChainAllJointGiLowVTwo}{82\%}

\newcommand{\capDocChainAllJointGiLowVTwoN}{143/174}
\newcommand{\capDocChainTwoCorpusCordFlareVsGiMedPHolmMinVTwo}{0.0016}
\newcommand{\capDocChainTwoCorpusCordFlareVsGiMedPHolmMaxVTwo}{0.18}

\newcommand{\capDocChainTwoCorpusCordFlareVsGiLowPHolmMinVTwo}{0.016}
\newcommand{\capDocChainTwoCorpusCordFlareVsGiLowPHolmMaxVTwo}{0.086}

\newcommand{\capDocChainScaleFourGiMedVTwo}{0.999}
\newcommand{\capDocChainScaleFourShareBelowNinetySevenGiMedVTwo}{15\%}
\newcommand{\capDocChainScaleFourFlareVTwo}{1.001}
\newcommand{\capDocChainScaleFourShareBelowNinetySevenFlareVTwo}{0\%}
\newcommand{\capDocChainScaleFourSunburstVTwo}{1.000}

\newcommand{\capDocChainScaleFourGiLowVTwo}{0.998}

\newcommand{\capChainStatelessGiMedLOneSurvivalVTwo}{95.6\%}

\newcommand{\capChainStatelessGiMedLTwoInPlaceVTwo}{98.3\%}

\newcommand{\capChainStatelessFlareNChainsVTwo}{39}
\newcommand{\capChainStatelessFlareNSourcePhotosVTwo}{23}
\newcommand{\capChainStatelessFlareLOneSurvivalVTwo}{94.1\%}

\newcommand{\capChainStatelessFlareLTwoInPlaceVTwo}{100.0\%}

\newcommand{\capChainStatelessSunburstLOneSurvivalVTwo}{95.6\%}

\newcommand{\capChainStatelessSunburstLTwoInPlaceVTwo}{100.0\%}

\newcommand{\capChainStatelessGiLowNChainsVTwo}{40}
\newcommand{\capChainStatelessGiLowNSourcePhotosVTwo}{23}
\newcommand{\capChainStatelessGiLowLOneSurvivalVTwo}{94.6\%}
\newcommand{\capChainStatelessGiLowLOneSurvivalVTwoLo}{91.4\%}
\newcommand{\capChainStatelessGiLowLOneSurvivalVTwoHi}{97.7\%}
\newcommand{\capChainStatelessGiLowLTwoInPlaceVTwo}{98.9\%}
\newcommand{\capChainStatelessGiLowLTwoInPlaceVTwoLo}{97.4\%}
\newcommand{\capChainStatelessGiLowLTwoInPlaceVTwoHi}{100.0\%}

\newcommand{\capChainConvGiMedLOneSurvivalVTwo}{98.8\%}

\newcommand{\capChainConvGiMedLTwoInPlaceVTwo}{38.1\%}

\newcommand{\capChainConvFlareNChainsVTwo}{19}
\newcommand{\capChainConvFlareNSourcePhotosVTwo}{19}
\newcommand{\capChainConvFlareLOneSurvivalVTwo}{95.0\%}

\newcommand{\capChainConvFlareLTwoInPlaceVTwo}{48.7\%}

\newcommand{\capChainConvSunburstLOneSurvivalVTwo}{95.2\%}

\newcommand{\capChainConvSunburstLTwoInPlaceVTwo}{57.0\%}

\newcommand{\capChainStatelessFlareMinusGiMedLOneSurvivalVTwo}{$-1.5$\%}
\newcommand{\capChainStatelessFlareMinusGiMedLOneSurvivalVTwoLo}{$-5.3$\%}
\newcommand{\capChainStatelessFlareMinusGiMedLOneSurvivalVTwoHi}{$+2.7$\%}

\newcommand{\capChainStatelessFlareMinusGiMedLOneSurvivalVTwoHolmSig}{no}

\newcommand{\capChainStatelessFlareMinusGiMedLOneSurvivalVTwoHolmSigGlobal}{no}
\newcommand{\capChainStatelessFlareMinusGiMedLTwoInPlaceVTwo}{$+1.7$\%}
\newcommand{\capChainStatelessFlareMinusGiMedLTwoInPlaceVTwoLo}{$0.0$\%}
\newcommand{\capChainStatelessFlareMinusGiMedLTwoInPlaceVTwoHi}{$+4.1$\%}

\newcommand{\capChainStatelessFlareMinusGiMedLTwoInPlaceVTwoHolmSig}{no}

\newcommand{\capChainStatelessFlareMinusGiMedLTwoInPlaceVTwoHolmSigGlobal}{no}
\newcommand{\capChainStatelessFlareMinusGiLowLOneSurvivalVTwo}{$-0.4$\%}
\newcommand{\capChainStatelessFlareMinusGiLowLOneSurvivalVTwoLo}{$-5.3$\%}
\newcommand{\capChainStatelessFlareMinusGiLowLOneSurvivalVTwoHi}{$+4.0$\%}

\newcommand{\capChainStatelessFlareMinusGiLowLOneSurvivalVTwoPHolm}{1}
\newcommand{\capChainStatelessFlareMinusGiLowLOneSurvivalVTwoHolmSig}{no}

\newcommand{\capChainStatelessFlareMinusGiLowLOneSurvivalVTwoHolmSigGlobal}{no}
\newcommand{\capChainStatelessFlareMinusGiLowLTwoInPlaceVTwo}{$+1.2$\%}
\newcommand{\capChainStatelessFlareMinusGiLowLTwoInPlaceVTwoLo}{$0.0$\%}
\newcommand{\capChainStatelessFlareMinusGiLowLTwoInPlaceVTwoHi}{$+2.6$\%}

\newcommand{\capChainStatelessFlareMinusGiLowLTwoInPlaceVTwoHolmSig}{no}

\newcommand{\capChainStatelessFlareMinusGiLowLTwoInPlaceVTwoHolmSigGlobal}{no}
\newcommand{\capChainStatelessSunburstMinusGiMedLOneSurvivalVTwo}{$0.0$\%}
\newcommand{\capChainStatelessSunburstMinusGiMedLOneSurvivalVTwoLo}{$-3.5$\%}
\newcommand{\capChainStatelessSunburstMinusGiMedLOneSurvivalVTwoHi}{$+3.7$\%}

\newcommand{\capChainStatelessSunburstMinusGiMedLOneSurvivalVTwoHolmSig}{no}

\newcommand{\capChainStatelessSunburstMinusGiMedLOneSurvivalVTwoHolmSigGlobal}{no}
\newcommand{\capChainStatelessSunburstMinusGiMedLTwoInPlaceVTwo}{$+1.7$\%}
\newcommand{\capChainStatelessSunburstMinusGiMedLTwoInPlaceVTwoLo}{$0.0$\%}
\newcommand{\capChainStatelessSunburstMinusGiMedLTwoInPlaceVTwoHi}{$+4.1$\%}

\newcommand{\capChainStatelessSunburstMinusGiMedLTwoInPlaceVTwoHolmSig}{no}

\newcommand{\capChainStatelessSunburstMinusGiMedLTwoInPlaceVTwoHolmSigGlobal}{no}
\newcommand{\capChainStatelessSunburstMinusGiLowLOneSurvivalVTwo}{$+1.1$\%}
\newcommand{\capChainStatelessSunburstMinusGiLowLOneSurvivalVTwoLo}{$-2.9$\%}
\newcommand{\capChainStatelessSunburstMinusGiLowLOneSurvivalVTwoHi}{$+4.7$\%}

\newcommand{\capChainStatelessSunburstMinusGiLowLOneSurvivalVTwoHolmSig}{no}

\newcommand{\capChainStatelessSunburstMinusGiLowLOneSurvivalVTwoHolmSigGlobal}{no}
\newcommand{\capChainStatelessSunburstMinusGiLowLTwoInPlaceVTwo}{$+1.2$\%}
\newcommand{\capChainStatelessSunburstMinusGiLowLTwoInPlaceVTwoLo}{$0.0$\%}
\newcommand{\capChainStatelessSunburstMinusGiLowLTwoInPlaceVTwoHi}{$+2.6$\%}

\newcommand{\capChainStatelessSunburstMinusGiLowLTwoInPlaceVTwoHolmSig}{no}

\newcommand{\capChainStatelessSunburstMinusGiLowLTwoInPlaceVTwoHolmSigGlobal}{no}
\newcommand{\capChainConvFlareMinusGiMedLOneSurvivalVTwo}{$-3.8$\%}
\newcommand{\capChainConvFlareMinusGiMedLOneSurvivalVTwoLo}{$-10.7$\%}
\newcommand{\capChainConvFlareMinusGiMedLOneSurvivalVTwoHi}{$+2.1$\%}

\newcommand{\capChainConvFlareMinusGiMedLTwoInPlaceVTwo}{$+10.6$\%}
\newcommand{\capChainConvFlareMinusGiMedLTwoInPlaceVTwoLo}{$+0.2$\%}
\newcommand{\capChainConvFlareMinusGiMedLTwoInPlaceVTwoHi}{$+20.8$\%}

\newcommand{\capChainConvFlareMinusGiMedLTwoInPlaceVTwoPHolmBothVariants}{0.74}
\newcommand{\capChainConvFlareMinusGiMedLTwoInPlaceVTwoPHolmInclExploratory}{1}
\newcommand{\capChainConvSunburstMinusGiMedLOneSurvivalVTwo}{$-3.6$\%}
\newcommand{\capChainConvSunburstMinusGiMedLOneSurvivalVTwoLo}{$-8.6$\%}
\newcommand{\capChainConvSunburstMinusGiMedLOneSurvivalVTwoHi}{$+1.1$\%}

\newcommand{\capChainConvSunburstMinusGiMedLTwoInPlaceVTwo}{$+18.9$\%}
\newcommand{\capChainConvSunburstMinusGiMedLTwoInPlaceVTwoLo}{$+4.4$\%}
\newcommand{\capChainConvSunburstMinusGiMedLTwoInPlaceVTwoHi}{$+33.3$\%}

\newcommand{\capChainConvSunburstMinusGiMedLTwoInPlaceVTwoP}{0.026}

\newcommand{\capChainConvSunburstMinusGiMedLTwoInPlaceVTwoPHolmBothVariants}{0.32}
\newcommand{\capChainConvSunburstMinusGiMedLTwoInPlaceVTwoPHolmInclExploratory}{0.95}

\newcommand{\capDetailBandCeilBoxH}{16}
\newcommand{\capDetailBandEdgeFalloffLo}{12}

\newcommand{\capDetailBandEdgeUpper}{22}

\newcommand{\capDetailBandExactAboveCeilFlareMax}{95\%}

\newcommand{\capDetailBandExactAboveCeilSunburstMax}{78\%}
\newcommand{\capDetailBandExactBelowTwelve}{5\%}

\newcommand{\capDetailBandExactTwelveToSixteen}{56\%}

\newcommand{\capDetailBandExactSixteenToTwentyTwo}{86\%}

\newcommand{\capDetailBandExactAboveTwentyTwo}{93\%}

\newcommand{\capDetailBandTierNBands}{60}

\newcommand{\capDetailBandTierNBandsFavourArm}{1}
\newcommand{\capDetailBandTierNBandsFavourControl}{2}

\newcommand{\capHolmNPerm}{20,000}
\newcommand{\capHolmPResolution}{$5.0\times10^{-5}$}
\newcommand{\capHolmNBoot}{2,000}
\newcommand{\capHolmNTestsLocalEditing}{8}

\newcommand{\capHolmNTestsMultiTurn}{40}

\newcommand{\capHolmNTestsReference}{4}

\newcommand{\capHolmNTestsDetail}{4}

\newcommand{\capHolmNTestsGlobal}{56}
\newcommand{\capHolmNRejectGlobal}{11}

\newcommand{\capHolmNExploratoryFamilies}{24}
\newcommand{\capHolmRefInSceneFlareVsGiLow}{$+36.1$\%}
\newcommand{\capHolmRefInSceneFlareVsGiLowLo}{$+26.1$\%}
\newcommand{\capHolmRefInSceneFlareVsGiLowHi}{$+45.4$\%}

\newcommand{\capHolmRefInSceneFlareVsGiLowHolmSig}{yes}

\newcommand{\capHolmRefInSceneFlareVsGiLowHolmSigGlobal}{yes}
\newcommand{\capHolmRefInSceneFlareVsGiMed}{$+41.2$\%}
\newcommand{\capHolmRefInSceneFlareVsGiMedLo}{$+31.1$\%}
\newcommand{\capHolmRefInSceneFlareVsGiMedHi}{$+50.4$\%}
\newcommand{\capHolmRefInSceneFlareVsGiMedN}{119}

\newcommand{\capHolmRefInSceneFlareVsGiMedHolmSig}{yes}

\newcommand{\capHolmRefInSceneFlareVsGiMedHolmSigGlobal}{yes}
\newcommand{\capHolmRefInSceneSunburstVsGiLow}{$+33.6$\%}
\newcommand{\capHolmRefInSceneSunburstVsGiLowLo}{$+23.5$\%}
\newcommand{\capHolmRefInSceneSunburstVsGiLowHi}{$+43.7$\%}

\newcommand{\capHolmRefInSceneSunburstVsGiLowHolmSig}{yes}

\newcommand{\capHolmRefInSceneSunburstVsGiLowHolmSigGlobal}{yes}
\newcommand{\capHolmRefInSceneSunburstVsGiMed}{$+38.7$\%}
\newcommand{\capHolmRefInSceneSunburstVsGiMedLo}{$+29.4$\%}
\newcommand{\capHolmRefInSceneSunburstVsGiMedHi}{$+47.9$\%}

\newcommand{\capHolmRefInSceneSunburstVsGiMedHolmSig}{yes}

\newcommand{\capHolmRefInSceneSunburstVsGiMedHolmSigGlobal}{yes}
\newcommand{\capHolmDetailFlareHighVsGiMedium}{$+2.9$\%}
\newcommand{\capHolmDetailFlareHighVsGiMediumLo}{$-4.8$\%}
\newcommand{\capHolmDetailFlareHighVsGiMediumHi}{$+11.8$\%}

\newcommand{\capHolmDetailFlareHighVsGiMediumHolmSig}{no}

\newcommand{\capHolmDetailFlareHighVsGiMediumHolmSigGlobal}{no}
\newcommand{\capHolmDetailSunburstHighVsGiMedium}{$-4.8$\%}
\newcommand{\capHolmDetailSunburstHighVsGiMediumLo}{$-12.4$\%}
\newcommand{\capHolmDetailSunburstHighVsGiMediumHi}{$+2.9$\%}

\newcommand{\capHolmDetailSunburstHighVsGiMediumHolmSig}{no}

\newcommand{\capHolmDetailSunburstHighVsGiMediumHolmSigGlobal}{no}
\newcommand{\capHolmDetailFlareMaxVsGiHigh}{$+13.2$\%}
\newcommand{\capHolmDetailFlareMaxVsGiHighLo}{$+4.1$\%}
\newcommand{\capHolmDetailFlareMaxVsGiHighHi}{$+24.0$\%}
\newcommand{\capHolmDetailFlareMaxVsGiHighN}{34}

\newcommand{\capHolmDetailFlareMaxVsGiHighPHolm}{0.054}
\newcommand{\capHolmDetailFlareMaxVsGiHighHolmSig}{no}
\newcommand{\capHolmDetailFlareMaxVsGiHighPHolmGlobal}{0.54}
\newcommand{\capHolmDetailFlareMaxVsGiHighHolmSigGlobal}{no}
\newcommand{\capHolmDetailSunburstMaxVsGiHigh}{$-0.4$\%}
\newcommand{\capHolmDetailSunburstMaxVsGiHighLo}{$-13.5$\%}
\newcommand{\capHolmDetailSunburstMaxVsGiHighHi}{$+12.9$\%}
\newcommand{\capHolmDetailSunburstMaxVsGiHighN}{35}

\newcommand{\capHolmDetailSunburstMaxVsGiHighHolmSig}{no}

\newcommand{\capHolmDetailSunburstMaxVsGiHighHolmSigGlobal}{no}

\title{ChatGPT Images 2.5 on Forgery Tasks:\\
Testing Advertised Improvements Against Known Answers}

\author{%
  Ankit Raj\textsuperscript{*} \quad Yuxin Zhang\textsuperscript{*} \quad Kidus Zewde\textsuperscript{*} \quad
  Tommy Duong\textsuperscript{*} \quad Jiaqi Gan\textsuperscript{*}\\[1pt]
  Xingyu Shen\textsuperscript{*} \quad Yuchen Zhou\textsuperscript{*} \quad Huaiyu Guo\textsuperscript{*} \quad
  Siyu Zhang\textsuperscript{*} \quad Simiao Ren\textsuperscript{\dag}\\[2pt]
  Scam.ai\\[1pt]
  {\footnotesize \textsuperscript{*}Equal contribution. \quad \textsuperscript{\dag}Corresponding author:
  \texttt{benren@scam.ai}}%
}
\date{}

\begin{document}
\maketitle
{\renewcommand{\thefootnote}{}\footnote{Authorship of this paper does not imply employment by or affiliation with Scam.AI (Reality Inc.); some authors are independent researchers who collaborated on this work.}\addtocounter{footnote}{-1}}

\begin{abstract}
OpenAI released ChatGPT Images~2.5 on \launchDate{}, advertising more precise local edits,
better consistency across edits, more faithful reference products and sharper detail.
We evaluate these claims on four forgery tasks with answers fixed in advance:
receipt-field alteration, repeated editing, product placement and small-print rendering.
GPT-Image-2 provides same-week baselines at a cheaper and a more expensive tier.
A limited improvement appears in receipt editing. After alignment, OCR detects changes
to surrounding text in \capDocRateFlareLTwoOtherVTwo{} of Flare outputs, against
\capDocRateGiLowLTwoOtherVTwo{} for the cheaper baseline. This gain is concentrated on
CORD receipts and sensitive to shifts of a pixel or less; the forged value itself is
no more often correct. Repeated editing and fine print show no measurable gain.
Product codes become more legible mainly because Images~2.5 draws the product larger.
Defence outcomes change little: localisation remains weak for both generations.
A detector that flags \wildXCtrlMean{} of controlled benchmark images flags only
\wildXWildRate{} of images posted online. Advertised improvements therefore transfer
unevenly to the tested forgery capabilities, while substantial detection limitations remain.
\end{abstract}

\section{Introduction}
\label{sec:intro}

Image editing supports both creative work and document or marketplace fraud.
Entrust's \emph{2025 Identity Fraud Report} records a 244\% year-over-year rise in
digital document forgeries, which now outnumber physical counterfeits
\cite{entrust2025fraud}. A model release that promises better editing therefore raises
two practical questions: does it make forgery easier, and do existing defences still work?

Our earlier AIForge-Doc v2 study evaluated \vTwoN{} receipt and form forgeries made with
GPT-Image-2. In a side-by-side comparison, human inspectors identified the forgery
\vTwoHuman{} of the time, at chance level. Three automated detectors achieved AUCs of
\vTwoSelf{}--\vTwoTrufor{}. Two of those detectors score \vTwoTruforCal{} and
\vTwoDoctamperCal{} on conventional tampering \cite{wu2026forgerjudge}. These findings
establish a detection problem, but leave the generator's editing capability largely unmeasured.

\begin{figure*}[tbp]
  \centering
  \includegraphics[width=\textwidth]{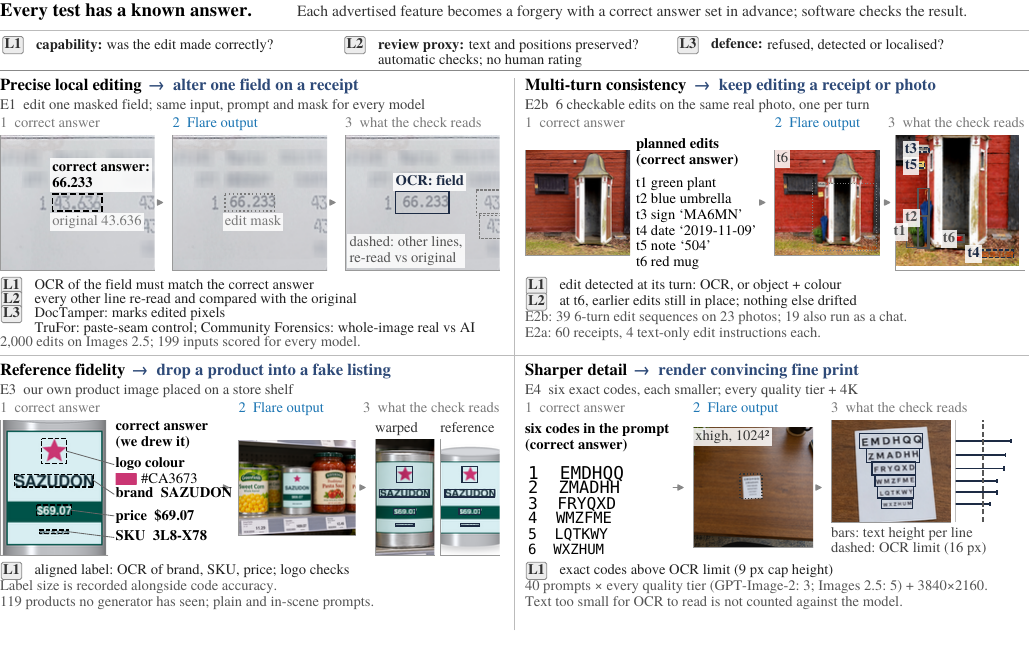}
  \caption{\textbf{Known-answer tasks and measurement layers.} Each panel shows the predetermined
  answer, a Flare output and the automatic check. We measure edit correctness (L1,
  capability), preservation of review-relevant content (L2, an automated review check),
  and refusal or forensic detection (L3, defence). Inputs and prompts are matched
  within each task; E1 also uses a matched mask. On E1, DocTamper localises edited
  pixels and TruFor tests sensitivity to the paste-back seam. Community Forensics
  classifies whole images. These examples explain the method; results appear in
  \S\ref{sec:capability} (L1--L2) and \S\ref{sec:defence} (L3).}
  \label{fig:known_answer}
\end{figure*}

\paragraph{The evaluation gap.} Public leaderboards rank models by pairwise human
preferences \cite{chiang2024chatbotarena,arena2026snapshot}. System cards report unsafe
outputs and refusals \cite{openai2026images25card}. Neither establishes whether a
forged receipt field is correct and the surrounding text preserved. The Images~2.5
system card includes answer-graded biological and cyber tests, but no equivalent
forgery tests. Forensic benchmarks address a complementary question: whether an output
is detected. We evaluate both the requested edit and the response of a defence.

Contemporaneous comparisons also matter. Our earlier study used one model at one
quality tier, leaving model, tier and benchmark effects unresolved. Here we compare
generations, tiers and evaluation samples. These comparisons describe differences;
they do not identify their causes.

\paragraph{From release claims to testable tasks.} On \launchDate{}, OpenAI released
two Images~2.5 API models, Flare and Sunburst \cite{openai2026images25launch}.
The release promises precise local editing and better preservation of earlier edits.
It also promises greater reference fidelity and sharper detail, with new
\texttt{xhigh} and \texttt{max} quality tiers and 4K output. Flare is advertised as faster.

We map the four editing claims to receipt-field alteration, repeated editing, product
placement in a fake listing and small-print rendering. Each task has a correct answer
fixed in advance (Figure~\ref{fig:known_answer}). Both new models are compared with
GPT-Image-2, re-run in the same week at a cheaper and a more expensive tier
(\S\ref{sec:method-setup}).

\paragraph{Research questions.}
\begin{enumerate}[nosep,leftmargin=*,labelindent=0pt,widest=4,label=\textbf{RQ\arabic*.}]
\item \textbf{Capability.} Does Images~2.5 perform the requested forgery edits more
accurately than GPT-Image-2 (\S\ref{sec:cap-local}--\S\ref{sec:cap-detail})?
\item \textbf{Preservation.} Do automatic checks find fewer unintended changes to
surrounding content and earlier edits (\S\ref{sec:cap-local}, \S\ref{sec:cap-multiturn})?
\item \textbf{Defence.} How do refusal, detection and localisation vary with generator
and quality tier (\S\ref{sec:defence-refusal}--\S\ref{sec:defence-wild})?
\item \textbf{Transfer.} Do benchmark detection rates hold for images posted online
(\S\ref{sec:defence-wild})?
\end{enumerate}

\paragraph{Contributions.} The four tasks match inputs and prompts across models
within each condition, with matched masks for receipt-field edits. Quality settings,
routing and run times differ as documented in \S\ref{sec:method}. The call ledger,
per-row scores and analysis code support recomputation. We contribute:
\begin{enumerate}[nosep,leftmargin=*,label=(\arabic*)]
\item an \textbf{evaluation with checkable ground truth}, applied to AIForge-Doc v3:
\docsAttempts{} masked receipt edits, receipt and photo edit sequences, synthetic
product images and a fine-print probe. Automatic checks use text recognition,
geometry, colour and object detection, with reported calibration and validation
limits. Public detectors measure defence outcomes; no OpenAI model serves as a judge;
\item \textbf{two same-week GPT-Image-2 baselines} on identical inputs, one cheaper and one more expensive than the
Images~2.5 setting we test;
\item \textbf{one test family per claim}: cluster permutation tests with Holm's correction within each advertised
claim, a stricter correction across all \capHolmNTestsGlobal{} primary tests as a check, and every other analysis
labelled exploratory.
\end{enumerate}
Appendix~\ref{app:dualuse} describes the planned public release and access to generated images on request.

\section{Related Work}
\label{sec:related}

\paragraph{Document tampering and AIForge-Doc.}
Forensic corpora pair tampered images with localisation masks. DocTamper
\cite{qu2023doctamper} provides 170,000 document images and a document-specific
detector. OSTF \cite{qu2025ostf} extends evaluation to scene text altered by eight
editing models, including unseen forgery types.

AIForge-Doc v1 \cite{wu2026aiforgedoc} contains \vOneN{} diffusion-generated
financial and form forgeries with pixel-precise masks. Version 2 retains the
specifications but uses GPT-Image-2; human two-alternative forced-choice accuracy
falls to \vTwoHuman{}, at chance level \cite{wu2026forgerjudge}.
Version 3 retains the catalogue and source corpora while adding the newer generator.
The present evaluation measures requested-edit correctness and collateral
preservation alongside forensic detection.

\paragraph{Preference and ground-truth evaluation.}
Blind-vote arenas aggregate human preferences into Elo-style ratings
\cite{chiang2024chatbotarena,arena2026snapshot}. These ratings are prominent evidence
for Images~2.5, but do not establish whether a forged value is correct.
GEditBench~v2 \cite{geditbenchv2} trains a visual-consistency judge on pairwise human
preferences. ImgEdit-Bench \cite{ye2025imgedit} uses GPT-4o to rate instruction
adherence, editing quality and preservation on a five-point scale. Such model-based
evaluations can favour a judge's own generations \cite{panickssery2024selfpreference}.

Text-rendering benchmarks provide closer methodological precedents.
MARIO-Eval \cite{chen2023textdiffuser} and AnyText \cite{tuo2024anytext} use OCR to
compare generated text with specified content; AnyText also supports editing.
Our contribution combines field correctness, collateral preservation and defence
measurements on forgery tasks. DINO and CLIP-I reference metrics
\cite{ruiz2023dreambooth} instead measure embedding similarity and do not verify
small label text.

\paragraph{Public forensic tools.}
TruFor \cite{guillaro2023trufor} combines RGB evidence with a learned noise
fingerprint to produce localisation and integrity scores. DocTamper is the
public document localiser used here. Community Forensics
\cite{park2025communityforensics} is a generator-agnostic detector trained on
images from thousands of generators. As in our earlier study, we omit
reconstruction-based AEROBLADE \cite{ricker2024aeroblade}, which is unsuitable for
these local edits because most of each image remains untouched.

\section{Evaluation Design}
\label{sec:method}

We define the model comparisons, scoring instruments and four forgery tasks below.
Several primary contrasts and the statistical framework were selected after results
were known. Section~\ref{sec:method-timing} records that timeline.

\subsection{Models and baselines}
\label{sec:method-setup}

We compare Flare and Sunburst (\texttt{gpt-image-2.5-flare} and
\texttt{gpt-image-2.5-sunburst}) with GPT-Image-2. AIForge-Doc v1 and v2 are earlier
\emph{dataset} releases \cite{wu2026aiforgedoc,wu2026forgerjudge}; v3 is the release
used here. For v3, we re-ran GPT-Image-2 in the same week as Images~2.5
(\docsControlRunMonth{}).

\textbf{Price-bracketing comparisons.} The receipt, repeated-editing and reference
experiments (E1--E3) run Images~2.5 at \texttt{medium}. OpenAI's \texttt{quality}
parameter determines price. No GPT-Image-2 tier has the same price as this setting:
\texttt{low} is the nearest cheaper tier, while \texttt{medium} costs four times as
much. We use both as baselines. The small-print experiment (E4) and paired timings
instead use equal-cost comparisons. E4 pairs GPT-Image-2 \texttt{medium} with
Images~2.5 \texttt{high}, and GPT-Image-2 \texttt{high} with Images~2.5 \texttt{max}
(Appendix~\ref{app:models}).

E1 included both baselines from the start. For E2 and E3, we added \texttt{low}
after the main runs under an addendum fixed before generating its images.
Stimuli, prompts, masks, edit orders, seeds and scoring rules were unchanged.
The added images were generated about 11 hours later than their comparators
(\S\ref{sec:method-timing}). Both baseline contrasts enter each claim's test family.
All calls used one OpenAI API account. The ledger records successes, refusals and
failures; Appendix~\ref{app:design} documents routing and execution.

\subsection{What we measure: capability, review check and defence}
\label{sec:method-layers}

Each task has a correct result fixed before generation (Figure~\ref{fig:known_answer}).
The figures and tables distinguish three measurement layers:
\begin{enumerate}[nosep,leftmargin=*,label=\textbf{L\arabic*.}]
\item \textbf{Capability:} whether the requested edit is correct.
\item \textbf{Review check:} whether automatic checks find changes to other text or
previous edits. This is a preservation proxy, not a measurement of human acceptance.
\item \textbf{Defence:} whether the request is refused, the output detected, or the
edited region localised by a public tool.
\end{enumerate}
We also measure Flare's advertised speed advantage (\S\ref{sec:defence-price}).

\textbf{Scoring instruments.} Capability and preservation checks compare outputs with
known answers using OCR (easyocr \cite{jaidedaieasyocr}), geometric alignment, colour
distance and object detection. Alignment uses SIFT \cite{lowe2004sift} or ORB
\cite{rublee2011orb} with RANSAC \cite{fischler1981ransac}; colour is measured by
$\Delta E_{00}$ \cite{sharma2005ciede2000}; object checks combine OWLv2
\cite{minderer2023owlv2} with a colour test. CLIP \cite{radford2021clip} and DINOv2
\cite{oquab2024dinov2} embedding similarity cannot establish whether label text is
correct, so they do not determine the capability verdicts.

Defence tests use DocTamper \cite{qu2023doctamper}, TruFor
\cite{guillaro2023trufor} and Community Forensics \cite{park2025communityforensics}.
No OpenAI model judges an OpenAI output \cite{panickssery2024selfpreference}.
In one E2 variant, \capChainConvTextModel{} relays instructions to the image model
but performs no scoring.

\textbf{Validation limits.} An AI coding agent (Claude) spot-checked earlier scorer
versions on blind sheets. These checks were not human ratings, and their agreement
counts carry no inferential weight. The final registered E1 token set has no
independent validation.

Two calibration quantities limit interpretation. The E1 \emph{false-change rate}
comes from rereading unedited source replicas. It is a reference rate, not a lower
bound or an estimate of measurement error on generated outputs. The \emph{OCR ceiling}
is the smallest glyph size read reliably on clean renders: \capDetailCeilCapPx{}\,px
cap height in E4, and an output-specific threshold in E3. Misses below that threshold
cannot be attributed to the generator. Appendix~\ref{app:locality} reports a separate
pixel-level measure of change outside the edit.

\subsection{Four forgery tasks}
\label{sec:method-tasks}

\textbf{E1: Change one receipt field.} We apply AIForge-Doc specifications to CORD
\cite{park2019cord} and WildReceipt \cite{sun2021wildreceipt}, using identical prompts
and alpha masks across models. Crops must satisfy the API's aspect-ratio constraints
(Appendix~\ref{app:sizeguard}). The analysis uses the \capDocNSpecs{} specifications
rendered by every model.

Capability is exact OCR agreement with the forged value, restricted to source fields
that OCR can read. The review check detects changes to other text in the crop.
Before rereading, we align each output to its source: median displacement is
\capDocRegDispMedGiMedVTwo{}\,px for GPT-Image-2 \texttt{medium} and
\capDocRegDispMedFlareVTwo{}--\capDocRegDispMedSunburstVTwo{}\,px for Images~2.5.
Even aligned unedited copies produce an OCR false-change rate of
\capDocFloorAnyOtherRegVTwo{}. Shift-stratified results assess sensitivity.
Defence tests use DocTamper for localisation, TruFor for the paste-back seam control,
and Community Forensics for whole-image detection (Appendix~\ref{app:e1}).

\textbf{E2: Edit the same image repeatedly.} The receipt experiment (E2a) applies
\capDocChainTurns{} edits to each of \capDocChainAllReceipts{} receipts:
\capDocChainCordReceipts{} from CORD and \capDocChainWildReceipts{} from WildReceipt.
One field is edited per turn in a seeded order, with each turn sent as a fresh request.
We check whether the target is correct at its own turn and remains intact at the last
turn, and whether never-edited fields remain unchanged. A joint endpoint, added post
hoc as co-primary, counts all requested edits that both succeed and survive.

The photo experiment (E2b) uses \capChainStatelessFlareNSourcePhotosVTwo{} photographs
without people, forming \capChainStatelessFlareNChainsVTwo{} edit sequences.
Each turn adds either a short code, checked by OCR, or a coloured object, checked by
an object detector. Turns are sent as fresh requests or within one chat conversation
where a text model relays instructions. We measure survival of successful edits at
the last turn and preservation of their requested position, requiring box overlap of
at least \capChainInPlaceIou{} (Appendix~\ref{app:e2}).

\textbf{E3: Place a reference product in a fake listing.} Each of \capRefPNRefs{}
synthetic references carries a nonsense brand, price, small SKU code and two-colour
logo. Programmatic references avoid giving any generator a home-field advantage.
Each model places the product in a front-on \emph{hero} view and an \emph{in-scene}
view among other items. We rectify the label and measure exact brand, SKU and price
OCR, logo colour distance and logo shape overlap.

Larger products can make the same label easier to read. We therefore measure label
area and relative glyph size, with exploratory comparisons at matched glyph height
and product size (Appendix~\ref{app:e3}). The documented \texttt{input\_fidelity}
parameter is rejected by every tested model, so all edits use default fidelity
(Appendix~\ref{app:inputfidelity}).

\textbf{E4: Render small print.} Each of \capDetailNPrompts{} prompts requests
\capDetailNLines{} exact random codes at decreasing sizes. We test every tier of each
model and 4K output. The primary outcome is the per-prompt fraction of lines read
exactly above the OCR ceiling. We report excluded prompts with no eligible line in
either configuration. Intention-to-treat rates and matched-glyph-height contrasts are
exploratory.

Latency is measured separately on prompt-paired synchronous runs of E1 edits and a
photoreal core generated at every tier (Appendices~\ref{app:e4} and \ref{app:tiers}).

\subsection{Statistical analysis}
\label{sec:method-stats}

\textbf{Tests and intervals.} Every contrast is paired: Images~2.5 minus a baseline
on the same units. Clusters are specifications in E1, receipts in E2a, source photos
in E2b, references in E3, and prompts in E4 and latency tests. Two-sided cluster
sign-flip tests use \capHolmNPerm{} permutations. Unadjusted percentile bootstrap
95\% confidence intervals use \capHolmNBoot{} cluster resamples. Exact McNemar tests
provide an E1 sensitivity check. We report minimum detectable differences where
computable, and equivalence bounds for defence comparisons interpreted as showing
little change (Appendix~\ref{app:stats}).

\textbf{Multiple comparisons.} Holm's correction \cite{holm1979holm} is applied within
each advertised claim, across layers, models and baselines. The four families are:
\begin{itemize}[nosep,leftmargin=*]
\item \emph{Local editing:} E1 target correctness and surrounding-text changes
(\capHolmNTestsLocalEditing{} tests).
\item \emph{Multi-turn consistency:} E2a success, survival, never-edited fields and
the joint endpoint by corpus, plus E2b survival and position
(\capHolmNTestsMultiTurn{} tests).
\item \emph{Reference fidelity:} E3 in-scene SKU exactness
(\capHolmNTestsReference{} tests).
\item \emph{Sharper detail:} E4 exact lines at the two highest matched settings
(\capHolmNTestsDetail{} tests).
\end{itemize}
A stricter correction across all \capHolmNTestsGlobal{} primary tests provides a
sensitivity check. Other analyses are exploratory and do not determine primary
verdicts. E3 programmatic and rendered references and E4 sensitivities have their own
Holm corrections in \capHolmNExploratoryFamilies{} families; other exploratory results
are unadjusted.

\subsection{Analysis plan and decision timeline}
\label{sec:method-timing}

The E1--E4 metrics were documented on 10~September~2026, before E1 scoring and E2--E4
generation. Primary contrasts in E1, E3 and E4 were selected after their results were
available. The statistical framework was also adopted after all results were known.
Both decisions are therefore post hoc.

E2 rules were fixed before full-run E2a estimates existed, but after a pilot had shown
Flare's results. Chat-mode E2b had already been scored in full, so its contrasts remain
exploratory. Appendix~\ref{app:timing} records the complete sequence and source documents.

\section{Do the Advertised Improvements Hold?}
\label{sec:capability}

\begin{figure*}[tbp]
  \centering
  \includegraphics[width=\textwidth]{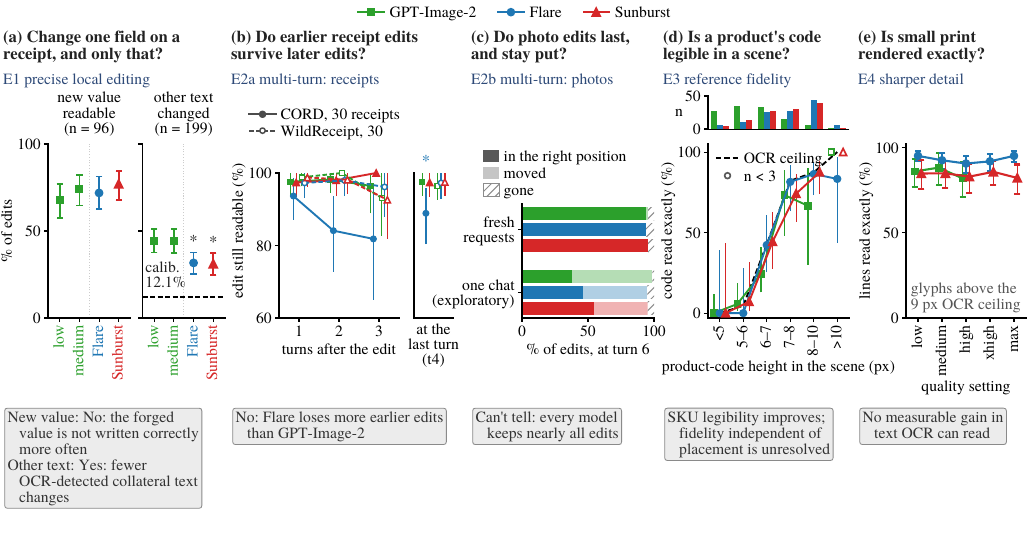}
  \caption{\textbf{Capability outcomes across the four tasks.} Flare: blue circles;
  Sunburst: red triangles; GPT-Image-2: green squares. Whiskers show 95\% intervals;
  $*$ marks a Holm-corrected difference from GPT-Image-2. Grey labels summarise the
  findings, with qualifications discussed in this section.
  \textbf{(a)}~Surrounding-text changes after one receipt edit; dashed lines show
  calibration false-change rates.
  \textbf{(b)}~Receipt edits still readable after 1, 2 or 3 further turns. The narrow
  right-hand axes show final-turn survival, the primary endpoint. Filled markers and
  solid lines denote CORD; hollow markers and dashed lines denote WildReceipt.
  \textbf{(c)}~Photo edits at the final turn: retained in place, retained elsewhere,
  or lost, summing to 100\%. Fresh requests are primary; chat sessions are exploratory
  and carry no significance markers.
  \textbf{(d)}~Exact product-code reads by glyph height. The histogram gives sample
  counts; the dashed OCR ceiling marks the limit below which failures cannot be
  attributed to the generator.
  \textbf{(e)}~Exact small-print reads by quality tier, an exploratory within-model
  comparison. Tier names are not price-matched: GPT-Image-2 \texttt{medium} costs as
  much as Images~2.5 \texttt{high}. The primary equal-cost contrasts are in
  Table~\ref{tab:results}.}
  \label{fig:answers}
\end{figure*}

\begin{table*}[tbp]
\centering
\begingroup\lccode`\~=`\-\lowercase{\endgroup\providecommand{\tabsigns}{\mathcode`\-="8000 \def~{\mbox{\textendash}}}}
\begingroup\lccode`\~=`\+\lowercase{\endgroup\providecommand{\tabnoplus}{\mathcode`\+="8000 \def~{}}}
\providecommand{\tabv}[1]{{\let\%\relax\tabsigns#1}}
\providecommand{\tabci}[2]{{\let\%\relax\tabsigns\tabnoplus[#1,\,#2]}}
\def\tabyes{yes}
\providecommand{\tabstar}[1]{\edef\tabtmp{#1}\ifx\tabtmp\tabyes$^{*}$\fi}
\providecommand{\tabdag}[1]{\edef\tabtmp{#1}\ifx\tabtmp\tabyes$^{\dagger}$\fi}
\providecommand{\tabmarks}[2]{\makebox[1.15em][l]{\tabstar{#1}\tabdag{#2}}}
\caption{\textbf{Task outcomes and matched GPT-Image-2 comparisons.} Flare and Sunburst are the Images~2.5 models. L1 measures task correctness; L2 is an automated preservation proxy, not human acceptance. Interpretation and task-specific limitations are in \S\ref{sec:capability}. $\Delta$: Images~2.5 $-$ GPT-Image-2 in pp (label area: ratio), unadjusted 95\% bootstrap CI clustered by specification, receipt, source photo, reference or prompt. \texttt{low}, \texttt{med.}, \texttt{high}: GPT-Image-2 tiers. E1--E3 compare Images~2.5 \texttt{medium} with GPT-Image-2 \texttt{low} and \texttt{medium}, which cost less and four times more (\S\ref{sec:method-setup}); the \texttt{low} runs of E2a, E2b and E3 were added post hoc (Appendix~\ref{app:lowarm}). E4 compares settings of equal cost: GPT-Image-2 \texttt{med.} with Images~2.5 \texttt{high}, and GPT-Image-2 \texttt{high} with Images~2.5 \texttt{max}. Rows marked expl.\ are exploratory; every other $\Delta$ is a primary endpoint. $^{*}$\,significant after Holm correction within the claim; $^{\dagger}$\,also after correction across all \capHolmNTestsGlobal{} primary tests (Appendix~\ref{app:stats}). WildReceipt and conversational chains: Table~\ref{tab:results-appendix}; L3: Table~\ref{tab:defence-results}.}
\label{tab:results}
\footnotesize
\setlength{\tabcolsep}{1.6pt}
\renewcommand{\arraystretch}{1.05}
\providecommand{\tabms}{\setlength{\baselineskip}{\arraystretch\baselineskip}}
\begin{tabular*}{\textwidth}{@{\extracolsep{\fill}}p{0.11\textwidth}ll>{\raggedright\arraybackslash}p{0.2\textwidth}l@{\extracolsep{0pt}\hspace{2pt}}r@{\extracolsep{\fill}\hspace{5pt}}rr@{\extracolsep{\fill}\hspace{5pt}}r@{\extracolsep{0pt}\hspace{2pt}}l@{\extracolsep{\fill}\hspace{5pt}}r@{\extracolsep{0pt}\hspace{2pt}}l@{}}
\toprule
Claim $\to$ & Exp. & Layer & Metric (unit, n) & \multicolumn{2}{c}{GPT-Image-2} & Flare & Sunburst & \multicolumn{4}{c}{$\Delta$ Images~2.5 $-$ GPT-Image-2 [95\% CI]} \tabularnewline
\cmidrule(lr){9-12}
fraud task & & & & & & & & \multicolumn{2}{c}{Flare} & \multicolumn{2}{c}{Sunburst} \tabularnewline
\midrule
\multirow[t]{6}{\hsize}{\tabms\raggedright \textbf{Precise local editing} $\to$ alter one field on a receipt\par}
& E1
& L1
& Target value read (\%)
& \texttt{low}
& \tabv{\capDocRateGiLowLOneTargetVTwo}
& \tabv{\capDocRateFlareLOneTargetVTwo}
& \tabv{\capDocRateSunburstLOneTargetVTwo}
& \tabv{\capDocFlareMinusGiLowLOneTargetVTwo}\tabmarks{\capDocFlareMinusGiLowLOneTargetVTwoHolmSig}{\capDocFlareMinusGiLowLOneTargetVTwoHolmSigGlobal}
& \tabci{\capDocFlareMinusGiLowLOneTargetVTwoLo}{\capDocFlareMinusGiLowLOneTargetVTwoHi}
& \tabv{\capDocSunburstMinusGiLowLOneTargetVTwo}\tabmarks{\capDocSunburstMinusGiLowLOneTargetVTwoHolmSig}{\capDocSunburstMinusGiLowLOneTargetVTwoHolmSigGlobal}
& \tabci{\capDocSunburstMinusGiLowLOneTargetVTwoLo}{\capDocSunburstMinusGiLowLOneTargetVTwoHi}
\tabularnewline

&
&
& \quad n\,=\,\capDocFlareMinusGiLowLOneTargetVTwoN{} readable fields
& \texttt{med.}
& \tabv{\capDocRateGiMedLOneTargetVTwo}
&
&
& \tabv{\capDocFlareMinusGiMedLOneTargetVTwo}\tabmarks{\capDocFlareMinusGiMedLOneTargetVTwoHolmSig}{\capDocFlareMinusGiMedLOneTargetVTwoHolmSigGlobal}
& \tabci{\capDocFlareMinusGiMedLOneTargetVTwoLo}{\capDocFlareMinusGiMedLOneTargetVTwoHi}
& \tabv{\capDocSunburstMinusGiMedLOneTargetVTwo}\tabmarks{\capDocSunburstMinusGiMedLOneTargetVTwoHolmSig}{\capDocSunburstMinusGiMedLOneTargetVTwoHolmSigGlobal}
& \tabci{\capDocSunburstMinusGiMedLOneTargetVTwoLo}{\capDocSunburstMinusGiMedLOneTargetVTwoHi}
\tabularnewline

&
& L2
& Other text changed, aligned (\%)
& \texttt{low}
& \tabv{\capDocRateGiLowLTwoOtherVTwo}
& \tabv{\capDocRateFlareLTwoOtherVTwo}
& \tabv{\capDocRateSunburstLTwoOtherVTwo}
& \tabv{\capDocFlareMinusGiLowLTwoOtherVTwo}\tabmarks{\capDocFlareMinusGiLowLTwoOtherVTwoHolmSig}{\capDocFlareMinusGiLowLTwoOtherVTwoHolmSigGlobal}
& \tabci{\capDocFlareMinusGiLowLTwoOtherVTwoLo}{\capDocFlareMinusGiLowLTwoOtherVTwoHi}
& \tabv{\capDocSunburstMinusGiLowLTwoOtherVTwo}\tabmarks{\capDocSunburstMinusGiLowLTwoOtherVTwoHolmSig}{\capDocSunburstMinusGiLowLTwoOtherVTwoHolmSigGlobal}
& \tabci{\capDocSunburstMinusGiLowLTwoOtherVTwoLo}{\capDocSunburstMinusGiLowLTwoOtherVTwoHi}
\tabularnewline

&
&
& \quad n\,=\,\capDocFlareMinusGiLowLTwoOtherVTwoN; calibration \tabv{\capDocFloorAnyOtherRegVTwo}
& \texttt{med.}
& \tabv{\capDocRateGiMedLTwoOtherVTwo}
&
&
& \tabv{\capDocFlareMinusGiMedLTwoOtherVTwo}\tabmarks{\capDocFlareMinusGiMedLTwoOtherVTwoHolmSig}{\capDocFlareMinusGiMedLTwoOtherVTwoHolmSigGlobal}
& \tabci{\capDocFlareMinusGiMedLTwoOtherVTwoLo}{\capDocFlareMinusGiMedLTwoOtherVTwoHi}
& \tabv{\capDocSunburstMinusGiMedLTwoOtherVTwo}\tabmarks{\capDocSunburstMinusGiMedLTwoOtherVTwoHolmSig}{\capDocSunburstMinusGiMedLTwoOtherVTwoHolmSigGlobal}
& \tabci{\capDocSunburstMinusGiMedLTwoOtherVTwoLo}{\capDocSunburstMinusGiMedLTwoOtherVTwoHi}
\tabularnewline
\cmidrule[\lightrulewidth]{1-12}
\multirow[t]{8}{\hsize}{\tabms\raggedright \textbf{Multi-turn consistency} $\to$ keep editing the receipt\par}
& E2a
& L1
& Edit success, CORD (\%)
& \texttt{med.}
& \tabv{\capDocChainCordSuccGiMedVTwo}
& \tabv{\capDocChainCordSuccFlareVTwo}
& \tabv{\capDocChainCordSuccSunburstVTwo}
& \tabv{\capDocChainCordSuccDiffFlareVsGiMedVTwo}\tabmarks{\capDocChainCordSuccDiffFlareVsGiMedVTwoHolmSig}{\capDocChainCordSuccDiffFlareVsGiMedVTwoHolmSigGlobal}
& \tabci{\capDocChainCordSuccDiffFlareVsGiMedVTwoLo}{\capDocChainCordSuccDiffFlareVsGiMedVTwoHi}
& \tabv{\capDocChainCordSuccDiffSunburstVsGiMedVTwo}\tabmarks{\capDocChainCordSuccDiffSunburstVsGiMedVTwoHolmSig}{\capDocChainCordSuccDiffSunburstVsGiMedVTwoHolmSigGlobal}
& \tabci{\capDocChainCordSuccDiffSunburstVsGiMedVTwoLo}{\capDocChainCordSuccDiffSunburstVsGiMedVTwoHi}
\tabularnewline

&
&
& \quad \capDocChainCordReceipts{} receipts
& \texttt{low}
& \tabv{\capDocChainCordSuccGiLowVTwo}
&
&
& \tabv{\capDocChainCordSuccDiffFlareVsGiLowVTwo}\tabmarks{\capDocChainCordSuccDiffFlareVsGiLowVTwoHolmSig}{\capDocChainCordSuccDiffFlareVsGiLowVTwoHolmSigGlobal}
& \tabci{\capDocChainCordSuccDiffFlareVsGiLowVTwoLo}{\capDocChainCordSuccDiffFlareVsGiLowVTwoHi}
& \tabv{\capDocChainCordSuccDiffSunburstVsGiLowVTwo}\tabmarks{\capDocChainCordSuccDiffSunburstVsGiLowVTwoHolmSig}{\capDocChainCordSuccDiffSunburstVsGiLowVTwoHolmSigGlobal}
& \tabci{\capDocChainCordSuccDiffSunburstVsGiLowVTwoLo}{\capDocChainCordSuccDiffSunburstVsGiLowVTwoHi}
\tabularnewline

&
& L1
& Edit intact at last turn (\%)
& \texttt{med.}
& \tabv{\capDocChainCordSurvFourGiMedVTwo}
& \tabv{\capDocChainCordSurvFourFlareVTwo}
& \tabv{\capDocChainCordSurvFourSunburstVTwo}
& \tabv{\capDocChainCordSurvFourDiffFlareVsGiMedVTwo}\tabmarks{\capDocChainCordSurvFourDiffFlareVsGiMedVTwoHolmSig}{\capDocChainCordSurvFourDiffFlareVsGiMedVTwoHolmSigGlobal}
& \tabci{\capDocChainCordSurvFourDiffFlareVsGiMedVTwoLo}{\capDocChainCordSurvFourDiffFlareVsGiMedVTwoHi}
& \tabv{\capDocChainCordSurvFourDiffSunburstVsGiMedVTwo}\tabmarks{\capDocChainCordSurvFourDiffSunburstVsGiMedVTwoHolmSig}{\capDocChainCordSurvFourDiffSunburstVsGiMedVTwoHolmSigGlobal}
& \tabci{\capDocChainCordSurvFourDiffSunburstVsGiMedVTwoLo}{\capDocChainCordSurvFourDiffSunburstVsGiMedVTwoHi}
\tabularnewline

&
&
& \quad
& \texttt{low}
& \tabv{\capDocChainCordSurvFourGiLowVTwo}
&
&
& \tabv{\capDocChainCordSurvFourDiffFlareVsGiLowVTwo}\tabmarks{\capDocChainCordSurvFourDiffFlareVsGiLowVTwoHolmSig}{\capDocChainCordSurvFourDiffFlareVsGiLowVTwoHolmSigGlobal}
& \tabci{\capDocChainCordSurvFourDiffFlareVsGiLowVTwoLo}{\capDocChainCordSurvFourDiffFlareVsGiLowVTwoHi}
& \tabv{\capDocChainCordSurvFourDiffSunburstVsGiLowVTwo}\tabmarks{\capDocChainCordSurvFourDiffSunburstVsGiLowVTwoHolmSig}{\capDocChainCordSurvFourDiffSunburstVsGiLowVTwoHolmSigGlobal}
& \tabci{\capDocChainCordSurvFourDiffSunburstVsGiLowVTwoLo}{\capDocChainCordSurvFourDiffSunburstVsGiLowVTwoHi}
\tabularnewline

&
& L1
& Succeeded and intact (\%)
& \texttt{med.}
& \tabv{\capDocChainCordJointGiMedVTwo}
& \tabv{\capDocChainCordJointFlareVTwo}
& \tabv{\capDocChainCordJointSunburstVTwo}
& \tabv{\capDocChainCordJointDiffFlareVsGiMedVTwo}\tabmarks{\capDocChainCordJointDiffFlareVsGiMedVTwoHolmSig}{\capDocChainCordJointDiffFlareVsGiMedVTwoHolmSigGlobal}
& \tabci{\capDocChainCordJointDiffFlareVsGiMedVTwoLo}{\capDocChainCordJointDiffFlareVsGiMedVTwoHi}
& \tabv{\capDocChainCordJointDiffSunburstVsGiMedVTwo}\tabmarks{\capDocChainCordJointDiffSunburstVsGiMedVTwoHolmSig}{\capDocChainCordJointDiffSunburstVsGiMedVTwoHolmSigGlobal}
& \tabci{\capDocChainCordJointDiffSunburstVsGiMedVTwoLo}{\capDocChainCordJointDiffSunburstVsGiMedVTwoHi}
\tabularnewline

&
&
& \quad
& \texttt{low}
& \tabv{\capDocChainCordJointGiLowVTwo}
&
&
& \tabv{\capDocChainCordJointDiffFlareVsGiLowVTwo}\tabmarks{\capDocChainCordJointDiffFlareVsGiLowVTwoHolmSig}{\capDocChainCordJointDiffFlareVsGiLowVTwoHolmSigGlobal}
& \tabci{\capDocChainCordJointDiffFlareVsGiLowVTwoLo}{\capDocChainCordJointDiffFlareVsGiLowVTwoHi}
& \tabv{\capDocChainCordJointDiffSunburstVsGiLowVTwo}\tabmarks{\capDocChainCordJointDiffSunburstVsGiLowVTwoHolmSig}{\capDocChainCordJointDiffSunburstVsGiLowVTwoHolmSigGlobal}
& \tabci{\capDocChainCordJointDiffSunburstVsGiLowVTwoLo}{\capDocChainCordJointDiffSunburstVsGiLowVTwoHi}
\tabularnewline

&
& L2
& Never-edited fields kept (\%)
& \texttt{med.}
& \tabv{\capDocChainCordUntouchedGiMedVTwo}
& \tabv{\capDocChainCordUntouchedFlareVTwo}
& \tabv{\capDocChainCordUntouchedSunburstVTwo}
& \tabv{\capDocChainCordUntouchedDiffFlareVsGiMedVTwo}\tabmarks{\capDocChainCordUntouchedDiffFlareVsGiMedVTwoHolmSig}{\capDocChainCordUntouchedDiffFlareVsGiMedVTwoHolmSigGlobal}
& \tabci{\capDocChainCordUntouchedDiffFlareVsGiMedVTwoLo}{\capDocChainCordUntouchedDiffFlareVsGiMedVTwoHi}
& \tabv{\capDocChainCordUntouchedDiffSunburstVsGiMedVTwo}\tabmarks{\capDocChainCordUntouchedDiffSunburstVsGiMedVTwoHolmSig}{\capDocChainCordUntouchedDiffSunburstVsGiMedVTwoHolmSigGlobal}
& \tabci{\capDocChainCordUntouchedDiffSunburstVsGiMedVTwoLo}{\capDocChainCordUntouchedDiffSunburstVsGiMedVTwoHi}
\tabularnewline

&
&
& \quad
& \texttt{low}
& \tabv{\capDocChainCordUntouchedGiLowVTwo}
&
&
& \tabv{\capDocChainCordUntouchedDiffFlareVsGiLowVTwo}\tabmarks{\capDocChainCordUntouchedDiffFlareVsGiLowVTwoHolmSig}{\capDocChainCordUntouchedDiffFlareVsGiLowVTwoHolmSigGlobal}
& \tabci{\capDocChainCordUntouchedDiffFlareVsGiLowVTwoLo}{\capDocChainCordUntouchedDiffFlareVsGiLowVTwoHi}
& \tabv{\capDocChainCordUntouchedDiffSunburstVsGiLowVTwo}\tabmarks{\capDocChainCordUntouchedDiffSunburstVsGiLowVTwoHolmSig}{\capDocChainCordUntouchedDiffSunburstVsGiLowVTwoHolmSigGlobal}
& \tabci{\capDocChainCordUntouchedDiffSunburstVsGiLowVTwoLo}{\capDocChainCordUntouchedDiffSunburstVsGiLowVTwoHi}
\tabularnewline
\cmidrule[\lightrulewidth]{1-12}
\multirow[t]{5}{\hsize}{\tabms\raggedright \textbf{Multi-turn consistency} $\to$ keep editing a photo\par}
& E2b
& L1
& Edit intact at last turn, stateless (\%)
& \texttt{med.}
& \tabv{\capChainStatelessGiMedLOneSurvivalVTwo}
& \tabv{\capChainStatelessFlareLOneSurvivalVTwo}
& \tabv{\capChainStatelessSunburstLOneSurvivalVTwo}
& \tabv{\capChainStatelessFlareMinusGiMedLOneSurvivalVTwo}\tabmarks{\capChainStatelessFlareMinusGiMedLOneSurvivalVTwoHolmSig}{\capChainStatelessFlareMinusGiMedLOneSurvivalVTwoHolmSigGlobal}
& \tabci{\capChainStatelessFlareMinusGiMedLOneSurvivalVTwoLo}{\capChainStatelessFlareMinusGiMedLOneSurvivalVTwoHi}
& \tabv{\capChainStatelessSunburstMinusGiMedLOneSurvivalVTwo}\tabmarks{\capChainStatelessSunburstMinusGiMedLOneSurvivalVTwoHolmSig}{\capChainStatelessSunburstMinusGiMedLOneSurvivalVTwoHolmSigGlobal}
& \tabci{\capChainStatelessSunburstMinusGiMedLOneSurvivalVTwoLo}{\capChainStatelessSunburstMinusGiMedLOneSurvivalVTwoHi}
\tabularnewline

&
&
& \quad \capChainStatelessFlareNChainsVTwo{} chains on \capChainStatelessFlareNSourcePhotosVTwo{} photos
& \texttt{low}
& \tabv{\capChainStatelessGiLowLOneSurvivalVTwo}
&
&
& \tabv{\capChainStatelessFlareMinusGiLowLOneSurvivalVTwo}\tabmarks{\capChainStatelessFlareMinusGiLowLOneSurvivalVTwoHolmSig}{\capChainStatelessFlareMinusGiLowLOneSurvivalVTwoHolmSigGlobal}
& \tabci{\capChainStatelessFlareMinusGiLowLOneSurvivalVTwoLo}{\capChainStatelessFlareMinusGiLowLOneSurvivalVTwoHi}
& \tabv{\capChainStatelessSunburstMinusGiLowLOneSurvivalVTwo}\tabmarks{\capChainStatelessSunburstMinusGiLowLOneSurvivalVTwoHolmSig}{\capChainStatelessSunburstMinusGiLowLOneSurvivalVTwoHolmSigGlobal}
& \tabci{\capChainStatelessSunburstMinusGiLowLOneSurvivalVTwoLo}{\capChainStatelessSunburstMinusGiLowLOneSurvivalVTwoHi}
\tabularnewline

&
& L2
& Intact edits in the right position (\%)
& \texttt{med.}
& \tabv{\capChainStatelessGiMedLTwoInPlaceVTwo}
& \tabv{\capChainStatelessFlareLTwoInPlaceVTwo}
& \tabv{\capChainStatelessSunburstLTwoInPlaceVTwo}
& \tabv{\capChainStatelessFlareMinusGiMedLTwoInPlaceVTwo}\tabmarks{\capChainStatelessFlareMinusGiMedLTwoInPlaceVTwoHolmSig}{\capChainStatelessFlareMinusGiMedLTwoInPlaceVTwoHolmSigGlobal}
& \tabci{\capChainStatelessFlareMinusGiMedLTwoInPlaceVTwoLo}{\capChainStatelessFlareMinusGiMedLTwoInPlaceVTwoHi}
& \tabv{\capChainStatelessSunburstMinusGiMedLTwoInPlaceVTwo}\tabmarks{\capChainStatelessSunburstMinusGiMedLTwoInPlaceVTwoHolmSig}{\capChainStatelessSunburstMinusGiMedLTwoInPlaceVTwoHolmSigGlobal}
& \tabci{\capChainStatelessSunburstMinusGiMedLTwoInPlaceVTwoLo}{\capChainStatelessSunburstMinusGiMedLTwoInPlaceVTwoHi}
\tabularnewline

&
&
& \quad
& \texttt{low}
& \tabv{\capChainStatelessGiLowLTwoInPlaceVTwo}
&
&
& \tabv{\capChainStatelessFlareMinusGiLowLTwoInPlaceVTwo}\tabmarks{\capChainStatelessFlareMinusGiLowLTwoInPlaceVTwoHolmSig}{\capChainStatelessFlareMinusGiLowLTwoInPlaceVTwoHolmSigGlobal}
& \tabci{\capChainStatelessFlareMinusGiLowLTwoInPlaceVTwoLo}{\capChainStatelessFlareMinusGiLowLTwoInPlaceVTwoHi}
& \tabv{\capChainStatelessSunburstMinusGiLowLTwoInPlaceVTwo}\tabmarks{\capChainStatelessSunburstMinusGiLowLTwoInPlaceVTwoHolmSig}{\capChainStatelessSunburstMinusGiLowLTwoInPlaceVTwoHolmSigGlobal}
& \tabci{\capChainStatelessSunburstMinusGiLowLTwoInPlaceVTwoLo}{\capChainStatelessSunburstMinusGiLowLTwoInPlaceVTwoHi}
\tabularnewline
\cmidrule[\lightrulewidth]{1-12}
\multirow[t]{6}{\hsize}{\tabms\raggedright \textbf{Reference fidelity} $\to$ drop a product into a fake listing\par}
& E3
& L1
& SKU code read exactly, in scene (\%)
& \texttt{med.}
& \tabv{\capRefPInSceneSkuExactGi}
& \tabv{\capRefPInSceneSkuExactFlare}
& \tabv{\capRefPInSceneSkuExactSunburst}
& \tabv{\capHolmRefInSceneFlareVsGiMed}\tabmarks{\capHolmRefInSceneFlareVsGiMedHolmSig}{\capHolmRefInSceneFlareVsGiMedHolmSigGlobal}
& \tabci{\capHolmRefInSceneFlareVsGiMedLo}{\capHolmRefInSceneFlareVsGiMedHi}
& \tabv{\capHolmRefInSceneSunburstVsGiMed}\tabmarks{\capHolmRefInSceneSunburstVsGiMedHolmSig}{\capHolmRefInSceneSunburstVsGiMedHolmSigGlobal}
& \tabci{\capHolmRefInSceneSunburstVsGiMedLo}{\capHolmRefInSceneSunburstVsGiMedHi}
\tabularnewline

&
&
& \quad n\,=\,\capHolmRefInSceneFlareVsGiMedN{} references
& \texttt{low}
& \tabv{\capRefPInSceneSkuExactGiLowRate}
&
&
& \tabv{\capHolmRefInSceneFlareVsGiLow}\tabmarks{\capHolmRefInSceneFlareVsGiLowHolmSig}{\capHolmRefInSceneFlareVsGiLowHolmSigGlobal}
& \tabci{\capHolmRefInSceneFlareVsGiLowLo}{\capHolmRefInSceneFlareVsGiLowHi}
& \tabv{\capHolmRefInSceneSunburstVsGiLow}\tabmarks{\capHolmRefInSceneSunburstVsGiLowHolmSig}{\capHolmRefInSceneSunburstVsGiLowHolmSigGlobal}
& \tabci{\capHolmRefInSceneSunburstVsGiLowLo}{\capHolmRefInSceneSunburstVsGiLowHi}
\tabularnewline

&
&
& Label area in scene (\% frame; $\Delta$ ratio, expl.)
& \texttt{med.}
& \tabv{\capRefPlProgInSceneAreaPctGi}
& \tabv{\capRefPlProgInSceneAreaPctFlare}
& \tabv{\capRefPlProgInSceneAreaPctSunburst}
& $\times$\tabv{\capRefPlProgInSceneFlareVsGiAreaRatio}\tabmarks{}{}
&
& $\times$\tabv{\capRefPlProgInSceneSunburstVsGiAreaRatio}\tabmarks{}{}
&
\tabularnewline

&
&
&
& \texttt{low}
& \tabv{\capRefPlProgInSceneAreaPctGiLow}
&
&
& $\times$\tabv{\capRefPlProgInSceneFlareVsGiLowAreaRatio}\tabmarks{}{}
&
& $\times$\tabv{\capRefPlProgInSceneSunburstVsGiLowAreaRatio}\tabmarks{}{}
&
\tabularnewline

&
&
& \quad at matched product size (expl.)
& \texttt{med.}
&
&
&
& \tabv{\capRefPlProgInSceneFlareVsGiMatchedPlace}\tabmarks{}{}
& \tabci{\capRefPlProgInSceneFlareVsGiMatchedPlaceLo}{\capRefPlProgInSceneFlareVsGiMatchedPlaceHi}
& \tabv{\capRefPlProgInSceneSunburstVsGiMatchedPlace}\tabmarks{}{}
& \tabci{\capRefPlProgInSceneSunburstVsGiMatchedPlaceLo}{\capRefPlProgInSceneSunburstVsGiMatchedPlaceHi}
\tabularnewline

&
&
&
& \texttt{low}
&
&
&
& \tabv{\capRefPlProgInSceneFlareVsGiLowMatchedPlace}\tabmarks{}{}
& \tabci{\capRefPlProgInSceneFlareVsGiLowMatchedPlaceLo}{\capRefPlProgInSceneFlareVsGiLowMatchedPlaceHi}
& \tabv{\capRefPlProgInSceneSunburstVsGiLowMatchedPlace}\tabmarks{}{}
& \tabci{\capRefPlProgInSceneSunburstVsGiLowMatchedPlaceLo}{\capRefPlProgInSceneSunburstVsGiLowMatchedPlaceHi}
\tabularnewline
\cmidrule[\lightrulewidth]{1-12}
\multirow[t]{5}{\hsize}{\tabms\raggedright \textbf{Sharper detail} $\to$ render legible fine print\par}
& E4
& L1
& OCR-readable lines exact, med.\ vs high (\%)
& \texttt{med.}
& \tabv{\capDetailExactAttGiMedium}
& \tabv{\capDetailExactAttFlareHigh}
& \tabv{\capDetailExactAttSunburstHigh}
& \tabv{\capHolmDetailFlareHighVsGiMedium}\tabmarks{\capHolmDetailFlareHighVsGiMediumHolmSig}{\capHolmDetailFlareHighVsGiMediumHolmSigGlobal}
& \tabci{\capHolmDetailFlareHighVsGiMediumLo}{\capHolmDetailFlareHighVsGiMediumHi}
& \tabv{\capHolmDetailSunburstHighVsGiMedium}\tabmarks{\capHolmDetailSunburstHighVsGiMediumHolmSig}{\capHolmDetailSunburstHighVsGiMediumHolmSigGlobal}
& \tabci{\capHolmDetailSunburstHighVsGiMediumLo}{\capHolmDetailSunburstHighVsGiMediumHi}
\tabularnewline

&
&
& \quad high vs max (n\,=\,\capHolmDetailFlareMaxVsGiHighN--\capHolmDetailSunburstMaxVsGiHighN{} prompts)
& \texttt{high}
& \tabv{\capDetailExactAttGiHigh}
& \tabv{\capDetailExactAttFlareMax}
& \tabv{\capDetailExactAttSunburstMax}
& \tabv{\capHolmDetailFlareMaxVsGiHigh}\tabmarks{\capHolmDetailFlareMaxVsGiHighHolmSig}{\capHolmDetailFlareMaxVsGiHighHolmSigGlobal}
& \tabci{\capHolmDetailFlareMaxVsGiHighLo}{\capHolmDetailFlareMaxVsGiHighHi}
& \tabv{\capHolmDetailSunburstMaxVsGiHigh}\tabmarks{\capHolmDetailSunburstMaxVsGiHighHolmSig}{\capHolmDetailSunburstMaxVsGiHighHolmSigGlobal}
& \tabci{\capHolmDetailSunburstMaxVsGiHighLo}{\capHolmDetailSunburstMaxVsGiHighHi}
\tabularnewline
\bottomrule
\end{tabular*}
\end{table*}

Only local editing shows a qualified improvement on the tested tasks: OCR detects
fewer changes to text surrounding a forged receipt field. The tests do not establish
gains in multi-turn consistency, reference fidelity or fine-print rendering.
Table~\ref{tab:results} reports the estimates; Figure~\ref{fig:answers} shows their
distributions. Differences are paired Images~2.5 minus GPT-Image-2, in percentage
points with 95\% confidence intervals. Unless stated otherwise, $p$-values use the
within-claim correction (\S\ref{sec:method-stats}, Appendix~\ref{app:stats}).

\subsection{Local editing: fewer changes to surrounding receipt text}
\label{sec:cap-local}

\textbf{Surrounding text.} After alignment, OCR flags a change to surrounding text in
\capDocRateFlareLTwoOtherVTwo{} of Flare outputs and
\capDocRateSunburstLTwoOtherVTwo{} of Sunburst outputs, over
\capDocFlareMinusGiLowLTwoOtherVTwoN{} edits. The rates are
\capDocRateGiLowLTwoOtherVTwo{} for GPT-Image-2 \texttt{low} and
\capDocRateGiMedLTwoOtherVTwo{} for \texttt{medium}; these equal rates occur on
different receipts. All four paired differences, from
\capDocFlareMinusGiLowLTwoOtherVTwo{} to \capDocSunburstMinusGiLowLTwoOtherVTwo{},
survive Holm correction ($p\le$\,\capDocFlareMinusGiMedLTwoOtherVTwoPHolm{}).
Exact McNemar tests agree. For comparison, OCR flags changes in
\capDocFloorAnyOtherRegVTwo{} of unedited calibration copies. This is a reference
false-change rate, not a guaranteed floor for generated outputs.

The improvement is concentrated on CORD receipts, with differences of
\capDocCordSunburstMinusGiLowLTwoOtherVTwo{} to
\capDocCordFlareMinusGiMedLTwoOtherVTwo{}. WildReceipt shows no significant difference.
The result also depends on alignment. Restricting displacement to at most 1\,px leaves
similar estimates (\capDocShiftLocLeOneSunburstMinusGiMedLTwoOtherVTwo{} to
\capDocShiftLocLeOneSunburstMinusGiLowLTwoOtherVTwo{}), but only the contrasts against
\texttt{low} are significant before correction. Without alignment, none is significant.
Under the stricter all-test correction, Flare versus \texttt{medium} also loses
significance ($p$\,=\,\capDocFlareMinusGiMedLTwoOtherVTwoPHolmGlobal{}).
The analysis covers only API-compatible receipt shapes, and the final endpoint lacks
independent validation (Appendices~\ref{app:e1shift}, \ref{app:e1ocr} and
\ref{app:pipeline}).

\textbf{Forged-field correctness.} On \capDocFlareMinusGiLowLOneTargetVTwoN{}
receipts with OCR-readable source fields, exact forged-value rates range from
\capDocRateGiLowLOneTargetVTwo{} to \capDocRateSunburstLOneTargetVTwo{}.
No model contrast is significant. The largest is Sunburst versus \texttt{low}:
\capDocSunburstMinusGiLowLOneTargetVTwo{}
[\capDocSunburstMinusGiLowLOneTargetVTwoLo{}, \capDocSunburstMinusGiLowLOneTargetVTwoHi{}],
with $p$\,=\,\capDocSunburstMinusGiLowLOneTargetVTwoPHolm{}.
Minimum detectable differences are \capDocMdeLOneMinVTwo{}--\capDocMdeLOneMaxVTwo{}\,pp
at \capDocMdeLOnePowerVTwo{} power, so smaller gains remain possible.
The evidence thus supports a limited preservation gain, not greater target-field accuracy
or demonstrated success against human review.

\subsection{Multi-turn editing: lower receipt consistency for Flare}
\label{sec:cap-multiturn}

\textbf{Receipt sequences (E2a).} After \capDocChainTurns{} edits, Flare retains fewer
of its successful CORD edits than either GPT-Image-2 baseline. Final-turn survival is
\capDocChainCordSurvFourFlareVTwo{} for Flare, compared with
\capDocChainCordSurvFourGiLowVTwo{} for \texttt{low} and
\capDocChainCordSurvFourGiMedVTwo{} for \texttt{medium}.
The paired differences are \capDocChainCordSurvFourDiffFlareVsGiLowVTwo{}
[\capDocChainCordSurvFourDiffFlareVsGiLowVTwoLo{}, \capDocChainCordSurvFourDiffFlareVsGiLowVTwoHi{}]
and \capDocChainCordSurvFourDiffFlareVsGiMedVTwo{}
[\capDocChainCordSurvFourDiffFlareVsGiMedVTwoLo{}, \capDocChainCordSurvFourDiffFlareVsGiMedVTwoHi{}].
Both survive within-claim Holm correction
($p$\,=\,\capDocChainCordSurvFourDiffFlareVsGiLowVTwoPHolm{} and
\capDocChainCordSurvFourDiffFlareVsGiMedVTwoPHolm{}) and the stricter all-test correction.

The joint endpoint gives the same conclusion without conditioning on successful edits.
The fraction of all requested edits that both succeed and survive is lower for Flare by
\capDocChainCordJointDiffFlareVsGiLowVTwo{} and
\capDocChainCordJointDiffFlareVsGiMedVTwo{} against the two baselines
(Holm $p$\,=\,\capDocChainCordJointDiffFlareVsGiLowVTwoPHolm{} and
\capDocChainCordJointDiffFlareVsGiMedVTwoPHolm{}).
Flare's first-time success rate and preservation of never-edited fields show no
significant difference (Holm $p$\,=\,\capDocChainCordSuccDiffFlareVsGiLowVTwoPHolm{}).
Sunburst differs from neither baseline, and WildReceipt shows no significant differences
(Table~\ref{tab:results-appendix}). GPT-Image-2 \texttt{medium} shrinks receipts slightly
with each turn, making OCR harder; this can understate Flare's survival deficit against
that baseline (Appendix~\ref{app:e2}).

\textbf{Photo sequences (E2b).} All models retain nearly all tested edits, leaving
little separation. Final-turn survival is \capChainStatelessFlareLOneSurvivalVTwo{}
for Flare and \capChainStatelessSunburstLOneSurvivalVTwo{} for Sunburst, compared with
\capChainStatelessGiLowLOneSurvivalVTwo{} and \capChainStatelessGiMedLOneSurvivalVTwo{}
for GPT-Image-2 \texttt{low} and \texttt{medium}. Surviving edits remain in the requested
position for every model. No difference is significant (all Holm
$p$\,=\,\capChainStatelessFlareMinusGiLowLOneSurvivalVTwoPHolm{}).

The chat-session variant, where \capChainConvTextModel{} relays instructions, was
scored before the analysis rules were fixed and is exploratory. Images~2.5 retains
more edits in place in that variant, but the contrast would remain nonsignificant
if included in the primary family (Holm
$p$\,=\,\capChainConvSunburstMinusGiMedLTwoInPlaceVTwoPHolmInclExploratory{};
Appendix~\ref{app:e2}). Overall, these tests show a Flare deficit on CORD receipts
and cannot distinguish the models on the near-saturated photo task.

\subsection{Reference fidelity: greater legibility accompanies larger products}
\label{sec:cap-reference}

\textbf{Programmatic references.} In cluttered scenes, exact SKU reads are more than
twice as frequent for Images~2.5. Rates over \capRefPNRefs{} references are
\capRefPInSceneSkuExactFlare{} for Flare and \capRefPInSceneSkuExactSunburst{} for
Sunburst, against \capRefPInSceneSkuExactGiLowRate{} and \capRefPInSceneSkuExactGi{}
for GPT-Image-2 \texttt{low} and \texttt{medium}. All four differences survive both
Holm corrections. This improves code legibility in a fake listing, but does not by
itself establish more faithful copying.

Images~2.5 also draws the product larger. Despite a prompt requesting a product about
a tenth of the frame width, label areas occupy \capRefPlProgInSceneAreaPctFlare{} and
\capRefPlProgInSceneAreaPctSunburst{} of the frame, compared with
\capRefPlProgInSceneAreaPctGiLow{} and \capRefPlProgInSceneAreaPctGi{} for the baselines.
Glyph size relative to the label is similar across models. Product size is chosen by
the model, so matching on it cannot isolate a fidelity effect. Those comparisons are
exploratory and establish no such gain; logo-shape overlap is mixed
(Appendix~\ref{app:e3}).

\textbf{Rendered references.} An exploratory check on \capRefRNRefs{} physically
rendered products reproduces the legibility and placement pattern. Exact SKU rates
are \capRefRInSceneSkuExactFlare{} and \capRefRInSceneSkuExactSunburst{} for Images~2.5,
versus \capRefRInSceneSkuExactGiLow{} and \capRefRInSceneSkuExactGi{} for the baselines.
Against \texttt{low}, the differences are \capRefPlRendInSceneFlareVsGiLowSku{} and
\capRefPlRendInSceneSunburstVsGiLowSku{} (Holm $p$ within this exploratory family:
\capRefPlRendInSceneFlareVsGiLowSkuPHolmPrimary{}).
Label areas are again larger: \capRefPlRendInSceneAreaPctFlare{} and
\capRefPlRendInSceneAreaPctSunburst{}, versus \capRefPlRendInSceneAreaPctGiLow{} and
\capRefPlRendInSceneAreaPctGi{}.

The size-controlled residual does not replicate consistently. Against \texttt{low},
both matched-placement intervals include zero:
\capRefPlRendInSceneFlareVsGiLowMatchedPlace{}
[\capRefPlRendInSceneFlareVsGiLowMatchedPlaceLo{}, \capRefPlRendInSceneFlareVsGiLowMatchedPlaceHi{}]
and \capRefPlRendInSceneSunburstVsGiLowMatchedPlace{}
[\capRefPlRendInSceneSunburstVsGiLowMatchedPlaceLo{}, \capRefPlRendInSceneSunburstVsGiLowMatchedPlaceHi{}].
Against \texttt{medium}, gains remain at matched glyph height:
\capRefPlRendInSceneFlareVsGiMatchedAbs{}
[\capRefPlRendInSceneFlareVsGiMatchedAbsLo{}, \capRefPlRendInSceneFlareVsGiMatchedAbsHi{}]
and \capRefPlRendInSceneSunburstVsGiMatchedAbs{}
[\capRefPlRendInSceneSunburstVsGiMatchedAbsLo{}, \capRefPlRendInSceneSunburstVsGiMatchedAbsHi{}].
Because the residual appears against different baselines in the two reference sets,
it does not establish improved fidelity. Sunburst's rendered-reference runs were all
synchronous, a further execution difference (Appendix~\ref{app:design}).

\subsection{Fine print: no detectable gain above the OCR limit}
\label{sec:cap-detail}

OCR reads the clean calibration text reliably only above
\capDetailCeilCapPx{}\,px cap height. That threshold lies above the steep decline in
legibility, so the primary test covers only the flatter upper part of the curve
(Appendix~\ref{app:e4}). Scores there still range from
\capDetailBandExactAboveCeilSunburstMax{} to \capDetailBandExactAboveCeilFlareMax{},
leaving room for measurable improvement. No primary contrast is significant.
The closest is Flare \texttt{max} versus GPT-Image-2 \texttt{high}:
\capHolmDetailFlareMaxVsGiHigh{}
[\capHolmDetailFlareMaxVsGiHighLo{}, \capHolmDetailFlareMaxVsGiHighHi{}] more exact lines,
with Holm $p$\,=\,\capHolmDetailFlareMaxVsGiHighPHolm{}
(\capHolmDetailFlareMaxVsGiHighPHolmGlobal{} under the all-test correction).

Eligibility differs between models. This comparison excludes
\capDetailVtwoDropFlareMaxVsGiHighN{} prompts, including
\capDetailVtwoDropFlareMaxVsGiHighCtl{} on the GPT-Image-2 side. It also scores different
lines: \capDetailBelowCeilGiHigh{} of GPT-Image-2 \texttt{high} lines are below the
threshold, versus \capDetailBelowCeilFlareMax{} for Flare. Counting such lines as
failures gives \capDetailVtwoIttFlareMaxVsGiHigh{}, but combines generator performance
with the OCR instrument's failures.

Exploratory within-model tests find no higher tier better than \texttt{low} on
OCR-readable text (\capDetailVtwoTierAttPermNReject{} significant comparisons out of
\capDetailVtwoTierM{}). A height-by-height scan likewise identifies no band with better
reads at a higher tier (Appendix~\ref{app:e4}). These results establish no measurable
gain above the OCR threshold. Performance at the sizes where fine print becomes
illegible remains unresolved by this instrument.

\section{Detection, Localisation and Refusal}
\label{sec:defence}

\begin{figure*}[tbp]
  \centering
  \includegraphics[width=\textwidth]{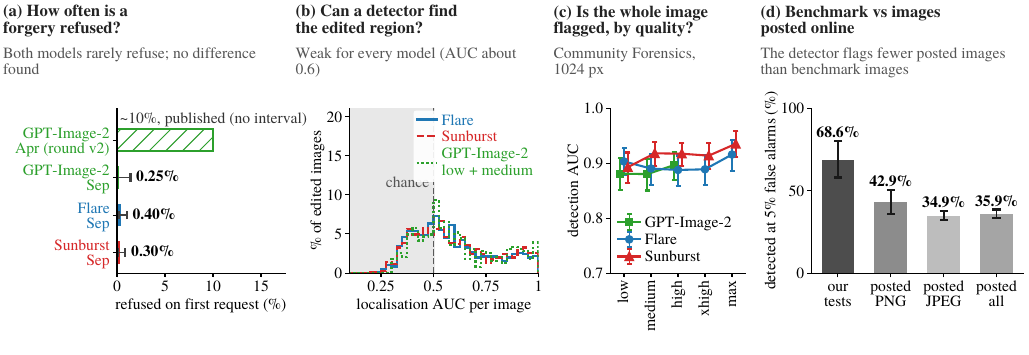}
  \caption{\textbf{Defence outcomes.} Panel subtitles summarise
  Table~\ref{tab:defence-results}.
  \textbf{(a)}~First-pass refusal rates with Wilson 95\% intervals. The hatched band
  is the earlier GPT-Image-2 result \cite{wu2026forgerjudge}.
  \textbf{(b)}~DocTamper's per-image localisation AUC; chance is 0.5.
  \textbf{(c)}~Community Forensics whole-image AUC against real photographs, by
  quality setting.
  \textbf{(d)}~Flag rates at a \wildXfpr{} false-positive rate. Controlled cells show
  their mean and min--max range; posted images show a Wilson 95\% interval.}
  \label{fig:defence}
\end{figure*}

We evaluate refusal, localisation and whole-image detection, with contemporaneous
GPT-Image-2 baselines (Figure~\ref{fig:defence}, Table~\ref{tab:defence-results}).
These comparisons show how outcomes differ between the tested configurations; they
do not identify the causes of those differences.

\subsection{Refusal remains rare in both generations}
\label{sec:defence-refusal}

Images~2.5 refused \docsRefused{} of \docsAttempts{} masked receipt edits on first
pass: \docsRefusalPct{} [\docsRefusalWilsonLo{}, \docsRefusalWilsonHi{}].
With the same requests, prompts and alpha masks, GPT-Image-2 refused
\docsControlRefused{} of \docsControlAttempts{} (\docsControlPct{}).
Both rates are far below the \vTwoRefusal{} reported for GPT-Image-2 in our earlier
study \cite{wu2026forgerjudge}. The low contemporaneous rates for both generations
do not support attributing the historical decline to Images~2.5 alone.

On the \refPairNShared{} specifications sent to every model, refusal counts are
\refPairFlareRefused{} for Flare, \refPairSunRefused{} for Sunburst,
\refPairGitwoLowRefused{} for GPT-Image-2 \texttt{low} and
\refPairGitwoMedRefused{} for \texttt{medium}. These sparse events do not distinguish
the models: every $p$\,=\,\refPairFlareVsGitwoLowPMcNemar{}.
A detectable difference would require at least \refPairMinDiscordant{}
specifications refused by only one model. Changes in request format confound the
historical comparison, alongside the change in date (Appendix~\ref{app:refusals}).

\subsection{Localisation remains weak}
\label{sec:defence-localisation}

DocTamper's mean per-image AUC is \detDtFlarePixAuc{}
[\detDtFlarePixAucLo{}, \detDtFlarePixAucHi{}] for Flare and
\detDtSunPixAuc{} [\detDtSunPixAucLo{}, \detDtSunPixAucHi{}] for Sunburst.
The GPT-Image-2 AUCs are \detDtGitwoLowPixAuc{} at \texttt{low} and
\detDtGitwoMedPixAuc{} at \texttt{medium}. All are around 0.6, with intervals above
the chance value of 0.5 (Appendix~\ref{app:equivalence}).

Flare is slightly harder to localise than GPT-Image-2 \texttt{medium}, with a paired
AUC difference of \detPairDtFlareMinusGitwoMedDelta{}
[\detPairDtFlareMinusGitwoMedLo{}, \detPairDtFlareMinusGitwoMedHi{}].
All model-to-model differences are within
$\pm$\eqDtPairFlareMinusGitwoMedMinMargin{} AUC; Appendix~\ref{app:equivalence}
gives the margins and tests.

These results apply to a crop-edit-and-paste-back pipeline. Editing a whole page
would not introduce the same crop boundary. TruFor's higher AUCs
(\detTfPixAucLo{}--\detTfPixAucHi{}) respond to that boundary rather than the edit,
and its successful runs cover only pages up to \tfCovFlareMaxScoredLongEdge{}\,px
on the long edge (Appendix~\ref{app:seam}). Community Forensics provides no
localisation map. Appendices~\ref{app:progression} and \ref{app:chains} report
cross-release localisation and detection over eight re-editing rounds, respectively.

\subsection{Speed depends on the baseline}
\label{sec:defence-price}

Flare generated a receipt edit \latForgeDocsGitwoMedOverFlareRatio{}$\times$
faster than GPT-Image-2 \texttt{medium}
[\latForgeDocsGitwoMedOverFlareLo{}, \latForgeDocsGitwoMedOverFlareHi{}].
Against \texttt{low}, the ratio was \latForgeDocsGitwoLowOverFlareRatio{}$\times$
[\latForgeDocsGitwoLowOverFlareLo{}, \latForgeDocsGitwoLowOverFlareHi{}].
Sunburst was about as fast as \texttt{low}
(\latForgeDocsGitwoLowOverSunRatio{}$\times$).
These are descriptive timings from one account over one week. Runs were not
interleaved, and only synchronous calls provide usable latency measurements
(Appendix~\ref{app:latency}).

\subsection{Detection rates fall on images posted online}
\label{sec:defence-wild}

Community Forensics shows no clear detectability trend across the tested 1024-px
quality settings: AUC ranges from \tierAucFlatLo{} to \tierAucFlatHi{}.
Equivalence across those settings is not established (Appendix~\ref{app:tiers}).

At a \wildXfpr{} false-positive rate calibrated on \wildXAuthN{} authentic images,
the detector flags an average of \wildXCtrlMean{} across \wildXCells{} controlled
Images~2.5 cells. This is an unweighted cell mean, with a 95\% bootstrap interval of
\wildXCtrlMeanBootLo{}--\wildXCtrlMeanBootHi{} and a cell range of
\wildXCtrlLo{}--\wildXCtrlHi{}.

On \wildXWildN{} posted images attributed by their authors to Images~2.5, the flag
rate is \wildXWildRate{} (Wilson 95\% interval:
\wildXWildRateLo{}--\wildXWildRateHi{}). The lower rate is an observational
comparison, not an isolated generator effect. Attribution rests on self-report:
the sampled delivery paths strip C2PA manifests, although some other platforms
preserve them \cite{paperwild2026}.

The posted-image intervals assume independent images and a fixed threshold.
They neither cluster posts by author nor propagate threshold uncertainty.
Posted PNGs have a flag rate of \wildXPngRate{}
(\wildXPngRateLo{}--\wildXPngRateHi{}), versus \wildXJpgRate{}
(\wildXJpgRateLo{}--\wildXJpgRateHi{}) for JPEGs. These format groups are not
content-matched (Appendix~\ref{app:wild}).

\section{Discussion}
\label{sec:discussion}

\subsection{Capability gains and defence outcomes differ}
\label{sec:disc-decoupling}

The four release claims translate unevenly to the tested forgery tasks.
The clearest legibility gain, in product codes, accompanies larger product placement;
it does not establish more faithful copying. Receipt editing provides a different
result: automatic checks detect fewer changes to surrounding text after alignment.
That preservation endpoint is directly relevant to comparison with an original,
whereas public preference votes do not measure it.

The receipt result remains narrow. It is concentrated on CORD, depends on
registration and token selection, and lacks independent validation of the final
endpoint. It therefore supports an instrument-defined preservation gain, not a claim
that human reviewers would accept the forgeries.

Defence limitations persist across both same-week generations. Refusals are rare,
and DocTamper localisation is weak despite a small Flare-versus-\texttt{medium}
difference. Community Forensics shows no clear trend across 1024-px quality tiers;
4K cells score lower for both Images~2.5 models, but equivalence across settings is
not established. Historical refusal differences also confound request format,
routing and sampling. Separate capability and defence measurements cannot establish
whether better editing causes a change in detectability.

The detector flags fewer posted images than controlled images at the evaluated
threshold. Content, format and provenance all differ between these samples, so the
gap cannot be attributed to any one factor. This result concerns whole-image
detection and does not establish transfer of document-localisation performance.

\subsection{Practical implications}
\label{sec:disc-actions}

\textbf{Fraud and trust-and-safety teams.} Provider refusal did not reliably block
the tested receipt edits. Forensic maps also require care: DocTamper weakly separates
edited pixels, while TruFor can respond to a pasted-crop boundary. Independently
verifiable information---issuer records, line-item totals and earlier submissions---
may complement image-based checks. Our outputs still contain incorrect values,
misplaced edits and lost earlier edits, but we did not test the operational
performance of those checks. Detection should be validated on representative
traffic, and a higher generation tier should not be assumed to evade the detector.

\textbf{Model vendors.} Release evaluation should include tasks with predetermined
answers alongside preference ratings. Comparisons should use matched inputs and
report both quality settings and their costs. Refusal rates need named misuse
categories and contemporaneous controls. Provenance metadata should identify the
generator: the three tested models emit the same C2PA generator string, preventing
that field from distinguishing Images~2.5 from GPT-Image-2
(Appendix~\ref{app:c2pa}). Appendix~\ref{app:limitations} gives the remaining design
and measurement limitations.

\section{Conclusion}
\label{sec:conclusion}

We tested four improvements advertised for ChatGPT Images~2.5 on forgery tasks
with predetermined answers. The evidence supports one qualified gain: after
alignment, Flare and Sunburst disturb less surrounding receipt text than the
GPT-Image-2 baselines. This result is concentrated on CORD and lacks independent
validation of the final preservation endpoint.

The remaining tasks establish no general improvement. Product-code legibility
increases with larger placement, without a replicated size-controlled fidelity gain.
Flare retains fewer earlier edits on CORD receipts, while the photo task scarcely
separates the models. Fine print shows no measurable gain above the OCR threshold;
performance below it remains unresolved. Flare is
\latForgeDocsGitwoMedOverFlareRatio$\times$ faster than GPT-Image-2 \texttt{medium}
for a receipt edit, but only \latForgeDocsGitwoLowOverFlareRatio$\times$ faster than
\texttt{low}.

Both generations rarely refuse the tested requests, and localisation remains weak.
Detection equivalence across quality tiers is not established. At the evaluated
threshold, Community Forensics flags \wildXWildRate{} of self-reported posted images,
versus a \wildXCtrlMean{} mean across controlled cells. These findings support
task-specific evaluation of release claims and validation of defences on
representative traffic, with independent factual checks where available.

\section*{Data and Code Availability}
Supporting code and the non-restricted resources for this work are available from the Scam.AI
Research portal, \url{https://scam.ai/research}~\cite{scamai_research}, with document- and
PDF-forgery resources maintained at Trustable, \url{https://trustable.ai}~\cite{trustable_ai}.
As noted in the paper, the raw generated images used in this study are not publicly released; requests
and further details are handled through the Scam.AI Research portal.

{\small
\bibliographystyle{plainurl}
\bibliography{references}
}

\appendix

\section{Models, Pricing and API Setup}
\label{app:models}

This appendix lists the models and quality tiers we tested, what each costs, and how requests
were routed and logged.

We ran a smoke grid of \smokePerCell{} images per (model, tier) cell at 1024$\times$1024 through the direct
OpenAI API and read the usage block the API returns.

\paragraph{Tiers and output tokens.}
In the smoke grid, output-token counts are identical across models at matched
1024$\times$1024 tiers: \tokLow{} (\texttt{low}), \tokMed{}
(\texttt{medium}), \tokHigh{} (\texttt{high}), \tokXhigh{}
(\texttt{xhigh}) and \tokMax{} (\texttt{max}). The \nOkSmoke{}
successful calls show no variation within a tier. GPT-Image-2's three tiers
match Images~2.5's \texttt{low}, \texttt{high} and \texttt{max} in output-token
count. Thus GPT-Image-2 \texttt{medium} matches Images~2.5 \texttt{high} on
this billing measure. The \nRefusedSmoke{} unsuccessful calls requested
unsupported GPT-Image-2 tiers (Appendix~\ref{app:design}). Costs computed from
returned usage and published rates match across models at each tier
(Table~\ref{tab:tiers}).

\paragraph{Smoke-grid latency and 4K cost.}
At matched tokens, Flare is \latRatioHigh{}--\latRatioMax{}$\times$ faster
than GPT-Image-2 in the smoke grid. Sunburst is
\latRatioSunLo{}--\latRatioSunHi{}$\times$ slower than Flare across
\latRatioSunTiers{} shared tiers. These are smoke-grid means; the main speed
estimates use forgery-task ratios (\S\ref{sec:defence-price},
Appendix~\ref{app:latency}). The 3840$\times$2160 output at \texttt{high}
costs about \capPriceFourKOverHigh{}$\times$ the 1024-px \texttt{high} price
on both families.

\begin{table*}[tbp]
  \centering
  \caption{Measured tier grid at 1024$\times$1024 (\smokePerCell{} calls per cell, direct OpenAI API), with the
  4K rows at their billed token counts. Output tokens are deterministic per tier and identical across models at
  1024 px, so one row is one price and GPT-Image-2's tiers are printed on the rows whose token count they match.
  1024-px latency is the smoke-grid mean $\pm$ SD; 4K latency is the mean over the photoreal-core 4K cells. The
  4K rows bill about \capPriceFourKOverHigh{}$\times$ the 1024 \texttt{high} price of the same family.}
  \label{tab:tiers}
  \footnotesize
  \begin{tabular}{rrllrrr}
    \toprule
    Out.\ tokens & \$/image & Images 2.5 tier & GPT-Image-2 tier & Flare s & Sunburst s & GPT-Image-2 s \\
    \midrule
    \tokLow{}   & \usdLow{}   & \texttt{low}    & \texttt{low}    & $\latFlareLow{} \pm \latSdFlareLow{}$     & $\latSunLow{} \pm \latSdSunLow{}$     & $\latGitwoLow{} \pm \latSdGitwoLow{}$   \\
    \tokMed{}   & \usdMed{}   & \texttt{medium} & ---             & $\latFlareMed{} \pm \latSdFlareMed{}$     & $\latSunMed{} \pm \latSdSunMed{}$     & ---                                     \\
    \tokHigh{}  & \usdHigh{}  & \texttt{high}   & \texttt{medium} & $\latFlareHigh{} \pm \latSdFlareHigh{}$   & $\latSunHigh{} \pm \latSdSunHigh{}$   & $\latGitwoMed{} \pm \latSdGitwoMed{}$   \\
    \tokXhigh{} & \usdXhigh{} & \texttt{xhigh}  & ---             & $\latFlareXhigh{} \pm \latSdFlareXhigh{}$ & $\latSunXhigh{} \pm \latSdSunXhigh{}$ & ---                                     \\
    \tokMax{}   & \usdMax{}   & \texttt{max}    & \texttt{high}   & $\latFlareMax{} \pm \latSdFlareMax{}$     & $\latSunMax{} \pm \latSdSunMax{}$     & $\latGitwoHigh{} \pm \latSdGitwoHigh{}$ \\
    \midrule
    \capPriceFourKTokFlare{}  & \capPriceFourKUsdFlare{} & 4K \texttt{high} & --- & \capPriceFourKLatFlare{} & \capPriceFourKLatSun{} & --- \\
    \capPriceFourKTokGitwo{}  & \capPriceFourKUsdGitwo{} & --- & 4K \texttt{high} & --- & --- & \capPriceFourKLatGitwo{} \\
    \bottomrule
  \end{tabular}
\end{table*}

\paragraph{The two baselines and what they cost.} E1--E3 run Images~2.5 at \texttt{medium} (\tokMed{} output
tokens), a price at which GPT-Image-2 has no tier, and compare it with GPT-Image-2 \texttt{low} (\tokLow{}) and
\texttt{medium} (\tokHigh{}): fewer than, and four times as many as, the Images~2.5 configurations, so the two
baselines bracket the Images~2.5 price rather than matching it. E4 and the paired timings compare tiers at equal
billed tokens (\S\ref{sec:method-setup}).

\paragraph{Route and call log.} Calls use the direct OpenAI API from one account; cost is read from the returned
usage, and \texttt{input\_fidelity} is accepted for no model. Every call, successful, refused or failed, appends one
row to an append-only ledger. Full E2--E4 runs used the Batch API with synchronous fill
(Table~\ref{tab:cells}), so latency uses synchronous rows only. Counts come from one snapshot (\ledgerLines{} rows,
SHA-256 \texttt{\ledgerSha{}}, \ledgerSnapshotDate{}; Appendix~\ref{app:design}).

\paragraph{The \texttt{input\_fidelity} parameter is not available.}
\label{app:inputfidelity}

The image-editing documentation associates reference preservation with
\texttt{input\_fidelity}. A direct request to \texttt{gpt-image-2.5-flare}
with \texttt{input\_fidelity="high"} returned HTTP 400 and
\texttt{invalid\_input\_fidelity}. The other two models have the same
configuration, but their behaviour is assumed rather than separately verified.

Cell builders pass the argument; the client removes it before sending a request
when \texttt{models.yaml} marks the model as not supporting it. Every edit
therefore uses default reference preservation. These results describe that
configuration, not an upper bound on reference fidelity.

\paragraph{Vendor-reported safety and third-party preference.}
OpenAI's system card \cite{openai2026images25card} reports unsafe-presented
rates of \scUnsafeSun{} for Sunburst, \scUnsafeFlare{} for Flare and
\scUnsafeGitwo{} for GPT-Image-2. Blocked rates are \scBlockedSun{},
\scBlockedFlare{} and \scBlockedGitwo{}, respectively. The evaluation does
not measure document fraud.

Our recorded \arenaDate{} arena snapshot \cite{arena2026snapshot} places
both new models above GPT-Image-2 on editing Elo: Sunburst \eloEditSun{}
($\pm$\eloCiEditSun{}, \votesEditSun{} votes), Flare \eloEditFlare{}
($\pm$\eloCiEditFlare{}, \votesEditFlare{} votes), and GPT-Image-2
\eloEditGitwo{} ($\pm$\eloCiEditGitwo{}, \votesEditGitwo{} votes).
The editing lead is \eloEditGain{} points, compared with \eloTtoiGain{}
on text-to-image.

These historical values are author-recorded transcriptions, supplied in
\texttt{anc/arena-snapshot.json}; the original page capture is unavailable.
The new models have fewer votes and wider intervals. Neither preference ranks
nor unsafe-output rates measure the forgery outcomes in
Figure~\ref{fig:known_answer}.

\section{Benchmark Design and Call Log}
\label{app:design}

This appendix describes how requests were issued, retried and logged.

\paragraph{Refusal bookkeeping.}
We record a request as \texttt{refused} when three consecutive attempts return
HTTP 400 with the same provider error code. This follows the round-v2 rule for
separating content rejection from outages \cite{wu2026forgerjudge}.
Transient failures are \texttt{error} rows. The \nRefusedSmoke{}
\texttt{invalid\_value} smoke rows concern unsupported GPT-Image-2 tiers;
they are content-neutral, unbilled and excluded from refusal statistics.

First-pass refusal uses the request's earliest row; persistent refusal uses
its latest row. We order rows by \texttt{created\_at}, which records row
construction rather than call completion (\ledgerOutOfOrder{} pairs are
out of order). The \texttt{row\_id} hashes request identity, excluding prompt
text and mask bytes. Re-runs skip requests already recorded as \texttt{ok}.
Refusals remain explicit ledger rows rather than missing observations.

\paragraph{Data flow.}
Cell builders emit jobs to a common retry, refusal and budget runner. Each call
produces a ledger row, with a raw PNG for successful outputs. Exporters pass
cells to the detector harness using \texttt{row\_id}; checkers join the results
to source images and known answers. Downstream stages read only files named in
the ledger.

\paragraph{Cells, routes and budget.} Table~\ref{tab:cells} lists the cells. The benchmark was capped at
\$\hardStop{}, enforced in code; at the ledger snapshot \$\ledgerSpend{} had been spent. Pilots, E1 and the
photoreal core ran synchronously. The full E2--E4 runs used the Batch API, billed at half the list price, with
synchronous fill where a batch did not complete (Table~\ref{tab:cells}).

E2b's conversational variant ran
synchronously throughout, and Flare's stateless E2b rows were filled mostly synchronously
(\routeArmBatchPhotoChainFlare{} batch against \routeArmSyncPhotoChainFlare{} synchronous, where GPT-Image-2
\texttt{medium} had \routeArmBatchPhotoChainGitwoMed{} and \routeArmSyncPhotoChainGitwoMed{}, GPT-Image-2
\texttt{low} \routeArmBatchPhotoChainGitwoLow{} and \routeArmSyncPhotoChainGitwoLow{}, and Sunburst
\routeArmBatchPhotoChainSun{} and \routeArmSyncPhotoChainSun{}).

Sunburst's rendered E3 rows all ran
synchronously (\routeArmSyncRefRSun{} of \routeArmOkRefRSun{}), so the rendered set has no batch-only check for
Sunburst. Batch rows carry no usable generation latency, so
every latency in this paper comes from synchronous rows.

\begin{table*}[tbp]
  \centering
  \caption{Benchmark cells. E1 counts are distinct requests by latest status (the documents rows print
  persistent refusals; first-pass counts are in \S\ref{sec:defence-refusal}). E2--E4 give the analysed units; the
  E2b refusals are chain turns blocked by moderation, all on one seed. Route columns: accepted ledger rows sent
  through the Batch API / synchronously, per model and configuration; the GPT-Image-2 \texttt{low} configuration is
  the baseline added after review, which ran about 11 hours after the others. E4 Images~2.5 counts pool the five
  quality tiers; its GPT-Image-2 column is the \texttt{high} baseline (the \texttt{medium} and \texttt{low} tiers ran
  \routeArmBatchDetailGitwoMed{}/\routeArmSyncDetailGitwoMed{} and
  \routeArmBatchDetailGitwoLow{}/\routeArmSyncDetailGitwoLow{}).}
  \label{tab:cells}
  \footnotesize
  \setlength{\tabcolsep}{3pt}
  \begin{tabular}{llrrrrrr}
    \toprule
    & & & & \multicolumn{4}{c}{Batch/sync rows} \\
    \cmidrule(lr){5-8}
    Cell & Claim, layer & Units & Ref. & Flare & Sunburst & GPT-Image-2 & GPT-Image-2 \texttt{low} \\
    \midrule
    smoke                     & tiers, cost, latency   & \nOkSmoke{} calls            & --- & \multicolumn{4}{c}{sync} \\
    E1 documents              & local editing, L1--L3  & \nOkDocs{} edits             & \nRefusedDocs{} & \multicolumn{4}{c}{sync} \\
    \quad GPT-Image-2 control & E1 control, L1--L3     & \nOkDocsControl{} edits      & \nRefusedDocsControl{} & \multicolumn{4}{c}{sync} \\
    E2a receipt chains        & multi-turn, L1--L2     & \capDocChainAllReceipts{} receipts & \capDocChainAllRefusedFlare{}
      & \routeArmBatchDocChainFlare{}/\routeArmSyncDocChainFlare{} & \routeArmBatchDocChainSun{}/\routeArmSyncDocChainSun{}
      & \routeArmBatchDocChainGitwoMed{}/\routeArmSyncDocChainGitwoMed{} & \routeArmBatchDocChainGitwoLow{}/\routeArmSyncDocChainGitwoLow{} \\
    E2b photo chains, stateless & multi-turn, L1--L2   & \capChainStatelessNChains{} chains & \capChainNRefusedChainTurns{}
      & \routeArmBatchPhotoChainFlare{}/\routeArmSyncPhotoChainFlare{} & \routeArmBatchPhotoChainSun{}/\routeArmSyncPhotoChainSun{}
      & \routeArmBatchPhotoChainGitwoMed{}/\routeArmSyncPhotoChainGitwoMed{} & \routeArmBatchPhotoChainGitwoLow{}/\routeArmSyncPhotoChainGitwoLow{} \\
    \quad conversational      & multi-turn (exploratory) & \capChainConvNChains{} chains & ---
      & \routeArmBatchPhotoChainConvFlare{}/\routeArmSyncPhotoChainConvFlare{} & \routeArmBatchPhotoChainConvSun{}/\routeArmSyncPhotoChainConvSun{}
      & \routeArmBatchPhotoChainConvGitwoMed{}/\routeArmSyncPhotoChainConvGitwoMed{} & --- \\
    E3 packshots, programmatic & reference, L1--L2     & \capRefPNCalls{} calls        & ---
      & \routeArmBatchRefPFlare{}/\routeArmSyncRefPFlare{} & \routeArmBatchRefPSun{}/\routeArmSyncRefPSun{}
      & \routeArmBatchRefPGitwoMed{}/\routeArmSyncRefPGitwoMed{} & \routeArmBatchRefPGitwoLow{}/\routeArmSyncRefPGitwoLow{} \\
    \quad rendered            & reference (exploratory) & \capRefRNCalls{} calls       & ---
      & \routeArmBatchRefRFlare{}/\routeArmSyncRefRFlare{} & \routeArmBatchRefRSun{}/\routeArmSyncRefRSun{}
      & \routeArmBatchRefRGitwoMed{}/\routeArmSyncRefRGitwoMed{} & \routeArmBatchRefRGitwoLow{}/\routeArmSyncRefRGitwoLow{} \\
    E4 fine print             & detail, L1, L3         & \capDetailNImages{} images    & \capDetailFailedCalls{}
      & \routeArmBatchDetailFlare{}/\routeArmSyncDetailFlare{} & \routeArmBatchDetailSun{}/\routeArmSyncDetailSun{}
      & \routeArmBatchDetailGitwoHigh{}/\routeArmSyncDetailGitwoHigh{} & --- \\
    photoreal core            & price, L3              & \nOkPhotoreal{} images        & --- & \multicolumn{4}{c}{sync} \\
    laundering chains         & repeated editing, L3   & \nChainsRand{} rows           & --- & \multicolumn{4}{c}{sync} \\
    \bottomrule
  \end{tabular}
\end{table*}

\section{Statistics and Analysis Timeline}
\label{app:stats}

\paragraph{Intervals and paired tests.}
Unless otherwise stated, intervals are 95\% percentile bootstrap intervals
from \capHolmNBoot{} resamples. The cluster unit is the specification for
E1, receipt for E2a, source photo for E2b, reference for E3 and prompt for E4
(\S\ref{sec:method-stats}). Detector AUCs are clustered by specification
(\dtCiConvention{}); refusal rates use Wilson 95\% intervals.

Primary contrasts use cluster sign-flip permutation tests with
\capHolmNPerm{} permutations and a minimum attainable $p$ of
\capHolmPResolution{}. The null assumes exchangeability of the two arms
within paired clusters, rather than simply equal marginal rates. Paired binary
E1 endpoints also use exact McNemar tests.

\paragraph{Multiple comparisons.}
Holm correction applies separately to four primary claim families:
\begin{itemize}[leftmargin=*]
  \item \emph{Local editing:} \capHolmNTestsLocalEditing{} tests of target
  read and other text changed, for both models against both controls.
  \item \emph{Multi-turn editing:} \capHolmNTestsMultiTurn{} tests covering
  E2a success, survival, joint outcomes and never-edited fields on both corpora,
  plus E2b stateless survival and in-place outcomes. Both models are compared
  with \texttt{medium} and \texttt{low}.
  \item \emph{Reference fidelity:} \capHolmNTestsReference{} tests of
  in-scene SKU exactness against both controls.
  \item \emph{Sharper detail:} \capHolmNTestsDetail{} tests of exact lines
  above the OCR ceiling at \tokHigh{} and \tokMax{} tokens.
\end{itemize}
A global Holm correction over all \capHolmNTestsGlobal{} primary tests is
reported as sensitivity; \capHolmNRejectGlobal{} tests survive it.

\paragraph{Exploratory analyses and interpretation.}
Exploratory analyses do not determine the primary conclusions. E3 analyses of
both reference sets and E4 sensitivity analyses use Holm correction in
\capHolmNExploratoryFamilies{} separate families. These cover size matching,
logos, intention-to-treat, matched height and tier contrasts; E4's allowlisted
OCR variant is counted separately. Shift-stratified, corpus-split and other
exploratory analyses are otherwise unadjusted where not explicitly stated.

E2b conversational tests are outside every primary family. Their Holm value in
\S\ref{sec:cap-multiturn} shows what the correction would be if they were
added to the multi-turn family. When a round-v3 interval contains a round-v2
point estimate, we describe the results as not distinguished, without claiming
equality. Table~\ref{tab:results-appendix} reports secondary and L3 estimates.

\paragraph{Design document against paper.} The dated design document fixed metrics, not every endpoint.
\begin{itemize}[leftmargin=*,nosep]
  \item \emph{E1.} Collateral text was designed as a per-line character error rate above the floor; it is reported as
        the binary ``any other text changed'', re-read at registered coordinates in the final analysis. Tolerant
        locality, designed as an L1 metric, is secondary.
  \item \emph{E2a.} The three frozen endpoints are unchanged, and the joint endpoint was added as co-primary after
        review. The frozen rule made CORD the headline only if WildReceipt's $t_0$ baseline fell below half of the
        planned fields; it did not, so both corpora are in the family.
  \item \emph{E2b.} The frozen endpoints are unchanged; the conversational variant was made exploratory post hoc.
  \item \emph{E3.} The design listed brand, SKU and price OCR, logo $\Delta E_{00}$ and logo mask IoU, and all are
        reported (Tables~\ref{tab:results-appendix} and~\ref{tab:e3place}). The in-scene condition was added after
        the pilot, in-scene SKU exact was named primary after scoring, and the placement and size analyses were
        added post hoc.
  \item \emph{E4.} The design's smallest line read correctly cannot be measured below the OCR ceiling; the primary is
        the exact share above it, with intention-to-treat and matched-height analyses added post hoc. The OCR
        reading without an allowlist was declared primary after scoring.
  \item \emph{Controls.} The GPT-Image-2 \texttt{low} arms of E2a, E2b and E3 were added post hoc
        (Appendix~\ref{app:lowarm}).
\end{itemize}

\paragraph{Detector harness.} Scoring runs in a harness outside this repository with one directory per public method.
An exporter copies each cell into the harness layout with every file named by its \texttt{row\_id}, records the
(model, tier, prompt, turn) tuple behind each name, and refuses size-mismatched rows; a driver copies the per-image CSV
back, and every score joins a ledger row by \texttt{row\_id} alone. The public-method subset of the harness, with its
metrics code and exporter, is planned for release (Appendix~\ref{app:release}).

\begin{table*}[tbp]
\centering
\begingroup\lccode`\~=`\-\lowercase{\endgroup\providecommand{\tabsigns}{\mathcode`\-="8000 \def~{\mbox{\textendash}}}}
\begingroup\lccode`\~=`\+\lowercase{\endgroup\providecommand{\tabnoplus}{\mathcode`\+="8000 \def~{}}}
\providecommand{\tabv}[1]{{\let\%\relax\tabsigns#1}}
\providecommand{\tabci}[2]{{\let\%\relax\tabsigns\tabnoplus[#1,\,#2]}}
\def\tabyes{yes}
\providecommand{\tabstar}[1]{\edef\tabtmp{#1}\ifx\tabtmp\tabyes$^{*}$\fi}
\providecommand{\tabdag}[1]{\edef\tabtmp{#1}\ifx\tabtmp\tabyes$^{\dagger}$\fi}
\providecommand{\tabmarks}[2]{\makebox[1.15em][l]{\tabstar{#1}\tabdag{#2}}}
\caption{\textbf{Secondary capability rows, WildReceipt and conversational chains, continuing Table~\ref{tab:results}.} Columns and marks as in Table~\ref{tab:results}. E2a WildReceipt rows are primary endpoints; other rows are exploratory. Conversational E2b was scored before its analysis freeze. Older chains start from each model's own seed. Defence results are in Table~\ref{tab:defence-results}.}
\label{tab:results-appendix}
\footnotesize
\setlength{\tabcolsep}{1.6pt}
\renewcommand{\arraystretch}{1.05}
\providecommand{\tabms}{\setlength{\baselineskip}{\arraystretch\baselineskip}}
\begin{tabular*}{\textwidth}{@{\extracolsep{\fill}}p{0.11\textwidth}ll>{\raggedright\arraybackslash}p{0.2\textwidth}l@{\extracolsep{0pt}\hspace{2pt}}r@{\extracolsep{\fill}\hspace{5pt}}rr@{\extracolsep{\fill}\hspace{5pt}}r@{\extracolsep{0pt}\hspace{2pt}}l@{\extracolsep{\fill}\hspace{5pt}}r@{\extracolsep{0pt}\hspace{2pt}}l@{}}
\toprule
Claim $\to$ & Exp. & Layer & Metric (unit, n) & \multicolumn{2}{c}{GPT-Image-2} & Flare & Sunburst & \multicolumn{4}{c}{$\Delta$ Images~2.5 $-$ GPT-Image-2 [95\% CI]} \tabularnewline
\cmidrule(lr){9-12}
fraud task & & & & & & & & \multicolumn{2}{c}{Flare} & \multicolumn{2}{c}{Sunburst} \tabularnewline
\midrule
\multirow[t]{6}{\hsize}{\tabms\raggedright \textbf{Precise local editing} $\to$ alter one field on a receipt\par}
& E1
& L2
& Numeric text changed, v1 coordinates
& \texttt{low}
& \tabv{\capDocAnyNumGiLow}
& \tabv{\capDocAnyNumFlare}
& \tabv{\capDocAnyNumSunburst}
& \tabv{\capDocDiffAnyNumFlareGiLow}\tabmarks{}{}
& \tabci{\capDocDiffAnyNumFlareGiLowLo}{\capDocDiffAnyNumFlareGiLowHi}
& \tabv{\capDocDiffAnyNumSunburstGiLow}\tabmarks{}{}
& \tabci{\capDocDiffAnyNumSunburstGiLowLo}{\capDocDiffAnyNumSunburstGiLowHi}
\tabularnewline

&
&
& \quad(\%, n\,=\,\capDocNSpecs)
& \texttt{med.}
& \tabv{\capDocAnyNumGiMed}
&
&
& \tabv{\capDocDiffAnyNumFlareGiMed}\tabmarks{}{}
& \tabci{\capDocDiffAnyNumFlareGiMedLo}{\capDocDiffAnyNumFlareGiMedHi}
& \tabv{\capDocDiffAnyNumSunburstGiMed}\tabmarks{}{}
& \tabci{\capDocDiffAnyNumSunburstGiMedLo}{\capDocDiffAnyNumSunburstGiMedHi}
\tabularnewline

&
&
& Tolerant locality, median
& \texttt{low}
& \tabv{\capLocGitwoLowTolMed}
& \tabv{\capLocFlareTolMed}
& \tabv{\capLocSunTolMed}
& \tabv{\capLocPairFlareGitwoLowTolDelta}\tabmarks{}{}
& \tabci{\capLocPairFlareGitwoLowTolDeltaLo}{\capLocPairFlareGitwoLowTolDeltaHi}
& \tabv{\capLocPairSunGitwoLowTolDelta}\tabmarks{}{}
& \tabci{\capLocPairSunGitwoLowTolDeltaLo}{\capLocPairSunGitwoLowTolDeltaHi}
\tabularnewline

&
&
& \quad(\% crop, n\,=\,\capLocPairGitwoLowFlareTolN; floor \tabv{\capLocFloorTolMed})
& \texttt{med.}
& \tabv{\capLocGitwoMedTolMed}
&
&
& \tabv{\capLocPairFlareGitwoMedTolDelta}\tabmarks{}{}
& \tabci{\capLocPairFlareGitwoMedTolDeltaLo}{\capLocPairFlareGitwoMedTolDeltaHi}
& \tabv{\capLocPairSunGitwoMedTolDelta}\tabmarks{}{}
& \tabci{\capLocPairSunGitwoMedTolDeltaLo}{\capLocPairSunGitwoMedTolDeltaHi}
\tabularnewline
\cmidrule[\lightrulewidth]{1-12}
\multirow[t]{8}{\hsize}{\tabms\raggedright \textbf{Multi-turn consistency} $\to$ keep editing the receipt\par}
& E2a
& L1
& Edit success, WildReceipt (\%)
& \texttt{med.}
& \tabv{\capDocChainWildSuccGiMedVTwo}
& \tabv{\capDocChainWildSuccFlareVTwo}
& \tabv{\capDocChainWildSuccSunburstVTwo}
& \tabv{\capDocChainWildSuccDiffFlareVsGiMedVTwo}\tabmarks{\capDocChainWildSuccDiffFlareVsGiMedVTwoHolmSig}{\capDocChainWildSuccDiffFlareVsGiMedVTwoHolmSigGlobal}
& \tabci{\capDocChainWildSuccDiffFlareVsGiMedVTwoLo}{\capDocChainWildSuccDiffFlareVsGiMedVTwoHi}
& \tabv{\capDocChainWildSuccDiffSunburstVsGiMedVTwo}\tabmarks{\capDocChainWildSuccDiffSunburstVsGiMedVTwoHolmSig}{\capDocChainWildSuccDiffSunburstVsGiMedVTwoHolmSigGlobal}
& \tabci{\capDocChainWildSuccDiffSunburstVsGiMedVTwoLo}{\capDocChainWildSuccDiffSunburstVsGiMedVTwoHi}
\tabularnewline

&
&
& \quad \capDocChainCordReceipts{} receipts
& \texttt{low}
& \tabv{\capDocChainWildSuccGiLowVTwo}
&
&
& \tabv{\capDocChainWildSuccDiffFlareVsGiLowVTwo}\tabmarks{\capDocChainWildSuccDiffFlareVsGiLowVTwoHolmSig}{\capDocChainWildSuccDiffFlareVsGiLowVTwoHolmSigGlobal}
& \tabci{\capDocChainWildSuccDiffFlareVsGiLowVTwoLo}{\capDocChainWildSuccDiffFlareVsGiLowVTwoHi}
& \tabv{\capDocChainWildSuccDiffSunburstVsGiLowVTwo}\tabmarks{\capDocChainWildSuccDiffSunburstVsGiLowVTwoHolmSig}{\capDocChainWildSuccDiffSunburstVsGiLowVTwoHolmSigGlobal}
& \tabci{\capDocChainWildSuccDiffSunburstVsGiLowVTwoLo}{\capDocChainWildSuccDiffSunburstVsGiLowVTwoHi}
\tabularnewline

&
& L1
& Edit intact at last turn (\%)
& \texttt{med.}
& \tabv{\capDocChainWildSurvFourGiMedVTwo}
& \tabv{\capDocChainWildSurvFourFlareVTwo}
& \tabv{\capDocChainWildSurvFourSunburstVTwo}
& \tabv{\capDocChainWildSurvFourDiffFlareVsGiMedVTwo}\tabmarks{\capDocChainWildSurvFourDiffFlareVsGiMedVTwoHolmSig}{\capDocChainWildSurvFourDiffFlareVsGiMedVTwoHolmSigGlobal}
& \tabci{\capDocChainWildSurvFourDiffFlareVsGiMedVTwoLo}{\capDocChainWildSurvFourDiffFlareVsGiMedVTwoHi}
& \tabv{\capDocChainWildSurvFourDiffSunburstVsGiMedVTwo}\tabmarks{\capDocChainWildSurvFourDiffSunburstVsGiMedVTwoHolmSig}{\capDocChainWildSurvFourDiffSunburstVsGiMedVTwoHolmSigGlobal}
& \tabci{\capDocChainWildSurvFourDiffSunburstVsGiMedVTwoLo}{\capDocChainWildSurvFourDiffSunburstVsGiMedVTwoHi}
\tabularnewline

&
&
& \quad
& \texttt{low}
& \tabv{\capDocChainWildSurvFourGiLowVTwo}
&
&
& \tabv{\capDocChainWildSurvFourDiffFlareVsGiLowVTwo}\tabmarks{\capDocChainWildSurvFourDiffFlareVsGiLowVTwoHolmSig}{\capDocChainWildSurvFourDiffFlareVsGiLowVTwoHolmSigGlobal}
& \tabci{\capDocChainWildSurvFourDiffFlareVsGiLowVTwoLo}{\capDocChainWildSurvFourDiffFlareVsGiLowVTwoHi}
& \tabv{\capDocChainWildSurvFourDiffSunburstVsGiLowVTwo}\tabmarks{\capDocChainWildSurvFourDiffSunburstVsGiLowVTwoHolmSig}{\capDocChainWildSurvFourDiffSunburstVsGiLowVTwoHolmSigGlobal}
& \tabci{\capDocChainWildSurvFourDiffSunburstVsGiLowVTwoLo}{\capDocChainWildSurvFourDiffSunburstVsGiLowVTwoHi}
\tabularnewline

&
& L1
& Succeeded and intact (\%)
& \texttt{med.}
& \tabv{\capDocChainWildJointGiMedVTwo}
& \tabv{\capDocChainWildJointFlareVTwo}
& \tabv{\capDocChainWildJointSunburstVTwo}
& \tabv{\capDocChainWildJointDiffFlareVsGiMedVTwo}\tabmarks{\capDocChainWildJointDiffFlareVsGiMedVTwoHolmSig}{\capDocChainWildJointDiffFlareVsGiMedVTwoHolmSigGlobal}
& \tabci{\capDocChainWildJointDiffFlareVsGiMedVTwoLo}{\capDocChainWildJointDiffFlareVsGiMedVTwoHi}
& \tabv{\capDocChainWildJointDiffSunburstVsGiMedVTwo}\tabmarks{\capDocChainWildJointDiffSunburstVsGiMedVTwoHolmSig}{\capDocChainWildJointDiffSunburstVsGiMedVTwoHolmSigGlobal}
& \tabci{\capDocChainWildJointDiffSunburstVsGiMedVTwoLo}{\capDocChainWildJointDiffSunburstVsGiMedVTwoHi}
\tabularnewline

&
&
& \quad
& \texttt{low}
& \tabv{\capDocChainWildJointGiLowVTwo}
&
&
& \tabv{\capDocChainWildJointDiffFlareVsGiLowVTwo}\tabmarks{\capDocChainWildJointDiffFlareVsGiLowVTwoHolmSig}{\capDocChainWildJointDiffFlareVsGiLowVTwoHolmSigGlobal}
& \tabci{\capDocChainWildJointDiffFlareVsGiLowVTwoLo}{\capDocChainWildJointDiffFlareVsGiLowVTwoHi}
& \tabv{\capDocChainWildJointDiffSunburstVsGiLowVTwo}\tabmarks{\capDocChainWildJointDiffSunburstVsGiLowVTwoHolmSig}{\capDocChainWildJointDiffSunburstVsGiLowVTwoHolmSigGlobal}
& \tabci{\capDocChainWildJointDiffSunburstVsGiLowVTwoLo}{\capDocChainWildJointDiffSunburstVsGiLowVTwoHi}
\tabularnewline

&
& L2
& Never-edited fields kept (\%)
& \texttt{med.}
& \tabv{\capDocChainWildUntouchedGiMedVTwo}
& \tabv{\capDocChainWildUntouchedFlareVTwo}
& \tabv{\capDocChainWildUntouchedSunburstVTwo}
& \tabv{\capDocChainWildUntouchedDiffFlareVsGiMedVTwo}\tabmarks{\capDocChainWildUntouchedDiffFlareVsGiMedVTwoHolmSig}{\capDocChainWildUntouchedDiffFlareVsGiMedVTwoHolmSigGlobal}
& \tabci{\capDocChainWildUntouchedDiffFlareVsGiMedVTwoLo}{\capDocChainWildUntouchedDiffFlareVsGiMedVTwoHi}
& \tabv{\capDocChainWildUntouchedDiffSunburstVsGiMedVTwo}\tabmarks{\capDocChainWildUntouchedDiffSunburstVsGiMedVTwoHolmSig}{\capDocChainWildUntouchedDiffSunburstVsGiMedVTwoHolmSigGlobal}
& \tabci{\capDocChainWildUntouchedDiffSunburstVsGiMedVTwoLo}{\capDocChainWildUntouchedDiffSunburstVsGiMedVTwoHi}
\tabularnewline

&
&
& \quad
& \texttt{low}
& \tabv{\capDocChainWildUntouchedGiLowVTwo}
&
&
& \tabv{\capDocChainWildUntouchedDiffFlareVsGiLowVTwo}\tabmarks{\capDocChainWildUntouchedDiffFlareVsGiLowVTwoHolmSig}{\capDocChainWildUntouchedDiffFlareVsGiLowVTwoHolmSigGlobal}
& \tabci{\capDocChainWildUntouchedDiffFlareVsGiLowVTwoLo}{\capDocChainWildUntouchedDiffFlareVsGiLowVTwoHi}
& \tabv{\capDocChainWildUntouchedDiffSunburstVsGiLowVTwo}\tabmarks{\capDocChainWildUntouchedDiffSunburstVsGiLowVTwoHolmSig}{\capDocChainWildUntouchedDiffSunburstVsGiLowVTwoHolmSigGlobal}
& \tabci{\capDocChainWildUntouchedDiffSunburstVsGiLowVTwoLo}{\capDocChainWildUntouchedDiffSunburstVsGiLowVTwoHi}
\tabularnewline
\cmidrule[\lightrulewidth]{1-12}
\multirow[t]{5}{\hsize}{\tabms\raggedright \textbf{Multi-turn consistency} $\to$ keep editing a photo\par}
& E2b
& L1
& Edit intact at last turn, in a chat (\%)
& \texttt{med.}
& \tabv{\capChainConvGiMedLOneSurvivalVTwo}
& \tabv{\capChainConvFlareLOneSurvivalVTwo}
& \tabv{\capChainConvSunburstLOneSurvivalVTwo}
& \tabv{\capChainConvFlareMinusGiMedLOneSurvivalVTwo}\tabmarks{}{}
& \tabci{\capChainConvFlareMinusGiMedLOneSurvivalVTwoLo}{\capChainConvFlareMinusGiMedLOneSurvivalVTwoHi}
& \tabv{\capChainConvSunburstMinusGiMedLOneSurvivalVTwo}\tabmarks{}{}
& \tabci{\capChainConvSunburstMinusGiMedLOneSurvivalVTwoLo}{\capChainConvSunburstMinusGiMedLOneSurvivalVTwoHi}
\tabularnewline

&
& L2
& \quad in place (\capChainConvFlareNChainsVTwo{} chains on \capChainConvFlareNSourcePhotosVTwo{} photos)
& \texttt{med.}
& \tabv{\capChainConvGiMedLTwoInPlaceVTwo}
& \tabv{\capChainConvFlareLTwoInPlaceVTwo}
& \tabv{\capChainConvSunburstLTwoInPlaceVTwo}
& \tabv{\capChainConvFlareMinusGiMedLTwoInPlaceVTwo}\tabmarks{}{}
& \tabci{\capChainConvFlareMinusGiMedLTwoInPlaceVTwoLo}{\capChainConvFlareMinusGiMedLTwoInPlaceVTwoHi}
& \tabv{\capChainConvSunburstMinusGiMedLTwoInPlaceVTwo}\tabmarks{}{}
& \tabci{\capChainConvSunburstMinusGiMedLTwoInPlaceVTwoLo}{\capChainConvSunburstMinusGiMedLTwoInPlaceVTwoHi}
\tabularnewline
\cmidrule[\lightrulewidth]{1-12}
\multirow[t]{5}{\hsize}{\tabms\raggedright \textbf{Multi-turn}, laundering chains\par}
& old
& L1
& Edit survival (\%, n\,=\,\capTurnSurvRateGptTwoN{} edits)
&
& \tabv{\capTurnSurvRateGptTwo}
& \tabv{\capTurnSurvRateFlare}
& \tabv{\capTurnSurvRateSunburst}
& \tabv{\capTurnDiffSurvRateFlare}\tabmarks{}{}
& \tabci{\capTurnDiffSurvRateFlareLo}{\capTurnDiffSurvRateFlareHi}
& \tabv{\capTurnDiffSurvRateSunburst}\tabmarks{}{}
& \tabci{\capTurnDiffSurvRateSunburstLo}{\capTurnDiffSurvRateSunburstHi}
\tabularnewline

&
&
& \quad preliminary: generated seeds,
&
&
&
&
&
&
&
&
\tabularnewline

&
&
& \quad stateless, own seed per model
&
&
&
&
&
&
&
&
\tabularnewline
\cmidrule[\lightrulewidth]{1-12}
\multirow[t]{9}{\hsize}{\tabms\raggedright \textbf{Reference fidelity} $\to$ drop a product into a fake listing\par}
& E3
&
& SKU exact at matched glyph height
& \texttt{med.}
&
&
&
& \tabv{\capRefPlProgInSceneFlareVsGiMatchedAbs}\tabmarks{}{}
& \tabci{\capRefPlProgInSceneFlareVsGiMatchedAbsLo}{\capRefPlProgInSceneFlareVsGiMatchedAbsHi}
& \tabv{\capRefPlProgInSceneSunburstVsGiMatchedAbs}\tabmarks{}{}
& \tabci{\capRefPlProgInSceneSunburstVsGiMatchedAbsLo}{\capRefPlProgInSceneSunburstVsGiMatchedAbsHi}
\tabularnewline

&
&
& \quad(pp; glyph px \capRefPInSceneSkuHeightGi, \capRefPInSceneSkuHeightFlare, \capRefPInSceneSkuHeightSunburst)
& \texttt{low}
&
&
&
& \tabv{\capRefPlProgInSceneFlareVsGiLowMatchedAbs}\tabmarks{}{}
& \tabci{\capRefPlProgInSceneFlareVsGiLowMatchedAbsLo}{\capRefPlProgInSceneFlareVsGiLowMatchedAbsHi}
& \tabv{\capRefPlProgInSceneSunburstVsGiLowMatchedAbs}\tabmarks{}{}
& \tabci{\capRefPlProgInSceneSunburstVsGiLowMatchedAbsLo}{\capRefPlProgInSceneSunburstVsGiLowMatchedAbsHi}
\tabularnewline

&
&
& Logo mask IoU, in scene
& \texttt{med.}
& \capRefPlProgInSceneLogoIouGi
& \capRefPlProgInSceneLogoIouFlare
& \capRefPlProgInSceneLogoIouSunburst
& \tabv{\capRefPlProgInSceneFlareVsGiLogoIou}\tabmarks{}{}
& \tabci{\capRefPlProgInSceneFlareVsGiLogoIouLo}{\capRefPlProgInSceneFlareVsGiLogoIouHi}
& \tabv{\capRefPlProgInSceneSunburstVsGiLogoIou}\tabmarks{}{}
& \tabci{\capRefPlProgInSceneSunburstVsGiLogoIouLo}{\capRefPlProgInSceneSunburstVsGiLogoIouHi}
\tabularnewline

&
&
&
& \texttt{low}
& \capRefPlProgInSceneLogoIouGiLow
&
&
& \tabv{\capRefPlProgInSceneFlareVsGiLowLogoIou}\tabmarks{}{}
& \tabci{\capRefPlProgInSceneFlareVsGiLowLogoIouLo}{\capRefPlProgInSceneFlareVsGiLowLogoIouHi}
& \tabv{\capRefPlProgInSceneSunburstVsGiLowLogoIou}\tabmarks{}{}
& \tabci{\capRefPlProgInSceneSunburstVsGiLowLogoIouLo}{\capRefPlProgInSceneSunburstVsGiLowLogoIouHi}
\tabularnewline

&
&
& Logo mask IoU, hero
& \texttt{med.}
& \capRefPlProgHeroLogoIouGi
& \capRefPlProgHeroLogoIouFlare
& \capRefPlProgHeroLogoIouSunburst
& \tabv{\capRefPlProgHeroFlareVsGiLogoIou}\tabmarks{}{}
& \tabci{\capRefPlProgHeroFlareVsGiLogoIouLo}{\capRefPlProgHeroFlareVsGiLogoIouHi}
& \tabv{\capRefPlProgHeroSunburstVsGiLogoIou}\tabmarks{}{}
& \tabci{\capRefPlProgHeroSunburstVsGiLogoIouLo}{\capRefPlProgHeroSunburstVsGiLogoIouHi}
\tabularnewline

&
&
&
& \texttt{low}
& \capRefPlProgHeroLogoIouGiLow
&
&
& \tabv{\capRefPlProgHeroFlareVsGiLowLogoIou}\tabmarks{}{}
& \tabci{\capRefPlProgHeroFlareVsGiLowLogoIouLo}{\capRefPlProgHeroFlareVsGiLowLogoIouHi}
& \tabv{\capRefPlProgHeroSunburstVsGiLowLogoIou}\tabmarks{}{}
& \tabci{\capRefPlProgHeroSunburstVsGiLowLogoIouLo}{\capRefPlProgHeroSunburstVsGiLowLogoIouHi}
\tabularnewline

&
&
& Hero logo $\Delta E_{00}$, white-balanced
& \texttt{med.}
& \capRefPHeroLogoDEwbGi
& \capRefPHeroLogoDEwbFlare
& \capRefPHeroLogoDEwbSunburst
& \tabv{\capRefPHeroLogoDEwbDiffFlare}\tabmarks{}{}
& \tabci{\capRefPHeroLogoDEwbDiffLoFlare}{\capRefPHeroLogoDEwbDiffHiFlare}
& \tabv{\capRefPHeroLogoDEwbDiffSunburst}\tabmarks{}{}
& \tabci{\capRefPHeroLogoDEwbDiffLoSunburst}{\capRefPHeroLogoDEwbDiffHiSunburst}
\tabularnewline

&
&
& Brand read in scene (\%)
& \texttt{med.}
& \tabv{\capRefPInSceneBrandExactGi}
& \tabv{\capRefPInSceneBrandExactFlare}
& \tabv{\capRefPInSceneBrandExactSunburst}
& \tabv{\capRefPInSceneBrandExactDiffFlare}\tabmarks{}{}
& \tabci{\capRefPInSceneBrandExactDiffLoFlare}{\capRefPInSceneBrandExactDiffHiFlare}
& \tabv{\capRefPInSceneBrandExactDiffSunburst}\tabmarks{}{}
& \tabci{\capRefPInSceneBrandExactDiffLoSunburst}{\capRefPInSceneBrandExactDiffHiSunburst}
\tabularnewline

&
&
& Price read in scene (\%)
& \texttt{med.}
& \tabv{\capRefPInScenePriceExactGi}
& \tabv{\capRefPInScenePriceExactFlare}
& \tabv{\capRefPInScenePriceExactSunburst}
& \tabv{\capRefPInScenePriceExactDiffFlare}\tabmarks{}{}
& \tabci{\capRefPInScenePriceExactDiffLoFlare}{\capRefPInScenePriceExactDiffHiFlare}
& \tabv{\capRefPInScenePriceExactDiffSunburst}\tabmarks{}{}
& \tabci{\capRefPInScenePriceExactDiffLoSunburst}{\capRefPInScenePriceExactDiffHiSunburst}
\tabularnewline
\cmidrule[\lightrulewidth]{1-12}
\multirow[t]{10}{\hsize}{\tabms\raggedright \textbf{Sharper detail} $\to$ render legible fine print\par}
& E4
& L1
& Matched glyph height, \tokMax{} tok (pp)
&
&
&
&
& \tabv{\capDetailVtwoMatchedFlareMaxVsGiHigh}\tabmarks{}{}
& \tabci{\capDetailVtwoMatchedFlareMaxVsGiHighLo}{\capDetailVtwoMatchedFlareMaxVsGiHighHi}
& \tabv{\capDetailVtwoMatchedSunburstMaxVsGiHigh}\tabmarks{}{}
& \tabci{\capDetailVtwoMatchedSunburstMaxVsGiHighLo}{\capDetailVtwoMatchedSunburstMaxVsGiHighHi}
\tabularnewline

&
&
& All planned lines, below ceiling failed (ITT), \tokHigh{} tok
& \texttt{med.}
& \tabv{\capDetailVtwoIttGiMedium}
& \tabv{\capDetailVtwoIttFlareHigh}
& \tabv{\capDetailVtwoIttSunburstHigh}
& \tabv{\capDetailVtwoIttFlareHighVsGiMedium}\tabmarks{}{}
& \tabci{\capDetailVtwoIttFlareHighVsGiMediumLo}{\capDetailVtwoIttFlareHighVsGiMediumHi}
& \tabv{\capDetailVtwoIttSunburstHighVsGiMedium}\tabmarks{}{}
& \tabci{\capDetailVtwoIttSunburstHighVsGiMediumLo}{\capDetailVtwoIttSunburstHighVsGiMediumHi}
\tabularnewline

&
&
& \quad\tokMax{} tok (\%; an instrument artefact)
& \texttt{high}
& \tabv{\capDetailVtwoIttGiHigh}
& \tabv{\capDetailVtwoIttFlareMax}
& \tabv{\capDetailVtwoIttSunburstMax}
& \tabv{\capDetailVtwoIttFlareMaxVsGiHigh}\tabmarks{}{}
& \tabci{\capDetailVtwoIttFlareMaxVsGiHighLo}{\capDetailVtwoIttFlareMaxVsGiHighHi}
& \tabv{\capDetailVtwoIttSunburstMaxVsGiHigh}\tabmarks{}{}
& \tabci{\capDetailVtwoIttSunburstMaxVsGiHighLo}{\capDetailVtwoIttSunburstMaxVsGiHighHi}
\tabularnewline

&
&
& Best tier $-$ \texttt{low}, same model (pp)
&
& \tabv{\capDetailTierBestDiffGi}
& \tabv{\capDetailTierBestDiffFlare}
& \tabv{\capDetailTierBestDiffSunburst}
&
&
&
&
\tabularnewline

&
&
& Detail energy (\%), \capSharpLatTokHigh{} tok
&
&
&
&
& \tabv{\capSharpMatchFlareHighLapPct}\tabmarks{}{}
& \tabci{\capSharpMatchFlareHighLapLo}{\capSharpMatchFlareHighLapHi}
& \tabv{\capSharpMatchSunHighLapPct}\tabmarks{}{}
& \tabci{\capSharpMatchSunHighLapLo}{\capSharpMatchSunHighLapHi}
\tabularnewline

&
&
& \quad\capSharpLatTokMax{} tok (\texttt{max} vs \texttt{high})
&
&
&
&
& \tabv{\capSharpMatchFlareMaxLapPct}\tabmarks{}{}
& \tabci{\capSharpMatchFlareMaxLapLo}{\capSharpMatchFlareMaxLapHi}
& \tabv{\capSharpMatchSunMaxLapPct}\tabmarks{}{}
& \tabci{\capSharpMatchSunMaxLapLo}{\capSharpMatchSunMaxLapHi}
\tabularnewline

&
&
& Noise $\sigma$ (\%, no CI), \capSharpLatTokHigh{} tok
&
&
&
&
& \tabv{\capSharpMatchFlareHighFlatNoisePct}\tabmarks{}{}
&
& \tabv{\capSharpMatchSunHighFlatNoisePct}\tabmarks{}{}
&
\tabularnewline

&
&
& \quad\capSharpLatTokMax{} tok
&
&
&
&
& \tabv{\capSharpMatchFlareMaxFlatNoisePct}\tabmarks{}{}
&
& \tabv{\capSharpMatchSunMaxFlatNoisePct}\tabmarks{}{}
&
\tabularnewline

&
&
& Edge width (\%), \capSharpLatTokHigh{} tok
&
&
&
&
& \tabv{\capSharpMatchFlareHighEdgeWPct}\tabmarks{}{}
& \tabci{\capSharpMatchFlareHighEdgeWLo}{\capSharpMatchFlareHighEdgeWHi}
& \tabv{\capSharpMatchSunHighEdgeWPct}\tabmarks{}{}
& \tabci{\capSharpMatchSunHighEdgeWLo}{\capSharpMatchSunHighEdgeWHi}
\tabularnewline

&
&
& \quad\capSharpLatTokMax{} tok
&
&
&
&
& \tabv{\capSharpMatchFlareMaxEdgeWPct}\tabmarks{}{}
& \tabci{\capSharpMatchFlareMaxEdgeWLo}{\capSharpMatchFlareMaxEdgeWHi}
& \tabv{\capSharpMatchSunMaxEdgeWPct}\tabmarks{}{}
& \tabci{\capSharpMatchSunMaxEdgeWLo}{\capSharpMatchSunMaxEdgeWHi}
\tabularnewline
\bottomrule
\end{tabular*}
\end{table*}

\begin{table*}[tbp]
\centering
\begingroup\lccode`\~=`\-\lowercase{\endgroup\providecommand{\tabsigns}{\mathcode`\-="8000 \def~{\mbox{\textendash}}}}
\begingroup\lccode`\~=`\+\lowercase{\endgroup\providecommand{\tabnoplus}{\mathcode`\+="8000 \def~{}}}
\providecommand{\tabv}[1]{{\let\%\relax\tabsigns#1}}
\providecommand{\tabci}[2]{{\let\%\relax\tabsigns\tabnoplus[#1,\,#2]}}
\def\tabyes{yes}
\providecommand{\tabstar}[1]{\edef\tabtmp{#1}\ifx\tabtmp\tabyes$^{*}$\fi}
\providecommand{\tabdag}[1]{\edef\tabtmp{#1}\ifx\tabtmp\tabyes$^{\dagger}$\fi}
\providecommand{\tabmarks}[2]{\makebox[1.15em][l]{\tabstar{#1}\tabdag{#2}}}
\caption{\textbf{Defence and receipt-edit latency.} Columns as in Table~\ref{tab:results}. Speed: ratio of median synchronous latency, with 95\% intervals clustered by specification; arms were not interleaved. Refusal: first-pass counts on shared specifications. DocTamper: mean per-image localisation AUC; the margin is the smallest symmetric band around chance containing the arm's 90\% interval. Tier results do not establish equivalence. Wild membership is self-reported; detection rates use one fixed threshold and are not content-matched. Detection-rate intervals are reported in \S\ref{sec:defence-wild}.}
\label{tab:defence-results}
\footnotesize
\setlength{\tabcolsep}{1.6pt}
\renewcommand{\arraystretch}{1.05}
\providecommand{\tabms}{\setlength{\baselineskip}{\arraystretch\baselineskip}}
\begin{tabular*}{\textwidth}{@{\extracolsep{\fill}}p{0.11\textwidth}ll>{\raggedright\arraybackslash}p{0.2\textwidth}l@{\extracolsep{0pt}\hspace{2pt}}r@{\extracolsep{\fill}\hspace{5pt}}rr@{\extracolsep{\fill}\hspace{5pt}}r@{\extracolsep{0pt}\hspace{2pt}}l@{\extracolsep{\fill}\hspace{5pt}}r@{\extracolsep{0pt}\hspace{2pt}}l@{}}
\toprule
Claim $\to$ & Exp. & Layer & Metric (unit, n) & \multicolumn{2}{c}{GPT-Image-2} & Flare & Sunburst & \multicolumn{4}{c}{$\Delta$ Images~2.5 $-$ GPT-Image-2 [95\% CI]} \tabularnewline
\cmidrule(lr){9-12}
fraud task & & & & & & & & \multicolumn{2}{c}{Flare} & \multicolumn{2}{c}{Sunburst} \tabularnewline
\midrule
\multirow[t]{4}{\hsize}{\tabms\raggedright \textbf{Faster} on the forgery task\par}
& E1
& L3
& Median s per edit; $\Delta$ GPT-Image-2\,/\,Images~2.5
& \texttt{med.}
& \latForgeDocsGitwoMedMedianS
& \latForgeDocsFlareMedianS
& \latForgeDocsSunMedianS
& $\times$\tabv{\latForgeDocsGitwoMedOverFlareRatio}\tabmarks{}{}
& \tabci{\latForgeDocsGitwoMedOverFlareLo}{\latForgeDocsGitwoMedOverFlareHi}
& $\times$\tabv{\latForgeDocsGitwoMedOverSunRatio}\tabmarks{}{}
& \tabci{\latForgeDocsGitwoMedOverSunLo}{\latForgeDocsGitwoMedOverSunHi}
\tabularnewline

&
&
& \quad(n\,=\,\latForgeDocsGitwoLowOverFlareN--\latForgeDocsGitwoMedOverFlareN{} specs)
& \texttt{low}
& \latForgeDocsGitwoLowMedianS
&
&
& $\times$\tabv{\latForgeDocsGitwoLowOverFlareRatio}\tabmarks{}{}
& \tabci{\latForgeDocsGitwoLowOverFlareLo}{\latForgeDocsGitwoLowOverFlareHi}
& $\times$\tabv{\latForgeDocsGitwoLowOverSunRatio}\tabmarks{}{}
& \tabci{\latForgeDocsGitwoLowOverSunLo}{\latForgeDocsGitwoLowOverSunHi}
\tabularnewline
\addlinespace[1.5pt]
\multirow[t]{3}{\hsize}{\tabms\raggedright Refusal\par}
& E1
& L3
& Refused of \refPairNShared{} shared specs
& \multicolumn{2}{r}{\texttt{low} \refPairGitwoLowRefused, \texttt{med.} \refPairGitwoMedRefused}
& \refPairFlareRefused
& \refPairSunRefused
&
&
&
&
\tabularnewline

&
&
& \quad exact McNemar $p$ vs \texttt{low}, \texttt{med.}
&
&
&
&
& \refPairFlareVsGitwoLowPMcNemar, \refPairFlareVsGitwoMedPMcNemar
&
& \refPairSunVsGitwoLowPMcNemar, \refPairSunVsGitwoMedPMcNemar
&
\tabularnewline
\addlinespace[1.5pt]
\multirow[t]{3}{\hsize}{\tabms\raggedright Localisation\par}
& E1
& L3
& DocTamper per-image AUC
& \texttt{low}
& \detDtGitwoLowPixAuc
& \detDtFlarePixAuc
& \detDtSunPixAuc
&
&
&
&
\tabularnewline

&
&
& \quad(n\,=\,\detDtGitwoLowPixN, \detDtGitwoMedPixN; \detDtFlarePixN, \detDtSunPixN)
& \texttt{med.}
& \detDtGitwoMedPixAuc
&
&
& \tabv{\detPairDtFlareMinusGitwoMedDelta}\tabmarks{}{}
& \tabci{\detPairDtFlareMinusGitwoMedLo}{\detPairDtFlareMinusGitwoMedHi}
& \tabv{\detPairDtSunMinusGitwoMedDelta}\tabmarks{}{}
& \tabci{\detPairDtSunMinusGitwoMedLo}{\detPairDtSunMinusGitwoMedHi}
\tabularnewline

&
&
& \quad smallest equivalence margin
& \multicolumn{2}{r}{\eqDtGitwoLowMinMargin, \eqDtGitwoMedMinMargin}
& \eqDtFlareMinMargin
& \eqDtSunMinMargin
&
&
&
&
\tabularnewline
\addlinespace[1.5pt]
\multirow[t]{4}{\hsize}{\tabms\raggedright Detection vs price\par}
& core
& L3
& Community Forensics AUC, 1024 px
& \multicolumn{4}{c}{\tierAucFlatLo--\tierAucFlatHi}
&
&
&
&
\tabularnewline

&
&
& \quad smallest equivalence margin
& \multicolumn{4}{c}{\eqTierAllMinMargin}
&
&
&
&
\tabularnewline

&
&
& \quad 4K cells (n\,=\,\tierFourKN{} each)
& \multicolumn{4}{c}{\tierFourKLo--\tierFourKHi}
&
&
&
&
\tabularnewline
\addlinespace[1.5pt]
\multirow[t]{4}{\hsize}{\tabms\raggedright Benchmark vs wild\par}
& wild
& L3
& Detected at \wildXfpr{} FPR (\%)
& \multicolumn{4}{c}{cells \tabv{\wildXCtrlMean}; posted \tabv{\wildXWildRate}}
&
&
&
&
\tabularnewline

&
&
& \quad(\wildXCells{} cells; n\,=\,\wildXWildN{} posted)
&
&
&
&
&
&
&
&
\tabularnewline
\bottomrule
\end{tabular*}
\end{table*}

\subsection{Timeline of pre-specified and post hoc decisions}
\label{app:timing}

Several analysis choices followed inspection of results. The sequence below
separates dated records from evidence that a decision preceded scoring.
\begin{itemize}[leftmargin=*]
  \item \emph{Metrics.} The design document is dated 10~September~2026,
  before E1 scoring and E2--E4 generation. The archived document and scorers
  are in commit \texttt{29b82d4}. That commit followed full-run scoring, so
  it establishes the documents' contents, not when each rule was fixed.

  \item \emph{E1, E3 and E4 primary contrasts.} These were selected on
  11~September~2026 after results were available. E1 re-scores images generated
  before the design document, with the earlier pixel-exact locality results
  already known to the team. E3's in-scene condition followed its pilot, and
  its primary endpoint followed full-run scoring. E4's no-allowlist OCR reading
  was also declared primary after scoring; the allowlisted reading is sensitivity.

  \item \emph{E2a.} The endpoints, contrasts and Holm family were recorded in
  the dated freeze section at 07:56~UTC on 11~September~2026. This followed a
  three-receipt pilot but preceded the first full-run estimate file at
  08:01~UTC. Full-run OCR had already started. The scorer was edited after
  estimates existed, without changing any estimate.

  \item \emph{E2b.} The rules were frozen at the same time, after results had
  been inspected. An interim conversational file already contained all
  \capChainConvFlareNChainsVTwo{} chains scored with the final scorer.
  A partial stateless pre-warm had also printed Flare's results. Detection
  thresholds were frozen on 10~September~2026. The umbrella colour-share
  threshold was lowered after one Flare conversational pilot miss, then frozen;
  the original threshold is reported as sensitivity (Appendix~\ref{app:e2}).
  The conversational variant is therefore exploratory.

  \item \emph{Post hoc analyses.} Registered E1 re-reading, the permutation
  framework, claim-wise and global Holm corrections, the E2a joint endpoint,
  E3 placement and size analyses, and E4 intention-to-treat and matched-height
  analyses all followed the main results. They use previously seen data.
  The GPT-Image-2 \texttt{low} arms of E2a, E2b and E3 followed the main runs.
  Their addendum was frozen at 18:13~UTC before any low-arm image existed;
  generation began minutes later, about 11 hours after the comparison arms.
  A price-description correction followed the low-arm results. It changes
  labels, not contrasts (Appendix~\ref{app:lowarm}).

  \item \emph{L3.} Refusal and DocTamper localisation retain round-v2
  conventions. All other L3 analyses, including equivalence tests, are exploratory.

  \item \emph{Human rating.} No outcome is human-rated. An AI coding agent
  (Claude) inspected blind sheets only to spot-check automatic scorers.
  Human replication remains future work.
\end{itemize}

\subsection{The GPT-Image-2 low setting (added post hoc)}
\label{app:lowarm}

The original E2a, E2b and E3 control used GPT-Image-2 \texttt{medium}, which
bills \tokHigh{} output tokens per 1024-px image. Images~2.5
\texttt{medium} bills \tokMed{} tokens, one quarter as many.
We added GPT-Image-2 \texttt{low} (\tokLow{} tokens), already used in E1,
to bracket that price. Neither control matches the Images~2.5 price because
GPT-Image-2 has no \tokMed{}-token tier.

The addendum was frozen at 18:13~UTC on 11~September~2026, before any low-arm
image was generated. It fixes the same stimuli, prompts, edit orders and seeds.
It covers E2a in full, stateless E2b and both E3 reference sets, retaining the
E2 thresholds and endpoint definitions. Each Images~2.5 $-$ \texttt{low}
contrast joins the existing claim-wise Holm family; the \texttt{medium}
contrast retains its status.

The low-arm images were generated about 11 hours after their comparison arms.
Within-day serving or moderation drift therefore remains a confound that the
interleaved \texttt{medium} arm does not share. The addendum initially described
\texttt{low} as price-matched. A dated correction written after the low-arm
results records that it is not. The completed rendered-reference analysis
appears in \S\ref{sec:cap-reference}.

\section{E1: Alter One Field on a Receipt}
\label{app:e1}

\subsection{Generation pipeline and prompt}
\label{app:pipeline}

E1 reuses the round-v1 forgery catalogue: source image, target field, original
and forged values, and bounding box \cite{wu2026aiforgedoc,wu2026forgerjudge}.
The catalogue entries are fixed while the generator changes.

For each specification, we expand the field box by 50\% per side, with a
100-pixel floor, to obtain a context crop. As in round~v2, we choose the
smallest legal output size near the crop's aspect ratio. The endpoint requires
dimensions divisible by 16, a minimum pixel count, an aspect ratio within
$[1{:}3,3{:}1]$, and height within $[128,2048]$. For a crop $(W_c,H_c)$,
we select the smallest legal $(W^*,H^*)$ satisfying
$W^*/H^*\approx W_c/H_c$. Returned images are resized to $(W_c,H_c)$
with Lanczos filtering.

These constraints caused deterministic failures for long receipts in round~v2
(\vTwoRejectPct{} of the round-v1 ceiling). This round admits only compatible
crops, so it is a filtered catalogue rather than a random sample.
Round~v2 marked the target with a green outline because it could not pass a
mask channel. The direct API instead accepts an opaque RGBA mask with a
transparent target field. Every request uses the following prompt, substituting
only the forged value:

\begin{quote}\small\ttfamily
Edit only the transparent (masked) region of this document image. Replace the text there with
exactly: "\{forged\_value\}". Match the original font, size, weight, color, alignment, ink
density and paper texture so the edit is indistinguishable from the rest of the document. Do not
change anything outside the masked region. Output the full image.
\end{quote}

\noindent The fixed template is shorter than round~v2's five-clause
prompt \cite{wu2026forgerjudge}, which explained the drawn rectangle.
There is no selection among prompt variants within this experiment, but the
changed template confounds the historical refusal comparison
(\S\ref{sec:defence-refusal}).

We resize the returned crop and paste it into the authentic document. Pixels
outside the crop are copied from the source and remain identical by
construction. The ground-truth mask is the tight field box at source resolution,
using the round-v1/v2 and DocTamper format \cite{qu2023doctamper}.

\paragraph{Control cohort.} The first \docsControlSpecs{} specifications, split evenly between the two corpora, are
also run on GPT-Image-2 through the same client, prompt and alpha mask in the same week, at \texttt{low}
(\tokLow{} tokens) and \texttt{medium} (\tokHigh{}), since no GPT-Image-2 tier bills the \tokMed{} tokens of the
Images~2.5 rows (\docsControlAttempts{} attempts, \docsControlOk{} accepted, \$\docsControlSpend{}).

\subsection{Excluding inputs with mismatched sizes}
\label{app:sizeguard}

We flag an output when its aspect ratio differs from the request by more than
2\%. Resizing it directly would distort the pasted region and compromise
pixel-level metrics. The flag applies to \docsSizeMismatch{}
(\docsSizeMismatchPct{}) document rows, which the detector exporter excludes.

The exclusion changes the evaluated population. Flag rates are higher on
WildReceipt than CORD (\smWildPct{} versus \smCordPct{}) and on long
rather than monetary fields (\smOtherPct{} versus \smAmountPct{}).
The most frequently flagged fields are \smTopFlaggedFields{}.
Flagged rows are \smFlaggedWildShare{} WildReceipt and
\smFlaggedAmountShare{} monetary, compared with \smRetainedWildShare{}
and \smRetainedAmountShare{} among retained rows.

E1 detector and locality estimates thus under-represent photographed receipts
with long, wide fields. The OCR analysis retains flagged rows and excludes
them only in sensitivity analysis, leaving \capDocNSpecsNoMismatch{}
of \capDocNSpecs{} specifications.

\subsection{OCR checks and calibration}
\label{app:e1ocr}

OCR (easyocr) reads the target field and other source-crop text lines in each
output. \emph{Target read} (L1) compares the recognised target with the forged
value, ignoring case, spacing and punctuation. It is scored only when OCR reads
the original field correctly: \capDocBaseN{} of \capDocNSpecs{}
specifications, including \capDocBaseCord{} CORD and \capDocBaseWild{}
WildReceipt fields. The small eligible WildReceipt subset limits its
contribution to L1.

\emph{Any other text changed} (L2) records whether a stable token changes or
new text appears. We register each output to its source using the
tolerant-locality transformation, then re-read at registered coordinates.
The primary analysis has \capDocNStableTokRegVTwo{} stable tokens;
the earlier unregistered analysis used \capDocNStableTokVOneVTwo{} and
re-read at identical coordinates. Absolute rates depend on the token set,
although the paired contrasts are robust to this choice
(Appendix~\ref{app:e1shift}).

Calibration uses source-crop noise replicas: a resampling round trip, a
sub-pixel shift and a high-quality JPEG. A token is eligible if all but one
replica reproduce the source reading. The held-out replica estimates the
OCR-only false-change rate. Under registration, the rate for any other text is
\capDocFloorAnyOtherRegVTwo{} (\capDocFloorAnyOtherCordRegVTwo{}
CORD; \capDocFloorAnyOtherWildRegVTwo{} WildReceipt), compared with
\capDocFloorAnyOther{} at identical coordinates.

Pairing controls the source specification, not all measurement error.
The replicas do not match each arm's shift distribution, so differential OCR
error can remain.

\paragraph{AI-agent spot check of the scorer (not a human rater).} An AI coding agent (Claude) judged blind sheets of source and
output crops without seeing the scorer's verdict. In the second of two rounds it agreed with the target-read verdict on
\capDocValTwoFieldK{} of \capDocValTwoFieldN{} items and with the collateral verdict on \capDocValTwoOtherK{} of
\capDocValTwoOtherN{}. In the first round, after which the scorer was revised once, it agreed on
\capDocValOneFieldK{} of \capDocValOneFieldN{} and \capDocValOneOtherK{} of \capDocValOneOtherN{}. Both rounds predate
registered re-reading, so no spot check covers the registered token set that is now primary. These counts describe scorer agreement only; no statistic is computed from them.

\paragraph{Exact tests, power and the corpus split.} Under registration every collateral contrast survives Holm with
both the permutation test and exact McNemar tests (McNemar Holm $p$ \capDocFlareMinusGiLowLTwoOtherVTwoPHolmMcn{},
\capDocFlareMinusGiMedLTwoOtherVTwoPHolmMcn{}, \capDocSunburstMinusGiLowLTwoOtherVTwoPHolmMcn{} and
\capDocSunburstMinusGiMedLTwoOtherVTwoPHolmMcn{} for Flare and Sunburst against \texttt{low} and \texttt{medium}), so no
decision depends on the test. Under the unregistered metric, Flare against \texttt{low} had Holm $p$
\capDocVOneMetricFlareMinusGiLowLTwoOtherVTwoPHolm{} by permutation and \capDocVOneMetricFlareMinusGiLowLTwoOtherVTwoPHolmMcn{}
by McNemar.

No target-read contrast survives under either test (Sunburst against \texttt{low}:
\capDocSunburstMinusGiLowLOneTargetVTwoPHolm{} and \capDocSunburstMinusGiLowLOneTargetVTwoPHolmMcn{}; against
\texttt{medium}: \capDocSunburstMinusGiMedLOneTargetVTwoPHolm{} and \capDocSunburstMinusGiMedLOneTargetVTwoPHolmMcn{}).

With \capDocMdeLOneNVTwo{} conditional edits the minimum detectable difference at \capDocMdeLOnePowerVTwo{} power is
\capDocMdeLOneMinVTwo{}--\capDocMdeLOneMaxVTwo{}\,pp, or \capDocMdeLOneHolmMinVTwo{}--\capDocMdeLOneHolmMaxVTwo{}\,pp at
the Holm-adjusted level.

The same corpus rule applies to both layers. On CORD all four collateral contrasts survive their
exploratory Holm (\capDocCordSunburstMinusGiLowLTwoOtherVTwo{} to \capDocCordFlareMinusGiMedLTwoOtherVTwo{}; McNemar
Holm $p\le$\,\capDocCordSunburstMinusGiLowLTwoOtherVTwoPHolmMcn{}) and no target-read contrast does (the largest,
Sunburst against \texttt{low}, \capDocCordSunburstMinusGiLowLOneTargetVTwo{}, has Holm
$p$\,=\,\capDocCordSunburstMinusGiLowLOneTargetVTwoPHolm{}). Under the permutation framework the unregistered metric gives
Sunburst's CORD target read against \texttt{medium} Holm $p$\,=\,\capDocVOneMetricCordSunburstMinusGiMedLOneTargetVTwoPHolm{}.

On WildReceipt no contrast survives at either layer (collateral text \capDocWildSunburstMinusGiLowLTwoOtherVTwo{} to
\capDocWildFlareMinusGiMedLTwoOtherVTwo{}).

\subsection{Registration and shift sensitivity}
\label{app:e1shift}

GPT-Image-2 \texttt{medium} displaces its returned crop more than any other arm. The share of specifications registered
with a shift above 1\,px is \capDocRegShareGtOneGiMedVTwo{} and above 2\,px \capDocRegShareGtTwoGiMedVTwo{}, against
\capDocRegShareGtOneGiLowVTwo{} and \capDocRegShareGtTwoGiLowVTwo{} for \texttt{low}, \capDocRegShareGtOneFlareVTwo{}
and \capDocRegShareGtTwoFlareVTwo{} for Flare, \capDocRegShareGtOneSunburstVTwo{} and \capDocRegShareGtTwoSunburstVTwo{}
for Sunburst, and \capDocRegShareGtOneFloorVTwo{} and \capDocRegShareGtTwoFloorVTwo{} for the noise replicas (median
shifts \capDocRegDispMedGiMedVTwo{}, \capDocRegDispMedGiLowVTwo{}, \capDocRegDispMedFlareVTwo{},
\capDocRegDispMedSunburstVTwo{} and \capDocRegDispMedFloorVTwo{}\,px).

Table~\ref{tab:e1shift} restricts both arms of
each contrast to rows whose locality registration shift is within 2 or 1\,px. Restricting on shift conditions on a
post-treatment variable and shrinks $n$, so these estimates are reported only as sensitivity analyses.

Registered re-reading keeps
the reduction at shifts of 1\,px or less, nominally significant against \texttt{low} and not against \texttt{medium};
re-reading the unregistered analysis's token set at registered coordinates, or using the unregistered metric, leaves no nominally
significant contrast there. The L2 result is therefore displacement-sensitive at shifts of 1\,px or less.

\begin{table*}[tbp]
  \centering
  \caption{E1 other text changed, Images~2.5 $-$ GPT-Image-2 in pp (unadjusted permutation $p$), with both arms
  restricted by locality registration shift. Registered: tokens re-read at registered coordinates, the primary metric.
  Old token set: the unregistered analysis's stable tokens re-read at registered coordinates. Unregistered: tokens
  read at identical coordinates, without registration. $n$: specifications with both arms at $\le$1\,px.}
  \label{tab:e1shift}
  \footnotesize
  \setlength{\tabcolsep}{3pt}
  \begin{tabular}{lrrrrrr}
    \toprule
    & Registered & Registered & Registered & Old token set & Unregistered & \\
    Contrast & all rows & $\le$2\,px & $\le$1\,px & $\le$1\,px & $\le$1\,px & $n$ \\
    \midrule
    Flare $-$ \texttt{low} & \capDocFlareMinusGiLowLTwoOtherVTwo{} (\capDocFlareMinusGiLowLTwoOtherVTwoP{}) & \capDocShiftLocLeTwoFlareMinusGiLowLTwoOtherVTwo{} (\capDocShiftLocLeTwoFlareMinusGiLowLTwoOtherVTwoP{}) & \capDocShiftLocLeOneFlareMinusGiLowLTwoOtherVTwo{} (\capDocShiftLocLeOneFlareMinusGiLowLTwoOtherVTwoP{}) & \capDocTokSetShiftLocLeOneFlareMinusGiLowLTwoOtherVTwo{} (\capDocTokSetShiftLocLeOneFlareMinusGiLowLTwoOtherVTwoP{}) & \capDocVOneShiftLocLeOneFlareMinusGiLowLTwoOtherVTwo{} (\capDocVOneShiftLocLeOneFlareMinusGiLowLTwoOtherVTwoP{}) & \capDocShiftLocLeOneFlareMinusGiLowLTwoOtherVTwoN{} \\
    Flare $-$ \texttt{medium} & \capDocFlareMinusGiMedLTwoOtherVTwo{} (\capDocFlareMinusGiMedLTwoOtherVTwoP{}) & \capDocShiftLocLeTwoFlareMinusGiMedLTwoOtherVTwo{} (\capDocShiftLocLeTwoFlareMinusGiMedLTwoOtherVTwoP{}) & \capDocShiftLocLeOneFlareMinusGiMedLTwoOtherVTwo{} (\capDocShiftLocLeOneFlareMinusGiMedLTwoOtherVTwoP{}) & \capDocTokSetShiftLocLeOneFlareMinusGiMedLTwoOtherVTwo{} (\capDocTokSetShiftLocLeOneFlareMinusGiMedLTwoOtherVTwoP{}) & \capDocVOneShiftLocLeOneFlareMinusGiMedLTwoOtherVTwo{} (\capDocVOneShiftLocLeOneFlareMinusGiMedLTwoOtherVTwoP{}) & \capDocShiftLocLeOneFlareMinusGiMedLTwoOtherVTwoN{} \\
    Sunburst $-$ \texttt{low} & \capDocSunburstMinusGiLowLTwoOtherVTwo{} (\capDocSunburstMinusGiLowLTwoOtherVTwoP{}) & \capDocShiftLocLeTwoSunburstMinusGiLowLTwoOtherVTwo{} (\capDocShiftLocLeTwoSunburstMinusGiLowLTwoOtherVTwoP{}) & \capDocShiftLocLeOneSunburstMinusGiLowLTwoOtherVTwo{} (\capDocShiftLocLeOneSunburstMinusGiLowLTwoOtherVTwoP{}) & \capDocTokSetShiftLocLeOneSunburstMinusGiLowLTwoOtherVTwo{} (\capDocTokSetShiftLocLeOneSunburstMinusGiLowLTwoOtherVTwoP{}) & \capDocVOneShiftLocLeOneSunburstMinusGiLowLTwoOtherVTwo{} (\capDocVOneShiftLocLeOneSunburstMinusGiLowLTwoOtherVTwoP{}) & \capDocShiftLocLeOneSunburstMinusGiLowLTwoOtherVTwoN{} \\
    Sunburst $-$ \texttt{medium} & \capDocSunburstMinusGiMedLTwoOtherVTwo{} (\capDocSunburstMinusGiMedLTwoOtherVTwoP{}) & \capDocShiftLocLeTwoSunburstMinusGiMedLTwoOtherVTwo{} (\capDocShiftLocLeTwoSunburstMinusGiMedLTwoOtherVTwoP{}) & \capDocShiftLocLeOneSunburstMinusGiMedLTwoOtherVTwo{} (\capDocShiftLocLeOneSunburstMinusGiMedLTwoOtherVTwoP{}) & \capDocTokSetShiftLocLeOneSunburstMinusGiMedLTwoOtherVTwo{} (\capDocTokSetShiftLocLeOneSunburstMinusGiMedLTwoOtherVTwoP{}) & \capDocVOneShiftLocLeOneSunburstMinusGiMedLTwoOtherVTwo{} (\capDocVOneShiftLocLeOneSunburstMinusGiMedLTwoOtherVTwoP{}) & \capDocShiftLocLeOneSunburstMinusGiMedLTwoOtherVTwoN{} \\
    \bottomrule
  \end{tabular}
\end{table*}

\subsection{Pixel-level versus tolerant measures of unintended change}
\label{app:locality}

Tolerant locality registers each output crop to its source (sub-pixel, with a \capLocGuardPx{}-px guard band and
\capLocBorderPx{}-px border excluded) and counts an off-field pixel as changed only if it differs by more than
\capLocThresh{} from every source pixel within \capLocTolPx{}. The resampling floor applies the same test to the
source after a round trip and a random shift of up to \capLocFloorShiftPx{}. Table~\ref{tab:locality} gives the
medians. The pixel-exact column is the metric of an earlier draft on the same rows: its floor with a sub-pixel
shift is \capLocFloorPaperMed{}, so most of what it counted is misregistration, and it is reported only to
reconcile the two.

Against GPT-Image-2 \texttt{low}, the paired tolerant differences hold on both corpora:
Flare \capLocPairGitwoLowFlareCordTolDelta{} [\capLocPairGitwoLowFlareCordTolDeltaLo{},
\capLocPairGitwoLowFlareCordTolDeltaHi{}] and Sunburst \capLocPairGitwoLowSunCordTolDelta{}
[\capLocPairGitwoLowSunCordTolDeltaLo{}, \capLocPairGitwoLowSunCordTolDeltaHi{}] on CORD, and
\capLocPairGitwoLowFlareWildTolDelta{} [\capLocPairGitwoLowFlareWildTolDeltaLo{},
\capLocPairGitwoLowFlareWildTolDeltaHi{}] and \capLocPairGitwoLowSunWildTolDelta{}
[\capLocPairGitwoLowSunWildTolDeltaLo{}, \capLocPairGitwoLowSunWildTolDeltaHi{}] on WildReceipt, and Images~2.5
is the more local arm on \capLocPairGitwoLowFlareTolMoreLocal{} (Flare) and \capLocPairGitwoLowSunTolMoreLocal{}
(Sunburst) of specifications.

Against \texttt{medium} the gap is larger (\capLocPairGitwoMedFlareTolDelta{} and
\capLocPairGitwoMedSunTolDelta{}). It is concentrated in rows where \texttt{medium} displaced the crop beyond the tolerance:
its median registration shift is \capLocGitwoMedShiftMed{}\,px against \capLocFlareShiftMed{}--\capLocSunShiftMed{}\,px.
With both arms restricted to shifts of at most 1\,px, \texttt{medium} $-$ Flare is
\capDocLocShiftLeOneGiMedMinusFlareTolVTwo{} [\capDocLocShiftLeOneGiMedMinusFlareTolVTwoLo{},
\capDocLocShiftLeOneGiMedMinusFlareTolVTwoHi{}] ($n$\,=\,\capDocLocShiftLeOneGiMedMinusFlareTolVTwoN{}) against
\capDocLocShiftAllGiMedMinusFlareTolVTwo{} on all rows, while \texttt{low} $-$ Flare keeps
\capDocLocShiftLeOneGiLowMinusFlareTolVTwo{} [\capDocLocShiftLeOneGiLowMinusFlareTolVTwoLo{},
\capDocLocShiftLeOneGiLowMinusFlareTolVTwoHi{}] ($n$\,=\,\capDocLocShiftLeOneGiLowMinusFlareTolVTwoN{}). These results link the gap against \texttt{medium} to displacement beyond the tolerance.
Tolerant locality is therefore reported only as exploratory. Inside the field every arm changes a similar share of pixels beyond
tolerance, so Images~2.5 is not simply editing less.

\begin{table*}[tbp]
  \centering
  \caption{E1 edit locality, context-crop denominator: medians over specifications with 95\% bootstrap intervals
  clustered by specification. ``Tolerant'' is the share of off-field crop pixels changed beyond \capLocThresh{}
  after registration and a \capLocTolPx{} tolerance; ``pixel-exact'' is the superseded metric on the same rows,
  shown only for reconciliation; ``inside'' is the tolerant share within the field; ``shift'' is the median
  registration displacement. Size-mismatched rows are excluded ($n$ excluded in the last column).}
  \label{tab:locality}
  \footnotesize
  \setlength{\tabcolsep}{4pt}
  \begin{tabular}{lrlrrrrr}
    \toprule
    Arm & $n$ & Tolerant median [95\% CI] & Pixel-exact & Inside & Shift (px) & Registered & Excl. \\
    \midrule
    GPT-Image-2 \texttt{low}    & \capLocGitwoLowN{} & \capLocGitwoLowTolMed{} [\capLocGitwoLowTolMedLo{}, \capLocGitwoLowTolMedHi{}] & \capLocGitwoLowPaperMed{} & \capLocGitwoLowInMed{} & \capLocGitwoLowShiftMed{} & \capLocGitwoLowRegistered{} & \capLocGitwoLowExcl{} \\
    GPT-Image-2 \texttt{medium} & \capLocGitwoMedN{} & \capLocGitwoMedTolMed{} [\capLocGitwoMedTolMedLo{}, \capLocGitwoMedTolMedHi{}] & \capLocGitwoMedPaperMed{} & \capLocGitwoMedInMed{} & \capLocGitwoMedShiftMed{} & \capLocGitwoMedRegistered{} & \capLocGitwoMedExcl{} \\
    Flare                       & \capLocFlareN{}    & \capLocFlareTolMed{} [\capLocFlareTolMedLo{}, \capLocFlareTolMedHi{}]          & \capLocFlarePaperMed{}    & \capLocFlareInMed{}    & \capLocFlareShiftMed{}    & \capLocFlareRegistered{}    & \capLocFlareExcl{} \\
    Sunburst                    & \capLocSunN{}      & \capLocSunTolMed{} [\capLocSunTolMedLo{}, \capLocSunTolMedHi{}]                & \capLocSunPaperMed{}      & \capLocSunInMed{}      & \capLocSunShiftMed{}      & \capLocSunRegistered{}      & \capLocSunExcl{} \\
    \midrule
    Resampling floor            & \capLocFloorN{}    & \capLocFloorTolMed{} [\capLocFloorTolMedLo{}, \capLocFloorTolMedHi{}]          & \capLocFloorPaperMed{}    & ---                    & ---                       & ---                         & --- \\
    \bottomrule
  \end{tabular}
\end{table*}

\subsection{Key to the receipt pairs}
\label{app:figonekey}

\begin{figure*}[tbp]
  \centering
  \includegraphics[width=0.72\textwidth]{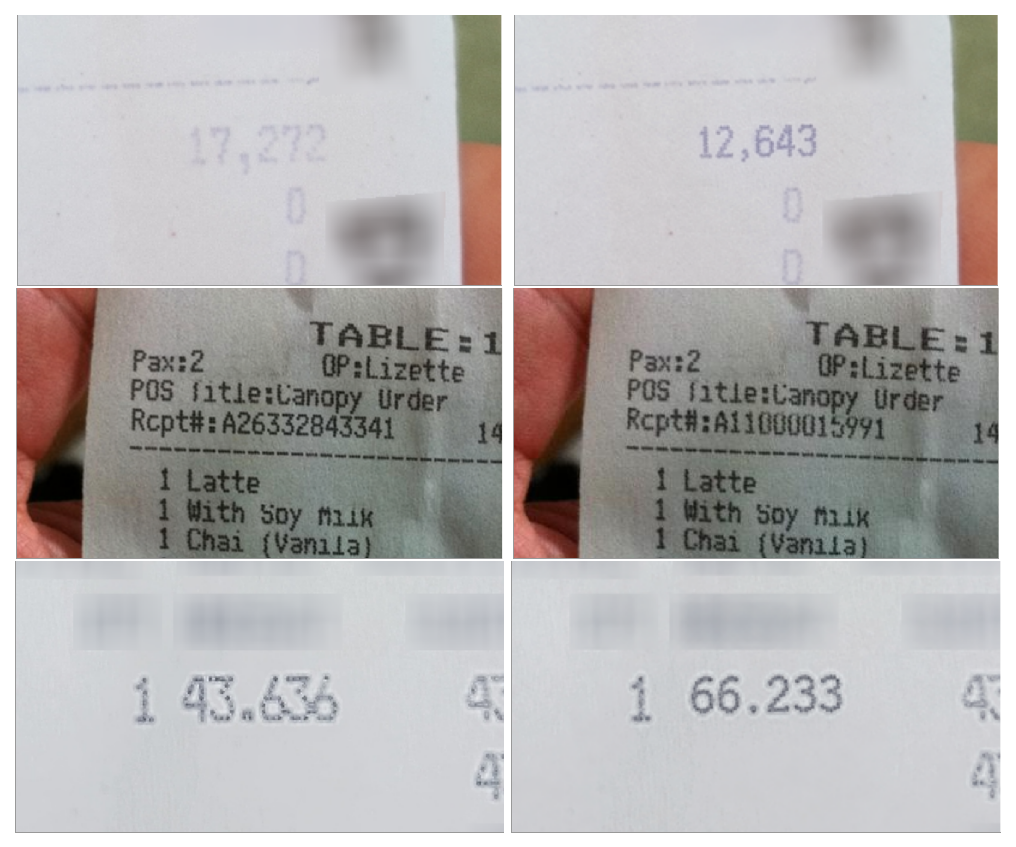}
  \caption{Three authentic--edited pairs of context crops from AIForge-Doc round~v3, produced by
  \texttt{gpt-image-2.5-flare} at \texttt{medium} for a median of \$\docsMedianCost{} and \docsMedianLatency{}~s
  each. In each pair one panel is authentic and one has a single field edited; which panel varies by row and is
  not marked. Key in Appendix~\ref{app:figonekey}. The pairs illustrate the edits; they are not a human discrimination test.}
  \label{fig:pairs}
\end{figure*}

Figure~\ref{fig:pairs}'s rows are authentic and edited context crops from \texttt{\figoneModel{}} at
\texttt{medium}. Edited panel by row: \figoneOnePanel{} (\figoneOneCorpus{}, \figoneOneField{}),
\figoneTwoPanel{} (\figoneTwoCorpus{}, \figoneTwoField{}) and \figoneThreePanel{} (\figoneThreeCorpus{},
\figoneThreeField{}). Forged values, identifiers and boxes are in the released answer key (Appendix~\ref{app:release}).

\section{E2: Keep Editing the Same Image}
\label{app:e2}

\paragraph{E2a procedure.} Each of \capDocChainAllReceipts{} receipts (\capDocChainCordReceipts{} CORD,
\capDocChainWildReceipts{} WildReceipt) is edited \capDocChainTurns{} times by each model (GPT-Image-2 at \texttt{medium} and, added post hoc,
\texttt{low}), statelessly, one field per
turn, with the unmasked instruction ``change `$o$' to `$t$'; keep everything else'' in a seeded field order shared
by all models. Targets keep the original's character layout and are unique on the receipt; never-edited control
fields are tracked too (\capDocChainCordPlannedFields{} planned fields on CORD, \capDocChainWildPlannedFields{} on
WildReceipt).

After each turn, OCR reads the receipt. An ORB and RANSAC similarity fit locates
each field's line. This task uses neither object detection nor a pixel-drift measure. Fields were pre-selected to be OCR-legible
at $t_0$, so the WildReceipt baseline (\capDocChainWildBaseline{}) is close to automatic and says nothing about OCR
reliability on photographed receipts.

Failure modes are reported outside the Holm family: the value written into
the wrong row (CORD: \capDocChainCordElsewhereGiTwo{} GPT-Image-2 \texttt{medium}, \capDocChainCordElsewhereFlare{} Flare,
\capDocChainCordElsewhereSunburst{} Sunburst) and near misses at edit distance one or two
(\capDocChainCordNearMissGiTwo{}, \capDocChainCordNearMissFlare{}, \capDocChainCordNearMissSunburst{}). No E2a
request was refused.

\paragraph{E2b procedure.} Every model starts from the same \capChainNSeedsPlan{} person-free authentic
photographs drawn from \capChainNSourcePhotosPlan{} source photos (IMD2020 \cite{novozamsky2020imd2020} and the
Columbia set \cite{hsu2006columbia}). One edit per turn, in a seeded order, adds either text (a sign code, a date
stamp, a sticky-note number) read by OCR, or an object (a red mug, a blue umbrella, a potted plant) found by OWLv2 and checked by a Lab colour test.

Object thresholds $\tau$ were set on \capChainCalibNPool{} authentic
candidate crops, giving pool false-positive rates of \capChainFprMugPool{} (mug, $\tau$\,=\,\capChainTauMug{}),
\capChainFprUmbrellaPool{} (umbrella, \capChainTauUmbrella{}) and \capChainFprPlantPool{} (plant,
\capChainTauPlant{}).

An edit is \emph{in place} if its box at the last turn overlaps the box at its own turn with
IoU $\ge$\capChainInPlaceIou{}; \emph{relocated} if present only elsewhere; \emph{lost} if absent. Outside-edit
drift is the registered \capLocTolPx{} changed share outside the union of edit boxes at the final turn against
$t_0$, reported with in-box edit strength.

The conversational variant relays each instruction through
\capChainConvTextModel{} in one Responses-API conversation \cite{openai2026responsesapi}; that model can rephrase
placement, so the variants differ in more than state.

Seed \texttt{pcv2-020} was blocked by moderation
(\capChainNRefusedChainTurns{} refused chain turns in total), leaving incomplete chains \capChainStatelessIncompleteList{}
(stateless) and \capChainConvIncompleteList{} (conversational); they are listed, not silently dropped, and units are
complete chains. Table~\ref{tab:e2b} gives every endpoint.

\begin{table*}[tbp]
  \centering
  \caption{E2b endpoints per variant and model, CIs clustered by source photo. Success: edit detected at its own
  turn. Present: among successes, present anywhere at the last turn (L1 survival; lost is its complement). In
  place: among survivors, still in place (L2). Relocated: share of successes present only elsewhere. Drift: changed
  share outside all edit boxes, final turn against $t_0$; strength: changed share inside the edit boxes. The
  GPT-Image-2 \texttt{low} arm, added post hoc, was scored on the primary endpoints only. Primary contrasts are in
  Table~\ref{tab:results}; the conversational variant is exploratory.}
  \label{tab:e2b}
  \footnotesize
  \setlength{\tabcolsep}{4pt}
  \begin{tabular}{llrrrrrrrr}
    \toprule
    Variant & Model & Chains & Success & Present [95\% CI] & & In place [95\% CI] & Relocated & Drift & Strength \\
    \midrule
    Stateless & GPT-Image-2 \texttt{med.} & \capChainStatelessGiNChains{} & \capChainStatelessGiLOneSuccess{} & \capChainStatelessGiLOneSurvival{} [\capChainStatelessGiLOneSurvivalLo{}, \capChainStatelessGiLOneSurvivalHi{}] & & \capChainStatelessGiLTwoInPlace{} [\capChainStatelessGiLTwoInPlaceLo{}, \capChainStatelessGiLTwoInPlaceHi{}] & \capChainStatelessGiLTwoRelocation{} & \capChainStatelessGiLTwoDrift{} & \capChainStatelessGiLTwoStrength{} \\
              & GPT-Image-2 \texttt{low} & \capChainStatelessGiLowNChainsVTwo{} & --- & \capChainStatelessGiLowLOneSurvivalVTwo{} [\capChainStatelessGiLowLOneSurvivalVTwoLo{}, \capChainStatelessGiLowLOneSurvivalVTwoHi{}] & & \capChainStatelessGiLowLTwoInPlaceVTwo{} [\capChainStatelessGiLowLTwoInPlaceVTwoLo{}, \capChainStatelessGiLowLTwoInPlaceVTwoHi{}] & --- & --- & --- \\
              & Flare & \capChainStatelessFlareNChains{} & \capChainStatelessFlareLOneSuccess{} & \capChainStatelessFlareLOneSurvival{} [\capChainStatelessFlareLOneSurvivalLo{}, \capChainStatelessFlareLOneSurvivalHi{}] & & \capChainStatelessFlareLTwoInPlace{} [\capChainStatelessFlareLTwoInPlaceLo{}, \capChainStatelessFlareLTwoInPlaceHi{}] & \capChainStatelessFlareLTwoRelocation{} & \capChainStatelessFlareLTwoDrift{} & \capChainStatelessFlareLTwoStrength{} \\
              & Sunburst & \capChainStatelessSunburstNChains{} & \capChainStatelessSunburstLOneSuccess{} & \capChainStatelessSunburstLOneSurvival{} [\capChainStatelessSunburstLOneSurvivalLo{}, \capChainStatelessSunburstLOneSurvivalHi{}] & & \capChainStatelessSunburstLTwoInPlace{} [\capChainStatelessSunburstLTwoInPlaceLo{}, \capChainStatelessSunburstLTwoInPlaceHi{}] & \capChainStatelessSunburstLTwoRelocation{} & \capChainStatelessSunburstLTwoDrift{} & \capChainStatelessSunburstLTwoStrength{} \\
    \midrule
    Conversational & GPT-Image-2 \texttt{med.} & \capChainConvGiNChains{} & \capChainConvGiLOneSuccess{} & \capChainConvGiLOneSurvival{} [\capChainConvGiLOneSurvivalLo{}, \capChainConvGiLOneSurvivalHi{}] & & \capChainConvGiLTwoInPlace{} [\capChainConvGiLTwoInPlaceLo{}, \capChainConvGiLTwoInPlaceHi{}] & \capChainConvGiLTwoRelocation{} & \capChainConvGiLTwoDrift{} & \capChainConvGiLTwoStrength{} \\
              & Flare & \capChainConvFlareNChains{} & \capChainConvFlareLOneSuccess{} & \capChainConvFlareLOneSurvival{} [\capChainConvFlareLOneSurvivalLo{}, \capChainConvFlareLOneSurvivalHi{}] & & \capChainConvFlareLTwoInPlace{} [\capChainConvFlareLTwoInPlaceLo{}, \capChainConvFlareLTwoInPlaceHi{}] & \capChainConvFlareLTwoRelocation{} & \capChainConvFlareLTwoDrift{} & \capChainConvFlareLTwoStrength{} \\
              & Sunburst & \capChainConvSunburstNChains{} & \capChainConvSunburstLOneSuccess{} & \capChainConvSunburstLOneSurvival{} [\capChainConvSunburstLOneSurvivalLo{}, \capChainConvSunburstLOneSurvivalHi{}] & & \capChainConvSunburstLTwoInPlace{} [\capChainConvSunburstLTwoInPlaceLo{}, \capChainConvSunburstLTwoInPlaceHi{}] & \capChainConvSunburstLTwoRelocation{} & \capChainConvSunburstLTwoDrift{} & \capChainConvSunburstLTwoStrength{} \\
    \bottomrule
  \end{tabular}
\end{table*}

\paragraph{Umbrella colour-share sensitivity.} The umbrella's blue-share threshold was lowered from
\capChainUmbrellaShareSens{} to \capChainUmbrellaShare{} after a single Flare conversational pilot miss, and then
frozen. Re-scoring the whole primary family at the pre-revision \capChainUmbrellaShareSens{} leaves every stateless
decision unchanged (largest effect \capChainSensStatelessFlareMinusGiLTwoInPlace{}, Holm
$p$\,=\,\capChainSensStatelessFlareMinusGiLTwoInPlacePHolm{}).

In the conversational variant, Sunburst's in-place
gain is \capChainSensConvSunburstMinusGiLTwoInPlace{} [\capChainSensConvSunburstMinusGiLTwoInPlaceLo{},
\capChainSensConvSunburstMinusGiLTwoInPlaceHi{}] with Holm $p$\,=\,\capChainSensConvSunburstMinusGiLTwoInPlacePHolm{},
against \capChainConvSunburstMinusGiLTwoInPlace{} (Holm $p$\,=\,\capChainConvSunburstMinusGiLTwoInPlacePHolm{})
under the frozen threshold, so its per-variant Holm decision depended on the threshold; the conversational
variant is now exploratory and no verdict rests on it. Flare's conversational in-place difference
is \capChainSensConvFlareMinusGiLTwoInPlace{} [\capChainSensConvFlareMinusGiLTwoInPlaceLo{},
\capChainSensConvFlareMinusGiLTwoInPlaceHi{}] (Holm $p$\,=\,\capChainSensConvFlareMinusGiLTwoInPlacePHolm{}), and
no L1 survival contrast survives under either threshold.

\paragraph{E2a: all arms, family breadth and registration.} Table~\ref{tab:e2arms} gives every E2a endpoint for all
four arms. Survival is conditional on success, which leaves Flare the smaller denominator; the joint endpoint shows the
CORD deficit without that conditioning (\S\ref{sec:cap-multiturn}).

In a narrower family of the
\capDocChainNTestsTwoCorpusVTwo{} E2a tests alone, Flare's CORD contrasts have Holm $p$
\capDocChainTwoCorpusCordFlareVsGiMedPHolmMinVTwo{}--\capDocChainTwoCorpusCordFlareVsGiMedPHolmMaxVTwo{} against
\texttt{medium} and \capDocChainTwoCorpusCordFlareVsGiLowPHolmMinVTwo{}--\capDocChainTwoCorpusCordFlareVsGiLowPHolmMaxVTwo{}
against \texttt{low}, and the success deficit against \texttt{low} survives there
(\capDocChainCordSuccDiffFlareVsGiLowVTwoPHolmTwoCorpus{}) but not in the claim family.

Pooled over corpora, Flare's
survival difference is \capDocChainAllSurvFourDiffFlareVsGiMedVTwo{} against \texttt{medium} and
\capDocChainAllSurvFourDiffFlareVsGiLowVTwo{} against \texttt{low}, so the survival deficit is specific to CORD.

The
median scale of the final-turn page relative to the source is \capDocChainScaleFourGiMedVTwo{} for \texttt{medium},
\capDocChainScaleFourGiLowVTwo{} for \texttt{low}, \capDocChainScaleFourFlareVTwo{} for Flare and
\capDocChainScaleFourSunburstVTwo{} for Sunburst, but \capDocChainScaleFourShareBelowNinetySevenGiMedVTwo{} of
\texttt{medium}'s final receipts are shrunk by more than 3\%, against \capDocChainScaleFourShareBelowNinetySevenFlareVTwo{}
of Flare's. Smaller glyphs would tend to reduce the control's OCR score. This direction of
measurement bias does not explain Flare's lower score.

\begin{table*}[tbp]
  \centering
  \caption{E2a endpoints by arm: rate (count). Success: exact target on its own line at its own turn. Survival: among
  edits that succeeded at turns 1--3, still read at the last turn. Joint: success and survival together, without
  conditioning. Never-edited kept: control fields still read their original value at the last turn. GPT-Image-2
  \texttt{low} was added post hoc.}
  \label{tab:e2arms}
  \footnotesize
  \setlength{\tabcolsep}{4pt}
  \begin{tabular}{llrrrr}
    \toprule
    Corpus & Endpoint & GPT-Image-2 \texttt{medium} & GPT-Image-2 \texttt{low} & Flare & Sunburst \\
    \midrule
    CORD & success & \capDocChainCordSuccGiMedVTwo{} (\capDocChainCordSuccGiMedVTwoN{}) & \capDocChainCordSuccGiLowVTwo{} (\capDocChainCordSuccGiLowVTwoN{}) & \capDocChainCordSuccFlareVTwo{} (\capDocChainCordSuccFlareVTwoN{}) & \capDocChainCordSuccSunburstVTwo{} (\capDocChainCordSuccSunburstVTwoN{}) \\
     & survival & \capDocChainCordSurvFourGiMedVTwo{} (\capDocChainCordSurvFourGiMedVTwoN{}) & \capDocChainCordSurvFourGiLowVTwo{} (\capDocChainCordSurvFourGiLowVTwoN{}) & \capDocChainCordSurvFourFlareVTwo{} (\capDocChainCordSurvFourFlareVTwoN{}) & \capDocChainCordSurvFourSunburstVTwo{} (\capDocChainCordSurvFourSunburstVTwoN{}) \\
     & joint & \capDocChainCordJointGiMedVTwo{} (\capDocChainCordJointGiMedVTwoN{}) & \capDocChainCordJointGiLowVTwo{} (\capDocChainCordJointGiLowVTwoN{}) & \capDocChainCordJointFlareVTwo{} (\capDocChainCordJointFlareVTwoN{}) & \capDocChainCordJointSunburstVTwo{} (\capDocChainCordJointSunburstVTwoN{}) \\
     & never-edited kept & \capDocChainCordUntouchedGiMedVTwo{} (\capDocChainCordUntouchedGiMedVTwoN{}) & \capDocChainCordUntouchedGiLowVTwo{} (\capDocChainCordUntouchedGiLowVTwoN{}) & \capDocChainCordUntouchedFlareVTwo{} (\capDocChainCordUntouchedFlareVTwoN{}) & \capDocChainCordUntouchedSunburstVTwo{} (\capDocChainCordUntouchedSunburstVTwoN{}) \\
    \midrule
    WildReceipt & success & \capDocChainWildSuccGiMedVTwo{} (\capDocChainWildSuccGiMedVTwoN{}) & \capDocChainWildSuccGiLowVTwo{} (\capDocChainWildSuccGiLowVTwoN{}) & \capDocChainWildSuccFlareVTwo{} (\capDocChainWildSuccFlareVTwoN{}) & \capDocChainWildSuccSunburstVTwo{} (\capDocChainWildSuccSunburstVTwoN{}) \\
     & survival & \capDocChainWildSurvFourGiMedVTwo{} (\capDocChainWildSurvFourGiMedVTwoN{}) & \capDocChainWildSurvFourGiLowVTwo{} (\capDocChainWildSurvFourGiLowVTwoN{}) & \capDocChainWildSurvFourFlareVTwo{} (\capDocChainWildSurvFourFlareVTwoN{}) & \capDocChainWildSurvFourSunburstVTwo{} (\capDocChainWildSurvFourSunburstVTwoN{}) \\
     & joint & \capDocChainWildJointGiMedVTwo{} (\capDocChainWildJointGiMedVTwoN{}) & \capDocChainWildJointGiLowVTwo{} (\capDocChainWildJointGiLowVTwoN{}) & \capDocChainWildJointFlareVTwo{} (\capDocChainWildJointFlareVTwoN{}) & \capDocChainWildJointSunburstVTwo{} (\capDocChainWildJointSunburstVTwoN{}) \\
     & never-edited kept & \capDocChainWildUntouchedGiMedVTwo{} (\capDocChainWildUntouchedGiMedVTwoN{}) & \capDocChainWildUntouchedGiLowVTwo{} (\capDocChainWildUntouchedGiLowVTwoN{}) & \capDocChainWildUntouchedFlareVTwo{} (\capDocChainWildUntouchedFlareVTwoN{}) & \capDocChainWildUntouchedSunburstVTwo{} (\capDocChainWildUntouchedSunburstVTwoN{}) \\
    \midrule
    Pooled & success & \capDocChainAllSuccGiMedVTwo{} (\capDocChainAllSuccGiMedVTwoN{}) & \capDocChainAllSuccGiLowVTwo{} (\capDocChainAllSuccGiLowVTwoN{}) & \capDocChainAllSuccFlareVTwo{} (\capDocChainAllSuccFlareVTwoN{}) & \capDocChainAllSuccSunburstVTwo{} (\capDocChainAllSuccSunburstVTwoN{}) \\
     & survival & \capDocChainAllSurvFourGiMedVTwo{} (\capDocChainAllSurvFourGiMedVTwoN{}) & \capDocChainAllSurvFourGiLowVTwo{} (\capDocChainAllSurvFourGiLowVTwoN{}) & \capDocChainAllSurvFourFlareVTwo{} (\capDocChainAllSurvFourFlareVTwoN{}) & \capDocChainAllSurvFourSunburstVTwo{} (\capDocChainAllSurvFourSunburstVTwoN{}) \\
     & joint & \capDocChainAllJointGiMedVTwo{} (\capDocChainAllJointGiMedVTwoN{}) & \capDocChainAllJointGiLowVTwo{} (\capDocChainAllJointGiLowVTwoN{}) & \capDocChainAllJointFlareVTwo{} (\capDocChainAllJointFlareVTwoN{}) & \capDocChainAllJointSunburstVTwo{} (\capDocChainAllJointSunburstVTwoN{}) \\
     & never-edited kept & \capDocChainAllUntouchedGiMedVTwo{} (\capDocChainAllUntouchedGiMedVTwoN{}) & \capDocChainAllUntouchedGiLowVTwo{} (\capDocChainAllUntouchedGiLowVTwoN{}) & \capDocChainAllUntouchedFlareVTwo{} (\capDocChainAllUntouchedFlareVTwoN{}) & \capDocChainAllUntouchedSunburstVTwo{} (\capDocChainAllUntouchedSunburstVTwoN{}) \\
    \bottomrule
  \end{tabular}
\end{table*}

\paragraph{E2b: the \texttt{low} arm and the conversational variant.} The GPT-Image-2 \texttt{low} arm completed
\capChainStatelessGiLowNChainsVTwo{} stateless chains on \capChainStatelessGiLowNSourcePhotosVTwo{} source photos.
Against it, Flare's and Sunburst's survival differences are \capChainStatelessFlareMinusGiLowLOneSurvivalVTwo{} and
\capChainStatelessSunburstMinusGiLowLOneSurvivalVTwo{} and their in-place differences
\capChainStatelessFlareMinusGiLowLTwoInPlaceVTwo{} and \capChainStatelessSunburstMinusGiLowLTwoInPlaceVTwo{}, all with
Holm $p$\,=\,\capChainStatelessFlareMinusGiLowLOneSurvivalVTwoPHolm{}.

The conversational contrasts are exploratory.
Sunburst's in-place gain (\capChainConvSunburstMinusGiMedLTwoInPlaceVTwo{} [\capChainConvSunburstMinusGiMedLTwoInPlaceVTwoLo{},
\capChainConvSunburstMinusGiMedLTwoInPlaceVTwoHi{}], unadjusted permutation
$p$\,=\,\capChainConvSunburstMinusGiMedLTwoInPlaceVTwoP{}) has Holm $p$\,=\,\capChainConvSunburstMinusGiMedLTwoInPlaceVTwoPHolmBothVariants{}
in a family of both E2b variants and \capChainConvSunburstMinusGiMedLTwoInPlaceVTwoPHolmInclExploratory{} in the
multi-turn family; Flare's (\capChainConvFlareMinusGiMedLTwoInPlaceVTwo{}) has
\capChainConvFlareMinusGiMedLTwoInPlaceVTwoPHolmBothVariants{} and
\capChainConvFlareMinusGiMedLTwoInPlaceVTwoPHolmInclExploratory{}.

The variant has \capChainConvFlareNSourcePhotosVTwo{}
clusters, one per chain, so its bootstrap intervals may under-cover. The relay model's rewritten instructions were not
logged, so a placement difference may arise in the relay rather than in the image model, and no estimate excluding the
pilot seeds is reported.

\section{E3: Drop a Product into a Fake Listing}
\label{app:e3}
\label{app:reference}

\paragraph{Procedure.}
We create \capRefPNRefs{} programmatic packshots: boxes, bottles or cans with
a nonsense brand, price, small SKU code and two-colour logo. OCR verifies the
JSON ground truth on the render. Because no compared generator supplies the
references, the design avoids an advantage from reproducing its own outputs.

Each model inserts each reference in two settings: a front-on \emph{hero}
shot and an \emph{in-scene} shot with a small product among other objects.
Images~2.5 uses \texttt{medium}; GPT-Image-2 uses \texttt{medium} and the
post hoc \texttt{low} arm, for \capRefPNCalls{} calls in total.
A SIFT/RANSAC homography with fixed acceptance thresholds maps the label to
reference coordinates. Failed homographies count as non-preservation;
labels are found in \capRefPNLabelFound{} of \capRefPNCalls{} outputs.

L1 measures exact brand, SKU and price OCR, plus logo $\Delta E_{00}$ before
and after white balance. Hero-shot SKU exactness is near ceiling for every
model (\capRefPHeroSkuExactFlare{}--\capRefPHeroSkuExactSunburst{}).
In-scene brand and price exactness are \capRefPInSceneBrandExactGi{}
and \capRefPInScenePriceExactGi{} for GPT-Image-2,
\capRefPInSceneBrandExactFlare{} and \capRefPInScenePriceExactFlare{}
for Flare, and \capRefPInSceneBrandExactSunburst{} and
\capRefPInScenePriceExactSunburst{} for Sunburst.

These references are flat programmatic packshots. The rendered set with
photographic lighting provides an exploratory check
(\S\ref{sec:cap-reference}).

\paragraph{Size control against \texttt{medium} (initial analysis).} GPT-Image-2 draws the in-scene SKU smaller (median glyph height
\capRefPInSceneSkuHeightGi{}\,px against \capRefPInSceneSkuHeightFlare{} and \capRefPInSceneSkuHeightSunburst{}), and
OCR accuracy depends steeply on glyph height. We use three size-adjusted estimators to examine the pooled Images~2.5 $-$
GPT-Image-2 gap of \capRefPInSceneSizeRawGapTwoFive{} [\capRefPInSceneSizeRawGapLoTwoFive{},
\capRefPInSceneSizeRawGapHiTwoFive{}].

Within glyph-height bins (each arm $\ge$\capRefPSizeMinBinN{} outputs) the
exact-match gap is \capRefPInSceneSizeMatchedGapTwoFive{} [\capRefPInSceneSizeMatchedGapLoTwoFive{},
\capRefPInSceneSizeMatchedGapHiTwoFive{}]. Normalised by bin, it is \capRefPInSceneSizeMatchedGapNormTwoFive{}
[\capRefPInSceneSizeMatchedGapNormLoTwoFive{}, \capRefPInSceneSizeMatchedGapNormHiTwoFive{}]. A logistic model
with log glyph height gives \capRefPSensLogisticGapTwoFive{} [\capRefPSensLogisticGapLoTwoFive{},
\capRefPSensLogisticGapHiTwoFive{}]. They leave \capRefPSensSurvivingLogisticPctTwoFive{}--\capRefPInSceneSizeSurvivingPctTwoFive{}
of the raw gap.

The OCR ceiling estimates the resolution-related component of the gap. For each
output, we take the reference
reduced to the output's SKU height and blurred to its sharpness (Gaussian $\sigma$ up to \capRefPSizeCeilingSigma{}) and read it with the same OCR. The ceiling alone predicts a gap of \capRefPInSceneSizeCeilingGapTwoFive{}
[\capRefPInSceneSizeCeilingGapLoTwoFive{}, \capRefPInSceneSizeCeilingGapHiTwoFive{}], leaving a residual of
\capRefPInSceneSizeResidualGapTwoFive{} [\capRefPInSceneSizeResidualGapLoTwoFive{},
\capRefPInSceneSizeResidualGapHiTwoFive{}]. This accounts for the OCR-measured gap against \texttt{medium}; it does not establish
equal underlying fidelity.

Against \texttt{low}, the pooled
raw gap is \capRefPInSceneSizeRawGapTwoFiveVsGiLow{} [\capRefPInSceneSizeRawGapTwoFiveVsGiLowCILo{},
\capRefPInSceneSizeRawGapTwoFiveVsGiLowCIHi{}], the binned matched gap \capRefPInSceneSizeMatchedGapTwoFiveVsGiLow{}
[\capRefPInSceneSizeMatchedGapTwoFiveVsGiLowCILo{}, \capRefPInSceneSizeMatchedGapTwoFiveVsGiLowCIHi{}] and the
ceiling residual \capRefPInSceneSizeResidualGapTwoFiveVsGiLow{} [\capRefPInSceneSizeResidualGapTwoFiveVsGiLowCILo{},
\capRefPInSceneSizeResidualGapTwoFiveVsGiLowCIHi{}].

\paragraph{Placement and size (post hoc).} The in-scene prompt asks for a product about a tenth of the frame wide.
Table~\ref{tab:e3place} gives, per arm, the label's share of the frame, the product's width, the share of outputs
wider than a tenth of the frame, glyph height per unit label scale and logo mask IoU. Images~2.5 places the in-scene
label \capRefPlProgInSceneFlareVsGiAreaRatio{}$\times$ (Flare) and \capRefPlProgInSceneSunburstVsGiAreaRatio{}$\times$
(Sunburst) as large as \texttt{medium} does, and \capRefPlProgInSceneFlareVsGiLowAreaRatio{}$\times$ and
\capRefPlProgInSceneSunburstVsGiLowAreaRatio{}$\times$ as large as \texttt{low}. In hero shots placement is similar
(\capRefPlProgHeroFlareVsGiAreaRatio{}$\times$ and \capRefPlProgHeroSunburstVsGiAreaRatio{}$\times$) and SKU exactness
does not differ (Flare \capRefPlProgHeroFlareVsGiSku{} [\capRefPlProgHeroFlareVsGiSkuLo{},
\capRefPlProgHeroFlareVsGiSkuHi{}], Sunburst \capRefPlProgHeroSunburstVsGiSku{} [\capRefPlProgHeroSunburstVsGiSkuLo{},
\capRefPlProgHeroSunburstVsGiSkuHi{}] against \texttt{medium}).

Three size-matched estimators are exploratory. Matched
absolute glyph height gives \capRefPlProgInSceneFlareVsGiMatchedAbs{} [\capRefPlProgInSceneFlareVsGiMatchedAbsLo{},
\capRefPlProgInSceneFlareVsGiMatchedAbsHi{}] and \capRefPlProgInSceneSunburstVsGiMatchedAbs{}
[\capRefPlProgInSceneSunburstVsGiMatchedAbsLo{}, \capRefPlProgInSceneSunburstVsGiMatchedAbsHi{}] against \texttt{medium}
(unadjusted permutation $p$ \capRefPlProgInSceneFlareVsGiMatchedAbsPPerm{} and
\capRefPlProgInSceneSunburstVsGiMatchedAbsPPerm{}) and \capRefPlProgInSceneFlareVsGiLowMatchedAbs{}
[\capRefPlProgInSceneFlareVsGiLowMatchedAbsLo{}, \capRefPlProgInSceneFlareVsGiLowMatchedAbsHi{}] and
\capRefPlProgInSceneSunburstVsGiLowMatchedAbs{} [\capRefPlProgInSceneSunburstVsGiLowMatchedAbsLo{},
\capRefPlProgInSceneSunburstVsGiLowMatchedAbsHi{}] against \texttt{low} (\capRefPlProgInSceneFlareVsGiLowMatchedAbsPPerm{}
and \capRefPlProgInSceneSunburstVsGiLowMatchedAbsPPerm{}). Matched placement scale is in \S\ref{sec:cap-reference}.
Matched relative glyph height is uninformative, because relative size is constant by construction, and returns the raw
gap (\capRefPlProgInSceneFlareVsGiMatchedRel{} and \capRefPlProgInSceneSunburstVsGiMatchedRel{}).

Logo $\Delta E_{00}$
after white balance is lower on Images~2.5 (in scene against \texttt{low}: \capRefPlProgInSceneFlareVsGiLowLogoDEwb{}
[\capRefPlProgInSceneFlareVsGiLowLogoDEwbLo{}, \capRefPlProgInSceneFlareVsGiLowLogoDEwbHi{}] and
\capRefPlProgInSceneSunburstVsGiLowLogoDEwb{} [\capRefPlProgInSceneSunburstVsGiLowLogoDEwbLo{},
\capRefPlProgInSceneSunburstVsGiLowLogoDEwbHi{}]).

Excluding the \capRefPlProgInSceneFlareVsGiSkuBatchOnlyExcluded{}
references with synchronously filled rows leaves the raw in-scene contrasts at
\capRefPlProgInSceneFlareVsGiSkuBatchOnly{} and \capRefPlProgInSceneSunburstVsGiSkuBatchOnly{} against \texttt{medium}
and \capRefPlProgInSceneFlareVsGiLowSkuBatchOnly{} and \capRefPlProgInSceneSunburstVsGiLowSkuBatchOnly{} against
\texttt{low}.

\begin{table}[tbp]
  \centering
  \caption{E3 placement per arm. Area: label share of the frame. Width: product width as a share of the frame. Wider:
  share of outputs wider than a tenth of the frame. Rel.\ px: SKU glyph height per unit label scale. IoU: logo mask
  IoU. Medians except for shares.}
  \label{tab:e3place}
  \footnotesize
  \setlength{\tabcolsep}{3pt}
  \begin{tabular}{llrrrr}
    \toprule
    & Metric & \shortstack{GPT-Image-2\\\texttt{med.}} & \shortstack{GPT-Image-2\\\texttt{low}} & Flare & Sunburst \\
    \midrule
    In scene & Area & \capRefPlProgInSceneAreaPctGi{} & \capRefPlProgInSceneAreaPctGiLow{} & \capRefPlProgInSceneAreaPctFlare{} & \capRefPlProgInSceneAreaPctSunburst{} \\
     & Width & \capRefPlProgInSceneWidthPctGi{} & \capRefPlProgInSceneWidthPctGiLow{} & \capRefPlProgInSceneWidthPctFlare{} & \capRefPlProgInSceneWidthPctSunburst{} \\
     & Wider & \capRefPlProgInSceneWidthAboveTenthPctGi{} & \capRefPlProgInSceneWidthAboveTenthPctGiLow{} & \capRefPlProgInSceneWidthAboveTenthPctFlare{} & \capRefPlProgInSceneWidthAboveTenthPctSunburst{} \\
     & Rel.\ px & \capRefPlProgInSceneRelPxGi{} & \capRefPlProgInSceneRelPxGiLow{} & \capRefPlProgInSceneRelPxFlare{} & \capRefPlProgInSceneRelPxSunburst{} \\
     & IoU & \capRefPlProgInSceneLogoIouGi{} & \capRefPlProgInSceneLogoIouGiLow{} & \capRefPlProgInSceneLogoIouFlare{} & \capRefPlProgInSceneLogoIouSunburst{} \\
    \midrule
    Hero & Area & \capRefPlProgHeroAreaPctGi{} & \capRefPlProgHeroAreaPctGiLow{} & \capRefPlProgHeroAreaPctFlare{} & \capRefPlProgHeroAreaPctSunburst{} \\
     & Width & \capRefPlProgHeroWidthPctGi{} & \capRefPlProgHeroWidthPctGiLow{} & \capRefPlProgHeroWidthPctFlare{} & \capRefPlProgHeroWidthPctSunburst{} \\
     & Wider & \capRefPlProgHeroWidthAboveTenthPctGi{} & \capRefPlProgHeroWidthAboveTenthPctGiLow{} & \capRefPlProgHeroWidthAboveTenthPctFlare{} & \capRefPlProgHeroWidthAboveTenthPctSunburst{} \\
     & Rel.\ px & \capRefPlProgHeroRelPxGi{} & \capRefPlProgHeroRelPxGiLow{} & \capRefPlProgHeroRelPxFlare{} & \capRefPlProgHeroRelPxSunburst{} \\
     & IoU & \capRefPlProgHeroLogoIouGi{} & \capRefPlProgHeroLogoIouGiLow{} & \capRefPlProgHeroLogoIouFlare{} & \capRefPlProgHeroLogoIouSunburst{} \\
    \bottomrule
  \end{tabular}
\end{table}

\paragraph{AI-agent spot check of the scorer (not a human rating).} An AI coding agent (Claude) read \capRefPBlindNAll{} blind
in-scene crops and judged each SKU correct, wrong or illegible. For GPT-Image-2 it judged \capRefPBlindCorrectGi{}
correct, \capRefPBlindWrongGi{} wrong and \capRefPBlindIllegibleGi{} illegible of \capRefPBlindNGi{}; for Images~2.5,
\capRefPBlindCorrectTwoFive{}, \capRefPBlindWrongTwoFive{} and \capRefPBlindIllegibleTwoFive{} of
\capRefPBlindNTwoFive{}.

Within the glyph-height range both arms cover (\capRefPBlindMatchedRange{}\,px) the counts
are \capRefPBlindMatchedCorrectGi{} correct and \capRefPBlindMatchedWrongGi{} wrong of \capRefPBlindMatchedNGi{},
against \capRefPBlindMatchedCorrectTwoFive{} and \capRefPBlindMatchedWrongTwoFive{} of \capRefPBlindMatchedNTwoFive{},
and GPT-Image-2's crops remain the smaller within that range.

The agent read \capRefPBlindOcrFalseNegAll{} codes as
correct that OCR had scored wrong. These counts describe the scorer; no test is computed on them, and no verdict
rests on them.

\paragraph{Why embeddings could not test the claim.} An earlier design scored reference fidelity by CLIP and DINOv2
similarity between Flare-generated references and each model's output. Those encoders reward object identity and
are blind to label text: a wrong-label control scores AUC \capRefWrongLabelAucClip{} (CLIP) and
\capRefWrongLabelAucDino{} (DINOv2). The references were also made by one of the compared models. Its null result
was a ceiling of the instrument, not evidence about fidelity, and E3 replaces it.

\section{E4: Render Convincing Fine Print}
\label{app:e4}

\paragraph{Procedure.} Each of \capDetailNPrompts{} seeded prompts asks for \capDetailNLines{} exact random codes,
each line smaller than the last: half on a small card on a desk seen from standing height, half on a receipt seen
from afar, so camera distance sets glyph size. Every tier of every model runs (GPT-Image-2 \texttt{low} to
\texttt{high}; Images~2.5 \texttt{low} to \texttt{max}), plus the card prompts at 3840$\times$2160 \texttt{high}:
\capDetailNArms{} arms and \capDetailNImages{} images, with \capDetailFailedCalls{} failed calls.

Codes are matched
one-to-one to OCR detections by minimum character error; line height is the OCR box height per 1024 image rows. The
declared reading uses OCR without an allowlist; the allowlisted reading is a sensitivity analysis.

\paragraph{OCR ceiling.}
We calibrate the ceiling on clean card renders. The reliable cap height is
\capDetailCeilCapPx{}\,px: the smallest height at which easyocr reads at
least \capDetailCeilReadRate{} of codes exactly at that height and every
larger one. This corresponds to an OCR box height of
\capDetailCeilBoxHten{} per 1024 rows (\capDetailCeilBoxHtenFourk{}
at 4K). Lines below the ceiling do not enter the primary model score.

Two consequences limit the interpretation. First, the fitted 50\% legibility
height (h50) lies at or below the ceiling for \capDetailHfiftyAtCeilN{}
of \capDetailNArms{} arms. The experiment therefore does not measure the
legibility limit itself.

Second, eligibility is determined after generation. The shares of lines below
the ceiling are \capDetailBelowCeilGiHigh{} for GPT-Image-2 \texttt{high},
\capDetailBelowCeilFlareMax{} for Flare \texttt{max} and
\capDetailBelowCeilSunburstMax{} for Sunburst \texttt{max}. The arms are
thus scored on different sets of lines. At 4K, above-ceiling exactness is
\capDetailExactAttGiHighFourk{} for GPT-Image-2,
\capDetailExactAttFlareHighFourk{} for Flare and
\capDetailExactAttSunburstHighFourk{} for Sunburst.

\paragraph{Legibility by glyph height.} Pooled over every setting, lines shorter than
\capDetailBandEdgeFalloffLo{} are read exactly \capDetailBandExactBelowTwelve{} of the time and lines between
\capDetailBandEdgeFalloffLo{} and \capDetailBandCeilBoxH{} \capDetailBandExactTwelveToSixteen{}, against
\capDetailBandExactSixteenToTwentyTwo{} just above the ceiling and \capDetailBandExactAboveTwentyTwo{} above
\capDetailBandEdgeUpper{}. The ceiling therefore sits above the steep part of the legibility curve, and the range we
can score is its flat top. A height-by-height scan finds no band, above or below the ceiling, in which a higher tier
reads better: of \capDetailBandTierNBands{} tier bands, \capDetailBandTierNBandsFavourArm{} favours the higher tier
and \capDetailBandTierNBandsFavourControl{} favour \texttt{low}.

\paragraph{Permutation primary, intention to treat and matched glyph height.} The four primary
contrasts are tested post hoc by permutation over prompts (\capDetailVtwoNPerm{} permutations; Table~\ref{tab:results}). Each
drops prompts on which either image has no line above the ceiling: \capDetailVtwoDropFlareHighVsGiMediumN{} for both
\tokHigh{}-token contrasts (\capDetailVtwoDropFlareHighVsGiMediumCtl{} on the control side),
\capDetailVtwoDropFlareMaxVsGiHighN{} for Flare \texttt{max} (\capDetailVtwoDropFlareMaxVsGiHighCtl{} control,
\capDetailVtwoDropFlareMaxVsGiHighArm{} Flare) and \capDetailVtwoDropSunburstMaxVsGiHighN{} for Sunburst \texttt{max}
(\capDetailVtwoDropSunburstMaxVsGiHighCtl{} control). Missingness that falls on the control side is not at random, and two
analyses address it (Table~\ref{tab:e4v2}).

The intention-to-treat rate counts every planned line and scores sub-ceiling
lines as failures; it favours whichever arm renders larger, so its contrasts (Flare
\capDetailVtwoIttFlareHighVsGiMedium{} and \capDetailVtwoIttFlareMaxVsGiHigh{}, Sunburst
\capDetailVtwoIttSunburstHighVsGiMedium{} and \capDetailVtwoIttSunburstMaxVsGiHigh{}) mostly measure the instrument.

The matched-height analysis compares exact rates within glyph-height bins holding at least \capDetailVtwoMinBinLines{}
lines: Flare \capDetailVtwoMatchedFlareHighVsGiMedium{} and \capDetailVtwoMatchedFlareMaxVsGiHigh{}, Sunburst
\capDetailVtwoMatchedSunburstHighVsGiMedium{} and \capDetailVtwoMatchedSunburstMaxVsGiHigh{}, with only Flare
\texttt{max} surviving its exploratory Holm ($p$\,=\,\capDetailVtwoMatchedFlareMaxVsGiHighPHolm{}).

Under the
allowlisted OCR reading the Flare \texttt{max} primary contrast is \capDetailVtwoAllowAttPermFlareMaxVsGiHigh{} (Holm
$p$\,=\,\capDetailVtwoAllowAttPermFlareMaxVsGiHighPHolm{}) and its matched-height contrast
\capDetailVtwoAllowMatchedFlareMaxVsGiHigh{} ($p$\,=\,\capDetailVtwoAllowMatchedFlareMaxVsGiHighPHolm{}).

Within models,
\capDetailVtwoTierAttPermNReject{} of \capDetailVtwoTierM{} tier contrasts survive Holm above the ceiling and
\capDetailVtwoTierMatchedNReject{} at matched height; \capDetailVtwoTierIttNReject{} survive under intention to treat, the
same instrument effect.

\begin{table*}[tbp]
  \centering
  \caption{E4 per arm: exact share of code lines above the OCR ceiling and the intention-to-treat share of all planned
  lines (sub-ceiling lines failed), with 95\% intervals clustered by prompt, and images with no line above the
  ceiling.}
  \label{tab:e4v2}
  \footnotesize
  \setlength{\tabcolsep}{3pt}
  \begin{tabular}{lrrr}
    \toprule
    Arm (tokens) & Above ceiling & Intention to treat & No line \\
    \midrule
    GPT-Image-2 \texttt{medium} (\tokHigh{}) & \capDetailExactAttGiMedium{} [\capDetailExactAttGiMediumLo{}, \capDetailExactAttGiMediumHi{}] & \capDetailVtwoIttGiMedium{} [\capDetailVtwoIttGiMediumLo{}, \capDetailVtwoIttGiMediumHi{}] & \capDetailVtwoNoLineAboveGiMedium{} of \capDetailVtwoNImagesGiMedium{} \\
    Flare \texttt{high} (\tokHigh{}) & \capDetailExactAttFlareHigh{} [\capDetailExactAttFlareHighLo{}, \capDetailExactAttFlareHighHi{}] & \capDetailVtwoIttFlareHigh{} [\capDetailVtwoIttFlareHighLo{}, \capDetailVtwoIttFlareHighHi{}] & \capDetailVtwoNoLineAboveFlareHigh{} of \capDetailVtwoNImagesFlareHigh{} \\
    Sunburst \texttt{high} (\tokHigh{}) & \capDetailExactAttSunburstHigh{} [\capDetailExactAttSunburstHighLo{}, \capDetailExactAttSunburstHighHi{}] & \capDetailVtwoIttSunburstHigh{} [\capDetailVtwoIttSunburstHighLo{}, \capDetailVtwoIttSunburstHighHi{}] & \capDetailVtwoNoLineAboveSunburstHigh{} of \capDetailVtwoNImagesSunburstHigh{} \\
    GPT-Image-2 \texttt{high} (\tokMax{}) & \capDetailExactAttGiHigh{} [\capDetailExactAttGiHighLo{}, \capDetailExactAttGiHighHi{}] & \capDetailVtwoIttGiHigh{} [\capDetailVtwoIttGiHighLo{}, \capDetailVtwoIttGiHighHi{}] & \capDetailVtwoNoLineAboveGiHigh{} of \capDetailVtwoNImagesGiHigh{} \\
    Flare \texttt{max} (\tokMax{}) & \capDetailExactAttFlareMax{} [\capDetailExactAttFlareMaxLo{}, \capDetailExactAttFlareMaxHi{}] & \capDetailVtwoIttFlareMax{} [\capDetailVtwoIttFlareMaxLo{}, \capDetailVtwoIttFlareMaxHi{}] & \capDetailVtwoNoLineAboveFlareMax{} of \capDetailVtwoNImagesFlareMax{} \\
    Sunburst \texttt{max} (\tokMax{}) & \capDetailExactAttSunburstMax{} [\capDetailExactAttSunburstMaxLo{}, \capDetailExactAttSunburstMaxHi{}] & \capDetailVtwoIttSunburstMax{} [\capDetailVtwoIttSunburstMaxLo{}, \capDetailVtwoIttSunburstMaxHi{}] & \capDetailVtwoNoLineAboveSunburstMax{} of \capDetailVtwoNImagesSunburstMax{} \\
    \bottomrule
  \end{tabular}
\end{table*}

\paragraph{Sharpness proxies.} On the photoreal core (\capSharpNBase{} prompts per 1024-px cell, \capSharpNMax{} at
\texttt{max}, \capSharpNFourK{} at 4K) we measure Laplacian and high-frequency energy, flat-region noise and edge
width; none is a perceptual judgement. Calibrated on GPT-Image-2 output, a Gaussian blur removes
\capSharpCalibBlurLapPct{} of Laplacian energy and \capSharpCalibBlurAcutPct{} of edge acutance, while bilateral
denoising moves edge width by only \capSharpCalibDenoiseEdgeWPct{}.

Within Flare, higher tiers lose detail energy
(\texttt{xhigh} against \texttt{low}: Laplacian \capSharpFlareXhighLapPct{} [\capSharpFlareXhighLapLo{},
\capSharpFlareXhighLapHi{}], high-frequency \capSharpFlareXhighHfPct{} [\capSharpFlareXhighHfLo{},
\capSharpFlareXhighHfHi{}]; \texttt{max} \capSharpFlareMaxHfPct{}), with noise \capSharpFlareXhighFlatNoisePct{}.
Higher tiers narrow edges on Sunburst (\capSharpSunXhighEdgeWPct{} at \texttt{xhigh}) as on GPT-Image-2 (\capSharpGitwoHighEdgeWPct{} [\capSharpGitwoHighEdgeWLo{},
\capSharpGitwoHighEdgeWHi{}] at \texttt{high}), alongside large noise drops. These proxy changes are consistent with denoising
or sharpening, without establishing added perceptual detail.

At matched tokens the Images~2.5 arms carry \capSharpMatchLapRangeLo{} to \capSharpMatchLapRangeHi{}
Laplacian energy relative to GPT-Image-2; the smallest of these, Sunburst \texttt{max}, has an interval that includes
zero ([\capSharpMatchSunMaxLapLo{}, \capSharpMatchSunMaxLapHi{}]). Flat-region noise is
\capSharpMatchFlareHighFlatNoisePct{} to \capSharpMatchSunMaxFlatNoisePct{}, and edge width moves only \capSharpMatchEdgeWRangeLo{} to
\capSharpMatchEdgeWRangeHi{} across the \capSharpMatchEdgeWNCells{} matched cells, every interval including zero. The pattern is consistent with denoising; these
non-perceptual proxies cannot establish perceived sharpness.

4K adds no
high-frequency detail per field of view: \capSharpFlareFourKVsOneKHfPct{} [\capSharpFlareFourKVsOneKHfLo{},
\capSharpFlareFourKVsOneKHfHi{}] Flare, \capSharpSunFourKVsOneKHfPct{} [\capSharpSunFourKVsOneKHfLo{},
\capSharpSunFourKVsOneKHfHi{}] Sunburst and \capSharpGitwoFourKVsOneKHfPct{} [\capSharpGitwoFourKVsOneKHfLo{},
\capSharpGitwoFourKVsOneKHfHi{}] GPT-Image-2.

\paragraph{AI-agent spot check of the scorer (not a human rating).} An AI coding agent (Claude) compared a sample of OCR line
matches with the rendered images and agreed with the scorer on every line of its sample; the sample size was not recorded, so the check carries no
weight and no statistic is computed from it.

\section{Refusals in Detail}
\label{app:refusals}

Because the runner skips only rows already recorded \texttt{ok}, a refused request is re-submitted on the cell's
next resume; \docsRefusedReattempted{} have been, \docsRefusedFlippedOk{} then accepted and \docsRefusedHeld{}
refused again, a persistent rate of \docsRefusalPersistentPct{} [\docsRefusalPersistentWilsonLo{},
\docsRefusalPersistentWilsonHi{}] (\docsRefusedNotReattempted{} not yet re-attempted, counted as persistent).

The
re-submission was a side effect of resuming the cell, not a designed protocol; the interval was
\docsRetryGapHours{}~hours; and that the request was identical rests on code history, not on the ledger
(Appendix~\ref{app:design}). These observations establish only that three consecutive same-code rejections did not predict
rejection \docsRetryGapHours{}~h later for \docsRefusedFlippedOk{} of \docsRefusedReattempted{} requests, and the
\docsPersistentSpecs{} persistent specifications recur across models (\docsPersistentBothModels{} refused by both).

The control cohort splits \docsControlLowRefused{}/\docsControlLowN{} at \texttt{low} and
\docsControlMedRefused{}/\docsControlMedN{} at \texttt{medium}. The accepted rows split \docsFlareCord{}/\docsFlareWild{}
(Flare, CORD/WildReceipt) and \docsSunCord{}/\docsSunWild{} (Sunburst), and the control
\docsControlCordN{}/\docsControlWildN{} across its two tiers, so cross-model comparisons are made on the matched
specification set and stratified by corpus.

\paragraph{Paired comparison on shared specifications.} The GPT-Image-2 control ran the first \refPairNShared{}
specifications (\refPairSharedCord{} CORD, \refPairSharedWild{} WildReceipt), which every Images~2.5 arm also ran within
its \refPairAllCord{} CORD and \refPairAllWild{} WildReceipt specifications. On that shared set Flare refused
\refPairFlareRefused{}, Sunburst \refPairSunRefused{}, GPT-Image-2 \texttt{low} \refPairGitwoLowRefused{} and
\texttt{medium} \refPairGitwoMedRefused{}.

Against \texttt{low}, Flare has \refPairFlareVsGitwoLowArmOnly{} and
\refPairFlareVsGitwoLowCtlOnly{} discordant pairs (arm only, control only; \refPairFlareVsGitwoLowBoth{} refused by
both) and Sunburst \refPairSunVsGitwoLowArmOnly{} and \refPairSunVsGitwoLowCtlOnly{}; against \texttt{medium}, Flare has
\refPairFlareVsGitwoMedArmOnly{} and \refPairFlareVsGitwoMedCtlOnly{} and Sunburst \refPairSunVsGitwoMedArmOnly{} and
\refPairSunVsGitwoMedCtlOnly{}.

Every exact McNemar $p$ is \refPairFlareVsGitwoLowPMcNemar{}. With all discordant pairs
in one direction an exact McNemar test reaches $p<0.05$ only at \refPairMinDiscordant{} or more pairs, so the design
cannot separate rates this low.

Of the Images~2.5 refusals, \refPairTwoFiveRefusedWild{} were WildReceipt and
\refPairTwoFiveRefusedCord{} CORD, and \refPairTwoFiveRefusedOutside{} fell outside the shared set. The unpaired rates in
Table~\ref{tab:refusals} compare differently composed sets and are descriptive.

\begin{table*}[tbp]
  \centering
  \caption{Refusal behaviour on the same specification catalogue, with the axes on which the measurements differ.
  Round~v2 is quoted from \cite{wu2026forgerjudge}; round-v3 rows are first-pass rates measured here with Wilson 95\%
  intervals (persistent rates in the text). The control row runs round~v2's generator, GPT-Image-2, through the
  round-v3 request in the same week, so route, mask, prompt, corpora and date are fixed and only the generation
  moves. Its rate is of the same order as the Images~2.5 arms; the comparison cannot resolve small differences.
  Date and request format remain confounded with each other.}
  \label{tab:refusals}
  \footnotesize
  \setlength{\tabcolsep}{4pt}
  \begin{tabular}{llllllrrl}
    \toprule
    Round (arm) & Generator & Route & Mask & Prompt & Run & Attempts & Refused (95\% CI) & Isolates \\
    \midrule
    v2 & GPT-Image-2 & gateway & drawn box & 5 clauses & Apr 2026 & \vTwoSpecs{} & \vTwoRefusal{} & --- \\
    \midrule
    v3 (Flare arm) & \texttt{2.5-flare} & direct & alpha & 4 sentences & Sep 2026 & \docsArmFlareN{} & \docsArmFlarePct{} [\docsArmFlareWilsonLo{}, \docsArmFlareWilsonHi{}] & --- \\
    v3 (Sunburst arm) & \texttt{2.5-sunburst} & direct & alpha & 4 sentences & Sep 2026 & \docsArmSunN{} & \docsArmSunPct{} [\docsArmSunWilsonLo{}, \docsArmSunWilsonHi{}] & --- \\
    \midrule
    v3 (control arm) & GPT-Image-2 & direct & alpha & 4 sentences & Sep 2026 & \docsControlN{} & \docsControlPct{} [\docsControlWilsonLo{}, \docsControlWilsonHi{}] & generation \\
    \bottomrule
  \end{tabular}
\end{table*}

\section{Localisation in Detail}
\label{app:localisation}

\subsection{Localisation across our three dataset releases}
\label{app:progression}

All three rounds forge the same specifications on the same source images, so a specification present in all three
supports a comparison holding document, field, box and target value fixed while the generator changes, although the
route and mask also changed between rounds. The intersection is \progN{} specifications, small because the round-v1
and round-v2 cells are staged in the harness as fixed subsets, and \progDtPairedN{} of them carry a score in every
arm. The localiser is \progDetectorName{} \cite{qu2023doctamper} at its released checkpoint (\dtCheckpoint{}).

\begin{figure}[tbp]
  \centering
  \includegraphics[width=\columnwidth]{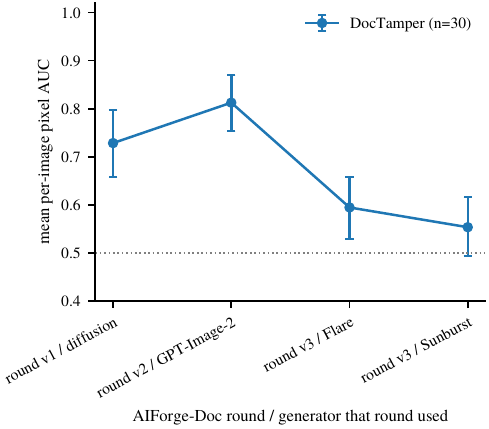}
  \caption{Per-image DocTamper localisation (pixel) AUC by AIForge-Doc round, paired over the \progDtPairedN{}
  specifications scored in all three rounds, with 95\% intervals; dotted line at chance. These round-v3 values are
  the paired subset, not the full-arm values of \S\ref{sec:defence-localisation}: the denominators differ, the two
  arms' order reverses between them, and the Sunburst interval contains 0.5. Suggestive only.}
  \label{fig:progression}
\end{figure}

Mean per-image localisation AUC is \progDtVone{} [\progDtVoneLo{}, \progDtVoneHi{}] on round~v1 and \progDtVtwo{}
[\progDtVtwoLo{}, \progDtVtwoHi{}] on round~v2, and \progDtVthreeFlare{} [\progDtVthreeFlareLo{},
\progDtVthreeFlareHi{}] on Flare and \progDtVthreeSun{} [\progDtVthreeSunLo{}, \progDtVthreeSunHi{}] on Sunburst in
round~v3 (Figure~\ref{fig:progression}). On these \progDtPairedN{} specifications the earlier rounds localise better
than round~v3, but the sample is small, the route and mask changed with the round, and the full-arm control of
\S\ref{sec:defence-localisation} shows GPT-Image-2 localised just as weakly through the round-v3 route. The
panel is therefore consistent with weak localisation being a property of the round-v3 editing route rather
than of the newest models, and supports no stronger reading. Locality is not measurable on the staged round-v1 and
round-v2 cells, which are re-encoded whole pages: a median \progStagedChangedPct{} of their pixels differ from the
source even outside the edited field.

\paragraph{Edit success and localisation (exploratory).} On the \capDecoupleDtNSpecs{} E1 specifications
shared with the control (\capDecoupleDtNImages{} Images~2.5 images), DocTamper's per-image localisation AUC does not
detectably depend on whether the forgery succeeded: clean edits (target right, nothing else changed) score \capDecoupleDtClean{}
[\capDecoupleDtCleanLo{}, \capDecoupleDtCleanHi{}] against \capDecoupleDtRest{} for the rest, a difference of
\capDecoupleDtCleanMinusRest{} [\capDecoupleDtCleanMinusRestLo{}, \capDecoupleDtCleanMinusRestHi{}]
($n$\,=\,\capDecoupleDtCleanMinusRestN{}), and AUC is uncorrelated with tolerant locality ($\rho$\,=\,\capDecoupleDtRho{}
[\capDecoupleDtRhoLo{}, \capDecoupleDtRhoHi{}], $n$\,=\,\capDecoupleDtRhoN{}). The intervals resolve differences of
about $\pm$\capDecoupleDtResolution{} AUC; this is not an equivalence test. The one visible pattern is legibility:
images whose source field OCR could not read localise worse (\capDecoupleDtUnreadable{} [\capDecoupleDtUnreadableLo{},
\capDecoupleDtUnreadableHi{}]). TruFor, whose pixel AUC is seam-inflated, shows only small, inconsistent couplings
(clean $-$ rest \capDecoupleTfCleanMinusRest{} [\capDecoupleTfCleanMinusRestLo{}, \capDecoupleTfCleanMinusRestHi{}];
$\rho$\,=\,\capDecoupleTfRho{}), and the decoupling reading is not extended to it.

\subsection{The seam left by pasting edits back}
\label{app:seam}

The pipeline pastes the edited crop back into the authentic page, so the rectangular discontinuity sits at the crop
boundary, not at the edited field. Measuring \seamDetector{} on \seamN{} Flare rows, mean response is
\seamFieldResp{} inside the field, \seamRingResp{} across the rest of the pasted crop and \seamOutResp{} outside it,
and scoring the maps against the crop rectangle does as well as against the field box (pixel AUC \seamAucCrop{}
against \seamAucField{}; Sunburst, \seamSunN{} rows: \seamSunAucCrop{} against \seamSunAucField{}).

Within the crop
the maps still rank field pixels above the rest (\seamAucFieldWithinCrop{} Flare, \seamSunAucFieldWithinCrop{}
Sunburst), a residual we have not separated from crop geometry. The same check on the control arms gives
field-within-crop AUC \seamCtrlGitwoLowAucFieldWithinCrop{} (\texttt{low}, \seamCtrlGitwoLowN{} rows) and
\seamCtrlGitwoMedAucFieldWithinCrop{} (\texttt{medium}, \seamCtrlGitwoMedN{} rows), and crop-rectangle AUC
\seamCtrlGitwoLowAucCrop{} and \seamCtrlGitwoMedAucCrop{}, so the seam signal is the same for both generations.

A splice localiser can therefore score well on AIForge-Doc round~v3 without localising the
edit. DocTamper's weak localisation is unaffected, but any positive localisation number on round~v3 must be read
against the crop rectangle first. On the E1 specifications TruFor's pixel AUC is \detTfGitwoLowPixAuc{} and
\detTfGitwoMedPixAuc{} on the GPT-Image-2 control arms against \detTfFlarePixAuc{} and \detTfSunPixAuc{}, the same
seam response on every arm (paired differences \detPairTfFlareMinusGitwoMedDelta{} to
\detPairTfSunMinusGitwoLowDelta{}).

Community Forensics has no localisation map; its detection AUC on the same pages
is \detCfImgAucLo{}--\detCfImgAucHi{} across the four arms, with GPT-Image-2 no lower (\detCfGitwoLowImgAuc{} and
\detCfGitwoMedImgAuc{} against \detCfFlareImgAuc{} and \detCfSunImgAuc{}).

\paragraph{TruFor coverage.} TruFor did not score every page: \tfCovFlareMissPct{} of Flare, \tfCovSunMissPct{} of
Sunburst, \tfCovGitwoLowMissPct{} and \tfCovGitwoMedMissPct{} of the two control arms and \tfCovAuthMissPct{} of
authentic pages are missing. The missing pages are the largest. No page with a long edge under 2.7k\,px is missing on
any arm (\tfCovFlareMissPctLtTwoSevenK{}), while \tfCovFlareMissPctGeTwoSevenK{}--\tfCovGitwoLowMissPctGeTwoSevenK{} of
larger pages are, and the largest scored long edge is \tfCovGitwoLowMaxScoredLongEdge{}--\tfCovFlareMaxScoredLongEdge{}\,px.
Every TruFor number is therefore conditioned on page size, equally across arms.

\subsection{Equivalence tests}
\label{app:equivalence}

We test equivalence by two one-sided tests at the 5\% level, that is, by whether a 90\% cluster-bootstrap interval lies
inside the margin. For DocTamper, each arm's mean per-image localisation AUC is tested against chance at margins of
$\pm$\eqDtMarginLo{} and $\pm$\eqDtMarginHi{}, and each paired Images~2.5 $-$ GPT-Image-2 difference against zero;
the reported margin is the smallest the interval fits in.

For Community Forensics, a model's AUC counts as flat over its price
tiers if every pairwise tier difference has a 90\% interval inside $\pm$\eqTierMarginLo{} or $\pm$\eqTierMarginHi{} (an
intersection-union test), resampling prompts and real images.

Table~\ref{tab:equivalence} gives the results. No DocTamper
arm lies within $\pm$\eqDtMarginLo{} of chance, and \eqDtNEquivTen{} of \eqDtNArms{} (Flare) lies within
$\pm$\eqDtMarginHi{}. Flatness fails at $\pm$\eqTierMarginLo{} for every model; at $\pm$\eqTierMarginHi{} it holds only
for GPT-Image-2's \eqTierGitwoNCells{} cells, and across all \eqTierAllNCells{} 1024-px cells the smallest supported
margin is \eqTierAllMinMargin{} (\eqTierAllFourKMinMargin{} with the 4K cells). Localisation is therefore weak rather
than absent, and detection shows no trend with price without being shown flat.

\begin{table*}[tbp]
  \centering
  \caption{Equivalence tests. Left: DocTamper per-image localisation AUC against chance per arm, and paired
  Images~2.5 $-$ GPT-Image-2 differences against zero, with 90\% intervals clustered by specification. Right: Community
  Forensics AUC across price tiers; range is the largest minus smallest cell AUC, with its bootstrap upper bound.
  Margin: the smallest symmetric margin the test supports.}
  \label{tab:equivalence}
  \footnotesize
  \setlength{\tabcolsep}{3pt}
  \begin{tabular}[t]{lrrcc}
    \toprule
    DocTamper & AUC [90\% CI] & Margin & $\pm$\eqDtMarginLo{} & $\pm$\eqDtMarginHi{} \\
    \midrule
    Flare & \detDtFlarePixAuc{} [\eqDtFlareCiNinetyLo{}, \eqDtFlareCiNinetyHi{}] & \eqDtFlareMinMargin{} & \eqDtFlareEquivFive{} & \eqDtFlareEquivTen{} \\
    Sunburst & \detDtSunPixAuc{} [\eqDtSunCiNinetyLo{}, \eqDtSunCiNinetyHi{}] & \eqDtSunMinMargin{} & \eqDtSunEquivFive{} & \eqDtSunEquivTen{} \\
    GPT-Image-2 \texttt{low} & \detDtGitwoLowPixAuc{} [\eqDtGitwoLowCiNinetyLo{}, \eqDtGitwoLowCiNinetyHi{}] & \eqDtGitwoLowMinMargin{} & \eqDtGitwoLowEquivFive{} & \eqDtGitwoLowEquivTen{} \\
    GPT-Image-2 \texttt{med.} & \detDtGitwoMedPixAuc{} [\eqDtGitwoMedCiNinetyLo{}, \eqDtGitwoMedCiNinetyHi{}] & \eqDtGitwoMedMinMargin{} & \eqDtGitwoMedEquivFive{} & \eqDtGitwoMedEquivTen{} \\
    \midrule
    Flare $-$ \texttt{low} & \detPairDtFlareMinusGitwoLowDelta{} [\eqDtPairFlareMinusGitwoLowCiNinetyLo{}, \eqDtPairFlareMinusGitwoLowCiNinetyHi{}] & \eqDtPairFlareMinusGitwoLowMinMargin{} & & \\
    Flare $-$ \texttt{med.} & \detPairDtFlareMinusGitwoMedDelta{} [\eqDtPairFlareMinusGitwoMedCiNinetyLo{}, \eqDtPairFlareMinusGitwoMedCiNinetyHi{}] & \eqDtPairFlareMinusGitwoMedMinMargin{} & & \\
    Sunburst $-$ \texttt{low} & \detPairDtSunMinusGitwoLowDelta{} [\eqDtPairSunMinusGitwoLowCiNinetyLo{}, \eqDtPairSunMinusGitwoLowCiNinetyHi{}] & \eqDtPairSunMinusGitwoLowMinMargin{} & & \\
    Sunburst $-$ \texttt{med.} & \detPairDtSunMinusGitwoMedDelta{} [\eqDtPairSunMinusGitwoMedCiNinetyLo{}, \eqDtPairSunMinusGitwoMedCiNinetyHi{}] & \eqDtPairSunMinusGitwoMedMinMargin{} & & \\
    \bottomrule
  \end{tabular}\hfill
  \begin{tabular}[t]{lrrrrcc}
    \toprule
    Community Forensics & Cells & Range & Upper & Margin & $\pm$\eqTierMarginLo{} & $\pm$\eqTierMarginHi{} \\
    \midrule
    Flare & \eqTierFlareNCells{} & \eqTierFlareRange{} & \eqTierFlareRangeUpper{} & \eqTierFlareMinMargin{} & \eqTierFlareEquivThree{} & \eqTierFlareEquivFive{} \\
    Sunburst & \eqTierSunNCells{} & \eqTierSunRange{} & \eqTierSunRangeUpper{} & \eqTierSunMinMargin{} & \eqTierSunEquivThree{} & \eqTierSunEquivFive{} \\
    GPT-Image-2 & \eqTierGitwoNCells{} & \eqTierGitwoRange{} & \eqTierGitwoRangeUpper{} & \eqTierGitwoMinMargin{} & \eqTierGitwoEquivThree{} & \eqTierGitwoEquivFive{} \\
    All, 1024 px & \eqTierAllNCells{} & \eqTierAllRange{} & \eqTierAllRangeUpper{} & \eqTierAllMinMargin{} & \eqTierAllEquivThree{} & \eqTierAllEquivFive{} \\
    All, with 4K & \eqTierAllFourKNCells{} & \eqTierAllFourKRange{} & \eqTierAllFourKRangeUpper{} & \eqTierAllFourKMinMargin{} & \eqTierAllFourKEquivThree{} & \eqTierAllFourKEquivFive{} \\
    \bottomrule
  \end{tabular}
\end{table*}

\section{Quality Settings, Cost, Speed and Provenance}
\label{app:tiers}

\paragraph{Design.} The photoreal core is 100 text-to-image prompts over six content classes (portrait, street,
interior, food, product and photographs of documents), written to be photographic rather than stylised. Every
prompt runs at 1024$\times$1024 on all three models at every tier that model offers, and a 20-prompt sub-grid adds
3840$\times$2160 at \texttt{high}. Scored cells hold \tierCellNLo{}--\tierCellNHi{} images at 1024 px and
\tierFourKN{} at 4K. Tokens are a price proxy, not a compute measure (\S\ref{sec:method-setup}).
Table~\ref{tab:auc_tokens} is the grid behind \S\ref{sec:defence-wild}.

\paragraph{Detection across quality tiers.} Community Forensics' AUC against \tierRealN{} real photographs stays
within \tierAucFlatLo{}--\tierAucFlatHi{} across \tierAucFlatCells{} 1024-px cells spanning a \tierTokFold{}-fold
token range (Table~\ref{tab:auc_tokens}). Part of every AUC may be format, since the real images are JPEG and TIFF.
No trend test was run, and flatness is not established: equivalence at $\pm$\eqTierMarginLo{} fails for every
model, and the smallest margin supported across all 1024-px cells is \eqTierAllMinMargin{}. At 4K, billed at about
\capPriceFourKOverHigh{}$\times$ the 1024 \texttt{high} price, AUC is lower for Flare (\aucFlareFourK{} against
\aucFlareHigh{}) and Sunburst (\aucSunFourK{} against \aucSunHigh{}), on unmatched prompts.

\begin{table*}[tbp]
  \centering
  \caption{\tierDetector{} image-level AUC on the photoreal core by billed output tokens (price proxy) and model,
  with 95\% bootstrap intervals and smoke-grid mean latency in parentheses. The negative class is \tierRealN{} real
  photographs from \tierRealPools{}, JPEG and uncompressed TIFF against PNG generations, so part of any AUC may be
  format rather than generator. The 4K row is 3840$\times$2160 at \texttt{high}, billed at about
  \capPriceFourKOverHigh{}$\times$ the 1024 \texttt{high} price (Table~\ref{tab:tiers}); its cells hold
  \tierFourKN{} prompts and are not prompt-matched to the 1024 cells. Equivalence tests are in
  Appendix~\ref{app:equivalence}.}
  \label{tab:auc_tokens}
  \footnotesize
  \setlength{\tabcolsep}{4pt}
  \begin{tabular}{rrrrr}
    \toprule
    Out.\ tokens & \$/image & GPT-Image-2 & Flare & Sunburst \\
    \midrule
    \tokLow{}   & \usdLow{}   & \aucGitwoLow{} \aucCiGitwoLow{} (\latGitwoLow{}~s)   & \aucFlareLow{} \aucCiFlareLow{} (\latFlareLow{}~s)     & \aucSunLow{} \aucCiSunLow{} (\latSunLow{}~s) \\
    \tokMed{}   & \usdMed{}   & ---                                                  & \aucFlareMed{} \aucCiFlareMed{} (\latFlareMed{}~s)     & \aucSunMed{} \aucCiSunMed{} (\latSunMed{}~s) \\
    \tokHigh{}  & \usdHigh{}  & \aucGitwoMed{} \aucCiGitwoMed{} (\latGitwoMed{}~s)    & \aucFlareHigh{} \aucCiFlareHigh{} (\latFlareHigh{}~s)  & \aucSunHigh{} \aucCiSunHigh{} (\latSunHigh{}~s) \\
    \tokXhigh{} & \usdXhigh{} & ---                                                  & \aucFlareXhigh{} \aucCiFlareXhigh{} (\latFlareXhigh{}~s) & \aucSunXhigh{} \aucCiSunXhigh{} (\latSunXhigh{}~s) \\
    \tokMax{}   & \usdMax{}   & \aucGitwoHigh{} \aucCiGitwoHigh{} (\latGitwoHigh{}~s)  & \aucFlareMax{} \aucCiFlareMax{} (\latFlareMax{}~s)     & \aucSunMax{} \aucCiSunMax{} (\latSunMax{}~s) \\
    \midrule
    \multicolumn{2}{l}{4K \texttt{high}} & \aucGitwoFourK{} \aucCiGitwoFourK{} & \aucFlareFourK{} \aucCiFlareFourK{} & \aucSunFourK{} \aucCiSunFourK{} \\
    \bottomrule
  \end{tabular}
\end{table*}

\paragraph{Attribution probe (exploratory).} Images~2.5 ships as two models under one brand, so we asked whether
pixels separate the three generators. A multinomial logistic regression on \attrFeatures{} radial power-spectrum
features, with standardised inputs under \attrCv{} cross-validation on a balanced three-class split with prompt and
tier held fixed, reaches \attrMedAcc{} [\attrMedCiLo{}, \attrMedCiHi{}] at \texttt{medium} ($n=\attrMedN{}$; chance
\attrChance{}), \attrLowAcc{} [\attrLowCiLo{}, \attrLowCiHi{}] at \texttt{low}, \attrHighAcc{} [\attrHighCiLo{},
\attrHighCiHi{}] at \texttt{high} and \attrFourKAcc{} [\attrFourKCiLo{}, \attrFourKCiHi{}] in the 4K cells
($n=\attrFourKN{}$). At \texttt{medium}, these spectral features do not reliably distinguish the generators; a probe on
detector embeddings would be the stronger test.

\paragraph{Content credentials do not name the generation.}
\label{app:c2pa}
The inspected raw PNGs from all three models contain C2PA content credentials \cite{c2pa2024spec}.
The inspected manifests declare the generator as \texttt{gpt-image} version \texttt{2.0} for GPT-Image-2, Flare and Sunburst
alike; that string is neither a model name nor our round~v2. That field therefore does not distinguish the tested versions. The inspected delivery
paths in our wild snapshot do not retain these manifests (Appendix~\ref{app:wild}).
Some files from other platforms retain them, with the same version-blind string
\cite{paperwild2026}.

\subsection{Speed on the forgery tasks}
\label{app:latency}

Latency comes from synchronous ledger rows only. For each task the ratio is GPT-Image-2's median seconds per accepted
call over the Images~2.5 arm's, with a bootstrap interval clustered by specification, receipt, source photo or reference
(Table~\ref{tab:latforge}). Only E1 ran every arm synchronously in full. The E2 and E3 full runs used the Batch API, so
their synchronous rows are pilots and fill, their $n$ is small and they have no \texttt{low} arm. Arms were not
interleaved and ran from one account in one week, so load at the time of the run is confounded with arm.

On
text-to-image generation at identical token counts, from the photoreal core with prompts paired within one serial run,
Flare was \capSharpLatFlareHighRatio{}$\times$ [\capSharpLatFlareHighLo{}, \capSharpLatFlareHighHi{}] faster than
GPT-Image-2 at \capSharpLatTokHigh{} tokens and \capSharpLatFlareMaxRatio{}$\times$ [\capSharpLatFlareMaxLo{},
\capSharpLatFlareMaxHi{}] at \capSharpLatTokMax{}; those requests overlapped other cells and measure neither edits nor
throughput.

\begin{table*}[tbp]
  \centering
  \caption{Forgery-task latency from synchronous rows: median seconds per accepted call and the GPT-Image-2\,/\,Images~2.5
  ratio of medians with 95\% bootstrap intervals. $n$: paired units (clusters).}
  \label{tab:latforge}
  \footnotesize
  \setlength{\tabcolsep}{4pt}
  \begin{tabular}{llrrrrrr}
    \toprule
    Task & Control & $n$ & GPT-Image-2 s & Flare s & Sunburst s & Ratio, Flare & Ratio, Sunburst \\
    \midrule
    E1 edits & \texttt{low} & \latForgeDocsGitwoLowOverFlareN{} (\latForgeDocsGitwoLowOverFlareNClusters{}) & \latForgeDocsGitwoLowMedianS{} & \latForgeDocsFlareMedianS{} & \latForgeDocsSunMedianS{} & \latForgeDocsGitwoLowOverFlareRatio{} [\latForgeDocsGitwoLowOverFlareLo{}, \latForgeDocsGitwoLowOverFlareHi{}] & \latForgeDocsGitwoLowOverSunRatio{} [\latForgeDocsGitwoLowOverSunLo{}, \latForgeDocsGitwoLowOverSunHi{}] \\
    E1 edits & \texttt{medium} & \latForgeDocsGitwoMedOverFlareN{} (\latForgeDocsGitwoMedOverFlareNClusters{}) & \latForgeDocsGitwoMedMedianS{} & \latForgeDocsFlareMedianS{} & \latForgeDocsSunMedianS{} & \latForgeDocsGitwoMedOverFlareRatio{} [\latForgeDocsGitwoMedOverFlareLo{}, \latForgeDocsGitwoMedOverFlareHi{}] & \latForgeDocsGitwoMedOverSunRatio{} [\latForgeDocsGitwoMedOverSunLo{}, \latForgeDocsGitwoMedOverSunHi{}] \\
    E2a receipt chains & \texttt{medium} & \latForgeDocChainGitwoMedOverFlareN{} (\latForgeDocChainGitwoMedOverFlareNClusters{}) & \latForgeDocChainGitwoMedMedianS{} & \latForgeDocChainFlareMedianS{} & \latForgeDocChainSunMedianS{} & \latForgeDocChainGitwoMedOverFlareRatio{} [\latForgeDocChainGitwoMedOverFlareLo{}, \latForgeDocChainGitwoMedOverFlareHi{}] & \latForgeDocChainGitwoMedOverSunRatio{} [\latForgeDocChainGitwoMedOverSunLo{}, \latForgeDocChainGitwoMedOverSunHi{}] \\
    E2b stateless & \texttt{medium} & \latForgePhotoChainGitwoMedOverFlareN{} (\latForgePhotoChainGitwoMedOverFlareNClusters{}) & \latForgePhotoChainGitwoMedMedianS{} & \latForgePhotoChainFlareMedianS{} & \latForgePhotoChainSunMedianS{} & \latForgePhotoChainGitwoMedOverFlareRatio{} [\latForgePhotoChainGitwoMedOverFlareLo{}, \latForgePhotoChainGitwoMedOverFlareHi{}] & \latForgePhotoChainGitwoMedOverSunRatio{} [\latForgePhotoChainGitwoMedOverSunLo{}, \latForgePhotoChainGitwoMedOverSunHi{}] \\
    E2b conversational & \texttt{medium} & \latForgePhotoChainConvGitwoMedOverFlareN{} (\latForgePhotoChainConvGitwoMedOverFlareNClusters{}) & \latForgePhotoChainConvGitwoMedMedianS{} & \latForgePhotoChainConvFlareMedianS{} & \latForgePhotoChainConvSunMedianS{} & \latForgePhotoChainConvGitwoMedOverFlareRatio{} [\latForgePhotoChainConvGitwoMedOverFlareLo{}, \latForgePhotoChainConvGitwoMedOverFlareHi{}] & \latForgePhotoChainConvGitwoMedOverSunRatio{} [\latForgePhotoChainConvGitwoMedOverSunLo{}, \latForgePhotoChainConvGitwoMedOverSunHi{}] \\
    E3 references & \texttt{medium} & \latForgeRefPGitwoMedOverFlareN{} (\latForgeRefPGitwoMedOverFlareNClusters{}) & \latForgeRefPGitwoMedMedianS{} & \latForgeRefPFlareMedianS{} & \latForgeRefPSunMedianS{} & \latForgeRefPGitwoMedOverFlareRatio{} [\latForgeRefPGitwoMedOverFlareLo{}, \latForgeRefPGitwoMedOverFlareHi{}] & \latForgeRefPGitwoMedOverSunRatio{} [\latForgeRefPGitwoMedOverSunLo{}, \latForgeRefPGitwoMedOverSunHi{}] \\
    \bottomrule
  \end{tabular}
\end{table*}

\section{Repeated Re-editing (``Laundering'') in Detail}
\label{app:chains}

A laundering chain repeatedly re-edits one generated image, as a forger would when reworking the same piece of
evidence, so that detection can be read at several turn positions. The cell is complete: \chainsRows{} randomised-order rows, \chainsChains{} chains, all \chainsAnalysed{} of them
complete eight-turn chains, with turns \chainsScoredTurns{} scored by \chainsDetector{} ($n=\chainGitwoComplete{}$
per arm at every scored turn). Each of \chainSeeds{} seeds is a \texttt{medium}-tier photoreal image generated by
the model that then edits it, and each goes through the same eight instruction edits on all three models: some
local, some global edits of the whole frame, and a camera-look pass that is the degradation step a forger would apply. The order
is a seeded random permutation per seed, identical across models, so that turn position and edit identity are
separable; the camera-look edit lands at turn positions 1--8 with counts \chainCameraPositions{}. An earlier
fixed-order pilot of \nChainsPilot{} rows is retired, and two control arms (an identity chain and a one-shot
camera-look on the seed) were not run.

\begin{figure*}[tbp]
  \centering
  \includegraphics[width=0.86\textwidth]{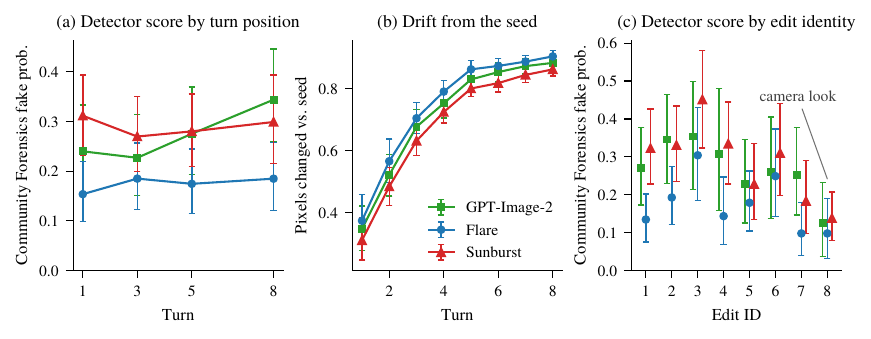}
  \caption{\chainsDetector{} score against turn index, all three arms, means with 95\% intervals over
  \chainGitwoComplete{} complete eight-turn chains per arm at turns \chainsScoredTurns{}. GPT-Image-2's mean rises
  from turn~1 to turn~8, but the rise is not significant after Holm correction across the three arms; both
  Images~2.5 arms are flat within their intervals. Turn-1 levels do not differ reliably between arms, and each arm
  starts from its own seed.}
  \label{fig:chains}
\end{figure*}

Per-turn means with 95\% intervals are plotted in Figure~\ref{fig:chains}: GPT-Image-2 runs \chainGitwoToneMean{},
\chainGitwoTthreeMean{}, \chainGitwoTfiveMean{} and \chainGitwoTeightMean{} at turns 1, 3, 5 and 8; Flare
\chainFlareToneMean{}, \chainFlareTthreeMean{}, \chainFlareTfiveMean{} and \chainFlareTeightMean{}; and Sunburst
\chainSunToneMean{}, \chainSunTthreeMean{}, \chainSunTfiveMean{} and \chainSunTeightMean{}.

Broken out by edit
identity, the camera-look pass ranks \chainGitwoEditCameraRank{} of eight on GPT-Image-2 (\chainGitwoEditCameraMean{}
against \chainGitwoEditOtherLo{}--\chainGitwoEditOtherHi{} for the other seven), \chainFlareEditCameraRank{} on Flare
(\chainFlareEditCameraMean{} against \chainFlareEditOtherLo{}--\chainFlareEditOtherHi{}) and \chainSunEditCameraRank{}
on Sunburst (\chainSunEditCameraMean{} against \chainSunEditOtherLo{}--\chainSunEditOtherHi{}).

\paragraph{Detector-score change over repeated edits.} GPT-Image-2's score rose from turn~1 to turn~8 by
\chainGitwoDeltaMean{} [\chainGitwoDeltaLo{}, \chainGitwoDeltaHi{}] (uncorrected $p$\,=\,\chainGitwoDeltaP{}), a rise
that is not significant after Holm correction across the \chainsHolmM{} arms ($p$\,=\,\chainGitwoDeltaPHolm{}); both
Images~2.5 intervals include zero (Flare \chainFlareDeltaMean{} [\chainFlareDeltaLo{}, \chainFlareDeltaHi{}], Sunburst
\chainSunDeltaMean{} [\chainSunDeltaLo{}, \chainSunDeltaHi{}]). Repeated editing therefore did not reliably move the
detector score on any arm, and the design supports nothing stronger: no between-arm test was run, each arm starts from
its own seed, and the Images~2.5 edits are weaker.

\paragraph{Why these chains cannot test consistency.} They are secondary to \S\ref{sec:cap-multiturn} and do not
test its claim. They edit each model's own generated seed, statelessly, with edits that cannot be checked
against a known answer. Scored by embedding over \capTurnNChains{} chains of \capTurnNTurns{} turns, edit survival
was \capTurnSurvRateGptTwo{} for GPT-Image-2, \capTurnSurvRateFlare{} for Flare and \capTurnSurvRateSunburst{} for
Sunburst: Flare \capTurnDiffSurvRateFlare{} [\capTurnDiffSurvRateFlareLo{}, \capTurnDiffSurvRateFlareHi{}] and
Sunburst \capTurnDiffSurvRateSunburst{} [\capTurnDiffSurvRateSunburstLo{}, \capTurnDiffSurvRateSunburstHi{}], the
latter \capTurnDiffSurvRateNoTOneSunburst{} [\capTurnDiffSurvRateNoTOneSunburstLo{},
\capTurnDiffSurvRateNoTOneSunburstHi{}] once turn-1 edits are excluded.

The Images~2.5 edits are weaker (per-edit
CLIP change \capTurnDiffStrengthClipFlare{} and \capTurnDiffStrengthClipSunburst{}), and drift per unit edit
strength is \capTurnDriftPerEditGptTwo{}, \capTurnDriftPerEditFlare{} and \capTurnDriftPerEditSunburst{}, so lower
drift on 2.5 reflects weaker edits rather than better consistency. That design motivated E2a and E2b.

\section{Comparison with Images Posted Online}
\label{app:wild}

We collected a new Images~2.5 sample using version-specific post filtering, extending the approach of our
April GPT-Image-2 dataset \cite{ren2026wild}; that earlier paper does not document this new sample. The collection
instrument and the full release are the subject of the companion paper \cite{paperwild2026}, whose release has
since grown beyond the snapshot scored here.

Our snapshot, taken on 11~September~2026, contains \wildStandalone{} standalone outputs, with posts dated
8~September~2026, 18:58:45~UTC through 10~September~2026, 02:00:30~UTC. Text must identify Images~2.5 or
Flare/Sunburst without also naming GPT-Image-2; comparison posts are excluded. Official X API rows additionally
pass a creation-language classifier, while gateway rows use the version gate alone. A vision-model filter retains
standalone outputs (the companion paper's screenshot filter).

The detection result uses exactly \wildXWildN{} image IDs in the original native-precision score file;
\wildXUnscoredN{} snapshot images have no score row and are outside this estimate. Membership in this snapshot
rests on self-report; the companion release has since added a host-attributed tier, whose provenance does not
depend on a caption.

We freeze the scored subset as \texttt{paper-b-wild-scored-20260911}, with image IDs, timestamps, platform,
file hashes and scores in \texttt{anc/wild-scored-manifest.csv}; its SHA-256 is
\texttt{b730b595\allowbreak 242dc0b2\allowbreak dd5884cc\allowbreak d1b1333b\allowbreak 5016cde6\allowbreak 55eff10c\allowbreak 912dc25c\allowbreak 4bf8bbd3}.
These fixed IDs, rather than the companion paper's later and larger release, define the reported sample; the
detection rate is not recomputed as that release grows.

The posted-image outcome is a flag rate at a threshold calibrated on authentic
images. Without verified per-image generator labels or authentic counterparts,
we do not report it as recall or AUC. On the path from our generations to posted JPEGs, about
\wildXGapContentPp{} points of the gap lie between our generations and un-re-encoded posted PNGs and
\wildXGapCodecPp{} between posted PNGs and posted JPEGs. The PNG subset is not content-matched (the one source that
preserves PNG is illustration-heavy), so this describes the gap rather than attributing it.

Pooled over images rather than cells, the controlled rate is \wildXCtrlPooledRate{} of \wildXCtrlPooledN{}
(Wilson 95\% \wildXCtrlPooledRateLo{}--\wildXCtrlPooledRateHi{}). The gap bears on whether the detection numbers
transfer to images posted online, not on the validity of the capability and review-check measurements.

\section{Limitations}
\label{app:limitations}

\paragraph{Scope.}
The study covers four forgery tasks, two Images~2.5 models and two
GPT-Image-2 quality baselines. Localisation refers to DocTamper on masked
receipt edits; whole-image detection uses Community Forensics. No test involves
faces. Results apply to one account and one week of API calls, with one output
per input because the API cannot be seeded. Parts of the analysis were
specified after results were known.

The principal limitations concern automatic measurement, crop-and-paste seams,
selected or synthetic inputs, and the analysis timeline. No human rated the
outputs. Claude's scorer spot checks do not cover the registered E1 OCR metric.
Photographed products remain untested, and the main E1 and E2a findings rely
largely on CORD because OCR coverage on WildReceipt is limited.

A nonsignificant result does not establish equality. This applies to refusal
rates, detection across quality tiers and the fine-print comparison. E4 measures
only lines above the OCR ceiling; those lines differ by arm and have high but
imperfect exact-read rates. One matched-height contrast does separate the
models, so the sharper-detail claim remains unsettled. Appendix~\ref{app:e4}
gives the glyph-height analysis; the detailed limitations below specify the
remaining boundaries.

\paragraph{Historical refusal comparison.}
Rounds v2 and v3 differ in generator, route, mask, prompt and corpus set.
The contemporaneous control holds the round-v3 procedure fixed, but it cannot
separate changes in request format from changes in serving or moderation since
round~v2. Both may contribute to the lower refusal rates. A distinguishing
condition, Images~2.5 with round~v2's drawn-box format, was not run.

The control has \docsControlAttempts{} attempts versus \docsAttempts{}
for Images~2.5, and its interval is the widest. The observed rates are of the
same order; the experiment does not establish equality
(Table~\ref{tab:refusals}).

\paragraph{Reduced judge panel and preliminary image-level AUC.} Round~v2's grid (non-expert humans,
TruFor, DocTamper and the generator as self-judge) was descoped for round~v3. A preliminary image-level DocTamper
number exists (\dtImgAuc{} [\dtImgAucCiLo{}, \dtImgAucCiHi{}] Flare, \dtSunImgAuc{} [\dtSunImgAucCiLo{},
\dtSunImgAucCiHi{}] Sunburst, against \dtImgNAuth{} authentic Images~2.5 document pages), but it aggregates each pixel
map by the 99th percentile of its pixel scores rather than by round~v2's rule, so it does not belong beside
round~v2's \vTwoDoctamper{}. Whether round-v3 forgeries are detected more or less often than round~v2's is open.

\paragraph{Detailed limitations.}
\begin{itemize}[leftmargin=*]
  \item \emph{Automatic measurement.} AI-agent spot checks do not cover the
  registered E1 token set. OCR, OWLv2 and homography checks have their own errors.
  OCR frequently fails below \capDetailCeilCapPx{}\,px, limiting E4's
  comparison of small text. E1 uses a calibrated false-change rate; E3 and E4
  use size-dependent OCR ceilings.
  \item \emph{Selection.} OCR reads only \capDocBaseWild{} E1 WildReceipt
  source fields, and E2a pre-selects fields legible at $t_0$. The main E1 and
  E2a findings therefore rely on CORD. E1's geometry filters apply before and
  after generation and change the evaluated population (Appendix~\ref{app:sizeguard}).
  \item \emph{References.} E3 uses programmatic packshots and an exploratory
  rendered set; photographed products and people are untested.
  \item \emph{Analysis timing.} E1, E3 and E4 primary contrasts and the
  statistical framework were selected after results were known. E1 images
  predate its design document. E2 rules preceded full-run E2a estimates but
  followed pilot results and completed conversational E2b scoring. The later
  \texttt{low}-arm addendum preceded its own data (Appendix~\ref{app:timing}).
  \item \emph{Post hoc additions.} The GPT-Image-2 \texttt{low} controls for
  E2a, E2b and E3 followed the main results, as did the sensitivity analyses
  listed in Appendix~\ref{app:timing}. E2b's umbrella threshold was revised
  after a pilot miss; only exploratory conversational decisions depend on it
  (Appendix~\ref{app:e2}).
  \item \emph{Price and timing.} No GPT-Image-2 tier bills \tokMed{} tokens.
  Its \texttt{low} tier (\tokLow{}) costs less and \texttt{medium}
  (\tokHigh{}) four times more. Low-arm E2a, E2b and E3 runs followed the
  comparison arms by about 11 hours.
  \item \emph{Execution route.} E2--E4 mix batch and synchronous calls,
  unevenly by model in stateless E2b (Table~\ref{tab:cells}). Latency uses
  synchronous calls only and does not measure concurrent throughput.
  E1 arms were not interleaved. The API results do not measure the ChatGPT
  interface; conversational E2b also uses \capChainConvTextModel{}, which
  can rewrite placement instructions.
  \item \emph{Equivalence.} DocTamper localisation is not equivalent to
  chance within $\pm$\eqDtMarginLo{} for any model. Community Forensics
  equivalence across tiers fails at $\pm$\eqTierMarginLo{}.
  \item \emph{TruFor coverage.} Only pages up to about
  \tfCovFlareMaxScoredLongEdge{}\,px on the long edge were scored.
  Missing shares are \tfCovFlareMissPct{}--\tfCovGitwoMedMissPct{}
  per configuration, so results are conditional on page size.
  \item \emph{Repeated re-editing.} The laundering cell has no control arms
  and is scored at only four turn positions.
  \item \emph{Format and transfer.} The photoreal real pool uses JPEG and
  TIFF against PNG generations, potentially contributing to AUCs of
  \tierAucFlatLo{}--\tierAucFlatHi{}. Posted images are self-attributed;
  their flag rate uses one calibrated threshold, and PNG/JPEG groups are not
  content-matched.
  \item \emph{Fine-print interpretation.} The above-ceiling band has high
  but imperfect OCR accuracy and different eligible lines by arm.
  Flare \texttt{max} separates from GPT-Image-2 \texttt{high} at matched
  glyph height (\capDetailVtwoMatchedFlareMaxVsGiHigh{}
  [\capDetailVtwoMatchedFlareMaxVsGiHighLo{},
  \capDetailVtwoMatchedFlareMaxVsGiHighHi{}]). The sharper-detail claim
  remains unsettled (Appendix~\ref{app:e4}).
\end{itemize}

\section{Data and Code Release}
\label{app:release}

\paragraph{Planned public release.}
The release will include E1 specifications and masks, E2 plans, E3 references
with ground truth and renderers, E4 prompts, and generation cells. It will also
include scorers, statistics and macro code, per-row outputs and cached OCR,
the de-identified call ledger, three public detectors' scores and harness code,
the Figure~\ref{fig:pairs} answer key, and dated design documents including
E2's freeze and the low-control addendum. Pointers and a fetch script will
provide CORD, WildReceipt, Columbia and IMD2020 inputs without re-hosting them.
These artifacts will support recomputation without new OCR or generation.

\paragraph{Planned access on request.} The generated images of E1--E4 and the photoreal core, and the paste-back pages the
detectors scored are planned to be shared with verified researchers under round~v2's dual-use policy. Requests should be directed to \href{mailto:benren@scam.ai}{benren@scam.ai}, stating affiliation and research purpose. The ledger export carries a SHA-256
hash per output. The API is not seedable, so regeneration will not reproduce these images; the planned per-row outputs will support recomputation of the reported estimates.

\paragraph{Not released.} Score files, logs and configurations of any detector other than the three public ones; API
keys, account identifiers and billing exports; internal drafts and audit notes. The generation cell accepts any
specification in the catalogue format, so restricting it to the fixed catalogue is a release policy, not a property of
the code. Table~\ref{tab:versions} pins model and instrument versions.

\begin{table}[tbp]
  \centering
  \caption{Model and instrument versions. The image API returns no dated snapshot; the call window is in the ledger.}
  \label{tab:versions}
  \footnotesize
  \setlength{\tabcolsep}{3pt}
  \begin{tabular}{@{}>{\raggedright\arraybackslash}p{0.27\columnwidth}>{\raggedright\arraybackslash}p{0.65\columnwidth}@{}}
    \toprule
    Component & Version \\
    \midrule
    Image models & \texttt{gpt-image-2}, \texttt{gpt-image-2.5-flare}, \texttt{gpt-image-2.5-sunburst} \\
    Relay (E2b conv.) & \texttt{gpt-5.4-nano-2026-03-17} \\
    OCR & easyocr 1.7.2, torch 2.10.0+cu128, English, GPU; E4 \texttt{mag\_ratio=1.0}; allowlist
          \texttt{ACDEFHKLMNPQRTUWXYZ} \\
    Grounding (E2b) & OWLv2 \path{google/owlv2-base-patch16-ensemble} \\
    TruFor & \texttt{trufor.pth.tar}, grip-unina/TruFor release v1.0 \\
    DocTamper & released checkpoint \texttt{\dtCheckpoint{}} \\
    Community Forensics & \texttt{buildborderless/}\allowbreak\texttt{CommunityForensics-}\allowbreak\texttt{DeepfakeDet-ViT}, input 224 \\
    \bottomrule
  \end{tabular}
\end{table}

\section{Release Scope and Measurement Window}
\label{app:dualuse}
\label{sec:ethics}

\paragraph{Release plan.}
Appendix~\ref{app:release} specifies public artifacts, access on request,
output hashes and exclusions under round~v2's dual-use policy.

\paragraph{Disclosure.} Before posting, we sent the refusal measurements and the same-week GPT-Image-2 comparison
to OpenAI.

\paragraph{Faces and identity.} No face data or photograph of an identifiable person is used, and the E2b
photographs contain no people. We did not test whether the models can place a real person's likeness into new
scenes; doing so would require consent from the people shown.

\paragraph{Measurement window.} These measurements record model behaviour during the test week. Subsequent changes to serving or safeguards may alter the results.

\end{document}